\documentclass[11pt,a4paper]{article}
\usepackage[utf8]{inputenc}

\usepackage{mathrsfs,amsthm,xcolor,verbatim,bbm,amsmath,amsfonts,amssymb,nicefrac,hyperref,bm,mathtools,xparse,etoolbox}
\usepackage[inline,shortlabels]{enumitem}
\usepackage[capitalise,sort]{cleveref} %

\crefname{enumi}{item}{items}
\usepackage[margin=1in]{geometry}
\usepackage[capitalise]{cleveref} 

\usepackage{stmaryrd} 

\crefname{enumi}{item}{items}
\crefname{equation}{}{}
\crefname{subsection}{Subsection}{Subsections}

\usepackage{xurl}

\hypersetup{
    colorlinks,
    linkcolor={red!80!black},
    citecolor={green},
    urlcolor={blue!80!black},
    linktocpage=true
}

\theoremstyle{plain}
\newtheorem{theorem}{Theorem}[section]

\newtheorem{prop}[theorem]{Proposition}

\newtheorem{setting}[theorem]{Setting}

\theoremstyle{definition}
\newtheorem{definition}[theorem]{Definition}

\DeclareMathAlphabet{\mathpzc}{OT1}{pzc}{m}{it}

\DeclareFontEncoding{LS1}{}{}
\DeclareFontSubstitution{LS1}{stix}{m}{n}
\DeclareMathAlphabet{\mathscr}{LS1}{stixscr}{m}{n}

\newcommand{\E}{\mathbb{E}}
\renewcommand{\P}{\mathbb{P}}

\newcommand{\R}{\mathbb{R}}
\newcommand{\N}{\mathbb{N}}
\newcommand{\Z}{\mathbb{Z}}

\newcommand{\Dens}{\mathscr{p}}

\newcommand{\inact}{\cI}

\renewcommand{\d}{ \mathrm{d}}

\renewcommand{\c}[1]{\mathfrak{c}^{#1}}
\newcommand{\neural}[4]{\mathcal{N}_{#1,#2}^{#3,#4}}
\newcommand{\neurali}[5]
{\mathcal{N}_{#1,#2,#3}^{#4,#5}}
\newcommand{\ssum}{\textstyle\sum}

\newcommand{\cLnri}[3]{\mathcal{L}_{#1}^{#2,#3}}

\newcommand{\cF}{\mathcal{F}}
\newcommand{\cG}{\mathcal{G}}

\newcommand{\cI}{\mathcal{I}}

\newcommand{\cL}{\mathcal{L}}

\newcommand{\bfc}{\mathbf{c}}

\newcommand{\bff}{\mathbf{f}}

\newcommand{\bfl}{\mathbf{l}}
\newcommand{\bfm}{\mathbf{m}}

\newcommand{\bfv}{\mathbf{v}}
\newcommand{\bfw}{\mathbf{w}}

\newcommand{\bfA}{\mathbf{A}}

\newcommand{\bfF}{\mathbf{F}}

\newcommand{\bfM}{\mathbf{M}}

\newcommand{\bfX}{\mathbf{X}}
\newcommand{\bfY}{\mathbf{Y}}

\newcommand{\scrA}{\mathscr{A}}
\newcommand{\scrB}{\mathscr{B}}
\newcommand{\scrC}{\mathscr{C}}

\newcommand{\scrM}{\mathscr{M}}

\newcommand{\scrR}{\mathscr{R}}

\newcommand{\boundexp}{\varrho}

\newcommand{\fA}{\mathfrak{A}}

\newcommand{\fG}{\mathfrak{G}}

\newcommand{\fJ}{\mathfrak{J}}

\newcommand{\fL}{\mathfrak{L}}
\newcommand{\fM}{\mathfrak{M}}

\newcommand{\fU}{\mathfrak{U}}

\newcommand{\fb}{\mathfrak{b}}

\newcommand{\fd}{\mathfrak{d}}

\newcommand{\fw}{\mathfrak{w}}

\newcommand{\bA}{\mathbb{A}}

\newcommand{\f}{f}
\newcommand{\F}{F}
\newcommand{\cki}{c}
\newcommand{\X}{X}
\newcommand{\Y}{Y}
\newcommand{\ZZ}{Z}
\newcommand{\T}{T}
\newcommand{\Snode}{S}
\newcommand{\scrc}{\mathscr{c}}

\def\mA{\mathcal A}
\def\mN{\mathcal N}

\newcommand{\nnode}{\mathbf{l}}
\newcommand{\arch}{l}
\newcommand{\InAct}{\mathrm{I}}
\newcommand{\InActL}[1]{\mathfrak{I}_{#1}}
\newcommand{\NNel}{\theta}

\newcommand{\NNelll}{\Theta}
\newcommand{\numf}{\delta}
\renewcommand{\emptyset}{\varnothing}

\newcommand{\W}{w}
\newcommand{\B}{B}

\newcommand{\permu}{\theta}

\newcommand{\defaultParamDim}{\mathfrak{d}}

\newcommand{\fnormal}{\psi}
\newcommand{\fcapital}{\Psi}
\newcommand{\fbold}{\phi}

\newcommand{\aaa}{b}
\newcommand{\bb}{a}

\DeclarePairedDelimiter{\abs}{\lvert}{\rvert}
\DeclarePairedDelimiter{\rbr}{(}{)}
\DeclarePairedDelimiter{\br}{[}{]}
\DeclarePairedDelimiter{\cu}{\{}{\}}
\DeclarePairedDelimiter{\spro}{\langle}{\rangle}

\newcommand{\ceil}[1]{ \left\lceil #1 \right\rceil}

\newcommand{\qandq}{\quad \text{and} \quad }
\newcommand{\qqandqq}{\qquad\text{and}\qquad}

\newcommand{\indicator}[1]{\mathbbm{1}_{\smash{#1}}}

\newcommand{\bbM}{\mathbb{M}}

\usepackage{cleveref}

\ExplSyntaxOn

\seq_new:N \g_cflist_loaded
\seq_new:N \g_cflist_pending

\NewDocumentCommand{\cfadd} { m } {
  \seq_if_in:NnF \g_cflist_loaded { #1 } {
    \seq_if_in:NnF \g_cflist_pending { #1 } {
      \seq_gput_right:Nn \g_cflist_pending { #1 }
    }
  }
}

\NewDocumentCommand{\cfconsiderloaded} { m } {
  \seq_gput_right:Nn \g_cflist_loaded {#1}
}

\NewDocumentCommand{\cfremove} { m } {
  \seq_gremove_all:Nn \g_cflist_pending { #1 }
}

\NewDocumentCommand{\cfload} { o } {
  \seq_if_empty:NTF \g_cflist_pending {
    \IfValueTF{#1}{\ignorespaces}{\unskip}
  } {
    (cf.\ \cref{\seq_use:Nn \g_cflist_pending {,}})\IfValueTF{#1}{#1~}{\unskip}
    \seq_gconcat:NNN \g_cflist_loaded \g_cflist_loaded \g_cflist_pending
    \seq_gclear:N \g_cflist_pending
    \IfValueT{#1}{\ignorespaces}
  }
}

\NewDocumentCommand{\cfclear} {} {
  \seq_gclear:N \g_cflist_loaded
  \seq_gclear:N \g_cflist_pending
}

\NewDocumentCommand{\cfout} { o } {
  \seq_if_empty:NTF \g_cflist_pending {\unskip\IfValueT{#1}{\ignorespaces}} {
    (cf.\ \cref{\seq_use:Nn \g_cflist_pending {,}})\IfValueTF{#1}{#1~}{\unskip}
    \seq_gclear:N \g_cflist_pending
    \IfValueT{#1}{\ignorespaces}
  }
}

\NewDocumentCommand{\ifnocf} { m } {
  \seq_if_empty:NT \g_cflist_pending { #1 }
}

\ExplSyntaxOff

\ExplSyntaxOn

\bool_new:N \g_noteobserve

\NewDocumentCommand{\setnote}{}{
  \bool_gset_true:N \g_noteobserve
}

\NewDocumentCommand{\setobserve}{}{
  \bool_gset_false:N \g_noteobserve
}

\NewDocumentCommand{\nobs}{ o }{
  \IfValueT{#1}{
    \str_if_eq:noTF {note} {#1} {
      \bool_gset_true:N \g_noteobserve
    } {
      \str_if_eq:noTF {Note} {#1} {
        \bool_gset_true:N \g_noteobserve
      } {
        \bool_gset_false:N \g_noteobserve
      }
    }
  }
  \bool_if:nTF { \g_noteobserve } {
    \bool_gset_false:N \g_noteobserve
    note
  } {
    \bool_gset_true:N \g_noteobserve
    observe
  }
  \IfValueF{#1}{~}
}

\NewDocumentCommand{\Nobs}{ o }{
  \IfValueT{#1}{
    \str_if_eq:noTF {note} {#1} {
      \bool_gset_true:N \g_noteobserve
    } {
      \str_if_eq:noTF {Note} {#1} {
        \bool_gset_true:N \g_noteobserve
      } {
        \bool_gset_false:N \g_noteobserve
      }
    }
  }
  \bool_if:nTF { \g_noteobserve } {
    \bool_gset_false:N \g_noteobserve
    Note
  } {
    \bool_gset_true:N \g_noteobserve
    Observe
  }
  \IfValueF{#1}{~}
}

\ExplSyntaxOff

\ExplSyntaxOn

\bool_new:N \g_hencetherefore

\NewDocumentCommand{\hence}{ o }{
  \IfValueT{#1}{
    \str_if_eq:noTF {hence} {#1} {
      \bool_gset_true:N \g_hencetherefore
    } {
      \str_if_eq:noTF {Hence} {#1} {
        \bool_gset_true:N \g_hencetherefore
      } {
        \bool_gset_false:N \g_hencetherefore
      }
    }
  }
  \bool_if:nTF { \g_hencetherefore } {
    \bool_gset_false:N \g_hencetherefore
    hence
  } {
    \bool_gset_true:N \g_hencetherefore
    therefore
  }
  \IfValueF{#1}{~}
}

\NewDocumentCommand{\Hence}{ o }{
  \IfValueT{#1}{
    \str_if_eq:noTF {hence} {#1} {
      \bool_gset_true:N \g_hencetherefore
    } {
      \str_if_eq:noTF {Hence} {#1} {
        \bool_gset_true:N \g_hencetherefore
      } {
        \bool_gset_false:N \g_hencetherefore
      }
    }
  }
  \bool_if:nTF { \g_hencetherefore } {
    \bool_gset_false:N \g_hencetherefore
    Hence,~we~obtain
  } {
    \bool_gset_true:N \g_hencetherefore
    Therefore,~we~obtain
  }
  \IfValueF{#1}{~}
}

\ExplSyntaxOff

\ExplSyntaxOn

\seq_const_from_clist:Nn \g_prove_mru {
  establish,
  demonstrate,
  prove,
  show,
  imply,
  ensure
}

\prop_new:N \l__verbs
\prop_put:Nnn \l__verbs {show} {shows}
\prop_put:Nnn \l__verbs {imply} {implies}
\prop_put:Nnn \l__verbs {demonstrate} {demonstrates}
\prop_put:Nnn \l__verbs {prove} {proves}
\prop_put:Nnn \l__verbs {establish} {establishes}
\prop_put:Nnn \l__verbs {ensure} {ensures}
\prop_put:Nnn \l__verbs {assure} {assures}

\tl_new:N \g_wordtmp
\seq_new:N \l_mytmps

\cs_generate_variant:Nn \str_if_in:nnTF { nVTF }
\cs_generate_variant:Nn \str_if_in:nnTF { xVTF }

\NewDocumentCommand{\prove}{ o }{
  \IfValueTF{#1}{
    \seq_clear:N \l_mytmps
    \seq_map_inline:Nn \g_prove_mru {
      \str_if_eq:nnTF {##1} {ensure} {
        \str_set:Nn \l_temps {n}
      } {
        \str_set:Nx \l_temps {\str_head_ignore_spaces:n {##1}}
      }
      \str_if_in:xVTF {#1} \l_temps {
        \seq_put_right:Nn \l_mytmps {##1}
      } { }
    }
    \seq_get_right:NN \l_mytmps \g_wordtmp
  } {
    \seq_get_right:NN \g_prove_mru \g_wordtmp
  }
  \tl_use:N \g_wordtmp
  \IfValueTF{#1}{}{~}
  \seq_gput_left:NV \g_prove_mru \g_wordtmp
  \seq_gremove_duplicates:N \g_prove_mru
}

\NewDocumentCommand{\proves}{ o }{
  \IfValueTF{#1}{
    \seq_clear:N \l_mytmps
    \seq_map_inline:Nn \g_prove_mru {
      \str_if_eq:nnTF {##1} {ensure} {
        \str_set:Nn \l_temps {n}
      } {
        \str_set:Nx \l_temps {\str_head_ignore_spaces:n {##1}}
      }
      \str_if_in:xVTF {#1} \l_temps {
        \seq_put_right:Nn \l_mytmps {##1}
      } { }
    }
    \seq_get_right:NN \l_mytmps \g_wordtmp
  } {
    \seq_get_right:NN \g_prove_mru \g_wordtmp
  }
  \str_set:NV \l_tmpa_str \g_wordtmp
  \prop_get:NVN \l__verbs \l_tmpa_str \l_tmpa_tl
  \tl_use:N \l_tmpa_tl
  \IfValueTF{#1}{}{~}
  \seq_gput_left:NV \g_prove_mru \g_wordtmp
  \seq_gremove_duplicates:N \g_prove_mru
}

\newcommand{\llabel}[1]{\savelabel{#1}\label{\loc.#1}\ignorespaces}

\clist_new:N \l_localreflist
\clist_new:N \l_reflist

\NewDocumentCommand{\lref} { m } {
  \clist_set:No \l_localreflist {#1}
  \clist_clear:N \l_reflist
  \clist_map_inline:Nn \l_localreflist { \clist_put_right:Nn \l_reflist {\loc.##1} }
  \cref{\l_reflist}
}

\NewDocumentCommand{\Lref} { m } {
  \clist_set:No \l_localreflist {#1}
  \clist_clear:N \l_reflist
  \clist_map_inline:Nn \l_localreflist { \clist_put_right:Nn \l_reflist {\loc.##1} }
  \Cref{\l_reflist}
}

\NewDocumentCommand{\itref}{ m m }{
  \clist_set:No \l_localreflist {#2}
  \clist_clear:N \l_reflist
  \clist_map_inline:Nn \l_localreflist { \clist_put_right:Nn \l_reflist {#1.##1} }
  \cref{\l_reflist}~in~\cref{#1}
}

\seq_new:N \l_enum_seq
\int_new:N \l_num_items

\bool_new:N \g_commaused_bool

\providecommand{\comma}{}

\cs_new:Nn \enum_it:nn {
  \int_case:nnF {\l_num_items - #1} {
    {0} {
      \renewcommand{\comma}{}
      #2\space
    }
    {1} {
      \bool_gset_false:N \g_commaused_bool
      \renewcommand{\comma}{,~\bool_gset_true:N \g_commaused_bool}
      #2
      \bool_if:NTF \g_commaused_bool {} {,~}
      and~
    }
  } {
    \bool_gset_false:N \g_commaused_bool
    \renewcommand{\comma}{,~\bool_gset_true:N \g_commaused_bool}
    #2
    \bool_if:NTF \g_commaused_bool {} {,~}
  }
}

\cs_new:Nn \enum_it_U:nn {
  \int_case:nnF {\l_num_items - #1} {
    {0} {
      \renewcommand{\comma}{}
      #2
      \space
    }
    {1} {
      \bool_gset_false:N \g_commaused_bool
      \renewcommand{\comma}{,~\bool_gset_true:N \g_commaused_bool}
      #2
      \bool_if:NTF \g_commaused_bool {} {,~}
      and~
    }
  } {
    \bool_gset_false:N \g_commaused_bool
    \renewcommand{\comma}{,~\bool_gset_true:N \g_commaused_bool}
    \int_compare:nTF {#1=1} {
      \text_titlecase_first:n {#2}
    } {
      #2
    }
    \bool_if:NTF \g_commaused_bool {} {,~}
  }
}

\cs_new:Nn \uc:n {\MFUsentencecase{#1}}
\cs_generate_variant:Nn \uc:n {x}

\cs_generate_variant:Nn \str_uppercase:n {x}
\cs_generate_variant:Nn \tl_if_eq:nnTF {onTF}

\cs_new:Nn \enum:nnnn {
  \seq_set_split:Nnn \l_enum_seq ; {#1}
  \seq_remove_all:Nn \l_enum_seq { }
  \seq_remove_all:Nn \l_enum_seq {#2}
  \seq_log:N \l_enum_seq
  \int_set:Nn \l_num_items {\seq_count:N \l_enum_seq}
  \int_log:N \l_num_items
  \int_case:nnF {\l_num_items} {
    { 0 } { 0 }
    { 1 } {
      \IfBooleanTF{#4} {
        \tl_set:Nn \l_text_case_exclude_arg_tl {\cref}
        \text_titlecase_first:n {\seq_use:Nn \l_enum_seq {}}
      } {
        \seq_use:Nn \l_enum_seq {}
      }
      \space
      \tl_if_eq:onTF{#3}{-}{}{
        \bool_if:NTF \l_plural_bool {
          \prove[#3]~
        } {
          \proves[#3]~
        }
      }
    }
    { 2 } {
      \IfBooleanTF{#4} {
        \tl_set:Nn \l_text_case_exclude_arg_tl {\cref}
        \text_titlecase_first:n {\seq_use:Nn \l_enum_seq {~and~}}
      } {
        \seq_use:Nn \l_enum_seq {~and~}
      }
      \space
      \tl_if_eq:onTF{#3}{-}{}{
        \prove[#3]~
      }
    }
  } {
    \IfBooleanTF{#4} {
      \tl_set:Nn \l_text_case_exclude_arg_tl {\cref}
      \seq_indexed_map_function:NN \l_enum_seq \enum_it_U:nn
    } {
      \seq_indexed_map_function:NN \l_enum_seq \enum_it:nn
    }
    \tl_if_eq:onTF{#3}{-}{}{
      \prove[#3]~
    }
  }
}

\cs_generate_variant:Nn \enum:nnnn {nxnn}
\cs_generate_variant:Nn \enum:nnnn {nxxn}

\NewDocumentCommand{\enum}{O{} m O{-} s}{
  \IfBooleanTF{#4}{
    \enum:nxnn {#2} {#1} {sindep} \BooleanFalse
  } {
    \enum:nxxn {#2} {#1} {#3} \BooleanFalse
  }
}

\NewDocumentCommand{\dott}{}{\ifnocf{.}\space}

\bool_new:N \g_arg_start_bool
\bool_gset_true:N \g_arg_start_bool

\NewDocumentCommand{\startnewargseq}{}{\bool_gset_true:N \g_arg_start_bool \tl_set:Nn \g_label_tl {}}

\cs_generate_variant:Nn \seq_if_in:NnTF {NxTF}
\cs_generate_variant:Nn \seq_remove_all:Nn {Nx}

\int_new:N \l_random_int

\cs_generate_variant:Nn \tl_if_head_eq_catcode:nNTF {oNTF}
\cs_generate_variant:Nn \tl_if_head_eq_catcode:nNTF {VNTF}

\cs_generate_variant:Nn \tl_if_head_eq_catcode:nNTF {eNTF}
\cs_generate_variant:Nn \tl_log:n {o}
\cs_generate_variant:Nn \tl_log:n {f}
\cs_generate_variant:Nn \tl_log:n {x}
\cs_generate_variant:Nn \tl_log:n {e}

\cs_generate_variant:Nn \tl_if_in:nnTF {onTF}
\cs_generate_variant:Nn \tl_if_in:NnTF {NeTF}

\cs_generate_variant:Nn \tl_if_head_eq_meaning:nNTF {VNTF}

\seq_const_from_clist:Nn \g_arg_mru_this {
  Ahpr,
  Tapr,
  Ctapr,
  H
}

\seq_const_from_clist:Nn \g_arg_mru_nothis {
  Ia,
  Nwc,
  N,
  Itns,
  Fm,
  Itnswc,
  Mo
}

\seq_new:N \l_arg_seq
\tl_new:N \l_cons_tl
\tl_new:N \l_dummy_tl

\bool_new:N \g_debug_bool
\bool_gset_false:N \g_debug_bool

\bool_new:N \l_insidearg_bool

\bool_new:N \g_firstargletter_bool

\sys_gset_rand_seed:n {0903}

\bool_new:N \l_plural_bool
\tl_new:N \l_arg_verbs_tl

\NewDocumentCommand{\argument}{mom}{
\color{black}
  \bool_set_false:N \l_plural_bool
  \tl_set:Nn \l_arg_verbs_tl {sindep}
  \keys_define:nn { benno/argument } {
    plural .value_forbidden:n = true,
    plural .code:n = {\bool_set_true:N \l_plural_bool},
    verbs .value_required:n = false,
    verbs .tl_set:N = \l_arg_verbs_tl,
  }
  \IfValueT{#2}{
    \keys_set:nn { benno/argument } {#2}
  }
  \bool_log:N \l_plural_bool
  \bool_gset_true:N \l_insidearg_bool
  \seq_set_split:Nnn \l_arg_seq ; {#1}
  \seq_remove_all:Nn \l_arg_seq { }
  \seq_log:N \l_arg_seq
  \tl_set:Nn \l_cons_tl {#3}
  \tl_trim_spaces:N \l_cons_tl
  \seq_if_in:NxTF \l_arg_seq {\lref{\g_label_tl}} {
    \seq_remove_all:Nx \l_arg_seq {\lref{\g_label_tl}}
    \seq_get_left:NNTF \l_arg_seq \l_dummy_tl {
      \tl_trim_spaces:N \l_dummy_tl
      \bool_gset_false:N \g_firstargletter_bool
      \tl_if_head_eq_catcode:VNTF \l_dummy_tl a {
        \bool_gset_true:N \g_firstargletter_bool
      } {
        \tl_if_head_eq_meaning:VNTF \l_dummy_tl {\cref} {
          \tl_set:Nx \l_tmpa_tl {\tl_tail:N \l_dummy_tl}
          \tl_set:Nx \l_tmpb_tl {\tl_head:N \l_tmpa_tl}
          \bool_gset_true:N \g_firstargletter_bool
          \tl_if_in:NeTF \l_tmpb_tl {lem\c_colon_str} {} {
            \tl_if_in:NeTF \l_tmpb_tl {thm\c_colon_str} {} {
              \tl_if_in:NeTF \l_tmpb_tl {prop\c_colon_str} {} {
                \tl_if_in:NeTF \l_tmpb_tl {cor\c_colon_str} {} {
                  \bool_gset_false:N \g_firstargletter_bool
                }
              }
            }
          }
        } {
        }
      }
      \bool_if:NTF \g_firstargletter_bool {
        \seq_set_eq:NN \l_tmpa_seq \g_arg_mru_this
        \seq_remove_all:Nn \l_tmpa_seq {H}
        \seq_get_right:NN \l_tmpa_seq \l_tmpa_tl
        \int_case:nnF {\seq_count:N \l_arg_seq} {
          {1} {
            \str_case:VnF {\l_tmpa_tl} {
              {Ahpr} {
                \bool_if:NT \g_debug_bool {C1.1}
                \seq_gput_left:Nn \g_arg_mru_this {Ahpr}
                \seq_gremove_duplicates:N \g_arg_mru_this
                \enum:nxnn {#1} {\lref{\g_label_tl}} {-} {\BooleanTrue}
                \hence~
                \bool_if:NTF \l_plural_bool {
                  \prove[\l_arg_verbs_tl]~\ignorespaces #3
                } {
                  \proves[\l_arg_verbs_tl]~\ignorespaces #3
                }
              }
              {Tapr} {
                \bool_if:NT \g_debug_bool {C1.2}
                \seq_gput_left:Nn \g_arg_mru_this {Tapr}
                \seq_gremove_duplicates:N \g_arg_mru_this
                \enum[\lref{\g_label_tl}]{
                  This;
                  #1
                }[\l_arg_verbs_tl]\ignorespaces #3
              }
              {Ctapr} {
                \bool_if:NT \g_debug_bool {C1.3}
                \seq_gput_left:Nn \g_arg_mru_this {Ctapr}
                \seq_gremove_duplicates:N \g_arg_mru_this
                Combining~
                \enum[\lref{\g_label_tl}]{
                  this;
                  #1
                } \proves[\l_arg_verbs_tl]~\ignorespaces #3
              }
            } {}
          }
        } {
          \str_case:VnF {\l_tmpa_tl} {
             {Ahpr} {
              \bool_if:NT \g_debug_bool {C2.1}
              \seq_gput_left:Nn \g_arg_mru_this {Ahpr}
              \seq_gremove_duplicates:N \g_arg_mru_this
              \enum:nxnn {#1} {\lref{\g_label_tl}} {-} {\BooleanTrue}
              \hence~
              \prove[\l_arg_verbs_tl]~\ignorespaces #3
            }
            {Tapr} {
              \bool_if:NT \g_debug_bool {C2.2}
              \seq_gput_left:Nn \g_arg_mru_this {Tapr}
              \seq_gremove_duplicates:N \g_arg_mru_this
              \enum[\lref{\g_label_tl}]{
                This;
                #1
              }[\l_arg_verbs_tl]\ignorespaces #3
            }
            {Ctapr} {
              \int_case:nn {\int_rand:nn {0} {1}} {
                {0} {
                  \bool_if:NT \g_debug_bool {C2.3}
                  \seq_gput_left:Nn \g_arg_mru_this {Ctapr}
                  \seq_gremove_duplicates:N \g_arg_mru_this
                  Combining~
                  \enum[\lref{\g_label_tl}]{
                    this;
                    #1
                  } \proves[\l_arg_verbs_tl]~\ignorespaces #3
                }
                {1} {
                  \bool_if:NT \g_debug_bool {C2.4}
                  \seq_gput_left:Nn \g_arg_mru_this {Ctapr}
                  \seq_gremove_duplicates:N \g_arg_mru_this
                  Combining~
                  \enum:nxnn {#1} {\lref{\g_label_tl}} {-} {\BooleanFalse}
                  \hence~
                  \proves[\l_arg_verbs_tl]~\ignorespaces #3
                }
              }
            }
          } {}
        }
      } {
        \seq_set_eq:NN \l_tmpa_seq \g_arg_mru_this
        \seq_remove_all:Nn \l_tmpa_seq {H}
        \seq_remove_all:Nn \l_tmpa_seq {Ahpr}
        \seq_get_right:NN \l_tmpa_seq \l_tmpa_tl
        \int_case:nnF {\seq_count:N \l_arg_seq} {
          {1} {
            \str_case:VnF {\l_tmpa_tl} {
              {Tapr} {
                \bool_if:NT \g_debug_bool {C3.1}
                \seq_gput_left:Nn \g_arg_mru_this {Tapr}
                \seq_gremove_duplicates:N \g_arg_mru_this
                \enum[\lref{\g_label_tl}]{
                  This;
                  #1
                }[\l_arg_verbs_tl]\ignorespaces #3
              }
              {Ctapr} {
                \bool_if:NT \g_debug_bool {C3.2}
                \seq_gput_left:Nn \g_arg_mru_this {Ctapr}
                \seq_gremove_duplicates:N \g_arg_mru_this
                Combining~
                \enum[\lref{\g_label_tl}]{
                  this;
                  #1
                } \proves[\l_arg_verbs_tl]~\ignorespaces #3
              }
            } {}
          }
        } {
          \str_case:VnF {\l_tmpa_tl} {
            {Tapr} {
              \bool_if:NT \g_debug_bool {C4.1}
              \seq_gput_left:Nn \g_arg_mru_this {Tapr}
              \seq_gremove_duplicates:N \g_arg_mru_this
              \enum[\lref{\g_label_tl}]{
                This;
                #1
              }[\l_arg_verbs_tl]\ignorespaces #3		
            }
            {Ctapr} {
              \int_case:nn {\int_rand:nn {0} {1}} {
                {0} {
                  \bool_if:NT \g_debug_bool {C4.2}
                  \seq_gput_left:Nn \g_arg_mru_this {Ctapr}
                  \seq_gremove_duplicates:N \g_arg_mru_this
                  Combining~
                  \enum[\lref{\g_label_tl}]{
                    this;
                    #1
                  } \proves[\l_arg_verbs_tl]~\ignorespaces #3		
                }
                {1} {
                  \bool_if:NT \g_debug_bool {C4.3}
                  \seq_gput_left:Nn \g_arg_mru_this {Ctapr}
                  \seq_gremove_duplicates:N \g_arg_mru_this
                  Combining~
                  \enum:nxnn {#1} {\lref{\g_label_tl}} {-} {\BooleanFalse}
                  \hence~
                  \proves[\l_arg_verbs_tl]~\ignorespaces #3    
                }
              }
            }
          } {}
        }
      }
    } {
      \tl_if_head_eq_catcode:oNTF \l_cons_tl a {
        \seq_set_eq:NN \l_tmpa_seq \g_arg_mru_this
        \seq_remove_all:Nn \l_tmpa_seq {Ctapr}
        \seq_remove_all:Nn \l_tmpa_seq {Ahpr}
        \seq_get_right:NN \l_tmpa_seq \l_tmpa_tl
        \str_case:VnF {\l_tmpa_tl} {
          {H} {
            \bool_if:NT \g_debug_bool {C5.1}
            \seq_gput_left:Nn \g_arg_mru_this {H}
            \seq_gremove_duplicates:N \g_arg_mru_this
            Hence,~we~obtain~\ignorespaces #3
          }
          {Tapr} {
            \bool_if:NT \g_debug_bool {C5.2}
            \seq_gput_left:Nn \g_arg_mru_this {Tapr}
            \seq_gremove_duplicates:N \g_arg_mru_this
            This~\proves[\l_arg_verbs_tl]~\ignorespaces #3
          }
        } {}
      } {
        \bool_if:NT \g_debug_bool {C6.1}
        \seq_gput_left:Nn \g_arg_mru_this {Tapr}
        \seq_gremove_duplicates:N \g_arg_mru_this
        This~\proves[\l_arg_verbs_tl]~\ignorespaces #3
      }
    } 
  } {
    \int_compare:nNnTF {\seq_count:N \l_arg_seq} = {0} {
      \bool_if:NTF \g_arg_start_bool {
        \bool_if:NT \g_debug_bool {C7.1}
        \Nobs\unskip
        #3
      } {
        \bool_if:NT \g_debug_bool {C7.2}
        \Moreover~
        #3
      }
    } {
      \bool_if:NTF \g_arg_start_bool {
        \bool_if:NT \g_debug_bool {C8.1}
        \tl_log:N \l_arg_verbs_tl
        \Nobs~that~
        \enum{
          #1
        }[\l_arg_verbs_tl]\ignorespaces #3
      } {
        \int_compare:nNnTF {\seq_count:N \l_arg_seq} = {1} {
          \seq_set_eq:NN \l_tmpa_seq \g_arg_mru_nothis
          \seq_remove_all:Nn \l_tmpa_seq {Nwc}
          \seq_remove_all:Nn \l_tmpa_seq {Itnswc}
          \seq_get_right:NN \l_tmpa_seq \l_tmpa_tl
        } {
          \seq_get_right:NN \g_arg_mru_nothis \l_tmpa_tl
        }
        \str_case:VnF {\l_tmpa_tl} {
          {Mo} {
            \bool_if:NT \g_debug_bool {C9.1}
            \seq_gput_left:Nn \g_arg_mru_nothis {Mo}
            \seq_gremove_duplicates:N \g_arg_mru_nothis
            Moreover,~\nobs~that~
            \enum{
              #1
            }[\l_arg_verbs_tl]\ignorespaces #3		
          }
          {Fm} {
            \bool_if:NT \g_debug_bool {C9.2}
            \seq_gput_left:Nn \g_arg_mru_nothis {Fm}
            \seq_gremove_duplicates:N \g_arg_mru_nothis
            Furthermore,~\nobs~that~
            \enum{
              #1
            }[\l_arg_verbs_tl]\ignorespaces #3		
          }
          {Ia} {
            \bool_if:NT \g_debug_bool {C9.3}
            \seq_gput_left:Nn \g_arg_mru_nothis {Ia}
            \seq_gremove_duplicates:N \g_arg_mru_nothis
            In~addition,~\nobs~that~
            \enum{
              #1
            }[\l_arg_verbs_tl]\ignorespaces #3		
          }
          {N} {
            \bool_if:NT \g_debug_bool {C9.4}
            \seq_gput_left:Nn \g_arg_mru_nothis {N}
            \seq_gremove_duplicates:N \g_arg_mru_nothis
            Next,~\nobs~that~
            \enum{
              #1
            }[\l_arg_verbs_tl]\ignorespaces #3		
          }
          {Itns} {
            \bool_if:NT \g_debug_bool {C9.5}
            \seq_gput_left:Nn \g_arg_mru_nothis {Itns}
            \seq_gremove_duplicates:N \g_arg_mru_nothis
            In~the~next~step~we~\nobs~that~
            \enum{
              #1
            }[\l_arg_verbs_tl]\ignorespaces #3		
          }
          {Nwc} {
            \bool_if:NT \g_debug_bool {C9.6}
            \seq_gput_left:Nn \g_arg_mru_nothis {Nwc}
            \seq_gremove_duplicates:N \g_arg_mru_nothis
            Next~we~combine~
            \enum{
              #1
            }to~obtain~\ignorespaces #3
          }
          {Itnswc} {
            \bool_if:NT \g_debug_bool {C9.7}
            \seq_gput_left:Nn \g_arg_mru_nothis {Itnswc}
            \seq_gremove_duplicates:N \g_arg_mru_nothis
            In~the~next~step~we~combine~
            \enum{
              #1
            }to~obtain~\ignorespaces #3
          }
        } {}
      }
    }
  }
  \bool_gset_false:N \g_arg_start_bool
  \bool_gset_false:N \l_insidearg_bool
  \cfload[.]
  \color{black}
}

\tl_new:N \g_label_tl
\tl_gset:Nn \g_label_tl { }

\NewDocumentCommand{\savelabel}{m}{
  \bool_if:NTF \l_insidearg_bool {
    \tl_gset:Nn \g_label_tl {#1}
  } {
    \tl_gset:Nn \g_label_tl { }
  }
}

\ExplSyntaxOff

\ExplSyntaxOn

\NewDocumentEnvironment {athm} {m m o} {
\str_if_eq:noTF {example} {#1} {
  \bool_gset_true:N \g_example_bool
} {
  \bool_gset_false:N \g_example_bool
}
\cfclear
\IfNoValueTF{#3}{
\begin{#1}\label{#2}\global\def\loc{#2}
}{
\begin{#1}[#3]\label{#2}\global\def\loc{#2}
}
}{
\end{#1}
}

\NewDocumentEnvironment {adef} {m} {
\begin{definition}\label{#1}\global\def\loc{#1}
}{
\end{definition}
}

\NewDocumentEnvironment{aproof} {} {
\bool_if:NTF \g_example_bool {
  \bool_gset_true:N \g_arg_start_bool
  \begin{proof}[Proof~for~\cref{\loc}]
} {
  \bool_gset_true:N \g_arg_start_bool
  \begin{proof}[Proof~of~\cref{\loc}]
}
\bool_gset_false:N \g_finishproof_bool
}{
\bool_if:NTF \g_finishproof_bool {}
{\finishproofthus}
\end{proof}
}

\NewDocumentCommand{\finishproofthus} {} {
  \bool_gset_true:N \g_finishproof_bool 
  \bool_if:NTF \g_example_bool {
    The~proof~for~\cref{\loc}~is~thus~complete.
  } {
    The~proof~of~\cref{\loc}~is~thus~complete.
  }
}
\NewDocumentCommand{\finishproofthis} {} {
  \bool_gset_true:N \g_finishproof_bool 
  \bool_if:NTF \g_example_bool {
    This~completes~the~proof~for~\cref{\loc}.
  } {
    This~completes~the~proof~of~\cref{\loc}.
  }
}

\ExplSyntaxOff

\NewDocumentEnvironment{cproof}{m}
{\begin{proof}[Proof of \cref{#1}]}%
{\noindent The proof of \cref{#1} is thus complete.
\end{proof}}

\NewDocumentEnvironment{cproof2}{m}
{\begin{proof}[Proof of \cref{#1}]}%
{\noindent This completes the proof of \cref{#1}.
\end{proof}}

\hypersetup{
    colorlinks,
    linkcolor={blue!80!black},
    citecolor={green},
    urlcolor={blue!80!black}
}

\makeatletter
\ExplSyntaxOn
\seq_new:N \g_abbrs
\prop_new:N \g_abbr_counts
\tl_new:N \l_abbr_count_tl

\ExplSyntaxOn

\bool_new:N \g_forexample

\NewDocumentCommand{\eg}{ o }{
	\IfValueT{#1}{
		\str_if_eq:noTF {fe} {#1} {
			\bool_gset_true:N \g_forexample
		} {\bool_gset_false:N \g_forexample}
	}
	\bool_if:nTF { \g_forexample } {
		\bool_gset_false:N \g_forexample
		for~example
	}{
		\bool_gset_true:N \g_forexample
		for~instance
	}
}

\NewDocumentCommand{\abbr}{m m O{#1} m m O{#4} m}{
	\expandafter\newcommand\csname#3\endcsname[1][]{
		\seq_if_in:NnTF \g_abbrs {#1} {
			\prop_get:NnN \g_abbr_counts {#1} \l_abbr_count_tl
			\prop_gput:Nnx \g_abbr_counts {#1} {\int_eval:n {\l_abbr_count_tl + 1}}
			\hyperref[#1]{#7}
		} {
			\seq_gput_left:Nn \g_abbrs {#1}
			\prop_gput:Nnn \g_abbr_counts {#1} {1}
			\expandafter\gdef\csname#1@def\endcsname{#2}
			\phantomsection\label{#1}
			\str_if_eq:nnTF{##1}{}{\emph{#2}}{##1}~(\hyperref[#1]{#7})
		}
	}
	\expandafter\newcommand\csname#6\endcsname[1][]{
		\seq_if_in:NnTF \g_abbrs {#1} {
			\prop_get:NnN \g_abbr_counts {#1} \l_abbr_count_tl
			\prop_gput:Nnx \g_abbr_counts {#1} {\int_eval:n {\l_abbr_count_tl + 1}}
			\hyperref[#1]{#4}
		} {
			\expandafter\gdef\csname#1@def\endcsname{#5}
			\seq_gput_left:Nn \g_abbrs {#1}
			\prop_gput:Nnn \g_abbr_counts {#1} {1}
			\phantomsection\label{#1}
			\str_if_eq:nnTF{##1}{}{\emph{#5}}{##1}~(\hyperref[#1]{#4})
		}
	}
}

\ExplSyntaxOff
\makeatother
\abbr{Adamax}{adaptive moment estimation maximum SGD}{Adamaxs}{Adaptive moment estimation maximum SGD}{Adamax}
\abbr{Adagrad}{adaptive gradient SGD}{Adagrads}{adaptive gradient SGD}{Adagrad}
\abbr{Adam}{adaptive moment estimation SGD}{Adams}{adaptive moment estimation SGD}{Adam}
\abbr{Nadam}{Nesterov accelerated adaptive moment estimation SGD}{Nadams}{Nesterov accelerated adaptive moment estimation SGD}{Nadam}
\abbr{RMSprop}{root mean square propagation SGD}{RMSprops}{the root mean square propagation}{RMSprop}
\abbr{ANN}{artificial neural network}{ANNs}{artificial neural networks}{ANN}
\abbr{DNN}{deep artificial neural network}{DNNs}{deep artificial neural networks}{DNN}
\abbr{SGD}{stochastic gradient descent}{SGDs}{stochastic gradient descent}{SGD}
\abbr{iid}{independent and identically distributed}{i.i.d}{independent and identically distributed}{i.i.d.}
\abbr{DL}{Deep learning}{DLs}{Deep learning}{DL}
\abbr{AI}{artificial intelligence}{AIs}{artificial intelligence}{AI}
\abbr{LLM}{large language model}{LLMs}{large language models}{LLM}
\abbr{PDE}{partial differential equation}{PDEs}{partial differential equations}{PDE}
\abbr{as}{almost surely}{ass}{almost surely}{a.s.}
\abbr{pas}{$\P$-almost surely}{pass}{almost surely}{$\P$-a.s.}
\abbr{ReLUs}{rectified linear unit}{ReLU}{rectified linear unit}
\makeatother

\title{
Non-convergence to global minimizers in data driven\\
supervised deep learning: Adam and stochastic
gradient descent\\ optimization provably fail to
converge to global minimizers in \\the training of
deep neural networks with ReLU activation
}
\author{Thang Do$^{1,2}$, Sonja Hannibal$^{3}$, and Arnulf Jentzen$^{4,5}$
	\bigskip
	\\
    \small{$^1$ School of Data Science, The Chinese University of Hong Kong, Shenzhen}
	\vspace{-0.1cm}\\
	\small{ (CUHK-Shenzhen), China, e-mail: \texttt{dmthang@cuhk.edu.cn}}
 \smallskip
	\\
 \small{$^2$ Department of Probability and Statistic, Institute of Mathematics,}
	\vspace{-0.1cm}\\
	\small{Vietnam Academy of Science and Technology, Vietnam, e-mail: \texttt{dmthang@math.ac.vn}}
	\smallskip
	\\
	\small{$^3$ Applied Mathematics: Institute for Analysis and Numerics, Faculty of Mathematics and}
	\vspace{-0.1cm}\\
	\small{Computer Science, University of M{\"u}nster, Germany, e-mail: \texttt{sonja.hannibal@uni-muenster.de}}
	\smallskip
	\\
	\small{$^4$ School of Data Science and Shenzhen Research Institute of Big Data, The Chinese University}
	\vspace{-0.1cm}\\
	\small{of Hong Kong, Shenzhen (CUHK-Shenzhen), China, e-mail: \texttt{ajentzen@cuhk.edu.cn}}
	\smallskip
	\\
 \small{$^5$ Applied Mathematics: Institute for Analysis and Numerics, Faculty of Mathematics and}
	\vspace{-0.1cm}\\
	\small{Computer Science, University of M{\"u}nster, Germany, e-mail: \texttt{ajentzen@uni-muenster.de}}
	\smallskip
	\\
}

\date{\today}
\allowdisplaybreaks 
\begin{document}
\maketitle
\begin{abstract}
    Deep learning (DL) methods -- consisting of a class of deep neural networks (DNNs) trained by a stochastic gradient descent (SGD) optimization method -- are nowadays key tools to solve data driven supervised learning problems. Despite the great success of SGD methods in the training of DNNs, it remains a fundamental open problem of research to explain the success and the limitations of such methods in rigorous theoretical terms. In particular, even in the standard setup of data driven supervised learning problems, it remained an open research problem to prove (or disprove) that SGD methods converge in the training of DNNs with the popular rectified linear unit (ReLU) activation function with high probability to global minimizers in the optimization landscape. In this work we answer this question negatively by proving that it does not hold that SGD methods converge with high probability to global minimizers of the objective function. Even stronger, in this work we prove for a large class of SGD methods that the considered optimizer does with high probability not converge to global minimizers of the optimization problem. It turns out that the probability to not converge to a global minimizer converges at least exponentially quickly to one as the width of the first hidden layer of the ANN (the number of neurons on the first hidden layer) and the depth of the ANN (the number of hidden layers), respectively, increase to infinity. The general non-convergence results of this work do not only apply to the plain vanilla standard SGD method but also to a large class of accelerated and adaptive SGD methods such as the momentum SGD, the Nesterov accelerated SGD, the Adagrad, the RMSProp, the Adam, the Adamax, the AMSGrad, and the Nadam optimizers. However, we would like to emphasize that the findings of this work do not imply that SGD methods do not succeed to train DNNs: it may still very well be the case that SGD methods provably succeed to train DNNs in data driven learning problems but, if this does hold, then the explanation for this can, due to the findings of this work, not be the convergence to global minimizers but instead can only be the convergence of the empirical risk to strictly suboptimal empirical risk levels, which should then be in some sense not too far away from the optimal empirical risk level (from the infimal value of the empirical risk function).
\end{abstract}
\tableofcontents
\section{Introduction}
\DL\ approximation methods -- typically based on an appropriate class of deep \ANNs\ trained through a suitable \SGD\ optimization method -- are these days the central components in a variety of \AI\ systems. \DL\ methods are, \eg, employed to create powerful \AI\ chatbot systems such as {\sc ChatGPT} (cf., \eg, \cite{Brownlanguagemodel2020,Yihengsummary2023}), {\sc Copilot}, and {\sc Gemini} (cf.\ \cite{geminiteam2024geminifamilyhighlycapable}), \DL\ methods are, \eg, used to build text-to-image models in \AI\ based image creation systems such as {\sc DALL-E} (cf.\ \cite{Rameshzeroshot}), {\sc Imagen} (cf.\ \cite{Sahariaimagen2022}), {\sc Midjourney}, and {\sc Stable Diffusions} (cf.\ \cite{Rombachstablediffusion2021}), 
and \DL\ methods are, \eg, also employed to approximately solve scientific models such as optimal control, optimal stopping,
and \PDE\ problems (cf., \eg, \cite{Beck_2023,WhatsNext,MR4356985,Gemain21,MR4795589,Ruf19}).  

The \DL\ approach to data driven supervised learning problems is usually to set up an optimization (minimization) problem in which a class of deep \ANNs\ aims to approximately match as best as possible the output data in relation to the input data and global minimizers in the landscape of the optimization problem then correspond to those deep \ANNs\ that, in an appropriate sense, best approximate the output data given the input data. The objective function of the optimization problem (the function one intends to minimize in the optimization problem) is referred to as empirical risk and, in this context, the minimization procedure is often also referred to as empirical risk minimization (cf., \eg, \cite{BerGroJent2020, 2001OnTM}). 

\SGD\ optimization methods are then employed to approximately compute global minimizers of the objective function of such optimization problems. In practically relevant learning problems, often not the plain vanilla standard \SGD\ optimization method is used for the training within the \DL\ scheme but instead more sophisticated suitably adaptive or accelerated variants of the standard \SGD\ method 
are employed to train the considered class of deep \ANNs. Adaptive or accelerated 
variants of the standard \SGD\ optimizer are, for example, 
\begin{itemize}
\item the momentum \SGD\ (cf.\ \cite{Polyak1964SomeMO}), 
\item the Nesterov accelerated \SGD\ (cf.\ \cite{Nesterov1983AMF,pmlr-v28-sutskever13}),
\item the \Adagrad\ (cf.\ \cite{JMLR:v12:duchi11a}),
\item the \RMSprop\ (cf.\ \cite{Hinton24_RMSprop}), 
\item the \Adam\ (cf.\ \cite{kingma2017adammethodstochasticoptimization}),
\item the \Adamax\ (cf.\ \cite{kingma2017adammethodstochasticoptimization}),
\item the AMSGrad (cf.\ \cite{ReddiKale2019}), and 
\item the \Nadam\ (cf.\ \cite{Dozat2015IncorporatingNM})
\end{itemize}
 optimizers. We also refer, \eg, to 
\cite{bach2024learning, Weinantoward2020, ArBePhi2024, Ruder2016AnOO, Sun2019OptimizationFD} for reviews, overview articles, and monographs 
treating such \SGD\ optimization methods. 

Despite the great success of \SGD\ optimization methods in the training of deep \ANNs\, it remains an open problem of research to explain the success (and the limitations) of such methods in rigorous theoretical terms. In particular, also in the standard setup of data driven supervised learning problems, it remained an open problem of research to prove (or disprove) that \SGD\ optimization methods converge in the training of deep \ANNs\ with the popular \ReLU\ activation function with high probability to global minimizers in the optimization landscape. In this work we answer this question negatively by proving that it does not hold that \SGD\ optimization methods converge with high probability to global minimizers of the objective function. Even stronger, in the main result of this work, see \cref{conjecture: multilayer2} in \cref{sec: non convergence of SGD method} below, we prove for a large class of \SGD\ optimization methods (including the plain vanilla standard \SGD\ optimization method and including each of the above named sophisticated \SGD\ optimization methods) that they do with high probability not converge to global minimizers of the optimization problem (cf.\ also \cref{cor: conjecture: multilayer} in \cref{sec: non convergence of SGD method} below).
In particular, it turns out that the probability to not converge to a global minimizer converges at least exponentially quickly to one as the width of the first hidden layer of the \ANN\ (the number of neurons on the first hidden layer) and the depth of the \ANN\ (the number of hidden layers), respectively, increase to infinity (see \cref{lem: estimate proba of nonconvergence: general case: rate of convergence 2} in \cref{sec: non convergence of SGD method} below).

To outline the contribution of this work within this short introductory section, we now present in the following result, \cref{main theorem} below, a special case of \cref{conjecture: multilayer2} (the main result of this work) in which we restrict ourselves, to simplify the presentation, to the plain vanilla standard \SGD\ optimization method with a simplified initialization procedure and we refer to \cref{conjecture: multilayer2} below for our more general non-convergence analysis covering the \Adam\ optimizer and each of the above named other more sophisticated \SGD\ optimization methods. The natural number $d \in \N = \{ 1, 2, 3, \dots \}$ in the first line of \cref{main theorem} represents the dimension of the input data of the supervised learning problem considered in \cref{main theorem}. We now present \cref{main theorem} in full mathematical details and, thereafter, intuitively explain the mathematical objects and the contribution of \cref{main theorem} in words.

\begin{athm}{theorem}{main theorem}
Let $d\in\N$,
$ a \in \R $, 
$ b \in (a, \infty)  $, let $ ( \Omega, \mathcal{F}, \P) $ be a probability space, for every $ m, n \in \N_0 $ 
let 
$ X^m_n \colon \Omega \to [a,b]^d $
and 
$ Y^m_n \colon \Omega \to \R $
be random variables, assume for all $i\in \N$, $j\in \N\backslash\{i\}$ that $\P( X_0^i=X_0^j)=0$, let $\f\colon\N\to \N$ be a function, for every $k\in \N_0$ let $\fd_k,L_k\in \N$, $\ell_k =(\ell_k^0,\ell_k^1,\dots,\ell_k^{L_k}) \in \N^{L_k+1}$ satisfy $\ell_k^0=d$, $\ell_k^{L_k}=1$, $\fd_k=\sum_{ i = 1 }^{L_k}\ell_k^i ( \ell_k^{i-1} + 1 )$, and $\max\{\ell_k^1,\ell_k^2,\dots,\ell_k^{L_k}\}\leq \f(\ell_k^1)$,  let $\mathbb{A}_r\colon \mathbb{R} \rightarrow \mathbb{R}$, $r \in [1,\infty]$, satisfy for all $r\in [1,\infty)$, $x\in (-\infty,r^{-1}]\cup[2 r^{-1},\infty)$, $y\in \R$ that
\begin{equation}\label{equation 1: main theorem}
    \mathbb A_r\in C^1(\R,\R),\qquad \mathbb A_r(x)=x\mathbbm 1_{(r^{-1},\infty)}(x),\qqandqq 0\leq \mathbb A_r(y)\leq \mathbb A_\infty(y)=\max\{y,0\},
    \end{equation}
assume $
  \sup_{ r \in [1,\infty) }
  \sup_{ x \in \R } 
  | ( \mathbb A_r )'( x ) | < \infty 
$, for every $r\in [1,\infty]$, $k\in\N_0$, $\theta=(\theta_1,\dots,\theta_{\fd_{k}})\in\R^{\fd_k}$ let $\mN^{ v, \theta }_{r,k}=(\mN^{ v, \theta }_{r,k,1},\dots,\mN^{ v, \theta }_{r,k,\ell_k^v}) \colon \R^{ d } \to \R^{ \ell_k^v} $, $v \in \{0,1,\dots,L_k\}$, satisfy for all $v\in \{0,1,\dots,L_k-1\}$, $x=(x_1,\dots, x_{d})\in \R^{d}$, $i\in\{1,2,\dots,\ell_k^{v+1}\}$ that
\begin{multline}\label{equation 2: main theorem}
  \mN^{v+1, \theta }_{r,k,i}( x ) = \theta_{\ell_k^{v+1}\ell_k^{v}+i+\sum_{h=1}^{v}\ell_k^h(\ell_k^{h-1}+1)}+\sum\limits_{j=1}^{\ell_k^{v}}\theta_{(i-1)\ell_k^{v}+j+\sum_{h=1}^{v}\ell_k^h(\ell_k^{h-1}+1)}\big(x_j\indicator{\{0\}}(v) \\ 
  +\mathbb A_{r^{1/\!\max\{v,1\}}}(\mN^{v,\theta}_{r,k,j}(x))\indicator{\N}(v) 
    \big),
\end{multline}
for every $k,n\in \N_0$ let $ M^{ k }_n \in  \N $, $\gamma_n^k\in \R$, for every $r\in [1,\infty]$, $k,n\in \N_0$ let
$ 
  \cLnri{n}{r}{k} \colon \R^{ \fd_{ k } } \times \Omega \to \R 
$
satisfy for all 
$ \theta \in \R^{ \fd_{ k} }$
that
\begin{equation}\label{equation 3: main theorem}
\displaystyle
  \cLnri{n}{r}{k}( \theta) 
  = 
  \frac{ 1 }{ M^{ k }_n } 
  \biggl[ \textstyle
  \sum\limits_{ m = 1 }^{ M^{ k }_n} 
    \abs{
      \mN^{ L_k,\theta }_{r,k}( X^m_n )
      - 
      Y^m_n 
    }^2
  \biggr]
  ,
\end{equation}
 for every $k,n\in \N_0$ let 
$ 
  \fG_n^{ k }  
  \colon \R^{ \fd_{ k} } \times \Omega \to \R^{ \fd_{ k } } 
$ 
satisfy for all $\omega\in \Omega$, $\theta\in \{\vartheta\in \R^{\fd_k}\colon (\nabla_{\vartheta} \cLnri{n}{r}{k}(\vartheta,\omega))_{r\in [1,\infty)}$ is convergent$\}$
that
\begin{equation}\label{equation 4: main theorem}
  \fG^{k}_n( \theta,\omega) 
  = 
  \lim_{r\to\infty}\bigl[\nabla_\theta \cLnri{n}{r}{k}(\theta,\omega)\bigr]
\end{equation}
and let 
	$
	\Theta_n^k 
	\colon \Omega  \to \R^{\fd_{ k } }
	$
	be a random variable, 
	 assume for all $k,n\in \N$ that 
	\begin{equation}\label{equation 5: main theorem}
		\Theta_{ n  } ^k
		= 
		\Theta_{n-1} ^k - \gamma_{n}^k
		\fG_n^k ( \Theta_{n-1}^k ),
	\end{equation} 
  assume $\liminf_{k\to \infty} \fd_{k}=\infty$ and $\liminf_{k \to \infty}\allowbreak \P\big(\inf_{\theta\in \R^{\fd_{k}}}\cL_0^{\infty,k}(\theta) >0\big)=1$, for every $k\in \N$ let $\cki_{k}\in (0,\infty)$, and assume for all $k\in \N$ that $\cki_k\Theta_0^k$ is standard normal. Then
\begin{equation}\label{equation 6: main theorem}
\textstyle
\liminf\limits_{ k \to \infty } \P\biggl(\inf\limits_{ n\in \N_0 }  \cLnri{0}{\infty}{k}(\NNelll^{k}_n)>\inf\limits_{\theta\in \R^{\fd_k}}\cLnri{0}{\infty}{k}(\theta)\biggr)=1.
\end{equation}
\end{athm}
\cref{main theorem} is an immediate consequence of \cref{conjecture: multilayer2} in \cref{sec: non convergence of SGD method} which is the main result of this work. In the following we provide some explanatory sentences regarding the mathematical objects and the contribution of \cref{main theorem}.

As mentioned above, the natural number $d \in \N$ in \cref{main theorem} specifies the dimension of the input data of the supervised learning problem in \cref{main theorem} and the real numbers $a \in \R$ and $b \in (a,\infty)$ in \cref{main theorem} describe the cube $[a,b]^d$ in which the input data of the supervised learning problem in \cref{main theorem} takes values in. The triple $( \Omega, \cF, \P )$ is the probability space on which the input and the output data of the supervised learning problem in \cref{main theorem} is defined. The random variables $X^m_n$, $(m,n) \in ( \N_0 )^2$, specify the input data of the considered supervised learning problem and the training input data for the optimization procedure in \cref{main theorem} and the random variables $Y^m_n \colon \Omega \to \R$, $(m,n) \in ( \N_0 )^2$, in \cref{main theorem} specify the associated output data of the considered supervised learning problem and the associated training output data for the optimization procedure in \cref{main theorem}. The assumption on input and output data in \cref{main theorem} are quite mild and, in particular, we do not even assume that the input-output data pairs $( X^m_0, Y^m_0 ) \colon \Omega \to [a,b]^d \times \R$, $m \in \N$, are \iid\ but we only assume that the input data is $\P$-almost surely pairwise distinct in the sense that we assume for all $i, j \in \N$ with $i \neq j$ that 
\begin{equation}\label{condition: different}
  \P( X^i_0 \neq X^j_0 ) = 1 .
\end{equation}
This assumption is usually satisfied in the context of deep learning methods for scientific models such as optimal control, optimal stopping, and \PDE\ problems and in our opinion there are also convincing arguments to believe that this assumption is also satisfied in models of typical data driven learning problems such as image classification tasks.

In \cref{main theorem} we show that the probability to converge in the training of deep \ANNs\ to a global minimizer in the optimization landscape converges to zero as the number of parameters of the \ANN\ increases to infinity (cf.\ also \cref{cor: conjecture: multilayer} in \cref{sec: non convergence of SGD method} below). To formulate this statement, we thus not consider one fixed \ANN\ architecture but instead in \cref{main theorem} we need to consider a whole sequence/family of \ANN\ architectures with increasing width/depth in the \ANN\ architectures so that the number of \ANN\ parameters converges to infinity. In \cref{main theorem} the natural numbers $L_k \in \N$, $k \in \N_0$, and the vectors $\ell_k = ( \ell_k^1,\dots,\ell_k^{L_k}) \in \N^{ L_k + 1 }$, $k \in \N_0$, of natural numbers precisely match this purpose. More specifically, we note for every $k \in \N_0$ that
\begin{itemize}
\item
we have that $L_k + 1$ specifies the number of layers of the $k\textsuperscript{th}$ sequence member of the considered sequence of \ANN\ architectures, 
\item
we have that $\ell_k^0$ specifies the dimension (the number of neurons) of the $1\textsuperscript{st}$ layer (the input layer) of the $k\textsuperscript{th}$ sequence member of the considered sequence of \ANN\ architectures,
\item 
we have that $\ell_k^1$ specifies the dimension (the number of neurons) of the $2\textsuperscript{nd}$ layer (the $1\textsuperscript{st}$ hidden layer) of the $k\textsuperscript{th}$ sequence member of the considered sequence of \ANN\ architectures,
\item 
we have that $\ell_k^2$ specifies the dimension (the number of neurons) of the $3\textsuperscript{rd}$ layer (the $2\textsuperscript{nd}$ hidden layer) of the $k\textsuperscript{th}$ sequence member of the considered sequence of \ANN\ architectures,
\item 
$\dots,$
\item 
we have that $\ell_k^{ L_k - 1 }$ specifies the dimension (the number of neurons) of the $(L_k)\textsuperscript{th}$ layer (the $(L_{k} - 1)\textsuperscript{th}$ hidden layer) of the $k\textsuperscript{th}$ sequence member of the considered sequence of \ANN\ architectures,
and 
\item 
we have that $\ell_k^{ L_k }$ specifies the dimension (the number of neurons) of the $(L_{k} + 1)\textsuperscript{th}$ layer (the output layer) of the $k\textsuperscript{th}$ sequence member of the considered sequence of \ANN\ architectures.
\end{itemize}
Furthermore, we observe for every $k \in \N_0$ that the natural number $\fd_k =\sum_{ i = 1 }^{L}\ell_k^i ( \ell_k^{i-1} + 1 )$  describes the number of \ANN\ parameters of the $k\textsuperscript{th}$ sequence member of the considered sequence of \ANN\ architectures. In \cref{main theorem} we also assume that the number of neurons on the $1\textsuperscript{st}$ hidden layer dominates in some mild sense the width of the \ANN\ in the sense that there exists an arbitrary possibly very rapidly growing function $f\colon \N \to \N$ such that for all $k \in \N_0$ it holds that
\begin{equation}
    \max\{\ell_k^1,\ell_k^2,\dots,\ell_k^{L_k}\}\leq \f(\ell_k^1).
\end{equation}
In \cref{main theorem} we study deep 
\ANNs\ with the popular \ReLU\ activation function. In \cref{main theorem} this \ReLU\ activation function is denoted by the function $\mathbb A_{ \infty } \colon \R \to \R$ that is specified in \cref{equation 1: main theorem} in \cref{main theorem}. The \ReLU\ activation $\mathbb A_{ \infty } \colon \R \to \R$ is not differentiable and this lack of smoothness transfers from the activation function to the the empirical risk function, that is, the objective function in the associated optimization problem. In view of this, in gradient based training procedures for such \ANNs\ with the \ReLU\ activation one needs to employ suitably generalized gradients as the usual gradients do in general not exist. In mathematical terms, the generalized gradients that are usually used in deep learning implementations can be described through a suitable approximation procedure in the sense that one considers continuously differentiable activation functions that converge to the \ReLU\ activation function and then defines the generalized gradients of the original objective function with the \ReLU\ activation as limits of the standard gradients of the objective functions with the continuously differentiable activations approximating the \ReLU\ activation (see, for example, \cite{Cheriditoconvegence2021,MaArconvergenceproof,ArAdconvergence2021}). In \cref{main theorem} the ReLU activation function is given by the function $\mathbb A_{ \infty } \colon \R \to \R$ in \cref{equation 1: main theorem} in \cref{main theorem} and the continuously differentiable activations approximating the \ReLU\ activation are given by the functions $\mathbb A_r \colon \R \to \R$, $r \in [1,\infty)$, in \cref{equation 1: main theorem} in \cref{main theorem}. For further details on this approximation approach to describe the generalized gradients in deep learning implementations we refer to \cite{MaArconvergenceproof} and \cite{ArAdconvergence2021}. Finally, using the sequence of \ANN\ architecture vectors $\ell_k = ( \ell_k^0,\ell_k^1,\dots\ell_k^{L_k}) \in \N^{ L_k + 1 }$, $k \in \N_0$, and the activation functions $\mathbb A_r \colon \R \to \R$, $r \in [1,\infty]$, we describe in \cref{equation 2: main theorem} in \cref{main theorem} the realization functions of the considered \ANNs. 

The natural numbers $M^k_n \in \N$, $(k,n) \in ( \N_0 )^2$, in \cref{main theorem} describe the number of the input-output data pairs of the considered supervised learning problem and the size of the mini-batches in the training procedures. More specifically, for every $k \in \N_0$ we have that $M^k_0$ specifies the number of input-output data pairs to describe the empirical risk for the $k\textsuperscript{th}$ sequence member of the sequence of optimization procedures and for every $k \in \N_0$, $n \in \N$ we have that $M^k_n$ specifies the number of input-output data pairs in the mini-batch of the $n\textsuperscript{th}$ \SGD\ step for the $k\textsuperscript{th}$ sequence member of the sequence of optimization procedures. The real numbers $\gamma^k_n \in \R$, $( k, n ) \in ( \N_0 )^2$, in \cref{main theorem} describe the learning rates (the step sizes) for the \SGD\ optimization processes (see \cref{equation 5: main theorem} in \cref{main theorem}). 

The random functions $\cL^{ r, k }_n \colon \R^{\fd_k} \times \Omega \to \R$, $(r,k,n) \in [1,\infty] \times ( \N_0 )^2$, in \cref{equation 3: main theorem} in \cref{main theorem} describe the overall objective functions of the considered optimization problems and the objective functions used in each step of the \SGD\ optimization procedures. In particular, these objective functions are used in \cref{equation 4: main theorem} in \cref{main theorem} to specify the generalized gradients according the above mentioned approximation approach. These generalized gradients are then employed in \cref{equation 5: main theorem} in \cref{main theorem} to specify the dynamics of the considered \SGD\ optimization processes. We note that only for simplicity we consider in \cref{main theorem} within this introductory section only the plain vanilla standard \SGD\ optimization method but our more general result in \cref{conjecture: multilayer2} proves non-convergence to global minimizers for each of the above named sophisticated \SGD\ optimization methods (see also \cref{sec: SGD optimization methods}).

Within this introductory section we also assume for simplicity that, at initial time, the \SGD\ optimization processes are, up to the scalar factors $( \cki_k )_{ k \in \N }$, standard normally distributed (see the assumption between \cref{equation 5: main theorem} and \cref{equation 6: main theorem} in \cref{main theorem}). Our more general non-convergence result in \cref{conjecture: multilayer2} covers, however, more general schemes for the initialization procedures with general probability density functions going much beyond the scaled standard normal case in \cref{main theorem}.

Roughly speaking, in \cref{equation 6: main theorem} in \cref{main theorem} we then prove that the probability to converge to the infimal value $\inf_{ \theta \in \R^{\fd_k} } \cL^{ \infty, k }_0( \theta )$ of the empirial risk function as the number of \SGD\ training steps $n$ increases to infinity converges to 1 as the depth/width of the \ANNs\ increases to infinity. In particular, we observe that \cref{equation 6: main theorem} in \cref{main theorem} ensures in the training of deep \ReLU\ \ANNs\ that with high probability the \SGD\ optimization method does not converge to a global minimizer in the optimization landscape (cf.\ \cref{cor: conjecture: multilayer} in \cref{sec: non convergence of SGD method} below). 

\subsection{Literature review}\label{subsec: Literature review}
In the scientific literature further lower bounds and non-convergence results for \SGD\ optimization methods can, for example, be found in \cite{CHERIDITO2021101540,Dereichnonconvergence2024,Dereichconvergencerate2024,gallon2022blowphenomenagradientdescent,Philiptheorytopractice2021,ArBePhi2024,ArAd2024,Lu_2020,ReddiKale2019}. In particular, in the situation of shallow \ReLU\ \ANNs\ with only one hidden layer we showed in our preliminary work \cite{ArAd2024} for a large class of \SGD\ optimization methods including, for example, the \Adam\ optimizer that the true risk of the optimization process converges with high probability not to the minimal value of the true risk function under suitable assumptions (on the target function and the density function of the probability distribution of the input data) that are sufficient to ensure that there exist global minimizers of the true risk function. These assumptions in \cite{ArAd2024} are based on the assumptions in \cite{Dereichexistenceshallow2023,Kassingexistence2023} in which the existence of global minimizers of the true risk function in the situation of shallow \ReLU\ \ANNs\ is established (see, \eg, also
\cite{Aradexistenceofglobmin,KAINEN2000695,QuocTung2023} for such existence of minimizers results).

Moreover, in certain simple situations of shallow \ANNs\ with one-dimensional input in which there do not exist minimizers in the optimization landscape in the work \cite{gallon2022blowphenomenagradientdescent} suitable lower bounds for some gradient based optimization procedures have been established that ensure divergence of the optimization process (cf.\ also \cite{MR4243432}). All works that establish the existence of minimizers in the optimisation landscape deal with sallow \ANNs\ with a single hidden layer and in the situation of deep \ANNs, as in the setup of \cref{main theorem} above, it remains an open question whether there exist minimizers in the optimisation landscape. Nonetheless, \cref{main theorem} shows that the empirical risk of the optimization method converges with high probability not to the infimal value of the empirical risk and \cref{cor: conjecture: multilayer} in \cref{sec: non convergence of SGD method} below also shows that the considered \SGD\ optimization method does with high probability not converge to any possible global minimizer in the optimization landscape. 

Furthermore, \eg, in \cite{CHERIDITO2021101540,Lu_2020} non-convergence in the training of certain very deep \ReLU\ \ANNs\ is investigated. Loosely speaking, the arguments in these works imply that the standard \SGD\ and related optimization methods do with strictly positive probability not converge to global minimizers of the risk (as they converge with strictly positive probability to local minimizers whose realization functions are constant functions). In contrast, in this work and in the above named reference \cite{ArAd2024}, respectively, it is shown that the considered optimization method converges with high probability not to global minimzers of the risk, that is, we show that the probability to not converge to global minimizers converges (rapidly) to one as the architecture of the considered deep \ANNs\ increases (in width or depth). 

Another direction of non-convergence results can, \eg, be found in \cite{Dereichnonconvergence2024} (see also \cite[Subsection 7.2.2.2]{ArBePhi2024}). In particular, in \cite{Dereichnonconvergence2024} it is shown for the \Adam\ optimizer and related adaptive \SGD\ optimization methods that the considered optimization process fails to converge to any possible (random) point in the opimization landscape (in particular, covering global minimizers) as soon as the learning rates $\gamma_n \in (0,\infty)$, $n \in \N$, are bounded away from zero in the sense that $\liminf_{ n \to \infty } \gamma_n > 0$. In contrast, in the situation where the learning rates do converge to zero with an appropriate speed of convergence, it is at least for certain simple convex optimization problems known that the plain-vanilla standard \SGD\ optimization method (cf., \eg, \cite{dereich2024learningrateadaptivestochastic,DereichKassingconvergence2021,Jentzen_2020,ArPhilowerbound,Tadicconvergence2015}) and also more sophisticated optimization methods such as the \Adam\ optimizer (cf., \eg, \cite{Barakat_2021_cvg,Xiangyiconvergence2018,Defossez2022,Dereichconvergencerate2024,ReddiKale2019}) converge with rates of convergence to the global minimizer of the optimization problem. In \cref{main theorem} there is not any assumption on the sequence of learning rates $( \gamma_n )_{ n \in \N }$ and, in particular, the learning rates $\gamma_n \in \R$, $n \in \N$, in \cref{main theorem} are arbitrary real numbers that may be zero, may be strictly negative, or may even diverge to infinity.

We also would like to emphasize that the findings of this work do not imply that \SGD\ optimization methods do not succeed to train \ANNs. More precisely, \cref{main theorem} and its generalizations and extensions in \cref{conjecture: multilayer2} and \cref{cor: conjecture: multilayer} in \cref{sec: non convergence of SGD method}, respectively, only assure that a large class of \SGD\ optimization methods can with high probability not converge to global minimizers in the training of \ANNs\ but they do not exclude the possibility that such \SGD\ methods do converge in the training of \ANNs\ to "good" non-global local minimizers or non-optimal critical points, whose empirical risks are in an appropriate sense not far away from the infimal value of the empirical risk function. In this context we refer, \eg, to \cite{MR4056927,DereichKassingconvergence2021,ArBePhi2024,Aradexistenceofglobmin,Tadicconvergence2015} and the references therein for convergence results to critical points for gradient based optimization procedures in the training of \ANNs\ and we refer, \eg, to
\cite{Gentile2022,Ibragimov2022,Welper2024,MR4765349} for convergence results to suitable "good" non-optimal critical points (whose risk values are not too far away from the infimal risk value) for gradient based optimization procedures in the training of \ANNs. Put it differently, it may still very well be the case that \SGD\ optimization methods provably succeed to train (deep) \ANNs\ in data driven supervised learning problems but, if this does hold, then the explanation for this can, due to the findings of this work, not be the convergence to global minimizers but instead can only be the convergence of the empirical risk to strictly suboptimal empirical risk levels, which should then be in some sense not too far away from the optimal empirical risk level (from the infimal value of the empirical risk function); cf., \eg, \cite{Gentile2022,Ibragimov2022,Welper2024,MR4765349}.

\subsection{Structure of this work}
The remainder of this work is organized as follows. In \cref{sec: number of inactive neurons} we establish suitable upper bounds for the probability to have a very small number of inactive neurons on the first hidden layer of the considered deep \ReLU\ \ANN\ after a random initialization of the parameters of the \ANN. Roughly speaking, here a neuron on 
the first hidden layer of the \ANN\ is understood to be inactive if the partial derivatives 
with respect to the \ANN\ parameters associated to this neuron all vanish (cf.\ \cref{eq:setting_definition_of_I_set} in \cref{setting:multilayer} for a rigorous introduction of the set of inactive neurons on the first hidden layer).

In \cref{sec: improve risk} we establish in \cref{cor: improve risk} the key ingredient of this work that the infimal value over 
the risk values of all \ReLU\ \ANNs\ with a given fixed architecture is strictly smaller than the infimal value 
over the risk values of all \ReLU\ \ANNs\ with this architecture and at least three inactive neurons 
on the first hidden layer. 

Loosely speaking, in \cref{sec: represent general grad} we provide in \cref{theo: explicit formula} explicit representations
for generalized gradients of the risk function describing the expected mean squared error 
between the output datum and the realization of an \ANN\ evaluated at the input datum. 
\cref{sec: represent general grad} extends and is strongly based on the findings in \cite[Section 2]{MaArconvergenceproof} where such 
explicit representations for generalized gradients of the risk function have been established 
in the smaller generality with a given target function instead of the output datum.

In \cref{sec: non convergence of SGD method} we combine the findings from \cref{sec: number of inactive neurons,sec: improve risk,sec: represent general grad} to establish \cref{conjecture: multilayer2}, which is 
the main result of this work. \cref{conjecture: multilayer2}, in particular, shows for a large class of \SGD\ optimization methods 
that in the training of deep \ReLU\ \ANNs\ we have that the empirical risk of the considered optimization 
method does with high probability not converge to the infimal value of the empirical risk function. 
\cref{main theorem} in this introductory section is an immediate consequence of \cref{conjecture: multilayer2}. 

In \cref{sec: SGD optimization methods} we verify that the general non-convergence result in \cref{conjecture: multilayer2} from \cref{sec: non convergence of SGD method}, 
in particular, applies to the plain vanilla standard \SGD\ (see \cref{subsec: standard SGD}), the momentum \SGD\ (see \cref{subsec: momentum SGD}), 
the Nesterov accelerated \SGD\ (see \cref{subsec: nesterov}), the \Adagrad\ (see \cref{subsec: Adagrad}),
the \RMSprop\ (see \cref{subsec: RMSprop}), the \Adam\ (see \cref{subsec: Adam}), the \Adamax\ (see \cref{subsec: Adamax}), the AMSGrad (see \cref{subsec: AMSgrad}), and the \Nadam\ optimizers (see \cref{subsec: Nadam}).

\section{Upper bounds for the probability to have a small number of inactive neurons}\label{sec: number of inactive neurons}
In the main results of this section, \cref{cor:estimate dead neuron hidden layer 1.1} and \cref{cor:estimate dead neuron hidden layer 1} in \cref{subsec: lower bound for inactive neuron} below, 
we establish, roughly speaking, suitable upper bounds for the probability 
to have a very small number of inactive neurons on the first hidden layer 
of the considered deep \ReLU\ \ANN\ after a random initialization 
of the parameters of the \ANN. Loosely speaking, here a neuron on the first hidden layer 
of the \ANN\ is understood to be inactive if the partial derivatives with respect 
to the ANN parameters associated to this neuron all vanish. We refer to \cref{eq:setting_definition_of_I_set} 
in \cref{setting:multilayer} for a rigorous introduction of the set of inactive neurons on 
the first hidden layer. 

We employ \cref{cor:estimate dead neuron hidden layer 1.1} and \cref{cor:estimate dead neuron hidden layer 1} in \cref{sec: non convergence of SGD method}, 
where we also present the proof of \cref{conjecture: multilayer2}, the main result of this work. 
\cref{cor:estimate dead neuron hidden layer 1.1} and \cref{cor:estimate dead neuron hidden layer 1} both employ the setup in \cref{setting:multilayer} in which 
we provide a short description of realization functions of deep \ReLU\ \ANNs\ 
with the \ANN\ parameters stored in a high-dimensional real vector 
(cf., \eg, \cite[Section 1.1]{ArBePhi2024}). 
Many of the arguments in this section are elementary and build up, \eg, 
on the estimates in \cite[Section 4.1]{ArAd2024}
(cf., \eg, also \cite{CHERIDITO2021101540}). 

Moreover, we recall in \cref{lem: binary1}, \cref{lem: binary}, and \cref{increasing property} some lower bounds 
for the probability to have sufficiently many successes in independent 
(not necessarily identically distributed) Bernoulli trials. 
Only for completeness we include in this section detailed proofs 
for \cref{lem: binary1}, \cref{lem: binary}, and \cref{increasing property}. 

\subsection{Mathematical description of deep artificial neural networks (ANNs)}
\begin{setting}
\label{setting:multilayer}
Let $d\in\N$, $ a \in \R $, 
$ b \in (a, \infty)  $, for every $ L\in \N$, $\ell=(\ell_0,\ell_1,\dots, \ell_L) \in \N^{L+1}$ let $\fd_\ell \in\N$ satisfy $ \fd_\ell= \sum_{ k = 1 }^{L}\ell_k ( \ell_{k-1} + 1 ) $, 
and for every 
 $L\in\N$, $\ell=(\ell_0,\ell_1,\dots,\ell_L)\in \N^{L+1}$, $\theta=(\theta_1,\dots,\theta_{\fd_\ell})\in\R^{\fd_\ell}$ let $\mN^{ k, \theta }_{\ell}=(\mN^{ k, \theta }_{\ell,1},\dots,\mN^{ k, \theta }_{\ell,\ell_k}) \colon \R^{ \ell_0 } \to \R^{ \ell_k} $, $k\in \{0,1,\dots,L\}$, satisfy for all $k\in \{0,1,\dots,L-1\}$, $x=(x_1,\dots, x_{\ell_0})\in \R^{\ell_0}$, $i\in\{1,2,\dots,\ell_{k+1}\}$ that
\begin{multline}\label{relization multi}
  \mN^{ k+1, \theta }_{\ell,i}( x ) = \theta_{\ell_{k+1}\ell_{k}+i+\sum_{h=1}^{k}\ell_h(\ell_{h-1}+1)}+\sum\limits_{j=1}^{\ell_{k}}\theta_{(i-1)\ell_{k}+j+\sum_{h=1}^{k}\ell_h(\ell_{h-1}+1)}\big(x_j\indicator{\{0\}}(k) \\ 
  +\max\!\big\{\mN^{k,\theta}_{\ell,j}(x),0\big\}\indicator{\N}(k)\big)
\end{multline}
and let $
  \inact_{ \ell}^{ \theta } 
  \subseteq \N 
$
satisfy
\begin{equation}
\label{eq:setting_definition_of_I_set}
  \inact_{ \ell }^{ \theta }
  = 
  \bigl\{ 
    i \in \{1,2,\dots,\ell_1\}
    \colon 
    \bigl(
      \forall \, x \in [a,b]^{\ell_0} \colon  
        \mathcal N^{1,\theta}_{\ell,i}(x)
      < 0
    \bigr)
  \bigr\}.
\end{equation}
\end{setting}
\subsection{Lower bounds for the probability to have sufficiently many successes}
\begin{athm}{lemma}{lem: binary1}
     Let $\alpha\in [0,1)$, $p \in (0,1)$.
 Then there exists $\boundexp\in (0,\infty)$ which satisfies\footnote{Note that for all $x\in \R$ it holds that $\lfloor x\rfloor=\max((-\infty,x]\cap\Z)$ and $\ceil x=\min([x,\infty)\cap\Z)$.} for all $k\in \N$ that
 \begin{equation}\llabel{conclude}
     \sum_{n=\ceil{k^\alpha}}^{k} \binom{k}{n}p^n(1-p)^{k-n}\geq 1-\exp(-\boundexp k).
 \end{equation}
\end{athm}
\begin{aproof}
\argument{the fact that $0<p<1$;}{that $\ln(\max\{p,1-p\})<0$\dott}
\argument{the fact that for all $k,m,n\in \N_0$ with $m<n\leq \frac k2$ it holds that $k-n\geq n>m$}{that for all $k,m,n\in \N_0$ with $m<n\leq \frac k2$ it holds that
\begin{equation}\llabel{arggg}
    \frac{n!(k-n)!}{m!(k-m)!}=\frac{\prod_{i=1}^{n-m} (m+i)}{\prod_{i=1}^{n-m} (k-n+i)}<1\dott
\end{equation}}
\argument{\lref{arggg}}{for all $k,m,n\in \N_0$ with $m<n\leq \frac k2$ that
\begin{equation}\llabel{arggg1}
     \binom{k}{m}=\frac{k!}{m!(k-m)!}<\frac{k!}{n!(k-n)!}=\binom{k}{n}.
\end{equation}}
    \argument{\lref{arggg1}; the fact that for all $x\in \R$ it holds that $\lfloor x\rfloor \geq \ceil{x}-1$}
{that for all $k\in \N$ with $k^\alpha\leq \frac k2$ it holds that
    \begin{equation}\llabel{cdistribution}
    \begin{split}
        &\sum_{n=0}^{\ceil{k^\alpha}-1} \binom{k}{n} p^n(1-p)^{k-n}\\
        &\leq \bigl(\ceil{k^\alpha}\bigr)\binom{k}{\lfloor k^\alpha\rfloor}(\max\{p,1-p\})^{k}
       \leq 2k^\alpha k^{(k^\alpha)}(\max\{p,1-p\})^{k}\dott
         \end{split}
    \end{equation}}
    \argument{\lref{cdistribution};} 
     {for all $k\in \N$ with $k\geq 2^{\frac{1}{1-\alpha}}$ that
     \begin{equation}\llabel{logcdistribution1}
    \begin{split}
    \ln\biggl(\textstyle\sum\limits_{n=0}^{\ceil{k^\alpha}-1} \binom{k}{n} p^n(1-p)^{k-n}\biggr)&\leq k\ln(\max\{p,1-p\})+  k^\alpha\ln(k)+\alpha\ln(k)+\ln(2)\\
      &\leq k\ln(\max\{p,1-p\})+k^\alpha\ln(k)+\alpha2^{\frac{-\alpha}{1-\alpha}}k^\alpha\ln(k)+\ln(2)\\
      &= k\ln(\max\{p,1-p\})+(1+\alpha 2^{\frac{-\alpha}{1-\alpha}})k^\alpha\ln(k)+\ln(2)\dott
      \end{split}
        \end{equation}}
        \argument{\lref{logcdistribution1};}{for all $k\in \N$ with $k\geq \max\bigl\{2^{\frac{1}{1-\alpha}},\frac{4\ln(2)}{-\ln(\max\{p,1-p\})}\bigr\}$ that
     \begin{equation}\llabel{logcdistribution}
    \begin{split}
    \ln\!\bigg(\textstyle\sum\limits_{n=0}^{\ceil{k^\alpha}-1} \binom{k}{n} p^n(1-p)^{k-n}\bigg) \leq\frac 34 k\ln(\max\{p,1-p\})+(1+\alpha 2^{\frac{-\alpha}{1-\alpha}})k^\alpha\ln(k)\dott
      \end{split}
        \end{equation}}
        \argument{the fact that $0\leq\alpha<1$;}{that \begin{equation}\llabel{argg1}
        \limsup_{x\to\infty}\biggl(\frac{|\ln(x)|}{x^{1-\alpha}}\biggr) =0\dott
        \end{equation}}
    \argument{\lref{argg1};the fact that $\ln(\max\{p,1-p\})<0$}{that there exists $\rho\in (0,\infty)$ which satisfies for all $k\in [\rho,\infty)$ that
    \begin{equation}\llabel{argg2}
        \ln(k)< \textstyle-\frac 12\bigl[1+\alpha 2^{\frac{-\alpha}{1-\alpha}}\bigr]^{-1}\bigl[\ln(\max\{p,1-p\})\bigr]k^{1-\alpha}\dott
    \end{equation}}
    \startnewargseq
    \argument{\lref{logcdistribution};\lref{argg2}}{that for all $k\in \N$ with $k\geq \max\bigl\{2^{\frac{1}{1-\alpha}}, \rho,\frac{4\ln(2)}{-\ln(\max\{p,1-p\})}\bigr\}$ it holds that
    \begin{equation}\llabel{argg3}
     \ln\biggl(\textstyle\sum\limits_{n=0}^{\ceil{k^\alpha}-1} \binom{k}{n} p^n(1-p)^{k-n}\biggr)\leq \frac 14 k \ln(\max\{p,1-p\})\dott
    \end{equation}}
    \argument{the fact that for all $x\in \R$ it holds that $1-x\leq \exp(-x)$}{ that for all $c\in (0,\infty)$, $k\in \N\cap[0,c]$ it holds that
    \begin{equation}\llabel{argg3'}
    \begin{split}
    &\ln\biggl(\textstyle\sum\limits_{n=0}^{\ceil{k^\alpha}-1} \binom{k}{n}p^n(1-p)^{k-n}\biggr)\leq \ln\biggl(\textstyle\sum\limits_{n=0}^{k-1} \binom{k}{n}p^n(1-p)^{k-n}\biggr)
        =\ln\Bigl(1-\binom{k}{k}p^k(1-p)^{k-k}\Bigr)\\
        &=\ln(1-p^k)\leq\ln(\exp(-p^k))=-p^k\leq -p^c=-\biggl[\frac{p^c}{k}\biggr]k\leq -\biggl[\frac{p^c}{c}\biggr]k\dott
        \end{split}
    \end{equation}}
    \argument{\lref{argg3};\lref{argg3'};the fact that $\ln(\max\{p,1-p\})<0$}{that there exists $\boundexp\in (0,\infty)$ which satisfies for all $k\in \N$ it holds that
    \begin{equation}\llabel{argg4}
    \ln\biggl(\textstyle\sum\limits_{n=0}^{\ceil{k^\alpha}-1} \binom{k}{n} p^n(1-p)^{k-n}\biggr)\leq -\boundexp k\dott
    \end{equation}}
    \startnewargseq
    \argument{\lref{argg4};}{that for all $k\in \N$ it holds that
    \begin{equation}\llabel{argg5}
       \sum\limits_{n=0}^{\ceil{k^\alpha}-1} \binom{k}{n} p^n(1-p)^{k-n}\leq \exp(-\boundexp k) \dott
    \end{equation}}
    \argument{\lref{argg5}; the fact that for all $k\in \N$ it holds that $\sum_{n=0}^{k} \binom{k}{n}p^n(1-p)^{k-n}=1$}[verbs=sep]{\lref{conclude}\dott}
\end{aproof}
\begin{athm}{lemma}{lem: binary}
 Let $\alpha\in [0,1)$, $p \in (0,1)$ and let $(\arch_k)_{k\in\N}\subseteq\N$ satisfy $\liminf_{k\to\infty} \arch_k=\infty$.
 Then 
 \begin{equation}\label{conclude: lem: estimate number of dead neuron layer 1}
     \liminf\limits_{k\to \infty}\textstyle\biggl[\sum\limits_{n=\ceil{(\arch_k)^\alpha}}^{\arch_k} \binom{\arch_k}{n}p^n(1-p)^{\arch_k-n}\biggr]=1.
 \end{equation}
\end{athm}
\begin{aproof}
\argument{\cref{lem: binary1};the assumption that $\liminf_{k\to\infty}\arch_k=\infty$}{\cref{conclude: lem: estimate number of dead neuron layer 1}\dott}
\end{aproof}
\begin{athm}{lemma}{increasing property}
 Let $m,n\in \N$ and let 
$f\colon \R^n\to \R$ satisfy for all $x=(x_1,\dots,x_n)\in \R^n$ that
 \begin{equation}\llabel{def: f}
     f(x)=\sum_{A\subseteq\{1,2,\dots,n\},\,\#(A)\geq m}\biggl(\biggl[\textstyle\prod\limits_{i\in A}x_i\biggr]\biggl[\prod\limits_{i\in \{1,2,\dots,n\}\backslash A}(1-x_i)\biggr]\biggr).
 \end{equation}
 Then
\begin{enumerate}[label=(\roman*)]
    \item \llabel{item 1} it holds for all $x=(x_1,\dots,x_n)$, $y=(y_1,\dots,y_n)\in [0,1]^n$ with $x_1\leq y_1$, $x_2\leq y_2$, $\dots$, $x_n\leq y_n$ that $f(x)\leq f(y)$ and
    \item \label{item 2: increasing property} it holds for all $p\in [0,1]$, $x=(x_1,\dots,x_n)\in [p,1]^n$ that 
        $\sum_{k=m}^{n} \binom{n}{k}p^k(1-p)^{n-k}\leq f(x)$.
\end{enumerate}
\end{athm}
\begin{aproof}
    \argument{\lref{def: f};}{that for all $j\in \{1,2,\dots,n\}$, $x=(x_1,\dots,x_n)\in \R^n$ it holds that
    \begin{equation}\llabel{eq1}
        \begin{split}
            \textstyle\frac{\partial }{\partial x_j}f(x)
            &=\sum_{A\subseteq\{1,2,\dots,n\},\,\#(A)\geq m,\,j\in A}\biggl(\biggl[\textstyle\prod\limits_{i\in (A\backslash\{j\})}x_i\biggr]\biggl[\prod\limits_{i\in \{1,2,\dots,n\}\backslash A}(1-x_i)\biggr]\biggr)\\
            &-\sum_{A\subseteq\{1,2,\dots,n\},\,\#(A)\geq m,\,j\notin A}\biggl(\biggl[\textstyle\prod\limits_{i\in A}x_i\biggr]\biggl[\prod\limits_{i\in (\{1,2,\dots,n\}\backslash (A\cup\{j\}))}(1-x_i)\biggr]\biggr)\\
            &=\sum_{A\subseteq(\{1,2,\dots,n\}\backslash\{j\}),\,\#(A)\geq m-1}\biggl(\biggl[\textstyle\prod\limits_{i\in A}x_i\biggr]\biggl[\prod\limits_{i\in (\{1,2,\dots,n\}\backslash (A\cup\{j\}))}(1-x_i)\biggr]\biggr)\\
            &-\sum_{A\subseteq(\{1,2,\dots,n\}\backslash\{j\}),\,\#(A)\geq m}\biggl(\biggl[\textstyle\prod\limits_{i\in A}x_i\biggr]\biggl[\prod\limits_{i\in (\{1,2,\dots,n\}\backslash (A\cup\{j\}))}(1-x_i)\biggr]\biggr)\\
            &=\sum_{A\subseteq(\{1,2,\dots,n\}\backslash\{j\}),\,\#(A)=m-1}\biggl(\biggl[\textstyle\prod\limits_{i\in A}x_i\biggr]\biggl[\prod\limits_{i\in (\{1,2,\dots,n\}\backslash (A\cup\{j\}))}(1-x_i)\biggr]\biggr).
        \end{split}
    \end{equation}}
    \argument{\lref{eq1};}{that for all $j\in \{1,2,\dots,n\}$, $x=(x_1,\dots,x_n)\in [0,1]^n$ it holds that
    \begin{equation}\llabel{eq2}
         \textstyle\frac{\partial }{\partial x_j}f(x)\geq 0.
    \end{equation}}
    \argument{\lref{eq2};the fundamental theorem of calculus}
    {that for all $x=(x_1,\dots,x_n)$, $y=(y_1,\dots,y_n)\in [0,1]^n$ with $x_1\leq y_1$, $x_2\leq y_2$, $\dots$, $x_n\leq y_n$ it holds that
  \begin{equation}\llabel{eqtg1}
  \begin{split}
      f(y)&=f(x)+\int_0^1 f'(x+r(y-x))(y-x)\,\d r\\
      &=f(x)+\sum_{j=1}^n\int_0^1 (y_j-x_j)\bigl(\textstyle\frac{\partial}{\partial x_j}f\bigr)(x+r(y-x))\,\d r\geq f(x)\dott
      \end{split}
  \end{equation}}
    \argument{\lref{eqtg1};}{\lref{item 1}\dott}
    \startnewargseq
    \argument{\lref{item 1}}{that for all $p\in [0,1]$, $x\in [p,1]^n$ it holds that
    \begin{equation}\llabel{eq3}
        f(p,p,\dots,p)\leq f(x)\dott
    \end{equation}}
    \argument{\lref{def: f}}{for all $p\in [0,1]$ that
    \begin{equation}\llabel{eq4}
    \begin{split}
&f(p,p,\dots,p)=\sum_{A\subseteq\{1,2,\dots,n\},\,\#(A)\geq m} p^{\#(A)}(1-p)^{n-\#(A)}\\
&=\sum_{k=m}^{n}\sum_{A\subseteq\{1,2,\dots,n\},\,\#(A)=k} p^k(1-p)^{n-k}=\sum\limits_{k=m}^{n} \binom{n}{k}p^k(1-p)^{n-k}\dott
\end{split}
    \end{equation}}
    \argument{\lref{eq4};\lref{eq3}}{\cref{item 2: increasing property}\dott}
\end{aproof}
\subsection{Scaled independent random variables}
\begin{athm}{lemma}{Lemma X}
    Let $(\Omega,\mathcal F,\P)$ be a probability space, let $n\in \N$, let $\Theta=(\Theta_1,\dots,\Theta_n)\colon \Omega\to \R^n$ be a random variable, let $\Dens\colon\R\to[0,\infty)$ be measurable, and let $\cki_1,\cki_2,\dots,\cki_n\in (0,\infty)$ satisfy for all $x_1,x_2,\dots,x_n\in\R$ that 
    \begin{equation}\llabel{NR777}
  \P\bigl( \cap_{i=1}^{n}
    \{
    \NNelll_i 
    < x_i\}
  \bigr)
  =
  \prod_{i=1}^{n}\biggl[\int_{ - \infty }^{\cki_{i}x_i} \Dens(y) \, \d y\biggr].
  \end{equation}
  Then
\begin{enumerate}[label=(\roman*)]
    \item \label{lemma X: item 1} it holds that $\int_{-\infty}^{\infty}\Dens(y)\,\d y=1$,
    \item \label{lemma X: item 2} it holds for all $k\in \{1,2,\dots,n\}$, $x\in \R$ that $\P(\Theta_k<x)=\int_{-\infty}^{\cki_kx}\Dens(y)\,\d y$, and
    \item \label{lemma X: item 3} it holds that $\cki_k\Theta_k$, $k\in \{1,2,\dots,n\}$, are \iid
\end{enumerate}
\end{athm}
\begin{aproof}
  \argument{\lref{NR777};continuity from below of the measure $\P$; the monotone convergence theorem; the fact that for all $k\in\{1,2,\dots,n\}$ it holds that $c_k>0$}{that 
  \begin{equation}\llabel{eq2}
      1=\P\bigl( \cap_{i=1}^{n}
    \{
    \NNelll_i 
    < \infty\}\bigr)= \prod_{i=1}^{n}\Biggl[\int_{ - \infty }^{\infty} \Dens(y) \, \d y\Biggr]=\Biggl[\int_{ - \infty }^{\infty} \Dens(y) \, \d y\Biggr]^n\dott
  \end{equation}}
  \argument{\lref{eq2};}[verbs=ip]{\cref{lemma X: item 1}\dott}
  \startnewargseq
  \argument{\lref{NR777};\cref{lemma X: item 1}; continuity from below of the measure $\P$; the monotone convergence theorem; the fact that for all $k\in \{1,2,\dots,n\}$ it holds that $c_k>0$}{that for all $k\in \{1,2,\dots,n\}$, $x\in \R$ it holds that
  \begin{equation}\llabel{eq3}
  \begin{split}
      \P(\Theta_k<x)&=\P\bigl((\cap_{i\in \{1,2,\dots,n\}\backslash\{k\}}\{\Theta_i<\infty\})\cap\{\Theta_k<x\}\bigr)\\
      &=\Bigg[\prod_{i\in \{1,2,\dots,n\}\backslash\{k\}}\Biggl[\int_{ - \infty }^{\infty} \Dens(y) \, \d y\Biggr]\Biggr] \Biggl[\int_{ - \infty }^{\cki_{k}x} \Dens(y) \, \d y\Biggr]=\int_{ - \infty }^{\cki_{k}x} \Dens(y) \, \d y\dott
        \end{split}
  \end{equation}}
  \argument{\lref{eq3};}{\cref{lemma X: item 2}\dott}
  \startnewargseq
  \argument{\lref{NR777};\lref{eq3};the fact that for all $k\in \{1,2,\dots,n\}$ it holds that $c_k>0$}{that for all $x_1,x_2,\dots,x_n\in \R$ it holds that
  \begin{equation}\llabel{eq4}
  \begin{split}
      &\P\bigl( \cap_{i=1}^{n}
    \{
    c_i\NNelll_i 
    < x_i\}
  \bigr)=\P\bigl( \cap_{i=1}^{n}
    \{
    \NNelll_i 
    <(c_i)^{-1} x_i\}
  \bigr)= \prod_{i=1}^{n}\biggl[\int_{ - \infty }^{c_i(c_i)^{-1}x_i} \Dens(y) \, \d y\biggr]\\
&=\prod_{i=1}^{n}\P\bigl(\Theta_i<(c_i)^{-1}x_i\bigr)=\prod_{i=1}^{n}\P(\cki_i\Theta_i<x_i)\dott
  \end{split}
  \end{equation}}
  \argument{\lref{eq4};}{\cref{lemma X: item 3}\dott}
\end{aproof}
\subsection{Lower bounds for the probability to have sufficiently many inactive neurons}\label{subsec: lower bound for inactive neuron}
\begin{athm}{lemma}{existence of eta}
    Let $ \Dens \colon \R \to [0,\infty) $ be measurable,
 assume 
\begin{equation}\llabel{assume}
    \sup_{ \eta \in (0,\infty) } \biggl(  \eta ^{ - 1 } \int_{ - \eta }^{ \eta } \allowbreak\mathbbm1_{ \R \backslash { \{0\} } }\allowbreak( \Dens( x ) ) \, \d x \biggr) \geq 2,
\end{equation} 
and let $\scrc,\scrC\in (0,\infty)$. Then there exists $\eta\in (0,\infty)$ such that
\begin{equation}\llabel{conclude}
    \min\biggl\{\int_{-\scrc\eta}^{\scrc\eta} \Dens ( x ) \, \d x  ,\int_{-\infty}^{ -\scrC\eta} \Dens ( x ) \, \d x\biggr\} >0.
\end{equation}
\end{athm}
\begin{aproof}
Throughout this proof let $\rho\in \R$ satisfy
\begin{equation}\llabel{def: rho}
    \rho=\inf\biggl\{\eta\in (0,\infty)\colon \int_{-\scrc\eta}^{\scrc\eta} \Dens ( x ) \, \d x > 0\bigg\}
\end{equation}
(cf.\ \lref{assume}).
\argument{\lref{assume};}{that
    \begin{equation}\llabel{eq1}
    \begin{split}
       & \sup_{ \eta \in (0,\infty) } \biggl(  \eta ^{ - 1 } \int_{ - \eta }^{ 0 } \allowbreak\mathbbm1_{\R\backslash\{0\}} \allowbreak( \Dens( x ) ) \, \d x \biggr)\\
       &= \biggl[\sup_{ \eta \in (0,\infty) } \biggl(\biggl(  \eta ^{ - 1 } \int_{ - \eta }^{ 0 } \allowbreak\mathbbm1_{\R\backslash\{0\}} \allowbreak( \Dens( x ) ) \, \d x\biggr)+\biggl(\eta^{-1}\int_{0}^\eta 1\,\d x \biggr)\biggr)\biggr]- 1\\
       &\geq \biggl[\sup_{ \eta \in (0,\infty) } \biggl(  \eta ^{ - 1 } \int_{ - \eta }^{ \eta } \allowbreak\mathbbm1_{\R\backslash\{0\}} \allowbreak( \Dens( x ) ) \, \d x\biggr)\biggr]- 1\geq 1\dott
       \end{split}
    \end{equation}}
    In the following  we distinguish our proof of $\lref{conclude}$ between the case $\rho=0$ and the case $\rho>0$. We first prove \lref{conclude} in the case
    \begin{equation}\llabel{case1}
        \rho=0.
    \end{equation}
    \startnewargseq
    \argument{\lref{def: rho};\lref{case1}}{that for all $\eta\in (0,\infty)$ it holds that
    \begin{equation}\llabel{case1: eq1}
         \int_{-\scrc\eta}^{\scrc\eta} \Dens ( x ) \, \d x > 0. 
    \end{equation}}
    \argument{\lref{eq1};the monotone convergence theorem}{that \begin{equation}\llabel{case1: eq2}
        \liminf_{\eta\searrow 0} \biggl(\int_{-\infty}^{ -\scrC\eta} \Dens ( x ) \, \d x\biggr)=\int_{-\infty}^{0} \Dens ( x ) \, \d x >0\dott
    \end{equation}}
    \argument{\lref{case1: eq1};\lref{case1: eq2}}[verbs=edp]{\lref{conclude} in the case $\rho=0$\dott}
    We now prove \lref{conclude} in the case
    \begin{equation}\llabel{case2}
        \rho>0.
    \end{equation}
    \startnewargseq
    \argument{\lref{def: rho};\lref{case2}}{that
    \begin{equation}\llabel{case2: eq1}
        \int_{\frac{-\rho\scrc}{2}}^{\frac{\rho\scrc}{2}} \Dens ( x ) \, \d x = 0.
    \end{equation}}
    \argument{\lref{case2: eq1};the fact that for all $x\in \R$ it holds that $\Dens(x)\geq 0$}{that
    \begin{equation}\llabel{case2: eq2}
        \int_{\frac{-\rho\scrc}{2}}^{\frac{\rho\scrc}{2}}\mathbbm 1_{\R\backslash\{0\}} (\Dens ( x )) \, \d x =0.
    \end{equation}}
    \argument{\lref{case2: eq2};}{ for all $\eta\in (0,\infty)$ that
    \begin{equation}\llabel{case2: eq3}
    \begin{split}
         &\eta ^{ - 1 } \int_{ - \eta }^{ \eta } \allowbreak\mathbbm1_{ \R \backslash { \{0\} } }\allowbreak( \Dens( x ) ) \, \d x \\
         &=\eta^{-1}\biggl(\int_{ - \eta }^{- \min\{\eta,\frac{\rho\scrc}{2}\} } \mathbbm1_{ \R \backslash { \{0\} } }\allowbreak( \Dens( x ) ) \, \d x\biggl)+\eta^{-1}\biggl(\int_{ - \min\{\eta,\frac{\rho\scrc}{2}\} }^{ \min\{\eta,\frac{\rho\scrc}{2}\} } \mathbbm1_{ \R \backslash { \{0\} } }\allowbreak( \Dens( x ) ) \, \d x\biggr)\\
         &+\eta^{-1}\biggl(\int_{  \min\{\eta,\frac{\rho\scrc}{2}\} }^{\eta } \mathbbm1_{ \R \backslash { \{0\} } }\allowbreak( \Dens( x ) ) \, \d x\biggr)\\
         &=\eta^{-1}\biggl(\int_{ - \eta }^{- \min\{\eta,\frac{\rho\scrc}{2}\} } \mathbbm1_{ \R \backslash { \{0\} } }\allowbreak( \Dens( x ) ) \, \d x\biggr)+\eta^{-1}\biggl(\int_{  \min\{\eta,\frac{\rho\scrc}{2}\} }^{ \eta } \mathbbm1_{ \R \backslash { \{0\} } }\allowbreak( \Dens( x ) ) \, \d x\biggr)\\
         &\leq 2\eta^{-1}\bigl(\eta-\min\bigl\{\textstyle\eta,\frac{\rho\scrc}{2}\bigr\}\bigr)=2-\min\bigl\{2,\frac{\rho\scrc}{\eta}\bigr\}\dott
             \end{split}
    \end{equation}}
    \argument{\lref{case2: eq3};\lref{case2}}{that for all $r\in (0,\infty)$ it holds that
    \begin{equation}\llabel{c2eq3.1}
        \sup_{\eta\in (0,r]}\biggl(\eta ^{ - 1 } \int_{ - \eta }^{ \eta } \allowbreak\mathbbm1_{ \R \backslash { \{0\} } }\allowbreak( \Dens( x ) ) \, \d x\biggr)\leq \sup_{\eta\in (0,r]}\bigl(2-\min\bigl\{2,\textstyle\frac{\rho \scrc}{\eta}\bigr\}\bigr)= 2-\min\bigl\{2,\textstyle\frac{\rho \scrc}{r}\bigr\}<2\dott
    \end{equation}}
    \argument{\lref{c2eq3.1};\lref{assume}}{that for all $r\in (0,\infty)$ it holds that
    \begin{equation}\llabel{c2eq3.1.1}
    \begin{split}
        2&\leq \sup_{\eta\in (0,\infty)}\biggl(\eta ^{ - 1 } \int_{ - \eta }^{ \eta } \allowbreak\mathbbm1_{ \R \backslash { \{0\} } }\allowbreak( \Dens( x ) ) \, \d x\biggr)\\
        &=\max\biggl\{\sup_{\eta\in (0,r]}\biggl(\eta ^{ - 1 } \int_{ - \eta }^{ \eta } \allowbreak\mathbbm1_{ \R \backslash { \{0\} } }\allowbreak( \Dens( x ) ) \, \d x\biggr),\sup_{\eta\in [r,\infty)}\biggl(\eta ^{ - 1 } \int_{ - \eta }^{ \eta } \allowbreak\mathbbm1_{ \R \backslash { \{0\} } }\allowbreak( \Dens( x ) ) \, \d x\biggr)\biggr\}
        \end{split}
    \end{equation}}
    \argument{\lref{c2eq3.1.1}}{that for all $r\in (0,\infty)$ it holds that
    \begin{equation}\llabel{c2eq3.2}
        \sup_{\eta\in [r,\infty)}\biggl(\eta ^{ - 1 } \int_{ - \eta }^{ \eta } \allowbreak\mathbbm1_{ \R \backslash { \{0\} } }\allowbreak( \Dens( x ) ) \, \d x\biggr)\geq 2\dott
    \end{equation}}
    \argument{\lref{c2eq3.2};}{that
    \begin{equation}\llabel{case2: eq4}
        \limsup_{ \eta \to\infty} \biggl(  \eta ^{ - 1 } \int_{ - \eta }^{ \eta } \allowbreak\mathbbm1_{ \R \backslash { \{0\} } }\allowbreak( \Dens( x ) ) \, \d x \biggr) \geq 2.
    \end{equation}}
    \argument{\lref{case2: eq4};}{that
    \begin{equation}\llabel{case2: eq5}
    \begin{split}
    &\liminf_{ \eta \to\infty} \biggl( \int_{ - \infty }^{ -\eta } \allowbreak\mathbbm1_{ \R \backslash { \{0\} } }\allowbreak( \Dens( x ) ) \, \d x \biggr)\\
    &=\liminf_{ \eta \to\infty} \biggl( \int_{ - \infty }^{ -\eta/2 } \allowbreak\mathbbm1_{ \R \backslash { \{0\} } }\allowbreak( \Dens( x ) ) \, \d x \biggr)=\limsup_{ \eta \to\infty} \biggl( \int_{ - \infty }^{ -\eta/2 } \allowbreak\mathbbm1_{ \R \backslash { \{0\} } }\allowbreak( \Dens( x ) ) \, \d x \biggr)\\
    &\geq \limsup_{ \eta \to\infty} \biggl( \int_{ - \eta}^{ -\eta/2 } \allowbreak\mathbbm1_{ \R \backslash { \{0\} } }\allowbreak( \Dens( x ) ) \, \d x \biggr)\geq \limsup_{ \eta \to\infty} \biggl(  \eta ^{ - 1 } \int_{ - \eta }^{ -\eta/2 } \allowbreak\mathbbm1_{ \R \backslash { \{0\} } }\allowbreak( \Dens( x ) ) \, \d x \biggr)\\
    &\geq \limsup_{ \eta \to\infty} \biggl( \biggl( \eta ^{ - 1 } \int_{ - \eta }^{ -\eta/2 } \allowbreak\mathbbm1_{ \R \backslash { \{0\} } }\allowbreak( \Dens( x ) ) \, \d x\biggr) +\biggl( \eta ^{ - 1 } \int_{ - \eta/2 }^{ \eta } \allowbreak[\mathbbm1_{ \R \backslash { \{0\} } }\allowbreak( \Dens( x ) )-1] \, \d x\biggr) \biggr)\\
        &= \limsup_{ \eta \to\infty} \biggl(  \biggl(\eta ^{ - 1 } \int_{ - \eta }^{ \eta } \allowbreak\mathbbm1_{ \R \backslash { \{0\} } }\allowbreak( \Dens( x ) ) \, \d x\biggr) -\frac 32 \biggr)\geq 2-\frac 32= \frac12>0.
        \end{split}
    \end{equation}}
    \argument{\lref{case2: eq5};}{for all $\eta\in \R$ that
    \llabel{case2: eq5'}
        $\int_{ - \infty }^{ \eta } \allowbreak\mathbbm1_{ \R \backslash { \{0\} } }\allowbreak( \Dens( x ) ) \, \d x>0$\dott}
    \argument{\lref{case2: eq5'};\lref{def: rho}}{that
    \begin{equation}\llabel{case2: eq6}
        \min\biggl\{\int_{-2\rho\scrc}^{2\rho\scrc} \Dens ( x ) \, \d x , \int_{ - \infty }^{ -2\rho\scrC} \allowbreak\mathbbm1_{ \R \backslash  \{0\} } (\Dens(x))\,\d x\biggr\}>0.
    \end{equation}}
    \argument{\lref{case2: eq6};}{\lref{conclude} in the case $\rho>0$\dott}
\end{aproof}
\begin{athm}{prop}{lem:estimate dead neuron hidden layer 1}
    Assume \cref{setting:multilayer}, let $\alpha \in (0,1)$, let $ ( \Omega, \mathcal{F}, \P) $ be a probability space, for every $k\in\N$ let $\arch_k\in \N$, $\ell_k\in (\cup_{L=1}^\infty( \{ d \} \times \{\arch_k\}\times \N^L ))$ and let $\NNelll^{k}=(\NNelll^{k}_1,\dots,\NNelll^{k}_{\fd_{\ell_k}})\colon \Omega \to \R^{\fd_{\ell_k}}$ be a random variable, 
let $ \Dens \colon \R \to [0,\infty) $ be measurable, 
assume $\sup_{ \eta \in (0,\infty) } \bigl(  \eta ^{ - 1 } \int_{ - \eta }^{ \eta } \mathbbm1_{ \R \backslash { \{0\} } }( \Dens( x ) ) \, \d x \bigr) \allowbreak\geq 2$, let $(\cki_{k,i})_{(k,i)\in \N^2}\subseteq(0,\infty)$ 
satisfy for all $k\in \N$, 
$ x_1,x_2,\dots,x_{\arch_kd+\arch_k} \in \R$ 
that
\begin{equation}\llabel{assume1}
  \P\bigl( \cap_{i=1}^{\arch_kd+\arch_k}
    \{
    \NNelll^{ k}_i 
    < x_i\}
  \bigr)
  =
  \prod_{i=1}^{\arch_kd+\arch_k}\biggl[\int_{ - \infty }^{\cki_{k,i}x_i} \Dens(y) \, \d y\biggr],
\end{equation}
 and assume $\sup_{k\in \N}\max_{i\in \{1,2,\dots,\arch_k\}}\max_{ j\in\{1,2,\dots,d\}}\allowbreak\bigl((c_{k,(i-1)d+j})^{-1}c_{k,\arch_kd +i}\bigr)<\infty$.
 Then there exists $p\in (0,1)$ which satisfies for all $k\in \N$ that
 \begin{equation}\llabel{conclude}
 \P\Big(\# \big(\inact_{ \ell_k }^{ \NNelll ^{k}}\big)\geq (\arch_k)^\alpha\Big)\geq \sum_{n=\ceil{(\arch_k)^{\alpha}} }^{\arch_k}\binom{\arch_k}{n}p^n(1-p)^{l_k-n}.
  \end{equation}
\end{athm}
\begin{aproof}
    Throughout this proof for every $k\in \N$, $i\in \{1,2,\dots,\arch_k\}$  let $ \fU_{ k,i } \subseteq \R^{\fd_{\ell_k}}$ satisfy 
\begin{equation}
\llabel{eq: def U_i: proof lem:estimate dead neuron hidden layer 1}
\begin{split}
  &\fU_{ k,i }=\{\theta\in \R^{\fd_{\ell_k}}\colon i\in \inact^{\theta}_{\ell_k}\},
\end{split}
\end{equation} 
 let $\bfc\in\R$ satisfy 
 \begin{equation}\llabel{def: c}
 \bfc=\sup_{k\in \N}\max_{i\in \{1,2,\dots,\arch_k\}}\max_{ j\in\{1,2,\dots,d\}}\bigl((c_{k,(i-1)d+j})^{-1}c_{k,\arch_kd +i}\bigr),
 \end{equation}
 and for every $k\in \N$ let $\Snode_k\subseteq\N$ satisfy $\Snode_k=\{1,2,\dots,l_k\}$.
\argument{\lref{eq: def U_i: proof lem:estimate dead neuron hidden layer 1};}{that for all $k\in\N$ it holds that
\begin{equation}\llabel{NRvvv}
   \begin{split}
       &\P\bigl(\#(\inact^{\Theta^k}_{\ell_k})\geq (l_k)^{\alpha}\bigr)\\
       &=\P\bigl(\cup_{A\subseteq\Snode_k,\, \#(A)\geq (l_k)^\alpha}\{\inact^{\Theta^k}_{\ell_k}=A\}\bigr)=\sum_{A\subseteq \Snode_k,\,\#(A)\geq (l_k)^\alpha}\P\bigl(\inact^{\Theta^k}_{\ell_k}=A\bigr)\\
&=\sum_{A\subseteq \Snode_k,\,\#(A)\geq (l_k)^\alpha}\P\Bigl(\bigl(\forall\, i\in A\colon i\in \inact^{\Theta^k}_{\ell_k}\bigr)\wedge\bigl(\forall\, i\in (\Snode_k\backslash A)\colon i\notin\inact^{\Theta^k}_{\ell_k}\bigr)\Bigr)\\
&=\sum_{A\subseteq \Snode_k,\,\#(A)\geq (l_k)^\alpha}\P\bigl((\forall\, i\in A\colon \Theta^k\in \fU_{k,i})\wedge(\forall\, i\in (\Snode_k\backslash A)\colon \Theta^k\notin \fU_{k,i})\bigr)\\
&=\sum_{A\subseteq \Snode_k,\,\#(A)\geq (l_k)^\alpha}\P\bigl(\bigl(\cap_{i\in A}\{\Theta^k\in \fU_{k,i}\}\bigr)\cap\bigl(\cap_{i\in (\Snode_k\backslash A)}\{\Theta^k\in (\R^d\backslash\fU_{k,i})\}\bigr)\bigr)\\
&=\sum_{A\subseteq \Snode_k,\,\#(A)\geq \ceil{(l_k)^\alpha}}\P\bigl(\bigl(\cap_{i\in A}\{\Theta^k\in \fU_{k,i}\}\bigr)\cap\bigl(\cap_{i\in (\Snode_k\backslash A)}(\Omega\backslash\{\Theta^k\in \fU_{k,i}\})\bigr)\bigr).
   \end{split} 
\end{equation}}
\argument{\cref{relization multi};\cref{eq:setting_definition_of_I_set};\lref{eq: def U_i: proof lem:estimate dead neuron hidden layer 1}}{for all $k\in \N$, $i\in \{1,2,\dots,l_k\}$ that
\begin{equation}\llabel{NRZZZ}
    \begin{split}
        &\fU_{k,i}\\
        & = 
  \bigl\{  
    \theta = ( \theta_1, \dots \theta_{ \fd_{ \ell_k } } ) 
    \in \R^{\fd_{ \ell_k } } 
    \colon  
    \bigl(
      \forall \, x_1,x_2,\dots,x_d \in [a,b] \colon 
      \theta_{ \arch_k d + i }  + \ssum_{j=1}^d \theta _{ ( i - 1 ) d + j } x_j < 0 
    \bigr)
  \bigr\} 
\\ & =
\textstyle 
  \bigl\{  
    \theta = ( \theta_1, \dots \theta_{ \fd_{ \ell_k } } ) 
    \in \R^{\fd_{ \ell_k } } 
    \colon  
    \sup_{ x_1,x_2,\dots,x_d \in [a,b] }
    \bigl(
      \theta_{ \arch_k d + i }  
      + 
      \ssum_{j=1}^d \theta _{ ( i - 1 ) d + j } x_j 
    \bigr)
    < 0 
  \bigr\} 
\\ & =
\textstyle 
  \bigl\{  
    \theta = ( \theta_1, \dots \theta_{ \fd_{ \ell_k } } ) 
    \in \R^{\fd_{ \ell_k } } 
    \colon  
    \bigl(
      \theta_{ \arch_k d + i }  
      + 
      \ssum_{ j = 1 }^d 
      \bigl[ 
        \sup_{ x \in [a,b] }
        \theta _{ ( i - 1 ) d + j } x 
      \bigr]
    \bigr)
    < 0 
  \bigr\} 
\\ & =
\textstyle 
  \bigl\{  
    \theta = ( \theta_1, \dots \theta_{ \fd_{ \ell_k } } ) 
    \in \R^{\fd_{ \ell_k } } 
    \colon  
    \bigl(
      \theta_{ \arch_k d + i }  
      + 
      \ssum_{ j = 1 }^d 
      \max_{ x \in \{ a, b \} }
      \bigl(
        \theta _{ ( i - 1 ) d + j } x
      \bigr)
    \bigr)
    < 0
  \bigr\}\\
  &=\bigl\{\theta = ( \theta_1, \dots \theta_{ \fd_{ \ell_k } } ) 
    \in \R^{\fd_{ \ell_k } } 
    \colon  \textstyle\sum\nolimits_{j=1}^d \max\{\theta_{(i-1)d+j}a,\theta_{(i-1)d+j}b\}<-\theta_{l_k d+i}\bigr\}.
    \end{split}
\end{equation}}
\argument{\lref{assume1};\cref{lemma X: item 3} in \cref{Lemma X}}{that \llabel{nr1} for all $k\in \N$ it holds that $\Theta^k_i$, $i\in \{1,2,\dots,l_k d+l_k\}$, are independent\dott}
\argument{\lref{nr1};}{that \llabel{nr1.5} for all $k\in \N$ it holds that $(\Theta^k_{(i-1)d+1},\Theta^k_{(i-1)d+2},\dots,\Theta^k_{(i-1)d+d},\Theta^k_{l_k d+i})$, $i\in \{1,2,\dots,l_k\}$, are independent\dott}
\argument{\lref{nr1.5};\lref{NRZZZ}}{that for all $k\in \N$ it holds that \llabel{nr2}$\{\Theta^k \in \fU_{k,i}\}\in \cF$, $i\in \Snode_k$, are independent\dott}
\argument{\lref{nr2};\lref{NRvvv};the fact that for all $N\in \N\backslash\{1\}$ and all independent $A_n\in \mathcal F$, $n\in \{1,2,\dots,N\}$, it holds that $\P((\cap_{n=1}^{N-1}A_n)\cap(\Omega\backslash A_N))=\E[\mathbbm 1_{(\cap_{n=1}^{N-1}A_n)}(1-\mathbbm 1_{A_N})]=(\prod_{n=1}^{N-1}\P(A_n))(1-\P(A_N))$}{that for all $k\in \N$ it holds that
\begin{equation}\llabel{eq19}
\begin{split}
    &\P(\#(\inact^k_{\ell_k})\geq (l_k)^\alpha)\\
    &=\sum_{A\subseteq \Snode_k,\,\#(A)\geq \ceil{(l_k)^\alpha}}\biggl(\biggl[\textstyle\prod\limits_{i\in A}\P(\Theta^k\in \fU_{k,i})\biggr]\biggl[\prod
    \limits_{i\in (\Snode_k\backslash A)}\P(\Omega\backslash\{\Theta^k\in \fU_{k,i}\})\biggr]\biggr)\\
    &=\sum_{A\subseteq \{1,2,\dots,l_k\},\,\#(A)\geq \ceil{(l_k)^\alpha}}\biggl(\biggl[\textstyle\prod\limits_{i\in A}\P(\Theta^k\in \fU_{k,i})\biggr]\biggl[\prod
    \limits_{i\in (\{1,2,\dots,l_k\}\backslash A)}[1-\P(\Theta^k\in \fU_{k,i})]\biggr]\biggr)
    \end{split}
\end{equation}}
\argument{\cref{lemma X: item 1} in \cref{Lemma X};\lref{assume1}}{that 
\begin{equation}\llabel{arggg1}
    \biggl[\int_{-\infty}^0 \Dens(x)\, \d x\biggr]+\biggl[\int_{0}^\infty \Dens(x)\, \d x\biggr]= \int_{-\infty}^\infty \Dens(x)\, \d x=1\dott
\end{equation}
}
\argument{\lref{arggg1}; the assumption that $\sup_{ \eta \in (0,\infty) } \bigl(  \eta ^{ - 1 } \int_{ - \eta }^{ \eta } \mathbbm1_{ \R \backslash { \{0\} } }( \Dens( x ) ) \, \d x \bigr) \allowbreak\geq 2$}{that 
\begin{equation}\llabel{need to prove 2}
    \int_{-\infty}^0\Dens(x)\,\d x <1\dott
\end{equation}}
\argument{\lref{def: c};\cref{existence of eta}}{that there exists $\eta\in (0,\infty)$ which satisfies 
\begin{equation}\llabel{def:eta}
   \min\biggl\{\int_{- \frac{\eta}{2 d\bfc \max \cu{\abs{a} , \abs{b} } } }^{\frac{\eta}{2 d\bfc \max \cu{\abs{a} , \abs{b} } } } \Dens ( x ) \, \d x ,\int_{-\infty}^{ -\eta/2} \Dens ( x ) \, \d x\biggr\} >0.
\end{equation}}
In the following let $p\in \R$ satisfy
\begin{equation}\llabel{NRPPP}
    p=\biggl[ \int_{-\infty}^{ -\eta/2} \Dens ( x ) \, \d x \biggr]
			\Biggl[ \int_{- \tfrac{\eta}{2  \bfc d \max \cu{\abs{a} , \abs{b} } } }^{\tfrac{\eta}{2  \bfc d\max \cu{\abs{a} , \abs{b} } } } \Dens ( x ) \, \d x \Biggr] ^d\dott
\end{equation}
\startnewargseq
\argument{\lref{arggg1};\lref{need to prove 2};\lref{def:eta};\lref{NRPPP}}{that
\begin{equation}\llabel{NRQQQ}
    0<p\leq \int_{-\infty}^{-\eta/2}\Dens(x)\,\d x\leq \int_{-\infty}^0\Dens(x)\,\d x<1\dott
    \end{equation}}
\argument{\lref{def: c}}{that for all $k\in \N$, $i\in \{1,2,\dots,\arch_k\}$, $j\in\{1,2,\dots,d\}$ it holds that
\begin{equation}\llabel{Eq1}
  c_{k,\arch_k d+i}\leq \bfc c_{k,(i-1)d+j} \dott
\end{equation}}
\argument{\lref{Eq1};\lref{NRZZZ}}{for all $k \in \N$, $i\in\{1,2,\dots,\arch_k\}$ that 
	\begin{equation}\llabel{P>0}
		\begin{split}
  &\P(\Theta^k\in \fU_{k,i})\\
			&= \P \rbr[\big]{\ssum_{j=1}^d \max \cu{\Theta^k_{ (i-1)d+j } a , \Theta^k_{  (i-1)d+j } b } < - \Theta ^k_{ \arch_k d + i } } \\
			&=  \P \rbr[\big]{\ssum_{j=1}^d \max \cu{c_{k,\arch_k d+i}\Theta^k_{  (i-1)d+j } a , c_{k,\arch_k d+i} \Theta^k_{ (i-1)d+ j } b } < - c_{k,\arch_kd+i} \Theta^k_ { \arch_k d + i } } \\
			& \ge \P \rbr[\big]{ \bigl\{ -c_{k,\arch_k d+i} \Theta^k_{ \arch_k d + i } >  \tfrac{\eta}{2} \bigr\} \cap \bigl\{\abs[\big]{\ssum_{j=1}^d \max \cu{c_{k,\arch_k d+i}\Theta^k_{  (i-1)d+j } a , c_{k,\arch_k d+i} \Theta^k_{  (i-1)d+j } b } } < \tfrac{\eta}{2} \bigr\}}  \\
   &\ge \P \rbr[\big]{ \bigl\{ c_{k,\arch_k d+i} \Theta^k_{ \arch_k d + i } < - \tfrac{\eta}{2} \bigr\} 
   \\
   &\quad \cap \bigl\{\ssum_{j=1}^d \max \cu{c_{k,\arch_k d+i}|\Theta^k_{  (i-1)d+j } a| , c_{k,\arch_k d+i} |\Theta^k_{  (i-1)d+j } b| } < \tfrac{\eta}{2} \bigr\}}  \\
   & \ge \P \rbr[\big]{ \bigl\{ c_{k,\arch_k d+i} \Theta^k_{ \arch_k d + i } < - \tfrac{\eta}{2}\bigr \}\\
   &\quad\cap \bigl\{\ssum_{j=1}^d \max \cu{\bfc c_{k,(i-1)d+j}|\Theta^k_{  (i-1)d+j } a|,\bfc c_{k,(i-1)d+j} |\Theta^k_{  (i-1)d+j } b| }  < \tfrac{\eta}{2} \bigr\} } \\
			& = \P \rbr[\big]{ \bigl\{ c_{k,\arch_k d+i} \Theta^k_{  \arch_k d + i } <- \tfrac{\eta}{2}  \bigr\} \\
   &\quad \cap \bigl\{\textstyle\sum_{j=1}^d \max \cu{ \abs{ c_{k,(i-1)d+j}\Theta^k_{ (i-1)d+ j } a  } ,\abs{ c_{k,(i-1)d+j} \Theta^k_{(i-1)d+ j } b } }  < \tfrac{\eta}{2 \bfc } \bigr\}} 
\end{split}
\end{equation} } 
\argument{\cref{lemma X: item 2,lemma X: item 3} in \cref{Lemma X};\lref{assume1};\lref{NRPPP}}{\llabel{independent} that for all $k \in \N$, $i\in \{1,2,\dots,l_k\}$ it holds that
\begin{equation}\llabel{NRXXX}
\begin{split}
    &\P(\Theta^k\in \fU_{k,i})\\
     &\ge \P \rbr[\big]{ \bigl\{ c_{k,\arch_k d+i} \Theta^k_{  \arch_k d + i } < - \tfrac{\eta}{2}  \bigr\} \\
   &\quad \cap \bigl\{\textstyle\max_{j\in \{1,2,\dots,d\}} \max \cu{ \abs{ c_{k,(i-1)d+j}\Theta^k_{ (i-1)d+ j } a  } ,\abs{ c_{k,(i-1)d+j} \Theta^k_{(i-1)d+ j } b } }  < \tfrac{\eta}{2 \bfc d} \bigr\}}\\
   & = \P \rbr[\big]{ \bigl\{ c_{k,\arch_k d+i} \Theta^k_{  \arch_k d + i }< - \tfrac{\eta}{2}  \bigr\} \\
   &\quad \cap \bigl\{\textstyle\max_{j\in \{1,2,\dots,d\}}  \abs{ c_{k,(i-1)d+j}\Theta^k_{ (i-1)d+ j }   }  < \tfrac{\eta}{2 \bfc d\max\{|a|,|b|\}} \bigr\}}\\
			& = \P \bigl( c_{k,\arch_k d+i}  \Theta^k_{\arch_kd + i } < - \tfrac{\eta}{2}\bigr) \br*{ \textstyle\prod_{j=1}^d \P\rbr[\big]{| c_{k,(i-1)d+j} \Theta^k_{ (i-1)d+j } |< \tfrac{\eta}{2  \bfc d\max \cu{\abs{a} , \abs{b} } }  } }
			= p\dott
   \end{split}
\end{equation}}
\argument{\lref{NRXXX};\lref{eq19};\cref{item 2: increasing property} in \cref{increasing property}}{that for all $k\in \N$ it holds that
\begin{equation}\llabel{eqqq1}
    \P\Big(\# \big(\inact_{ \ell_k }^{ \NNelll ^{k}}\big)\geq (\arch_k)^\alpha\Big)\geq \sum_{n=\ceil{(\arch_k)^{\alpha}} }^{\arch_k}\binom{\arch_k}{n}p^n(1-p)^{l_k-n}\dott
\end{equation}}
\argument{\lref{eqqq1};\lref{NRQQQ}}{\lref{conclude}\dott}
\end{aproof}
\begin{athm}{lemma}{lem:estimate dead neuron hidden layer 1.1}
    Assume \cref{setting:multilayer}, let $\alpha \in (0,1)$, let $ ( \Omega, \mathcal{F}, \P) $ be a probability space, for every $k\in\N$ let $\arch_k\in \N$, $\ell_k\in (\cup_{L=1}^\infty( \{ d \} \times \{\arch_k\}\times \N^L ))$ and let $\NNelll^{k}=(\NNelll^{k}_1,\dots,\NNelll^{k}_{\fd_{\ell_k}})\colon \Omega \to \R^{\fd_{\ell_k}}$ be a random variable, 
let $ \Dens \colon \R \to [0,\infty) $ be measurable, 
assume $\sup_{ \eta \in (0,\infty) } \bigl(  \eta ^{ - 1 } \int_{ - \eta }^{ \eta } \mathbbm1_{ \R \backslash { \{0\} } }\allowbreak( \Dens(\allowbreak x ) ) \, \d x \bigr) \geq 2$, let $(\cki_{k,i})_{(k,i)\in \N^2}\subseteq(0,\infty)$ 
satisfy for all $k\in \N$, 
$ x_1,x_2,\dots,x_{\arch_kd+\arch_k} \in \R$ 
that
\begin{equation}
  \P\bigl( \cap_{i=1}^{\arch_kd+\arch_k}
    \{
    \NNelll^{ k}_i 
    < x_i\}
  \bigr)
  =
  \prod_{i=1}^{\arch_kd+\arch_k}\biggl[\int_{ - \infty }^{\cki_{k,i}x_i} \Dens(y) \, \d y\biggr],
  \end{equation} and assume $\sup_{k\in \N}\allowbreak\max_{i\in \{1,2,\dots,\arch_k\}}\max_{ j\in\{1,2,\dots,d\}}\bigl((c_{k,(i-1)d+j})^{-1}c_{k,\arch_k d+i}\bigr)<\infty$. Then there exists $\boundexp\in (0,\infty)$ which satisfies for all $k\in \N$ that
    \begin{equation}\llabel{conclude}
\P\Big(\# \big(\inact_{ \ell_k }^{ \NNelll ^{k}}\big)< (\arch_k)^\alpha\Big) \leq \exp(-\boundexp\arch_k).
    \end{equation}
\end{athm}
\begin{aproof}
\argument{\cref{lem:estimate dead neuron hidden layer 1};}{that there exists $p\in (0,1)$ which satisfies for all $k\in \N$ that
\begin{equation}\llabel{NR1}
    \P\Big(\# \big(\inact_{ \ell_k }^{ \NNelll ^{k}}\big)\geq (\arch_k)^\alpha\Big)\geq \sum_{n=\ceil{(\arch_k)^{\alpha}} }^{\arch_k}\binom{\arch_k}{n}p^n(1-p)^{l_k-n}\dott
\end{equation}}
\startnewargseq
\argument{\cref{lem: binary1};}{that there exists $\boundexp\in (0,\infty)$ which satisfies for all $k\in \N$ that 
\begin{equation}\llabel{NR2}
    \sum_{n=\ceil{(\arch_k)^{\alpha}} }^{\arch_k}\binom{\arch_k}{n}p^n(1-p)^{l_k-n}\geq 1-\exp(-\boundexp \arch_k)\dott
\end{equation}}
\startnewargseq
   \argument{\lref{NR1};\lref{NR2}}{that for all $k\in \N$ it holds that
   \begin{equation}\llabel{eq1}
   \begin{split}
       &\P\Big(\# \big(\inact_{ \ell_k }^{ \NNelll ^{k}}\big)< (\arch_k)^\alpha\Big)=1-\P\Big(\# \big(\inact_{ \ell_k }^{ \NNelll ^{k}}\big)\geq (\arch_k)^\alpha\Big)\\
       &\leq 1-\textstyle\biggl[\sum\limits_{n=\ceil{(\arch_k)^{\alpha}} }^{\arch_k}\binom{\arch_k}{n}p^n(1-p)^{l_k-n}\biggr]
       \leq 1-(1-\exp(-\boundexp \arch_k))=\exp(-\boundexp \arch_k)\dott
     \end{split}
   \end{equation}
   }
   \argument{\lref{eq1};}{\lref{conclude}\dott}
\end{aproof}
\begin{athm}{cor}{cor:estimate dead neuron hidden layer 1.1}
     Assume \cref{setting:multilayer}, let $ ( \Omega, \mathcal{F}, \P) $ be a probability space, for every $k\in\N$ let $\arch_k\in \N$, $\ell_k\in (\cup_{L=1}^\infty( \{ d \} \times \{\arch_k\}\times \N^L ))$ and let $\NNelll^{k}=(\NNelll^{k}_1,\dots,\NNelll^{k}_{\fd_{\ell_k}})\colon \Omega \to \R^{\fd_{\ell_k}}$ be a random variable, 
let $ \Dens \colon \R \to [0,\infty) $ be measurable, 
assume $\sup_{ \eta \in (0,\infty) } \bigl(  \eta ^{ - 1 } \int_{ - \eta }^{ \eta } \mathbbm1_{ \R \backslash { \{0\} } }( \Dens( x ) ) \, \d x \bigr) \geq 2$,
 let $(c_{k,i})_{(k,i)\in \N^2}\subseteq(0,\infty)$ 
satisfy for all $k\in \N$, 
$ x_1,x_2,\dots,x_{\arch_kd+\arch_k}\in \R$ 
that
\begin{equation}
  \P\bigl( \cap_{i=1}^{\arch_kd+\arch_k}
    \{
    \NNelll^{ k}_i 
    < x_i\}
  \bigr)
  =
  \prod_{i=1}^{\arch_kd+\arch_k}\biggl[\int_{ - \infty }^{\cki_{k,i}x_i} \Dens(y) \, \d y\biggr],
\end{equation}
assume $\sup_{k\in \N}\allowbreak\max_{i\in \{1,2,\dots,\arch_k\}}\max_{ j\in\{1,2,\dots,d\}}\bigl((c_{k,(i-1)d+j})^{-1}c_{k,\arch_k d+i}\bigr)<\infty$, and let $\bfc\in \R$. Then there exists $\boundexp\in(0,\infty)$ which satisfies for all $k\in \N$ that
    \begin{equation}\llabel{conclude}
\P\Big(\# \big(\inact_{ \ell_k }^{ \NNelll ^{k}}\big)<  \bfc\Big) \leq \boundexp^{-1}\exp(-\boundexp\arch_k).
    \end{equation}
\end{athm}
\begin{aproof}
    \argument{\cref{lem:estimate dead neuron hidden layer 1.1};}{that there exists $\rho\in (0,\infty)$  which satisfies for all $k\in \N$ that 
    \begin{equation}\llabel{arg1}
        \P\Big(\# \big(\inact_{ \ell_k }^{ \NNelll ^{k}}\big)<  (\arch_k)^{1/2}\Big)\leq \exp(-\rho\arch_k)\dott   
    \end{equation}}
    \startnewargseq
 \argument{\lref{arg1}}{that for all $k\in \N$ with $\arch_k\geq \bfc^2$ it holds that 
 \begin{equation}\llabel{arg2}
\P\Big(\# \big(\inact_{ \ell_k }^{ \NNelll ^{k}}\big)<  \bfc\Big) \leq \exp(-\rho\arch_k)\dott
\end{equation}}
\argument{the fact that for all $A\in \cF$ it holds that $\P(A)\leq 1$}{that for all $k\in \N$ with $\arch_k\leq \bfc^2$ it holds that
\begin{equation}\llabel{argg1}
    \P\Big(\# \big(\inact_{ \ell_k }^{ \NNelll ^{k}}\big)<  \bfc\Big)\leq 1=\exp(\bfc^2)\exp(-\bfc^2)\leq \exp(\bfc^2)\exp(-\arch_k)\dott
\end{equation}}
\argument{\lref{argg1};\lref{arg2}}{ for all $k\in \N$ that
\begin{equation}\llabel{argg2}
\begin{split}
   & \P\Big(\# \big(\inact_{ \ell_k }^{ \NNelll ^{k}}\big)<  \bfc\Big)\leq \exp(\bfc^2)\exp(-\arch_k\min\{\rho,1\})\\
   &\leq \bigl[\exp(-\bfc^2)\bigr]^{-1}\exp(-\arch_k\min\{\exp(-\bfc^2),\rho,1\})\\
   &\leq \bigl[\min\{\exp(-\bfc^2),\rho\}\bigr]^{-1}\exp(-\arch_k\min\{\exp(-\bfc^2),\rho\})\dott
    \end{split}
\end{equation}}
\argument{\lref{argg2}}{that there exists $\boundexp\in(0,\infty)$ which satisfies for all $k\in \N$  that 
 \begin{equation}\llabel{arg3}
\P\Big(\# \big(\inact_{ \ell_k }^{ \NNelll ^{k}}\big)<  \bfc\Big) \leq \boundexp^{-1}\exp(-\boundexp\arch_k)\dott
\end{equation}}
\argument{\lref{arg3};}{\lref{conclude}\dott}
\end{aproof}
\begin{athm}{cor}{cor:estimate dead neuron hidden layer 1}
     Assume \cref{setting:multilayer}, let $ ( \Omega, \mathcal{F}, \P) $ be a probability space, for every $k\in\N$ let $\arch_k\in \N$, $\ell_k\in  (\cup_{L=1}^\infty( \{ d \} \times \{\arch_k\}\times \N^L ))$ and let $\NNelll^{k}=(\NNelll^{k}_1,\dots,\NNelll^{k}_{\fd_{\ell_k}})\colon \Omega \to \R^{\fd_{\ell_k}}$ be a random variable, 
let $ \Dens \colon \R \to [0,\infty) $ be measurable, 
assume $\sup_{ \eta \in (0,\infty) } \bigl(  \eta ^{ - 1 } \int_{ - \eta }^{ \eta } \mathbbm1_{ \R \backslash { \{0\} } }( \Dens( x ) ) \, \d x \bigr) \geq 2$,
let $(\cki_{k,i})_{(k,i)\in \N^2}\subseteq(0,\infty)$ 
satisfy for all $k\in \N
$,
$x_1,x_2,\dots,x_{\fd_{\ell_k}} \in \R$
that
\begin{equation}\llabel{assume}
  \P \bigl( \cap_{i=1}^{\fd_{\ell_k}}
    \big\{
    \NNelll^{k}_i 
    < x_i\big\}
  \bigr)
  =
  \prod_{i=1}^{\fd_{\ell_k}}\biggl[\int_{ - \infty }^{\cki_{k,i}x_i} \Dens(y) \, \d y\biggr],
\end{equation} 
assume $\sup_{k\in \N}\allowbreak\max_{i\in \{1,2,\dots,\arch_k\}}\max_{ j\in\{1,2,\dots,d\}}\allowbreak\bigl((c_{k,(i-1)d+j})^{-1}c_{k,\arch_k d+i}\bigr)<\infty$, and let $\bfc\in \R$, $\varepsilon\in (0,\infty)$. Then there exists $N\in \N$ which satisfies for all $k\in \N$ with $\arch_k>N$  that
    \begin{equation}\label{eq: conclude_cor: estimate dead neuron hidden layer 1}
\P\Big(\# \big(\inact_{ \ell_k }^{ \NNelll ^{k}}\big)<  \bfc\Big) <\varepsilon.
    \end{equation}
\end{athm}
\begin{aproof}
\argument{\cref{Lemma X};\lref{assume}}{that \llabel{arg1} 
\begin{enumerate}[label=(\roman*)]
     \item \llabel{item 1} it holds that $\int_{-\infty}^{\infty}\Dens(y)\,\d y=1$,
    \item \llabel{item 2} it holds for all $k\in \N$, $i\in \{1,2,\dots,\fd_{\ell_k}\}$, $x\in \R$ that $\P(\Theta^k_{i}<x)=\int_{-\infty}^{\cki_{k,i}x}\Dens(y)\,\d y$, and
    \item \llabel{item 3} it holds for all $k\in \N$ that $c_{k,i}\Theta_i^{k}$, $i\in \{1,2,\dots,\fd_{\ell_k}\}$, are \iid\ 
\end{enumerate}}
\startnewargseq
\argument{\lref{item 2, item 3};}{that for all $k\in \N$,
$ x_1,x_2,\dots,x_{\arch_kd+\arch_k} \in \R$ 
it holds that
\begin{equation}\llabel{eq0}
  \P\bigl( \cap_{i=1}^{\arch_kd+\arch_k}
    \{
    \NNelll^{k}_i 
    < x_i\}
  \bigr)
  =
\prod_{i=1}^{\arch_kd+\arch_k}\biggl[\int_{ - \infty }^{\cki_{k,i}x_i} \Dens(y) \, \d y\biggr]\dott
\end{equation}}
\argument{\lref{eq0};\cref{cor:estimate dead neuron hidden layer 1.1};}{that there exists $\boundexp\in(0,\infty)$ which satisfies for all $k\in \N$ that
    \begin{equation}\llabel{eq1}
\P\Big(\# \big(\inact_{ \ell_k }^{ \NNelll ^{k}}\big)<  \bfc\Big) \leq \boundexp^{-1}\exp(-\boundexp\arch_k)\dott
    \end{equation}}
    \startnewargseq
    \argument{\lref{eq1};}{that for all $k\in \N$ with $\arch_k>-\boundexp^{-1}\ln(\boundexp\varepsilon)$ it holds that 
    \begin{equation}\llabel{eqq1}
        \P\Big(\# \big(\inact_{ \ell_k }^{ \NNelll ^{k}}\big)<  \bfc\Big)< \boundexp^{-1}\exp(\boundexp\boundexp^{-1}\ln(\boundexp\varepsilon))=\varepsilon.
    \end{equation}}
    \argument{\lref{eqq1};}{\cref{eq: conclude_cor: estimate dead neuron hidden layer 1}\dott}
\end{aproof}
\section{Improving empirical risks in the training of deep ANNs}\label{sec: improve risk}
In the main result of this section, \cref{cor: improve risk} in \cref{subsec: improving general limiting data} below, 
we show that the infimal value over the risk values 
of all \ANNs\ with a given fixed architecture 
is strictly smaller than the infimal value over the risk values 
of all \ANNs\ with this architecture and at least three inactive neurons 
on the first hidden layer. 
In our proof of \cref{cor: improve risk} we make, among other things, use of some well-known properties 
of convex sets and convex hulls, respectively, which we recall in \cref{lemma Y}, \cref{lemma Z}, \cref{seperation theorem}, 
\cref{lemma W'}, and \cref{lemma W} in \cref{subsec: geometric preliminaries} below. Only for completeness we include in this section detailed proofs for the results in \cref{subsec: geometric preliminaries}.
\subsection{Mathematical description of empirical risks for data driven learning problems}
\begin{setting}\label{setting: improve risk}
Assume \cref{setting:multilayer}, let $M\in \N$, for every $m\in\{1,2,\dots,M\}$ let $\X^m=(\X^m_{1},\dots,\allowbreak\X^m_{d})\in [a,b]^d$, $\Y^m \in \R$, assume $\#\{\X^1,\X^2,\dots,\X^M\}=M$, let $\fL\in \N\backslash\{1\}$, $\bfl=(\bfl_0,\bfl_1,\dots,\bfl_{\fL})\in \N^{\fL+1}$ satisfy $\bfl_0=d$ and $\bfl_\fL=1$, let
$ 
  \cL \colon \R^{ \fd_{ \nnode } } \to \R 
$
satisfy for all 
$ \theta \in \R^{ \fd_{ \nnode} }$
that
\begin{equation}
\label{eq:empirical_risk_for_mini_batch-conj3:setting: improve risk}
  \cL( \theta) 
  = 
  \frac{ 1 }{ M } 
  \biggl[ 
 \textstyle \sum\limits_{ m = 1 }^{ M}  
    \abs{
      \mN^{\fL,\theta}_{\nnode}(\X^m)
      - 
      \Y^m
    }^2
  \biggr]
  ,
\end{equation}
 let $(\varTheta^n)_{n\in \N}\subseteq \R^{\fd_\nnode}$ satisfy $\limsup_{n\to\infty}\cL(\varTheta^n)>0$, let $\bfY^1,\bfY^2,\dots,\bfY^M\in \R$ satisfy   $\limsup_{n\to \infty}\allowbreak\sum_{m=1}^M|\mN^{\fL,\varTheta^n}_{\nnode}(\X^m)
      -\bfY^m|=0$,  and let $\bfA_z\subseteq\N$, $z\in \Z$, and $\bfM_z\subseteq\N$, $z\in \Z$, satisfy for all $z\in \Z\backslash\{0\}$ that 
       \begin{equation}\label{def: M}
         \bfM_z=\{m\in \{1,2,\dots,M\}\colon z\bfY^m>z\Y^m\}, \qquad \bfA_0=\{m\in \{1,2,\dots,M\}\colon \bfY^m=\Y^m\},
 \end{equation}
 \begin{equation}\label{def: A}
        \text{and} \qquad\bfA_z=\big\{m\in \{1,2,\dots,M\}\colon  z\bfY^m\geq \max\nolimits_{n\in\{1,2,\dots,M\}}(z\bfY^{n})\big\}.
        \end{equation}
\end{setting}
\begin{athm}{lemma}{estimate lim}
    Assume \cref{setting: improve risk}. Then
    \begin{equation}\label{conclude: estimate lim}
        \liminf_{n\to \infty}\cL(\varTheta^n)=\limsup_{n\to \infty}\cL(\varTheta^n)=\frac 1M \bigg[\textstyle\sum\limits_{m=1}^M|\bfY^m-Y^m|^2\bigg]>0\dott
    \end{equation}
\end{athm}
\begin{aproof}
\argument{\cref{eq:empirical_risk_for_mini_batch-conj3:setting: improve risk}; the assumption that $\limsup_{n\to \infty}\allowbreak\sum_{m=1}^M|\mN^{\fL,\varTheta^n}_{\nnode}(\X^m)
      -\bfY^m|=0$; the assumption that $\limsup_{n\to\infty}\cL(\varTheta^n)>0$}{that 
      \begin{equation}\llabel{eq1}
      \begin{split}
 0&<\limsup_{n\to\infty}\cL(\varTheta^n)\\
 &=\limsup_{n\to\infty}\bigg(\frac 1M \biggl[ 
 \textstyle \sum\limits_{ m = 1 }^{ M}  
    \abs{
      \mN^{\fL,\varTheta^n}_{\nnode}(\X^m)
      - 
      \Y^m
    }^2
  \biggr] \bigg)\\
  &=\lim_{n\to\infty}\bigg(\frac 1M \biggl[ 
 \textstyle \sum\limits_{ m = 1 }^{ M}  
    \abs{
      \mN^{\fL,\varTheta^n}_{\nnode}(\X^m)
      - 
      \Y^m
    }^2
  \biggr] \bigg)\\
  &=\frac 1M \bigg[\textstyle\sum\limits_{m=1}^M|\bfY^m-Y^m|^2\bigg]\dott
  \end{split}
      \end{equation}}
\end{aproof}
 \subsection{Interpolating indicator functions through ANNs}
\begin{athm}{prop}{prop: seperation}
    Let $M,d\in \N$, $\delta\in \R$, $x_1,x_2,\dots,x_M \in \R^d$, $k\in \{1,2,\dots,M\}$ and assume $\#\{x_1,x_2,\dots,x_M\}=M$. Then there exist $w\in \R^d$, $a_1,a_2,a_3,b_1,b_2,b_3\in \R$ such that for all $m\in\{1,2,\dots,M\}$ it holds that
         \begin{equation}\label{prop: seperation: 1 point property}
         \textstyle
           \sum\limits_{i=1}^3 \big(a_i\max\{\langle w,x_m\rangle+b_i,0\}\big)=\delta\mathbbm{1}_{\{k\}}(m).
           \end{equation}
\end{athm}
\begin{aproof}
Throughout this proof assume without loss of generality that $M>1$ (otherwise choose $w=0$, $a_1=a_2=b_1=b_2=0$, $a_3=\delta$, and $b_3=1$), for every $v\in \R^d$ let $v^\perp\subseteq\R^d$ satisfy $v^\perp=\{u\in \R^d\colon \spro{u,v}=0\}$, for every $i\in\{1,2,\dots,M-1\}$ let $v_i\in \R^d$ satisfy 
\begin{equation}\llabel{def:v_i}
v_i=x_{i+\mathbbm 1_{[k,M]}(i)}-x_k,
\end{equation}
and let $K\in \N_0$ satisfy
\begin{equation}\llabel{ARG2}
    K=\max\big\{m\in \{0,1,\dots,M-1\}\colon  (\cup_{i=1}^m (v_i)^{\perp})\neq \R^d\big\}\dott
\end{equation}
\argument{the assumption that $\#\{x_1,x_2,\dots,x_M\}=M$;}{that for all $m\in \{1,2,\dots,M\}\backslash\allowbreak\{k\}$ it holds that
\begin{equation}\llabel{arg'1}
x_m-x_k\neq 0\dott
\end{equation}}
\argument{\lref{arg'1};\lref{def:v_i}}{\llabel{arg''2} for all $i\in \{1,2,\dots,M-1\}$ that $v_i\neq 0$\dott}
\argument{\lref{arg''2};}{for all $i\in\{1,2,\dots,M-1\}$ that \llabel{arg''1}$v_i\notin (v_i)^\perp$\dott}
\argument{\lref{arg''1}}{for all $i\in \{1,2,\dots,M-1\}$ that
\begin{equation}\llabel{ARG1}
(v_i)^\perp \neq\R^d\dott
\end{equation}}
In the following we prove that
\begin{equation}\llabel{K<M}
    K=M-1.
\end{equation}
We prove \lref{K<M} by a contradiction. In the following we thus assume that 
\begin{equation}\llabel{assume}
    K<M-1.
\end{equation}
\startnewargseq
\argument{\lref{ARG2};\lref{assume}}{ that 
\begin{equation}\llabel{ARG2.5}
    (\cup_{i=1}^{K} (v_i)^{\perp})\neq\R^d=(\cup_{i=1}^{K+1} (v_i)^{\perp})\dott
\end{equation}}
\argument{\lref{ARG2.5}; \lref{ARG1}; the fact that $K+1\leq M-1$}{that
there exist $\bfv,\bfw\in \R^d$ which satisfy
\begin{equation}\llabel{ARG3}
    \bfv\notin (\cup_{i =1}^K (v_i)^{\perp}), \qquad  \bfv\in (v_{K+1})^{\perp},\qqandqq \bfw\notin (v_{K+1})^\perp\dott
\end{equation}}
\startnewargseq
\argument{\lref{ARG3};}{ \llabel{ARG3.5}for all $i\in\{1,2,\dots,K\}$ that $\spro{\bfv,v_i}\neq 0$\dott}
 \argument{\lref{ARG3.5};}{that there exists
$\alpha\in\R\backslash\{0\}$ which satisfies for all $i\in \{1,2,\dots,K\}$ that
\begin{equation}\llabel{ARG5}
\spro{\bfv+\alpha\bfw,v_i}=\spro{\bfv,v_i}+\alpha\spro{\bfw,v_i}\neq 0\dott
\end{equation}} 
\startnewargseq
\argument{\lref{ARG3}; the fact that $\alpha\neq 0$}{that 
\begin{equation}\llabel{ARG5'} 
\spro{\bfv+\alpha\bfw,v_{K+1}}=\alpha\spro{\bfw,v_{K+1}}\neq 0 \dott
\end{equation}}
\argument{\lref{ARG5'};\lref{ARG2.5};\lref{ARG5}}{that 
\begin{equation}\llabel{ARG6}
\bfv+\alpha\bfw\notin(\cup_{i=1}^{K+1} (v_i)^{\perp})=\R^d\dott
\end{equation}}
\startnewargseq
This contradiction \proves[ped] \lref{K<M}\dott
\startnewargseq
\argument{\lref{K<M}}{that \begin{equation}\llabel{arg'2}
    \cup_{m \in \{1,2,\dots,M\}\backslash \{k\}} ((x_m-x_k)^{\perp})\neq \R^d\dott
\end{equation}}
\argument{\lref{arg'2};}{ that there exists $w\in \R^d$ which satisfies 
\begin{equation}\llabel{arg'2.5}
w\notin\cup_{m \in \{1,2,\dots,M\}\backslash \{k\}} ((x_m-x_k)^{\perp})
\dott
\end{equation}}
\argument{\lref{arg'2.5}}{ for all $m\in \{1,2,\dots,M\}\backslash\{k\}$ that \llabel{arg'3}$\spro{\W,x_m-x_k}\neq 0$\dott}
\argument{\lref{arg'3};}{
     that there exist $\B\in \R$ which satisfies for all $m\in \{1,2,\dots,M\}\backslash\{k\}$ that
    \begin{equation}\llabel{proof prop: seperation: def w and b: 1 point}
   \spro{\W,x_m}+\B\neq 0=\spro{\W,x_k}+\B \dott
   \end{equation}}
   In the following let $\varepsilon \in (0,\min_{m\in \{1,2,\dots,M\}\backslash\{k\}}\abs{\spro{\W,x_m}+\B})$, $a_1,a_2,a_3,b_1,b_2,b_3\in \R$ satisfy
   \begin{equation}\llabel{def wab}
   \begin{split}
       &a_1=\frac{2\delta}{\varepsilon},\qquad a_2=\frac{10\delta}{\varepsilon},\qquad a_3=\frac{-12\delta}{\varepsilon},
       \end{split}
   \end{equation}
   \begin{equation}
       \llabel{def wab2}
       b_1=\B+\frac{\varepsilon}{2}, \qquad b_2=\B-\varepsilon,\qquad \text{and} \qquad
      b_3=\B-\frac{3\varepsilon}{4}. 
   \end{equation}
   \startnewargseq
    \argument{\lref{proof prop: seperation: def w and b: 1 point}; \lref{def wab}; \lref{def wab2}; the fact that $\varepsilon>0$} {that \llabel{property}$\spro{w,x_k}+b_1=\frac\varepsilon 2>0$, $\spro{w,x_k}+b_2=-\varepsilon<0$, and $\spro{w,x_k}+b_3=\frac{-3\varepsilon}{4}<0$\dott}
    \argument{\lref{property};\lref{def wab};\lref{def wab2}}
    {that 
    \begin{equation}\llabel{at k}
         \sum\limits_{i=1}^3 \big(a_i\max\{\langle w,x_k\rangle+b_i,0\}\big) =
        \left[\frac{2\delta}{\varepsilon}\right]\left[\frac{\varepsilon}{2}\right]=\delta\dott
    \end{equation}} 
   \argument{the fact that $\varepsilon<\min_{m\in \{1,2,\dots,M\}\backslash\{k\}}\abs{\spro{\W,x_m}+\B})$}
{that for all $m\in\{1,2,\dots,M\}\backslash\{k\}$ with $\spro{\W,x_m}+\B\geq 0$ it holds that \llabel{case 1: arg1} $\spro{\W,x_m}+\B=|\spro{\W,x_m}+\B|>\varepsilon$\dott}
\argument{\lref{case 1: arg1};}{that for all $m\in\{1,2,\dots,M\}\backslash\{k\}$ with $\spro{\W,x_m}+\B\geq 0$ it holds that 
\begin{equation}\llabel{case 1: arg2}
\spro{w,x_m}+b_1>0, \qquad \spro{w,x_m}+b_2>0, \qqandqq \spro{w,x_m}+b_3>0\dott
\end{equation}}
\argument{\lref{case 1: arg2};\lref{def wab};\lref{def wab2}}{that for all $m\in\{1,2,\dots,M\}\backslash\{k\}$ with $\spro{\W,x_m}+\B\geq 0$ it holds that
    \begin{equation}\llabel{case 1: eq1}
        \begin{split}
            & \sum\limits_{i=1}^3 \big(a_i\max\{\langle w,x_m\rangle+b_i,0\}\big)= \\
        &= \frac{2\delta}{\varepsilon}\Big[\spro{\W,x_m}+\B+\frac{\varepsilon}{2}\Big]+\frac{10\delta}{\varepsilon}\Big[\spro{\W,x_m}+\B-\varepsilon\Big]-\frac{12\delta}{\varepsilon}\Big[\spro{\W,x_m}+\B-\frac{3\varepsilon}{4}\Big]=0\dott
        \end{split}
    \end{equation}}
    \argument{the fact that $\varepsilon<\min_{m\in \{1,2,\dots,M\}\backslash\{k\}}\abs{\spro{\W,x_m}+\B})$} {that for all $m\in\{1,2,\dots,M\}\backslash\{k\}$ with $\spro{\W,x_m}+\B<0$ it holds that \llabel{case 2: arg1}$\spro{\W,x_m}+\B<-\varepsilon$\dott} 
    \argument{\lref{case 2: arg1};} {that for all $m\in\{1,2,\dots,M\}\backslash\{k\}$ with $\spro{\W,x_m}+\B<0$ it holds that \begin{equation}\llabel{case 2: arg2}
    \spro{w,x_m}+b_1<0, \qquad \spro{w,x_m}+b_2<0, \qqandqq \spro{w,x_m}+b_3<0\dott
    \end{equation}}
    \argument{\lref{case 2: arg2};}{that for all $m\in\{1,2,\dots,M\}\backslash\{k\}$ with $\spro{\W,x_m}+\B<0$ it holds that
    \begin{equation}\llabel{case2: eq1}
            \sum\limits_{i=1}^3 \big(a_i\max\{\langle w,x_m\rangle+b_i,0\}\big) =0\dott
    \end{equation}} 
    \argument{\lref{case2: eq1}; \lref{case 1: eq1}}{that for all $m\in \{1,2,\dots,M\}\backslash\{k\}$ it holds that 
    \begin{equation}\llabel{at different point}
        \sum
        \limits_{i=1}^3 \big(a_i\max\{\langle w,x_m\rangle+b_i,0\}\big)=\delta\mathbbm{1}_{\{k\}}(m) =0\dott
   \end{equation}}
    \argument{\lref{at k};\lref{at different point}}{
    \cref{prop: seperation: 1 point property}\dott}
    \end{aproof}
    \subsection{Interpolating ANNs and indicator functions through ANNs}
      \begin{athm}{prop}{prop: represent1}
          Assume \cref{setting: improve risk}, assume $\bfl_1>3$, assume $\{1,2,3\}\subseteq(\cap_{n=1}^\infty\inact_{ \nnode }^{ \varTheta^n})$, let $\mathfrak{p},p_1,p_2\in \{1,2,\dots,M\}$, and let $\ZZ\in \R$ satisfy $\bfY^{p_1}< \ZZ<\bfY^{p_2}$. Then there exists $N\in\N$ such that for all $n\in \N\cap[N,\infty)$ there exists $\vartheta\in \R^{\fd_\nnode}$ such that 
          \begin{enumerate}[label=(\roman{*})]
              \item it holds for all $m\in\{1,2,\dots,M\}\backslash\{\mathfrak{p}\} $  that $\mN^{\fL,\vartheta}_{\nnode}(\X^m)=\mN^{\fL,\varTheta^n}(\X^m)$ and
              \item it holds that $\mN^{\fL,\vartheta}_{\nnode}(\X^{\mathfrak{p}})=\ZZ$.
          \end{enumerate}
          \end{athm}
          \begin{aproof}
          Throughout this proof for every $\theta=(\theta_1,\dots,\theta_{\fd_\bfl})\in \R^{\fd_\bfl}$ let 
$\scrM^{ k, \theta }=(\scrM^{ k, \theta }_{1},\dots,\scrM^{ k, \theta }_{\nnode_k}) \colon \R \to \R^{ \nnode_k} $, $ k \in \{1,2,\dots,\fL\}$, satisfy for all $k\in \{1,2,\dots,\fL-1\}$, $i\in\{1,2,\dots,\allowbreak\nnode_{k+1}\}$, $t\in \R$ that
$
\scrM^{1,\theta}(t)=\mN^{1,\theta}_{\nnode}(\X^{p_1})$
and
\begin{equation}\llabel{eq: relization multi: prop: improve risk_case 2a:eq2}
\begin{split}
 & \scrM^{ k+1, \theta }_{i}(t) 
  = \mN^{2,\theta}_{\nnode,i}(\X^{p_{2}})t\mathbbm 1_{\{1\}}(k)\\
&+\bigg[\theta_{i+\nnode_{k+1}\nnode_{k}+\sum_{h=1}^{k}\nnode_h(\nnode_{h-1}+1)}+\sum\limits_{j=1}^{\nnode_{k}}\theta_{(i-1)\nnode_{k}+j+\textstyle\sum_{h=1}^{k}\nnode_h(\nnode_{h-1}+1)}  
  \max\{\scrM^{k,\theta}_{j}(t),0\}\bigg][1-t\mathbbm 1_{\{1\}}(k)]\dott
  \end{split}
\end{equation}
\startnewargseq
\argument{\lref{eq: relization multi: prop: improve risk_case 2a:eq2};}{that for all $\theta=(\theta_1,\dots,\theta_{\fd_\bfl})\in \R^{\fd_\bfl}$, $i\in \{1,2,\dots,\bfl_{2}\}$, $t\in \R$ it holds that
\begin{equation}\llabel{NRX}
    \begin{split}
       & \mathcal M^{2,\theta}_i(t)=\mathcal N^{2,\theta}_{\bfl,i}(\X^{p_2})t\\
&+\bigg[\theta_{i+\nnode_{2}\nnode_{1}+\nnode_1(\nnode_{0}+1)}+\textstyle\sum\limits_{j=1}^{\nnode_{1}}\theta_{(i-1)\nnode_{1}+j+\nnode_1(\nnode_{0}+1)} 
  \max\{\mN^{1,\theta}_{\bfl,j}(X^{p_1}),0\}\bigg][1-t]\dott
    \end{split}
\end{equation}}
\argument{\lref{eq: relization multi: prop: improve risk_case 2a:eq2};}{that for all $\theta=(\theta_1,\dots,\theta_{\fd_\bfl})\in \R^{\fd_\bfl}$, $k\in \N\cap[2,L)$, $i\in \{1,2,\dots,\bfl_{k+1}\}$, $t\in \R$ it holds that
\begin{equation}\llabel{NRV}
\begin{split}
  & \scrM^{ k+1, \theta }_{i}(t) 
=\bigg[\theta_{i+\nnode_{k+1}\nnode_{k}+\sum_{h=1}^{k}\nnode_h(\nnode_{h-1}+1)}+\sum\limits_{j=1}^{\nnode_{k}}\theta_{(i-1)\nnode_{k}+j+\textstyle\sum_{h=1}^{k}\nnode_h(\nnode_{h-1}+1)}  
  \max\{\scrM^{k,\theta}_{j}(t),0\}\bigg]\dott
  \end{split}
\end{equation}}
  \argument{the fact that $\bfY^{p_1}<\bfY^{p_2}$;}{\llabel{ArG1}that $\bfY^{p_1}\neq \bfY^{p_2}$\dott}
  \argument{\lref{ArG1};}{that \llabel{ArG2}$p_1\neq p_2$\dott}
  \argument{\lref{ArG2};}{that $M>1$\dott}
  \argument{ \eqref{eq:setting_definition_of_I_set};}{that for all $L\in\N$, $\ell\in \N^{L+1}$, $\theta\in \R^{\fd_\ell}$ it holds that
  \begin{equation}\llabel{eqttt2}
      \inact_\ell^{\theta}=\{i\in \{1,2,\dots,\ell_1\}\colon (\forall \, x\in [a,b]^d\colon \mathcal N^{1,\theta}_{\ell,i}(x)<0)\}\dott
  \end{equation}}
  \argument{\lref{eqttt2};}{that for all $\theta\in \R^{\fd_\bfl}$, $i\in \inact^{\theta}_\bfl$, $x\in [a,b]^d$ it holds that
  \begin{equation}\llabel{eqttt3}
      \max\{\mathcal N^{1,\theta}_{\bfl,i}(x),0\}=0\dott
  \end{equation}}
\argument{\lref{eqttt3};the fact that for all $n\in \N$ it holds that $\{1,2,3\}\subseteq \inact_{\nnode}^{\varTheta^n}$}{that for all $n\in \N$, $j\in\{1,2,3\}$, $x\in [a,b]^d$ it holds that
\begin{equation}\llabel{arg1}
\max\{\mN_{\nnode,j}^{1,\varTheta^n}(x),0\}=0\dott
\end{equation}}
\argument{\cref{relization multi};}{for all $n\in\N$, $i\in \{1,2,\dots,\bfl_2\}$, $x\in[a,b]^d$ that
\begin{equation}\llabel{ARG11}
     \mathcal N_{\bfl,i}^{2,\varTheta^n}(x)= \varTheta^n_{i+\nnode_2\nnode_1+\nnode_1(\nnode_0+1)}+\sum\limits_{j=1}^{\nnode_1}\varTheta^n_{(i-1)\nnode_1+j+\nnode_1(\nnode_0+1)}\max\{\mN^{1,\varTheta^n}_{\nnode,j}(x),0\}\dott
\end{equation}}
\argument{\lref{ARG11};\lref{arg1}}{for all $n\in \N$, $i\in \{1,2,\dots,\nnode_2\}$, $x\in [a,b]^d$ that
\begin{equation}\llabel{ARG1}
    \mathcal N_{\bfl,i}^{2,\varTheta^n}(x)= \varTheta^n_{i+\nnode_2\nnode_1+\nnode_1(\nnode_0+1)}+\sum\limits_{j=4}^{\nnode_1}\varTheta^n_{(i-1)\nnode_1+j+\nnode_1(\nnode_0+1)}\max\{\mN^{1,\varTheta^n}_{\nnode,j}(x),0\}\dott
\end{equation}}
\argument{\lref{NRX};\lref{ARG11}; the fact that for all $k\in\{1,2\}$ it holds that $\X^{p_k}\in[a,b]^d$}{that for all $n\in \N$, $i\in \{1,2,\dots,\bfl_2\}$, $t\in \R$ it holds that
\begin{equation}\llabel{eqttt4}
    \mathcal M^{2,\varTheta^n}_i(t)=t\mathcal N^{2,\varTheta^n}_{\bfl,i}(X^{p_2})+(1-t)\mathcal N^{2,\varTheta^n}_{\bfl,i}(X^{p_1})\dott
\end{equation}}
\argument{\lref{eqttt4};}{for all $n\in \N$, $t\in \R$ that
\begin{equation}\llabel{arg2}
    \mathcal M^{2,\varTheta^n}(t)=t\mathcal N^{2,\varTheta^n}_\bfl(X^{p_2})+(1-t)\mathcal N^{2,\varTheta^n}_\bfl(X^{p_1})\dott
\end{equation}}
In the following we prove that for all $k\in \{2,3,\dots,\fL\}$, $n\in \N$, $t\in \{0,1\}$ it holds that 
\begin{equation}\llabel{eq2}
\scrM^{k,\varTheta^n}(t)=\mN_{\nnode}^{k,\varTheta^n}(\X^{p_{1+t}})\dott
\end{equation}
We prove \lref{eq2} by induction on $k\in \{2,3,\dots,\fL\}$. For the base case $k=2$ observe that \lref{arg2} proves for all $n\in \N$, $t\in \{0,1\}$ that
\begin{equation}\llabel{arg2.5} 
\scrM^{2,\varTheta^n}(t)=\mN_{\nnode}^{2,\varTheta^n}(\X^{p_{1+t}})\dott
\end{equation}
This establishes \lref{eq2} in the base case $k=2$.
For the induction step we assume that there exists $k\in \N\cap[2,L)$ which satisfies for all $n\in \N$, $t\in \{0,1\}$ that
\begin{equation}\llabel{eq3}
\scrM^{k,\varTheta^n}(t)=\mN_{\nnode}^{k,\varTheta^n}(\X^{p_{1+t}})\dott
\end{equation}
\startnewargseq
\argument{\lref{eq3};}{for all $n\in \N$, $j\in \{1,2,\dots,\bfl_k\}$, $t\in\{0,1\}$ that
\begin{equation}\llabel{eq4}
    \scrM_{j}^{k,\varTheta^n}(t)=\mN_{\nnode,j}^{k,\varTheta^n}(\X^{p_{1+t}})\dott
\end{equation}}
\argument{\lref{eq4}; \cref{relization multi};\lref{NRV}}{ for all $n\in \N$, $i\in \{1,2,\dots,\bfl_{k+1}\}$, $t\in \{0,1\}$ that \llabel{arg2.875}
\begin{equation}
\begin{split}
\scrM_{i}^{k+1,\varTheta^n}(t)&=\varTheta^n_{i+\nnode_{k+1}\nnode_{k}+\sum_{h=1}^{k}\nnode_h(\nnode_{h-1}+1)}+\sum\limits_{j=1}^{\nnode_{k}}\varTheta^n_{(i-1)\nnode_{k}+j+\sum_{h=1}^{k}\nnode_h(\nnode_{h-1}+1)}  
  \max\{\scrM^{k,\varTheta^n}_{j}(t),0\}\\
&=\varTheta^n_{i+\bfl_{k+1}\bfl_{k}+\sum_{h=1}^{k}\bfl_h(\bfl_{h-1}+1)}+\sum\limits_{j=1}^{\bfl_{k}}\varTheta^n_{(i-1)\bfl_{k}+j+\sum_{h=1}^{k}\bfl_h(\bfl_{h-1}+1)}\max\{\mN^{k,\varTheta^n}_{\bfl,j}(X^{p_{1+t}}),0\}\\
& = \mN_{\nnode,i}^{k+1,\varTheta^n}(\X^{p_{1+t}})\dott
    \end{split}
\end{equation}}
\argument{\lref{arg2.875};}{for all $n\in \N$, $t\in \{0,1\}$ that
\begin{equation}\llabel{eq5}
\scrM^{k+1,\varTheta^n}(t)=\mN_{\nnode}^{k+1,\varTheta^n}(\X^{p_{1+t}})\dott
\end{equation}}
\argument{\lref{eq5};\lref{arg2.5};induction}[verbs=ep]{\lref{eq2}\dott}
In the following we prove that for all $k\in \{1,2,\dots,\fL\}$, $n\in\N$ it holds that 
\begin{equation}\llabel{eq6}
   \scrM^{k,\varTheta^n}\in C(\R,\R^{\bfl_k})\dott
\end{equation}  
We prove \lref{eq6} by induction on $k\in \{1,2,\ldots,\fL\}$. For the base case $k=1$ observe that \lref{eq: relization multi: prop: improve risk_case 2a:eq2} implies for all $n\in\N$ that 
\begin{equation}\llabel{eqt1}
\scrM^{1,\varTheta^n}\in C(\R,\R^{\bfl_k})\dott
\end{equation}
\startnewargseq
For the induction step we assume that there exists $k\in \{1,2,\dots,\fL-1\}$ which satisfies for all $n\in \N$ that 
\begin{equation}\llabel{eq7}
    \scrM^{k,\varTheta^n}\in C(\R,\R^{\bfl_k})\dott
\end{equation}  
\startnewargseq
\argument{\lref{eq7};}{ that for all $n\in\N$, $i\in \{1,2,\dots,\bfl_k\}$ it holds that \llabel{argg2}$\scrM_{i}^{k,\varTheta^n}$ is  continuous\dott}
\argument{\lref{argg2}; \lref{NRX};\lref{NRV}}{for all $i\in \{1,2,\dots,\bfl_{k+1}\}$, $n\in \N$ that \llabel{argg3} $\scrM_{i}^{k+1,\varTheta^n}$ is continuous\dott}
\argument{\lref{argg3}}{for all $n\in \N$ that \llabel{argg4}$\scrM^{k+1,\varTheta^n}$ is  continuous\dott}
\argument{\lref{argg4};\lref{eqt1};induction}[verbs=ep]{\lref{eq6}\dott}
\startnewargseq
\argument{\lref{eq2}; the fact that $\limsup_{n\to \infty}\allowbreak\sum_{m=1}^M|\mN^{\fL,\varTheta^n}_{\nnode}(\X^m)
      -\bfY^m|=0$}{that for all $t\in \{0,1\}$ it holds that 
\begin{equation}\llabel{argg5}\lim_{n\to\infty}\scrM^{\fL,\varTheta^n}(t)=\lim_{n\to \infty}\mN_{\nnode}^{\fL,\varTheta^n}(\X^{p_{1+t}})=\bfY^{p_{1+t}}\dott
\end{equation}}
\argument{\lref{argg5};the assumption that $\bfY^{p_1}< \ZZ<\bfY^{p_2}$}{there exists $N\in \N$ which satisfies for all $n\in \N\cap[N,\infty)$ that
\begin{equation}\llabel{arggg5}
    \scrM^{\fL,\varTheta^n}(0)<\ZZ<\scrM^{\fL,\varTheta^n}(1)\dott
\end{equation}}
\startnewargseq
\argument{\lref{eq6};\lref{arggg5}; the intermediate value theorem}{ that there exist $T=(T_n)_{n\in\N}\colon \N\to[0,1]$ which satisfies for all $n\in \N\cap[N,\infty)$ that
\begin{equation}\llabel{def: T_n}
\scrM^{\fL,\varTheta^n}(T_n)=\ZZ\dott
\end{equation}}
\startnewargseq
\argument{
 \cref{prop: seperation} (applied with $M\curvearrowleft M$, $d\curvearrowleft d$, $\delta\curvearrowleft 1$, $(x_m)_{m\in\{1,2,\dots,M\}}\curvearrowleft(X^m)_{m\in\{1,2,\dots,M\}}$, $k\curvearrowleft\mathfrak{p}$ in the notation of \cref{prop: seperation})} {that there exist 
 $w=(w_1,\dots,w_d)\in \R^d$, $a_1,a_2,a_3,b_1,b_2,b_3\in \R$ which satisfy for all $m\in\{1,2,\dots,M\}$ that
 \begin{equation}\llabel{def:v: case 2a: step 1}
    \begin{split}
\sum\limits_{i=1}^3\bigg[a_{i}\max\bigg\{\bigg[\textstyle\sum\limits_{k=1}^{\nnode_0}w_{k}\X^m_k\bigg]+b_{i},0\bigg\}\bigg]=\mathbbm 1_{\{\mathfrak{p}\}}(m).
        \end{split}
    \end{equation}}
In the following for every $n\in \N$ let $\vartheta^n=(\vartheta^n_1,\dots,\vartheta^n_{\fd_\nnode})\in \R^{\fd_\nnode}$ satisfy for all $s \in \{1,2,\dots,\fd_\nnode\}\backslash\big(\cup_{i=1}^{\bfl_2}\cup_{j=1}^3\cup_{k=1}^{\bfl_0} \{(i-1)\nnode_1+j+\nnode_1(\nnode_0+1),\nnode_1\nnode_0+j,(j-1)\nnode_0+k\}\big)$, $i\in \{1,2,\dots,\nnode_2\}$, $j\in \{1,2,3\}$, $k\in\{1,2,\dots,\nnode_0\}$ that
\begin{equation}\llabel{def:v: case 2a: step 1.2}
\vartheta^n_s=\varTheta^n_s,\qquad\vartheta^n_{(j-1)\nnode_0+k}=w_{k},\qquad \vartheta^n_{\nnode_1\nnode_0+j}=b_{j},
\end{equation}

\begin{equation}
  \llabel{def:v: case 2a: step 2}
    \text{and}\qquad\vartheta^n_{(i-1)\nnode_1+j+\nnode_1(\nnode_0+1)}
=a_{j}\bigl[\scrM^{2,\varTheta^n}_i(T_n)
-\mathcal N^{2,\varTheta^n}_{\bfl,i}(X^{\mathfrak{p}})\bigr]\dott
\end{equation}
\startnewargseq
\argument{\lref{def:v: case 2a: step 1.2};}{that for all $n\in \N$, $i\in \{1,2,\dots,\nnode_2\}$, $m\in\{ 1,2,\dots,M\}$ it holds that
\begin{equation}\llabel{tggg1}
\begin{split}
     &\textstyle\sum\limits_{j=1}^3\vartheta^n_{(i-1)\nnode_1+j+\nnode_1(\nnode_0+1)}\max\bigg\{\bigg[\textstyle\sum\limits_{k=1}^{\nnode_0}\vartheta^n_{(j-1)\nnode_0+k}\X^m_k\bigg]+\vartheta^n_{\nnode_1\nnode_0+j},0\bigg\}\\
&=\textstyle\sum\limits_{j=1}^3\vartheta^n_{(i-1)\nnode_1+j+\nnode_1(\nnode_0+1)}\max\bigg\{\bigg[\textstyle\sum\limits_{k=1}^{\nnode_0}w_{k}\X^m_k\bigg]+b_{j},0\bigg\}\dott
 \end{split}
\end{equation}}
\argument{\lref{tggg1};\lref{def:v: case 2a: step 1};\lref{def:v: case 2a: step 2}}{ that for all $n\in \N$, $i\in \{1,2,\dots,\nnode_2\}$, $m\in\{ 1,2,\dots,M\}$ it holds that
    \begin{equation}\llabel{def:v: case 2a}
    \begin{split}
&\sum\limits_{j=1}^3\vartheta^n_{(i-1)\nnode_1+j+\nnode_1(\nnode_0+1)}\max\bigg\{\bigg[\textstyle\sum\limits_{k=1}^{\nnode_0}\vartheta^n_{(j-1)\nnode_0+k}\X^m_k\bigg]+\vartheta^n_{\nnode_1\nnode_0+j},0\bigg\}\\
   & =\bigl[\scrM^{2,\varTheta^n}_i(T_n) 
-\mathcal N^{2,\varTheta^n}_{\bfl,i}(X^{\mathfrak{p}})\bigr]\bigg[\textstyle\sum\limits_{j=1}^3 a_{j}\max\bigg\{\bigg[\textstyle\sum\limits_{k=1}^{\nnode_0}w_{k}\X^m_k\bigg]+b_{j},0\bigg\}\bigg]\\
& =\big[\scrM^{2,\varTheta^n}_i(T_n) 
-\mathcal N^{2,\varTheta^n}_{\bfl,i}(X^{\mathfrak{p}})\big]\mathbbm 1_{\{\mathfrak{p}\}}(m)\dott
\end{split}
    \end{equation}}
\argument{\lref{def:v: case 2a};\cref{relization multi}}{\llabel{arg4} that for all  
$n\in \N$, $i\in\{1,2,\dots,\nnode_2\}$ it holds that 
\begin{equation}\llabel{EQ1'}
\begin{split}
\mN^{2,\vartheta^n}_{\nnode,i}(\X^{\mathfrak{p}})&=\vartheta^n_{i+\nnode_2\nnode_1+\nnode_1(\nnode_0+1)}+\sum\limits_{j=1}^{\nnode_1}\vartheta^n_{(i-1)\nnode_1+j+\nnode_1(\nnode_0+1)}\max\{\mN^{1,\vartheta^n}_{\nnode,j}(\X^{\mathfrak{p}}),0\}\\
&=\vartheta^n_{i+\nnode_2\nnode_1+\nnode_1(\nnode_0+1)}+\bigg[\textstyle\sum\limits_{j=4}^{\nnode_1}\vartheta^n_{(i-1)\nnode_1+j+\nnode_1(\nnode_0+1)}\max\{\mN^{1,\vartheta^n}_{\nnode,j}(\X^{\mathfrak{p}}),0\}\bigg]\\
&+ \bigg[\textstyle\sum\limits_{j=1}^{3}
\vartheta^n_{(i-1)\nnode_1+j+\nnode_1(\nnode_0+1)}\max\bigg\{\textstyle\bigg[\sum\limits_{k=1}^{\nnode_0}\vartheta^n_{(j-1)\nnode_0+k}\X^{\mathfrak{p}}_k\bigg]+\vartheta^n_{\nnode_1\nnode_0+j},0\bigg\}\bigg]\\
&=\varTheta^n_{i+\nnode_2\nnode_1+\nnode_1(\nnode_0+1)}+\biggl[\textstyle\sum\limits_{j=4}^{\nnode_1}\varTheta^n_{(i-1)\nnode_1+j+\nnode_1(\nnode_0+1)}\max\{\mN^{1,\varTheta^n}_{\nnode,j}(\X^{\mathfrak{p}}),0\}\biggr]\\
&
+\scrM^{2,\varTheta^n}_i(T_n) -\mathcal N^{2,\varTheta^n}_{\bfl,i}(X^{\mathfrak{p}})\dott
\end{split}
\end{equation}}
\argument{\lref{EQ1'};
\lref{ARG1}; the fact that $X^{\mathfrak{p}}\in[a,b]^d$}{that for all  
$n\in \N$, $i\in\{1,2,\dots,\nnode_2\}$ it holds that 
\begin{equation}\llabel{EQ1}
\begin{split}
   \mN^{2,\vartheta^n}_{\nnode,i}(\X^{\mathfrak{p}})&=\varTheta^n_{i+\nnode_2\nnode_1+\nnode_1(\nnode_0+1)}+\biggl[\textstyle\sum\limits_{j=4}^{\nnode_1}\varTheta^n_{(i-1)\nnode_1+j+\nnode_1(\nnode_0+1)}\max\{\mN^{1,\varTheta^n}_{\nnode,j}(\X^{\mathfrak{p}}),0\}\biggr]\\
&
+\scrM^{2,\varTheta^n}_i(T_n) -\mathcal N^{2,\varTheta^n}_{\bfl,i}(X^{\mathfrak{p}})\\
&=\scrM^{2,\varTheta^n}_{i}(T_n)\dott
   \end{split}
\end{equation}
}
\argument{\cref{relization multi};\lref{def:v: case 2a}}{that for all $n\in \N$, $i\in\{1,2,\dots,\nnode_2\}$, $m\in \{1,2,\dots,M\}\backslash\{\mathfrak{p}\}$ it holds that
\begin{equation}\llabel{EQ2'}
\begin{split}
   & \mN^{2,\vartheta^n}_{\nnode,i}(\X^m) \\  &=\vartheta^n_{i+\nnode_2\nnode_1+\nnode_1(\nnode_0+1)}+\sum\limits_{j=1}^{\nnode_1}\vartheta^n_{(i-1)\nnode_1+j+\nnode_1(\nnode_0+1)}\max\{\mN^{1,\vartheta^n}_{\nnode,j}(\X^{m}),0\}\\
&=\vartheta^n_{i+\nnode_2\nnode_1+\nnode_1(\nnode_0+1)}+\biggl[\textstyle\sum\limits_{j=4}^{\nnode_1}\vartheta^n_{(i-1)\nnode_1+j+\nnode_1(\nnode_0+1)}\max\{\mN^{1,\vartheta^n}_{\nnode,j}(\X^m),0\}\biggr]\\
&+ \bigg[\textstyle\sum\limits_{j=1}^{3}
\vartheta^n_{(i-1)\nnode_1+j+\nnode_1(\nnode_0+1)}\max\bigg\{\bigg[\textstyle\sum\limits_{k=1}^{\nnode_0}\vartheta^n_{(j-1)\nnode_0+k}\X^m_k\bigg]+\vartheta^n_{\nnode_1\nnode_0+j},0\bigg\}\bigg]\\
&=\varTheta^n_{i+\nnode_2\nnode_1+\nnode_1(\nnode_0+1)}+\sum\limits_{j=4}^{\nnode_1}\varTheta^n_{(i-1)\nnode_1+j+\nnode_1(\nnode_0+1)}\max\{\mN^{1,\varTheta^n}_{\nnode,j}(\X^{m}),0\}\dott
   \end{split}
\end{equation}}
\argument{\lref{EQ2'};\lref{ARG1}; the fact that for all $m\in \{1,2,\dots,M\}$ it holds that $\X^m\in [a,b]^d$}{that for all $n\in \N$, $i\in\{1,2,\dots,\nnode_2\}$, $m\in \{1,2,\dots,M\}\backslash\{\mathfrak{p}\}$ it holds that
\begin{equation}\llabel{EQ2}
    \begin{split}
         \mN^{2,\vartheta^n}_{\nnode,i}(\X^m)& =\varTheta^n_{i+\nnode_2\nnode_1+\nnode_1(\nnode_0+1)}+\sum\limits_{j=4}^{\nnode_1}\varTheta^n_{(i-1)\nnode_1+j+\nnode_1(\nnode_0+1)}\max\{\mN^{1,\varTheta^n}_{\nnode,j}(\X^{m}),0\}\\
         &=\mN^{2,\varTheta^n}_{\bfl,i}(\X^m)\dott
    \end{split}
\end{equation}
}
In the following we prove that for all $k\in \{2,3,\dots,\fL\}$, $m\in \{1,2,\dots,M\}\backslash\{\mathfrak{p}\}$, $n\in \N$ it holds that
\begin{equation}\llabel{induction: 3}
\mN^{k,\vartheta^n}_{\nnode}(\X^{\mathfrak{p}})=\scrM^{k,\varTheta^n}(T_n) \qqandqq \mN^{k,\vartheta^n}_{\nnode}(\X^m)=\mN^{k,\varTheta^n}_{\nnode}(\X^m)\dott
\end{equation}
We prove \lref{induction: 3} by induction on $k\in \{2,3,\dots,\fL\}$. 
\startnewargseq
For the base case $k=2$ observe that \lref{EQ1} and \lref{EQ2} ensure for all $m\in \{1,2,\dots,M\}\backslash\{\mathfrak{p}\}$, $n\in \N$, $i\in \{1,2,\dots,\bfl_2\}$ that 
\begin{equation}\llabel{induction: 3: eq1}
\mN^{2,\vartheta^n}_{\nnode,i}(\X^{\mathfrak{p}})=\scrM_{i}^{2,\varTheta^n}(T_n) \qqandqq \mN^{2,\vartheta^n}_{\nnode,i}(\X^m)=\mN^{2,\varTheta^n}_{\nnode,i}(\X^m)\dott
\end{equation}
This establishes \lref{induction: 3} in the base case $k=2$.
For the induction step we assume that there exists $k\in \N\cap[2,L)$ which satisfies for all $m\in \{1,2,\dots,M\}\backslash\{\mathfrak{p}\}$, $n\in \N$ that 
\begin{equation}\llabel{induction: 3: eq2}
\mN^{k,\vartheta^n}_{\nnode}(\X^{\mathfrak{p}})=\scrM^{k,\varTheta^n}(T_n) \qqandqq \mN^{k,\vartheta^n}_{\nnode}(\X^m)=\mN^{k,\varTheta^n}_{\nnode}(\X^m)\dott
\end{equation}
\startnewargseq
\argument{\lref{induction: 3: eq2};}{for all $m\in \{1,2,\dots,M\}\backslash\{\mathfrak{p}\}$, $n\in \N$, $j\in \{1,2,\dots,\bfl_k\}$ that 
\begin{equation}\llabel{induction: 3: eq3}
\mN^{k,\vartheta^n}_{\nnode,j}(\X^{\mathfrak{p}})=\scrM_{j}^{k,\varTheta^n}(T_n) \qqandqq \mN^{k,\vartheta^n}_{\nnode,j}(\X^m)=\mN^{k,\varTheta^n}_{\nnode,j}(\X^m)\dott
\end{equation}}
\argument{\lref{induction: 3: eq3};\cref{relization multi};\lref{NRV};\lref{def:v: case 2a: step 1.2}}{that for all $n\in \N$, $i\in \{1,2,\dots,\bfl_{k+1}\}$, $m\in \{1,2,\dots,M\}\backslash\{\mathfrak{p}\}$ it holds that 
\begin{equation}
    \begin{split}
        &\mN^{k+1,\vartheta^n}_{\nnode,i}(\X^{\mathfrak{p}})\\
&=\vartheta^n_{i+\bfl_{k+1}\bfl_{k}+\sum_{h=1}^{k}\bfl_h(\bfl_{h-1}+1)}+\sum\limits_{j=1}^{\bfl_{k}}\vartheta^n_{(i-1)\bfl_{k}+j+\sum_{h=1}^{k}\bfl_h(\bfl_{h-1}+1)}\max\{\mN^{k,\vartheta^n}_{\bfl,j}(X^{\mathfrak{p}}),0\}
\\&
=\varTheta^n_{i+\bfl_{k+1}\bfl_{k}+\sum_{h=1}^{k}\bfl_h(\bfl_{h-1}+1)}+\sum\limits_{j=1}^{\bfl_{k}}\varTheta^n_{(i-1)\bfl_{k}+j+\sum_{h=1}^{k}\bfl_h(\bfl_{h-1}+1)}\max\{\scrM_{j}^{k,\varTheta^n}(T_n),0\}
\\&
=\scrM_{i}^{k+1,\varTheta^n}(T_n)
    \end{split}
\end{equation}
and 
\begin{equation}\llabel{Eq3}
    \begin{split}
        &\mN^{k+1,\vartheta^n}_{\nnode,i}(\X^m)\\
&=\vartheta^n_{i+\bfl_{k+1}\bfl_{k}+\sum_{h=1}^{k}\bfl_h(\bfl_{h-1}+1)}+\sum\limits_{j=1}^{\bfl_{k}}\vartheta^n_{(i-1)\bfl_{k}+j+\sum_{h=1}^{k}\bfl_h(\bfl_{h-1}+1)}\max\{\mN^{k,\vartheta^n}_{\bfl,j}(X^m),0\}\\
&=\varTheta^n_{i+\bfl_{k+1}\bfl_{k}+\sum_{h=1}^{k}\bfl_h(\bfl_{h-1}+1)}+\sum\limits_{j=1}^{\bfl_{k}}\varTheta^n_{(i-1)\bfl_{k}+j+\sum_{h=1}^{k}\bfl_h(\bfl_{h-1}+1)}\max\{\mN^{k,\varTheta^n}_{\nnode,j}(\X^m),0\}\\
&=\mN^{k+1,\varTheta^n}_{\nnode,i}(\X^m)\dott
    \end{split}
\end{equation}
}
\argument{\lref{Eq3};\lref{induction: 3: eq1};induction}[verbs=ep]{\lref{induction: 3}\dott}
\startnewargseq
\argument{\lref{def: T_n};\lref{induction: 3}}{that for all $n\in \N\cap[N,\infty)$, $m\in \{1,2,\dots,M\}\backslash\{\mathfrak{p}\}$ it holds that 
\begin{equation}\llabel{arg5}\mN^{\fL,\vartheta^n}_{\nnode}(\X^{\mathfrak{p}})=\scrM^{\fL,\varTheta^n}(T_n)=\ZZ\qqandqq\mN^{\fL,\vartheta^n}_{\nnode}(\X^m)=\mN^{\fL,\varTheta^n}_{\nnode}(\X^m)\dott
\end{equation}}
    \end{aproof}
    \begin{athm}{cor}{prop: represent}
          Assume \cref{setting: improve risk}, assume $\bfl_1>3$, assume $\inf_{n\in \N}(\#(\inact_{\nnode }^{ \varTheta^n}))\geq 3$, let $\mathfrak{p},p_1,p_2\in \{1,2,\dots,M\}$, and let $\ZZ\in \R$ satisfy $\bfY^{p_1}< \ZZ<\bfY^{p_2}$. Then there exists $N\in\N$ such that for all $n\in \N\cap[N,\infty)$ there exists $\vartheta\in \R^{\fd_\nnode}$ such that 
          \begin{enumerate}[label=(\roman{*})]
              \item \label{represent item 1}it holds for all $m\in\{1,2,\dots,M\}\backslash\{\mathfrak{p}\} $  that $\mN^{\fL,\vartheta}_{\nnode}(\X^m)=\mN^{\fL,\varTheta^n}(\X^m)$ and
              \item \label{represent item 2}it holds that $\mN^{\fL,\vartheta}_{\nnode}(\X^{\mathfrak{p}})=\ZZ$.
          \end{enumerate}
          \end{athm}
          \begin{aproof}
              Throughout this proof for every $n\in \N$ let $a^n_1,a^n_2, a^n_3\in \N$ satisfy 
              \begin{equation}\llabel{def a}
              \#\{a^n_1,a^n_2,a^n_3\}=3 \qqandqq \{a^n_1,a^n_2,a^n_3\}\subseteq \inact_\bfl^{\varTheta^n},
              \end{equation}
              let $\varsigma^n\colon \{1,2,\dots,\bfl_1\}\to\{1,2,\dots,\bfl_1\}$ be a bijection which satisfies for all $i\in \{1,2,3\}$ that $\varsigma^n(a_i^n)=i$, and let $\permu^n=(\permu^n_1,\dots,\permu^n_{\fd_\bfl})\in \R^{\fd_\bfl}$ satisfy for all $i\in \{1,2,\dots,\bfl_2\}$, $j\in \{1,2,\dots,\bfl_1\}$, $k\in \{1,2,\dots,\bfl_0\}$ that
                  \begin{equation}\llabel{eq1}
                    \permu^n_{(j-1)\bfl_0+k}=\varTheta^n_{(\varsigma^n(j)-1)\bfl_0+k},\qquad\permu^n_{\bfl_1\bfl_0+j}=\varTheta^n_{\bfl_1\bfl_0+\varsigma^n(j)},
                  \end{equation}
                   \begin{equation}\llabel{eq2}
                     \text{and}\qquad \permu^n_{\bfl_1(\bfl_0+1)+(i-1)\bfl_1+j}=\varTheta^n_{\bfl_1(\bfl_0+1)+(i-1)\bfl_1+\varsigma^n(j)}
                  \end{equation}
                  and for all $s\in \{1,2,\dots,\fd_
                  \bfl\}\backslash (\cup_{i=1}^{\bfl_2}\cup_{j=1}^{\bfl_1}\cup_{k=1}^{\bfl_0}\{(j-1)\bfl_0+k,\bfl_1\bfl_0+j,\bfl_1(\bfl_0+1)+(i-1)\bfl_1+j\})$ that
                  \begin{equation}\llabel{eq3}
                      \permu^n_s=\varTheta^n_s.
                  \end{equation}
              \argument{\cref{relization multi};\lref{eq1};\lref{eq2}}{for all $n\in \N$, $x\in \R^{\bfl_0}$ that \llabel{arg1}$\mN^{2,\permu^n}_\bfl(x)=\mN^{2,\varTheta^n}_\bfl(x)$\dott}
              \argument{\cref{relization multi};\lref{arg1};\lref{eq3};induction}{for all $n\in \N$, $x\in \R^{\bfl_0}$ that
              \begin{equation}\llabel{eq4}
                  \mN^{\fL,\permu^n}_\bfl(x)=\mN^{\fL,\varTheta^n}_\bfl(x)\dott
              \end{equation}}
             \argument{\lref{eq1}; the fact that for all $n\in \N$ it holds that $a^n_1,a^n_2,a^n_3\in \inact_\bfl^{\varTheta^n}$, $\varsigma^n(a_1^n)=1$, $\varsigma^n(a_2^n)=2$, and $\varsigma^n(a_3^n)=3$}{that for all $n\in \N$ it holds that \llabel{arg2}$\{1,2,3\}\subseteq\inact_\bfl^{\permu^n}$\dott}
             \argument{\lref{arg2}; \lref{eq4};\cref{prop: represent1}}{that there exists $N\in \N$ such that for all $n\in \N\cap[N,\infty)$ there exists $\vartheta\in \R^{\fd_\bfl}$ such  that 
             \begin{enumerate}[label=(\Roman{*})]
                 \item \llabel{arg3} it holds for all $m\in\{1,2,\dots,M\}\backslash\{\mathfrak{p}\} $ that $\mN^{\fL,\vartheta}_{\nnode}(\X^m)=\mN_\nnode^{\fL,\permu^n}(\X^m)$ and
                 \item \llabel{arg4} it holds that $\mN^{\fL,\vartheta}_{\nnode}(\X^{\mathfrak{p}})=\ZZ$\dott
             \end{enumerate} }
             \argument{\lref{arg4};\lref{eq4}} {items \ref{represent item 1} and \ref{represent item 2}\dott}
          \end{aproof}
   \subsection{Geometric preliminaries
   }\label{subsec: geometric preliminaries}
   \begin{athm}{lemma}{lemma Y}
       Let $d,n\in \N$, $v_1,v_2,\dots,v_n\in \R^d$. Then
       \begin{enumerate}[label=(\roman*)]
       \item \label{lemma Y:item 1} it holds\footnote{Note that for every $d\in \N$ and every $A\subseteq\R^d$ it holds that $\operatorname{conv}(A)=\cup_{n\in \N}\cup_{a_1,a_2,\ldots,a_n\in A}\cup_{\lambda_1,\lambda_2,\ldots,\lambda_n\in[0,1],\, \sum_{i=1}^n\lambda_i=1}\{\sum_{i=1}^n\lambda_ia_i\}$.} that
       \begin{equation}\label{lemma Y:eq1}
           \operatorname{conv}(\{v_1,v_2,\dots,v_n\})=\cup_{\lambda_1,\lambda_2,\dots,\lambda_n\in [0,1],\, \sum_{i=1}^n\lambda_i=1}\{\textstyle\sum_{i=1}^n\lambda_iv_i\}
       \end{equation}
       and
       \item \label{lemma Y:ietm 2} it holds that $\operatorname{conv}(\{v_1,v_2,\dots,v_n\})$
        is compact.
        \end{enumerate}
   \end{athm}
   \begin{aproof}
       Throughout this proof let $\psi\colon \R^n\to \R^d$ satisfy for all $\lambda=(\lambda_1,\dots,\lambda_n)\in \R^n$ that
       \begin{equation}\llabel{eq1}
\psi(\lambda)=\lambda_1v_1+\lambda_2v_2+\ldots+\lambda_nv_n 
       \end{equation}
       and let $K\subseteq\R^n$ satisfy
       \begin{equation}\llabel{eq2}
           K=\{\lambda=(\lambda_1,\dots, \lambda_n)\in [0,1]^n\colon \lambda_1+\lambda_2+\ldots+\lambda_n=1\}\dott
       \end{equation}
       \argument{\lref{eq1};\lref{eq2}; the fact that 
       \begin{equation}\llabel{eq1.1}
        \operatorname{conv}(\{v_1,v_2,\dots,v_n\})=\cup_{N\in \N}\cup_{a_1,a_2,\dots,a_N\in \{1,2,\dots,n\}}\cup_{\lambda_1,\lambda_2,\dots,\lambda_N\in[0,1],\,\sum_{i=1}^N\lambda_i=1}\{\lambda_iv_{a_i}\}
       \end{equation}}{that
       \begin{equation}\llabel{NR}
           \operatorname{conv}(\{v_1,v_2,\dots,v_n\}=\psi(K)\dott
       \end{equation}}
       \argument{\lref{NR}}[verbs=ep]{\cref{lemma Y:item 1}\dott}
       \argument{the fact that $\psi$ is continuous; the fact that $K$ is compact}{that \llabel{arg1}$\psi(K)$ is compact\dott}
       \argument{\lref{arg1};\lref{NR}}[verbs=ep]{\cref{lemma Y:ietm 2}\dott}
   \end{aproof}
   \begin{athm}{lemma}{lemma Z}   
   Let $d,n\in \N$, $v_1,v_2,\dots,v_n\in \R^d$ and let $V\in \R^d$ be an extreme point\footnote{Note that for every $d\in \N$, every convex $C\subseteq\R^d$, and every $v\in \R^d$ it holds that $v$ is an extreme point of $C$ if and only if $v\in C\backslash(\cup_{x,y\in C,\, x\neq y}\cup_{\lambda\in (0,1)}\{\lambda x+(1-\lambda)y\})$.} of 
   \begin{equation}\label{lemma Z: eq1}
       \operatorname{conv}(\{v_1,v_2,\dots,v_n\}).
   \end{equation}
   Then 
   \begin{equation}
\label{lemma Z: eq2}
       V\in \{v_1,v_2,\dots,v_n\}.
   \end{equation}
   \end{athm}
   \begin{aproof}
       We prove \cref{lemma Z: eq2} by contradiction. In the following we thus assume that
       \begin{equation}\llabel{eq1}
           V\notin \{v_1,v_2,\dots,v_n\}.
       \end{equation}
       \argument{\cref{lemma Y:item 1} in \cref{lemma Y};\cref{lemma Z: eq1}}{that there exist $\lambda_1,\lambda_2,\ldots,\lambda_n\in [0,1]$ which satisfy  
       \begin{equation}\llabel{eq2}
\lambda_1+\lambda_2+\ldots+\lambda_n=1\qqandqq
V=\lambda_1v_1+\lambda_2v_2+\ldots+\lambda_nv_n\dott
       \end{equation}}
       \startnewargseq
       \argument{\lref{eq1};\lref{eq2}}{that there exists $i\in \{1,2,\dots,n\}$ which satisfies 
       \begin{equation}\llabel{arg1}  0<\lambda_i<1\dott
       \end{equation}}
       \startnewargseq
       \argument{\lref{eq1};\lref{eq2}; \lref{arg1}}{that there exists $j\in \{1,2,\dots,n\}$ which satisfies 
       \begin{equation}\llabel{argt1}
       v_j\neq v_i\qqandqq 0<\lambda_j<1\dott
       \end{equation}}
       In the following let $\varepsilon \in (0,\min\{\lambda_i,\lambda_j,1-\lambda_i,1-\lambda_j\})$ 
       and let $W_1,W_2\in \R^d$ satisfy for all $k\in \{1,2\}$ that
       \begin{equation}\llabel{eq3}
           W_k=\sum_{q=1}^n \Big[\lambda_q-\varepsilon(-1)^k\mathbbm 1_{\{i\}}(q)+\varepsilon(-1)^k\mathbbm{1}_{\{j\}}(q)\Big]v_q.
       \end{equation}
    \startnewargseq
    \argument{\lref{eq2};the fact that $i\neq j$; the fact that $0<\varepsilon<\min\{\lambda_i,\lambda_j,1-\lambda_i,1-\lambda_j\}$; the fact that $\lambda_1,\lambda_2,\dots,\lambda_n\in [0,1]$}{that for all $q\in \{1,2,\dots,n\}$, $k\in\{1,2\}$ it holds that  $0\leq\lambda_q-\varepsilon(-1)^k\mathbbm 1_{\{i\}}(q)+\varepsilon(-1)^k\mathbbm{1}_{\{j\}}(q)<1$ and
\begin{equation}\llabel{arg2}\sum_{r=1}^n(\lambda_r-\varepsilon(-1)^k\mathbbm 1_{\{i\}}(r)+\varepsilon(-1)^k\mathbbm{1}_{\{j\}}(r))=1\dott
    \end{equation}}
    \argument{\lref{arg2}; \lref{eq3}}{for all $k\in\{1,2\}$ that 
    \begin{equation}\llabel{eq5}
        W_k \in \operatorname{conv}(\{v_1,v_2,\dots,v_n\})\dott
    \end{equation}}
\argument{\lref{eq2};\lref{argt1};\lref{eq3}; the fact that $\varepsilon>0$}{that \begin{equation}\llabel{arg3}
W_1- W_2=2\varepsilon v_i-2\varepsilon v_j=2\varepsilon(v_i-v_j)\neq 0\qqandqq\frac{W_1+W_2}{2}=V\dott
    \end{equation}}
    \argument{\lref{arg3};\lref{eq5}}{that $V$ is not an extreme point of $\operatorname{conv}(\{v_1,v_2,\dots,v_n\})$\dott}
    This contradiction proves \cref{lemma Z: eq2}.
   \end{aproof}
   In the following result, \cref{seperation theorem} below, we recall a special case of the well-known hyperplane separation theorem. \cref{seperation theorem} is, \eg, proved as \cite[Example 2.20]{convexoptimization}.
   \begin{athm}{prop}{seperation theorem}[Special case of the hyperplane separation theorem]
   Let $d\in \N$, let $C\subseteq \R^d$ be a closed convex set, and let $v\in \R^d\backslash C$. Then there exist  $w\in \R^d\backslash\{0\}$, $b\in \R$ such that
   \begin{enumerate}[label=(\roman{*})]
       \item it holds that $v\in \{x\in \R^d\colon \spro{w,x}+b>0\}$ and 
       \item it holds that $C\subseteq \{x\in \R^d\colon \spro{w,x}+b< 0\}$.
   \end{enumerate}
   \end{athm}
   \begin{aproof}
       Throughout this proof assume without loss of generality that $C\neq \emptyset$, let $x\in C$, and let $D\subseteq\R^d$ satisfy
       \begin{equation}\llabel{def: D}
           D=\{y\in C\colon \|y-x\|\leq \|v-x\|\}\dott
       \end{equation}
       \argument{\lref{def: D};}{that $D$ is bounded\dott}
       \argument{the fact that $C$ is closed;}{that \llabel{arg1} $D$ is closed\dott}
       \argument{\lref{arg1};the fact that $D$ is bounded}{that
       \llabel{arg2} $D$ is compact\dott}
       \argument{\lref{arg2};}{that there exists $z\in D$ which satisfies 
       \begin{equation}\llabel{eq1}
           \|z-v\|=\textstyle\inf_{y\in D} \|y-v\|\dott
       \end{equation}}
       \startnewargseq
       \argument{\lref{def: D};\lref{eq1}}{that
       \begin{equation}\llabel{eq2}
           \|z-v\|=\textstyle\inf_{y\in C} \|y-v\|\dott
       \end{equation}}
       \argument{the fact that $C$ is convex;the fact that $z\in C$}{that for all $y\in C$, $\lambda\in [0,1]$ it holds that \llabel{arg3} $\lambda y+(1-\lambda)z\in C$\dott}
       \argument{\lref{arg3};\lref{eq2}}{for all $y\in C$, $\lambda\in (0,1]$ that
       \begin{equation}\llabel{eq3}
           \langle z-v,z-v\rangle\leq \langle \lambda y+(1-\lambda)z-v, \lambda y+(1-\lambda)z-v\rangle\dott
       \end{equation}}
       \argument{\lref{eq3};} {for all $y\in C$, $\lambda\in [0,1]$ that
       \begin{equation}\llabel{eq4}
          \langle z-v,z-v\rangle\leq \langle \lambda (y-v)+(1-\lambda)(z-v), \lambda (y-v)+(1-\lambda)(z-v)\rangle\dott
       \end{equation}}
       \argument{\lref{eq4};}{for all $y\in C$, $\lambda\in [0,1]$ that
       \begin{equation}\llabel{eq5}
           2\lambda(1-\lambda) \langle z-v,z-v\rangle\leq \lambda^2 \langle y-v,y-v\rangle+ 2\lambda(1-\lambda)\langle y-v,z-v\rangle \dott
       \end{equation}}
        \argument{\lref{eq5};}{for all $y\in C$, $\lambda\in [0,1]$ that
       \begin{equation}\llabel{eq6}
           2\lambda(1-\lambda) \langle z-y,z-v\rangle\leq \lambda^2 \langle y-v,y-v\rangle \dott
       \end{equation}}
               \argument{\lref{eq6};}{for all $y\in C$, $\lambda\in (0,1]$ that
       \begin{equation}\llabel{eq7}
          2(1-\lambda) \langle z-v,z-y\rangle\leq \lambda \langle y-v,y-v\rangle \dott
       \end{equation}}
        \argument{\lref{eq7};}{for all $y\in C$ that
       \begin{equation}\llabel{eq8}
            \langle v-z,z-y\rangle\geq 0\dott
       \end{equation}}
       \argument{the fact that $v\notin C$;}{that \llabel{arg6} $v\neq z$\dott}
       \argument{\lref{eq8};\lref{arg6}}{for all $y\in C$ that
       \begin{equation}
       \begin{split}
          & \langle v-z,y\rangle +\frac{\|z\|^2}{2}-\frac{\|v\|^2}{2}\\
          &\leq \langle v-z,y\rangle +\langle v-z, z-y\rangle+\frac{\|z\|^2}{2}-\frac{\|v\|^2}{2}=\langle v-z,z\rangle +\frac{\|z\|^2}{2}-\frac{\|v\|^2}{2}\\
         &=\langle v,z\rangle -\langle z,z\rangle+\frac{\|z\|^2}{2}-\frac{\|v\|^2}{2} =\frac 12 \bigl(-\|v\|^2+2\langle v,z\rangle -\|z\|^2\bigr)=\frac {-1}{2} \|v-z\|^2<0\dott
           \end{split}
       \end{equation}}
       \argument{the fact that $v\neq z$}{that
       \begin{equation}\llabel{eq9}
          \langle v-z,v\rangle +\frac{\|z\|}{2}-\frac{\|v\|}{2}=\frac 12 \bigl(\|v\|^2-2\langle v,z\rangle +\|z\|^2\bigr)=\frac 12 \|v-z\|^2>0\dott 
       \end{equation}}
       \argument{\lref{eq9};\lref{eq8}; the fact that $v\neq z$}{that
       \begin{enumerate}[label=(\Roman*)]
         \item it holds that $v\in \bigl\{x\in \R^d\colon \spro{v-z,x}+\frac{\|z\|^2}{2}-\frac{\|v\|^2}{2}>0\bigr\}$, 
       \item it holds that $C\subseteq \bigl\{x\in \R^d\colon \spro{v-z,x}+\frac{\|z\|^2}{2}-\frac{\|v\|^2}{2}< 0\bigr\}$, and
       \item it holds that $v-z\neq 0$.
       \end{enumerate}}
   \end{aproof}
   \begin{athm}{lemma}{lemma W'}
       Let $d,n\in\N$, $v_1,v_2,\dots,v_n\in\R^d$ satisfy for all $m\in \{1,2,\dots,n\}$ that
       \begin{equation}\label{lemma W: assumption}
           \operatorname{conv}(\{v_{1},v_{2},\dots,v_{n}\}\backslash \{v_{m}\})\neq\operatorname{conv}(\{v_{1},v_{2},\dots,v_{n}\})
       \end{equation}
       Then it holds for all $m\in \{1,2,\dots,n\}$ that $v_m$ is an extreme point of 
       \begin{equation}\label{lemma W: conclude}
           \operatorname{conv}(\{v_1,v_2,\dots,v_n\}).
       \end{equation}
   \end{athm}
   \begin{aproof}
Throughout this proof assume without loss of generality that $\#\{v_1,v_2,\allowbreak\dots,\allowbreak v_n\}=n$ (otherwise let $N\in \N$ satisfy $ N=\#\{v_1,v_2,\dots,v_n\}$, let $k_1,k_2,\dots,k_N\in \{1,2,\dots,n\}$ satisfy $\#\{v_{k_1},v_{k_2},\dots,v_{k_N}\}=N$, and let $V_1,V_2,\dots,V_N\in \R^d$ satisfy for all $i\in \{1,2,\dots,N\}$ that $V_i=v_{k_i}$).
We prove \cref{lemma W: conclude} by contradiction. In the following we thus assume that there exists $m\in \{1,2,\dots,n\}$ which satisfies that $v_m$ is not an extreme point of
\begin{equation}\llabel{eq4}
    \operatorname{conv}(\{v_{1},v_{2},\dots,v_{n}\}).
\end{equation}
\startnewargseq
\argument{\lref{eq4};}{that there exist $a,b\in (0,1)$, $U,V\in \operatorname{conv}(\{v_{1},v_{1},\dots,v_{n}\})$ which satisfy
\begin{equation}\llabel{eq4.1}
   U\neq V,\qquad a+b=1,\qqandqq aU+bV=v_{m}\dott
\end{equation}}
\startnewargseq
\argument{\cref{lemma Y:item 1} in \cref{lemma Y};the fact that $U,V\in \operatorname{conv}(\{v_{1},v_{2},\dots,v_{n}\})$;}{that there exist $\lambda_1,\lambda_2,\dots,\lambda_n, \mu_1,\allowbreak\mu_2,\dots,\mu_n\in [0,1]$ which satisfy
\begin{equation}\llabel{eq4.2}
    \sum_{i=1}^n\lambda_i=\sum_{i=1}^n\mu_i=1,
   \qquad \sum_{i=1}^n\mu_iv_{i}=U,\qqandqq \sum_{i=1}^n\lambda_iv_{i}=V\dott
\end{equation}}
\startnewargseq
\argument{\lref{eq4.1};\lref{eq4.2}}{that
\begin{equation}\llabel{eq4.3}
    \sum_{i=1}^n (a\mu_i+b\lambda_i)v_{i}=v_{m}\dott
    \end{equation}}
    \argument{\lref{eq4.1};\lref{eq4.2}}{that \llabel{arg1}$(\mu_{m},\lambda_{m})\neq(1,1)$\dott}
    \argument{\lref{arg1};the fact that $0\leq \mu_{m}\leq 1$; the fact that $0\leq \lambda_{m}\leq 1$;\lref{eq4.1}}{that
    \begin{equation}\llabel{eq4.4}
        a\mu_{m}+b\lambda_{m}<a+b=1\dott
    \end{equation}}
\argument{\lref{eq4.4};\lref{eq4.3}}{that
\begin{equation}\llabel{eq4.5}
    \sum_{i\in\{1,2,\dots,n\}\backslash\{m\}}\Bigl[\frac{a\mu_i+b\lambda_i}{1-a\mu_{m}-b\lambda_{m}}\Bigr]v_{i}=v_{m}\dott
\end{equation}}
\argument{\lref{eq4.1};\lref{eq4.2}}{that
\begin{equation}\llabel{eq4.6}
    \sum_{i=1}^n(a\mu_{i}+b\lambda_{i})= a\biggl[\textstyle\sum\limits_{i=1}^n\mu_{i}\biggr]+b\biggl[\textstyle\sum\limits_{i=1}^n\lambda_{i}\biggr]=a+b=1\dott
\end{equation}}
\argument{\lref{eq4.6};}{that \llabel{eq4.6'} $\sum_{i\in\{1,2,\dots,n\}\backslash\{m\}}(a\mu_i+b\lambda_i)=1-a\mu_m-b\lambda_m$\dott}
\argument{\lref{eq4.6'};}{that \begin{equation}\llabel{eq4.6''}
    \sum_{i\in\{1,2,\dots,n\}\backslash\{m\}}\biggl[\frac{a\mu_i+b\lambda_i}{1-a\mu_m-b\lambda_m}\biggr]=1\dott
\end{equation}}
\argument{\lref{eq4.6''};\lref{eq4.5}; the fact that for all $i\in \{1,2,\dots,n\}$ it holds that $\frac{a\mu_i+b\lambda_i}{1-a\mu_{m}-b\lambda_{m}}
\geq 0$}{that
\begin{equation}\llabel{eq5}
    v_m\in \operatorname{conv}(\cup_{i\in \{1,2,\dots,n\}\backslash\{m\}}\{v_i\})\dott
\end{equation}}
\argument{the assumption that $\#\{v_1,v_2,\dots,v_n\}=n$;}{that
\begin{equation}\llabel{eq5.}
    \cup_{i\in \{1,2,\dots,n\}\backslash\{m\}}\{v_i\}=\{v_{1},v_{2},\dots,v_{n}\}\backslash \{v_{m}\}\dott
\end{equation}}
\argument{\lref{eq5.};}{
that 
\begin{equation}\llabel{eq5'}
    \operatorname{conv}(\cup_{i\in \{1,2,\dots,n\}\backslash\{m\}}\{v_i\})=\operatorname{conv}(\{v_{1},v_{2},\dots,v_{n}\}\backslash \{v_{m}\})\dott
\end{equation}}
\argument{\lref{eq5'};\lref{eq5}}{that
\begin{equation}\llabel{eq5''}
    v_m\in \operatorname{conv}(\{v_{1},v_{2},\dots,v_{n}\}\backslash \{v_{m}\})\dott
\end{equation}}
\argument{\cref{lemma W: assumption};\lref{eq5''}; the fact that for every $C\subseteq \R^d$ and every $x\in \operatorname{conv}(C)$ it holds that $\operatorname{conv} (C\cup\{x\})=\operatorname{conv}(C)$}{that
\begin{equation}\llabel{eq6}
\begin{split}
    & \operatorname{conv}(\{v_{1},v_{2},\dots,v_{n}\})\neq \operatorname{conv}(\{v_{1},v_{2},\dots,v_{n}\}\backslash \{v_{m}\})\\
    &=\operatorname{conv}([\{v_1,v_2,\dots,v_n\}\backslash\{v_m\}]\cup\{v_m\})
    =\operatorname{conv}(\{v_{1},v_{2},\dots,v_{n}\})\dott
     \end{split}
\end{equation}}
This contradiction proves \cref{lemma W: conclude}.
\end{aproof}
\begin{athm}{cor}{lemma W}
     Let $d,n\in\N$, $v_1,v_2,\dots,v_n\in\R^d$.
       Then there exists $m\in\{1,2,\dots,M\}$ such that $v_m$ is an extreme point of
       \begin{equation}\label{lemma W : conclude}
           \operatorname{conv}(\{v_1,v_2,\dots,v_n\}).
       \end{equation}
\end{athm}
\begin{aproof}
      Throughout this proof let $N\in \N$ satisfy
      \begin{multline}\llabel{eq1}
        N=\min\!\big\{m\in \N\colon \big(\exists\, i_1,i_2,\dots,i_m \in \{1,2,\dots,n\}\colon \\\operatorname{conv}(\{v_{i_1},v_{i_2},\dots,v_{i_m}\})=\operatorname{conv}(\{v_1,v_2,\dots,v_n\})\big) \big\}
      \end{multline}
      and let $k_1,k_2,\dots,k_N\in \{1,2,\dots,n\}$ satisfy
      \begin{equation}\llabel{eq2}
          \operatorname{conv}(\{v_{k_1},v_{k_2},\dots,v_{k_N}\})=\operatorname{conv}(\{v_1,v_2,\dots,v_n\}).
      \end{equation}
\argument{\lref{eq1};\lref{eq2}}{that for all $i\in \{1,2,\dots,N\}$ it holds that
\begin{equation}\llabel{eq3}
    \operatorname{conv}(\{v_{k_1},v_{k_2},\dots,v_{k_N}\}\backslash\{v_{k_i}\})\subseteq \operatorname{conv}(\cup_{j\in \{1,2,\dots,N\}\backslash\{i\}}\{v_{k_j}\})\neq \operatorname{conv}(\{v_{k_1},v_{k_2},\dots,v_{k_N}\})\dott
\end{equation}}
\argument{\lref{eq3};\cref{lemma W'}}{that $v_{k_1}$ is an extreme point of 
\begin{equation}\llabel{eqqqq1}
    \operatorname{conv}(\{v_{k_1},v_{k_2},\dots,v_{k_N}\})\dott
\end{equation}}
\argument{\lref{eqqqq1};\lref{eq2}}{\cref{lemma W : conclude}\dott}
\end{aproof}
\subsection{Improving empirical risks for constant limiting output data}
\begin{athm}{lemma}{prop: improve risk_case2b2}
    Assume \cref{setting: improve risk} and assume \begin{equation}\label{min=max}
    \min\limits_{m\in \{1,2,\dots,M\}}\bfY^m= \max\limits_{m\in \{1,2,\dots,M\}}\bfY^m.
    \end{equation}
    Then there exists $\vartheta\in \R^{\fd_\nnode}$ such that 
    \begin{equation}\label{conclude: min=max}
    \cL(\vartheta)<\liminf_{n\to \infty}\cL(\varTheta^n).
\end{equation}
\end{athm}
\begin{aproof}
Throughout this proof let $Q\in \R$ satisfy \begin{equation}\llabel{def:Q}
\min\limits_{m\in \{1,2,\dots,M\}}\bfY^m= \max\limits_{m\in \{1,2,\dots,M\}}\bfY^m=Q
\end{equation}
and let $\mathcal P \subseteq \N$ satisfy
 \begin{equation}\llabel{def:bP}
     \mathcal P  =\bigl\{m \in \{1,2,\dots,M\} \colon Y^m \ne Q \bigr\}.
 \end{equation}
 \argument{
  \lref{def:bP};\cref{estimate lim};} {\llabel{arg1}that $\mathcal P\ne \emptyset$\dott}
 \argument{\lref{arg1};\cref{lemma W}}{that there exists $p\in \mathcal P$ which satisfies that $X^p$ is a extreme point of 
 \begin{equation}\llabel{arg1''}
     \operatorname{conv}(\cup_{i\in \mathcal P}\{X^i\})\dott
 \end{equation}}
 \startnewargseq
 \argument{\lref{arg1''}; the fact that for every convex $C\subseteq\R^d$ and every extreme point $x\in \R^d$ of $C$ it holds that $C\backslash\{x\}$ is convex;the assumption that $\#\{X^1,X^2,\dots,X^M\}=M$}{ that 
 \begin{equation}\llabel{arg1.5}
     X^p\notin ([\operatorname{conv}(\cup_{i\in \mathcal P\backslash\{p\}}\{X^i\})]\backslash\{\X^p\})\subseteq \operatorname{conv}(\cup_{i\in \mathcal P\backslash\{p\}}\{X^i\})\dott
 \end{equation}}
 \argument{\lref{arg1.5}; \cref{lemma Y};\cref{seperation theorem}}{that there exist $w=(w_1,\dots, w_{\nnode_0})\in \R^{\bfl_0}$, $b\in \R$ which satisfy that
\begin{enumerate}[label=(\roman{*})]
\item \llabel{at not p} it holds for all $m\in \mathcal P\backslash\{p\}$ that
$\spro{w,X^m}+b<0$
    and
\item \llabel{at p} it holds that
$\spro{w,X^p}+b>0$\dott
    \end{enumerate}}
Next let $\mathcal M\subseteq\{1,2,\dots,M\}$ satisfy
\begin{equation}\llabel{def:K}
\mathcal M=\{m\in \{1,2,\dots,M\}\colon \spro{w,X^m}+b>0\},
\end{equation}
let $a\in \R$ satisfy
\begin{equation}\llabel{def v1}
a=-\frac{\bigl[\sum_{m \in \mathcal M}[\max\{\spro{w,X^m}+b,0\}][Q-Y^m]\bigr]}{\bigl[\sum_{m\in \mathcal M }[\max\cu{\spro{w,X^m}+b,0}]^2\bigr]},
 \end{equation}
and let $\vartheta=(\vartheta_1,\ldots,\vartheta_{\fd_\bfl})\in \R^{\fd_\bfl}$ satisfy for all $k\in \{1,2,\dots,\fL-1\}$, $s\in \{1,2,\dots,\fd_\bfl\}\backslash(\cup_{k=1}^{\fL}\cup_{q=1}^{\bfl_0}\{\fd_\bfl,1+\sum_{h=1}^{k}\nnode_h(\nnode_{h-1}+1),q,\nnode_1\nnode_0+1\})$ that
\begin{equation} \llabel{def:wab}
(\vartheta_1,\vartheta_2,\dots,\vartheta_{\nnode_0})=w, \quad \vartheta_{\nnode_1\nnode_0+1}=b,\quad\vartheta_{\fd_\bfl}=Q,\quad\vartheta_{1+\sum_{h=1}^{k}\nnode_h(\nnode_{h-1}+1)}=1+(a-1)\mathbbm 1_{\{1\}}(k),
     \end{equation}
     and $ \vartheta_s=0$\dott
     \startnewargseq
     \argument{\lref{def:wab};}{for all $m\in \{1,2,\dots,M\}$ that 
     \begin{equation}\llabel{arg'1}
     \begin{split}
         \mathcal N^{\fL,\vartheta}_\bfl(\X^m)&= \vartheta_{1+\nnode_1(\nnode_0+1)}\max\big\{\spro{(\vartheta_1,\vartheta_2,\dots,\vartheta_{\nnode_0}),X^m}+\vartheta_{\nnode_1\nnode_0+1},0\big\}+Q\\
         &=Q+a\max\{\spro{w,X^m}+b,0\}\dott
         \end{split}
     \end{equation}}
     \argument{\lref{arg'1};\lref{at not p}; \lref{at p}}
     {\llabel{arg2}that 
     \begin{enumerate}[label=(\Roman*)]
     \item it holds for all $m\in \mathcal M$ that 
     \begin{equation}\llabel{eqp1}
     \mN^{\fL,\vartheta}_\bfl(X^m)=Q+a\max\{\spro{w,X^m}+b,0\}
     \end{equation} and 
     \item it holds for all $m\in \{1,2,\dots,M\} \backslash\mathcal M $ that 
     \begin{equation}\llabel{eqp2}
     \mN^{\fL,\vartheta}_\bfl(X^m)=Q\dott
     \end{equation}
     \end{enumerate}}
\argument{\cref{eq:empirical_risk_for_mini_batch-conj3:setting: improve risk};\lref{eqp1};\lref{eqp2}}{that
     \begin{equation}\llabel{tg1}
     \begin{split}
         \cL(\vartheta)&=\frac 1M\bigg(\textstyle\sum\limits_{m\in\mathcal M}|\mN^{\fL,\vartheta}_\bfl(X^m)-Y^m|^2\bigg)+\displaystyle\frac 1M \bigg(\textstyle\sum\limits_{m\in \{1,2,\dots,M\}\backslash \mathcal M}|\mN^{\fL,\vartheta}_\bfl(X^m)-Y^m|^2\bigg)\\
         &=\frac 1M\bigg(\textstyle\sum\limits_{m\in\mathcal M}|Q+a\max\{\spro{w,X^m}+b,0\}-Y^m|^2\bigg)+\displaystyle\frac 1M \bigg(\textstyle\sum\limits_{m\in \{1,2,\dots,M\}\backslash \mathcal M}|Q-Y^m|^2\bigg)\\
         &=\frac{1}{M}\biggl[ \textstyle\sum\limits_{m=1}^{M}\abs{Q-Y^m}^2\biggl]+
     \displaystyle\frac{2a}{M}\biggl[ \textstyle\sum\limits_{m\in \mathcal M }\max\cu{\spro{w,X^m}+b,0}\bigl[Q-Y^m\bigr]\biggl]\\
     &+
     \displaystyle \frac{a^2}{M}\biggl[\textstyle \sum\limits_{m\in \mathcal M }\bigl[\max\cu{\spro{w,X^m}+b,0}\bigr]^2\biggr]\dott
         \end{split}
     \end{equation}}
     \argument{\lref{tg1};\lref{def v1}}{that 
     \begin{equation}\llabel{tg1.1}
     \begin{split}
         \cL(\vartheta)&=\frac{1}{M}\biggl[ \textstyle\sum\limits_{m=1}^{M}\abs{Q-Y^m}^2\biggl]-\displaystyle\frac{2a^2}{M}\biggl[\textstyle\sum\limits_{m\in \mathcal M }\bigl[\max\cu{\spro{w,X^m}+b,0}\bigr]^2\biggr]\\
         &+\displaystyle\frac{a^2}{M}\biggl[\textstyle\sum\limits_{m\in \mathcal M }\bigl[\max\cu{\spro{w,X^m}+b,0}\bigr]^2\biggr]\\
         &=\frac{1}{M}\biggl[ \textstyle\sum\limits_{m=1}^{M}\abs{Q-Y^m}^2\biggl]-\displaystyle\frac{a^2}{M}\biggl[\textstyle\sum\limits_{m\in \mathcal M }\bigl[\max\cu{\spro{w,X^m}+b,0}\bigr]^2\biggr]\\
         &=\displaystyle\frac{1}{M}\biggl[ \textstyle\sum\limits_{m=1}^{M}\abs{Q-Y^m}^2\biggr]-\displaystyle\frac{\big[\textstyle\sum_{m\in \mathcal M}[\spro{w,X^m}+b][Q-Y^m]\big]^2}{M\bigl[\textstyle\sum_{m\in  \mathcal M }|\max\cu{\spro{w,X^m}+b,0}|^2\bigr]}\dott
     \end{split}
     \end{equation}}
     \argument{\lref{tg1.1}; \cref{estimate lim}; \lref{def:Q} }{that 
 \begin{equation}
 \llabel{transform}
 \begin{split}
     \cL(\vartheta)
     = \biggl[\liminf_{n\to \infty}\cL(\varTheta^n)\biggr]
     -\frac{\bigl[\textstyle\sum_{m\in \mathcal M}[\spro{w,X^m}+b][Q-Y^m]\bigr]^2}{M\bigl[\textstyle\sum_{m\in  \mathcal M }[\max\cu{\spro{w,X^m}+b,0}]^2\bigr]}\dott
  \end{split} 
 \end{equation}}
 \argument{the fact that $\mathcal M\cap\mathcal{P}=\{p\}$;}{that \llabel{tg3}$(\mathcal{M}\backslash\{p\})=(\mathcal M\backslash(\mathcal M\cap \mathcal P))\subseteq (\{1,2,\dots,M\}\backslash\mathcal P)$\dott}
 \argument{\lref{tg3};the fact that for all $m\in \{1,2,\dots,M\}\backslash\mathcal P$ it holds that $Y^m=Q$; the fact that $Y^p\neq Q$}{that
 \begin{equation} \llabel{tg4} 
 \sum\limits_{m\in \mathcal M}\bigl[\spro{w,X^m}+b\bigr]\bigl[Q-Y^m\bigl]=[\spro{w,X^p}+b]\big[Q-Y^p\big]\neq 0\dott
 \end{equation}}
 \argument{\lref{tg4};}{that
 \begin{equation}\llabel{tg5}
 \begin{split}
     &\frac{\bigl[\textstyle\sum_{m\in \mathcal M}[\spro{w,X^m}+b][Q-Y^m]\bigr]^2}{M\bigl[\textstyle\sum_{m\in  \mathcal M }\bigl|\max\cu{\spro{w,X^m}+b,0}\bigr|^2\bigr]}\geq \frac{\bigl[\textstyle\sum_{m\in \mathcal M}[\spro{w,X^m}+b][Q-Y^m]\bigr]^2}{M\big[|\max\{\spro{w,\X^p}+b,0\}|^2\big]}\\
    &  =\displaystyle\frac{\bigl|[\spro{w,X^p}+b][Q-Y^p]\bigr|^2}{M\big[|\spro{w,\X^p}+b|^2\big]}=\frac{|Q-\Y^p|^2}{M|
\spro{w,\X^p}+b|}   
    >0\dott
     \end{split}
 \end{equation}}
 \argument{\lref{transform};\lref{tg5}}{\cref{conclude: min=max}\dott}
\end{aproof}
\subsection{Improving empirical risks for non-constant limiting output data}
\begin{athm}{lemma}{prop: improve risk_case2a}
Assume \cref{setting: improve risk}, assume $\bfl_1>3$, assume $\inf_{n\in\N}(\# (\inact_{ \nnode}^{ \varTheta^n}))\geq 3$, and assume $(\cup_{z=-1}^1\bfA_z)\neq\{1,2,\dots,M\}$. Then there exists $\vartheta\in \R^{\fd_\nnode}$ such that \begin{equation}\label{case 2b2 conclude}
            \cL(\vartheta)<\liminf_{n\to \infty}\cL(\varTheta^n).
        \end{equation}
\end{athm}
\begin{aproof}
\argument{the assumption that $(\cup_{z=-1}^1\bfA_z)\neq\{1,2,\dots,M\}$; the fact that $\bfA_{-1}\neq \emptyset$; the fact that $\bfA_1\neq \emptyset$}{that there exist $\mathfrak{p},p_1,p_2\in \{1,2,\dots,M\}$ which satisfy 
\begin{equation}\label{def pp1p2}
\mathfrak{p}\notin(\cup_{z=-1}^1\bfA_z),\qquad p_1\in \bfA_{-1},\qqandqq p_2\in \bfA_{1}\dott 
\end{equation}
}
\startnewargseq
\argument{\cref{def: M}; \cref{def pp1p2}; the fact that for all $q\in\{1,2,\dots,M\}$ it holds that $q\in \bfA_{-1}$ if and only if $\bfY^q\leq \min_{m\in \{1,2,\dots,M\}}\bfY^m$;the fact that for all $q\in\{1,2,\dots,M\}$ it holds that $q\in \bfA_{1}$ if and only if $\bfY^q\geq \max_{m\in \{1,2,\dots,M\}}\bfY^m$}{ that 
\begin{equation}\llabel{eq: prop: improve risk_case2a: 1}
\bfY^{\mathfrak{p}}\neq \Y^\mathfrak{p}, \qquad \bfY^\mathfrak{p} >\min_{m\in \{1,2,\dots,M\}}\bfY^m\geq\bfY^{p_1},\qqandqq \bfY^\mathfrak{p} <\max_{m\in \{1,2,\dots,M\}}\bfY^m\leq\bfY^{p_2}\dott
\end{equation}}
\argument{\lref{eq: prop: improve risk_case2a: 1};}{that
\begin{equation}\llabel{eq: prop: improve risk_case2a: 2}
    \bfY^{\mathfrak{p}}\neq \Y^\mathfrak{p}\qqandqq \bfY^{p_1}<\bfY^{\mathfrak{p}}<\bfY^{p_2}\dott
\end{equation}}
\argument{\lref{eq: prop: improve risk_case2a: 2};}{that there exists $Z\in \R$ which satisfies 
\begin{equation}\llabel{defY}
\bfY^{p_1}<Z<\bfY^{p_2} \qqandqq \abs{\bfY^{\mathfrak{p}}-\Y^\mathfrak{p}}>\abs{Z-\Y^\mathfrak{p}}\dott
\end{equation}}
\startnewargseq
\argument{\lref{defY};\cref{prop: represent};} { that there exist $N\in\N$ and $\vartheta=(\vartheta^n)_{n\in\N}\colon\N\to\R^{\fd_\bfl}$ which satisfy that 
\begin{enumerate}[label=(\roman*)]
    \item \label{arg:1 lem constant} it holds for all $n\in \N\cap[N,\infty)$, $m\in\{1,2,\dots,M\}\backslash\{\mathfrak{p}\} $ that $\mN^{\fL,\vartheta^n}_{\nnode}(\X^m)=\mN_{\bfl}^{\fL,\varTheta^n}(\X^m)$ 
    and
    \item \label{arg1' lem constant}it holds for all $n\in 
\N\cap[N,\infty)$ that $\mN^{\fL,\vartheta^n}_{\nnode}(\X^\mathfrak{p})=Z$\dott
\end{enumerate}}
\startnewargseq
\argument{ \cref{eq:empirical_risk_for_mini_batch-conj3:setting: improve risk}; \lref{defY}; items \ref{arg:1 lem constant} and \ref{arg1' lem constant};\cref{estimate lim};the assumption that $\limsup_{n\to \infty}\allowbreak\sum_{m=1}^M|\mN^{\fL,\varTheta^n}_{\nnode}(\X^m)
      -\bfY^m|=0$}{that 
\begin{equation}\llabel{arg1.5}
\begin{split}
\liminf_{n\to \infty}\cL(\vartheta^n)&=\liminf_{n\to\infty}
\bigg(\frac{ 1 }{ M } 
  \biggl[ \textstyle
  \sum\limits_{ m = 1 }^{ M}  
    \abs{
      \mN^{\fL,\vartheta^n}_{\nnode}(\X^m)
      - 
      \Y^m
    }^2
  \biggr]\bigg)\\
  &=\liminf_{n\to\infty}\bigg(\frac{ 1 }{ M } 
  \biggl[ \bigg(
  \textstyle\sum\limits_{m\in \{1,2,\dots,M\}\backslash\{\mathfrak{p}\}}  
    \abs{
      \mN^{\fL,\varTheta^n}_{\nnode}(\X^m)
      - 
      \Y^m
    }^2\bigg)+|Z-\Y^{\mathfrak{p}}|^2
  \biggr]\bigg)\\
  &= \frac{ 1 }{ M } 
  \biggl[ \bigg(\textstyle
  \sum\limits_{ m\in \{1,2,\dots,M\}\backslash\{\mathfrak{p}\}}  
    \abs{
      \bfY^m
      - 
      \Y^m
    }^2\bigg)+|Z-\Y^{\mathfrak{p}}|^2
  \biggr]\\
 & <\frac{ 1 }{ M } 
  \biggl[\textstyle 
  \sum\limits_{ m = 1 }^{ M}  
    \abs{
      \bfY^m
      - 
      \Y^m
    }^2
  \biggr]
=\liminf\limits_{n\to\infty}\cL(\varTheta^n)\dott
  \end{split}
\end{equation}} 
\argument{\lref{arg1.5};}{that there exists $m\in\N$ such that $\cL (\vartheta^m)<\liminf_{n\to\infty}\cL(\varTheta^n)$\dott}
\end{aproof}
\begin{athm}{lemma}{prop: improve risk_case2b3}
Assume \cref{setting: improve risk}, assume $\bfl_1>3$, assume $\inf_{n\in\N}(\# (\inact_{ \nnode}^{ \varTheta^n}))\geq 3$, assume \begin{equation}
    \min\limits_{m\in \{1,2,\dots,M\}}\bfY^m<\max\limits_{m\in \{1,2,\dots,M\}}\bfY^m,
    \end{equation}
    and assume $(\cup_{z\in\{-1,1\}}(\bfA_{z}\cap\bfM_{z}))\neq \emptyset$. Then there exists $\vartheta\in \R^{\fd_\nnode}$ such that 
    \begin{equation}\llabel{improve risk}
            \cL(\vartheta)<\liminf_{n\to \infty}\cL(\varTheta^n).
        \end{equation}
\end{athm}
\begin{aproof}
\argument{the assumption that $(\cup_{z\in\{-1,1\}}(\bfA_{z}\cap\bfM_{z}))\neq \emptyset$;}{that there exists $z\in \{-1,1\}$ which satisfies
\begin{equation}\llabel{def: z}
(\bfA_z\cap\bfM_z)\neq \emptyset\dott
\end{equation}}
\startnewargseq
     \argument{\lref{def: z};the fact that $\bfA_{-1}\neq\emptyset$; the fact that $\bfA_1\neq \emptyset$}{ that there exist $\mathfrak{p},q\in \N$ which satisfy 
     \begin{equation}\llabel{def: p}
     \mathfrak{p}\in (\bfA_{z}\cap \bfM_{z})\qqandqq q\in \bfA_{-z}\dott
     \end{equation}} 
     \startnewargseq
     \argument{\cref{def: M};\cref{def: A};\lref{def: p}; the assumption that $\min_{m\in \{1,2,\dots,M\}}\bfY^m<\max_{m\in \{1,2,\dots,M\}}\bfY^m$}{that $ z\bfY^\mathfrak{p}>z\Y^\mathfrak{p}$ and
     \begin{equation}\llabel{arg1}
     \begin{split}
      z\bfY^{\mathfrak{p}}&\geq \max_{m\in\{1,2,\dots,M\}}(z\bfY^m)>\min_{m\in\{1,2,\dots,M\}}(z\bfY^m)=\min_{m\in\{1,2,\dots,M\}}(-(-z)\bfY^m)\\
     &=-\bigg[\max_{m\in \{1,2,\dots,M\}}((-z)\bfY^m)\bigg]\geq -\big[(-z)\bfY^q\big]= z\bfY^{q}\dott
      \end{split}
     \end{equation}} 
     \argument{\lref{arg1};}{ that there exists $Z\in \R$ which satisfies 
     \begin{equation}\llabel{arg2}
     z\bfY^{q}<zZ<z\bfY^{\mathfrak{p}}\qqandqq\abs{\bfY^{\mathfrak{p}}-\Y^\mathfrak{p}}=|z\bfY^{\mathfrak{p}}-z\Y^{\mathfrak{p}}|>|zZ-z\bfY^{\mathfrak{p}}|=|Z-\Y^{\mathfrak{p}}|\dott
    \end{equation}}
    \startnewargseq
     \argument{\lref{arg2};\cref{prop: represent}}{that there exist $N\in \N$ and $\vartheta=(\vartheta^n)_{n\in \N}\colon \N\to \R^{\fd_\bfl}$  which satisfy that
     \begin{enumerate}[label=(\roman*)]
     \item \llabel{def N: 1}it holds for all $n\in \N\cap[N,\infty)$, $m\in\{1,2,\dots,M\}\backslash\{\mathfrak{p}\} $  that $\mN^{\fL,\vartheta^n}_{\nnode}(\X^m)=\mN^{\fL,\varTheta^n}_{\nnode}(\X^m)$ and 
     \item \llabel{def N: 2} it holds for all $n\in \N\cap[N,\infty)$ that $\mN^{\fL,\vartheta^n}_{\nnode}(\X^\mathfrak{p})=Z$\dott
     \end{enumerate}}
     \startnewargseq
  \argument{\lref{def N: 1}; \lref{def N: 2}; \lref{arg2};the assumption that $\limsup_{n\to \infty}\sum_{m=1}^M \allowbreak|\mN_\bfl^{\fL,\varTheta^n}(X^m)-\bfY^m|$=0}
      {that
      \begin{equation}\llabel{arg4}
      \begin{split}
          \liminf_{n\to \infty}\cL(\vartheta^n)&=\displaystyle\frac 1M\bigg[\bigg(\textstyle\sum\limits_{m\in \{1,2,\dots,M\}\backslash\{\mathfrak{p}\}}|\bfY^m-\Y^m|^2\bigg)+|Z-\bfY^{\mathfrak{p}}|^2\bigg]\\
          &<\displaystyle\frac 1M\bigg[\textstyle\sum\limits_{m=1}^M|\bfY^m-\Y^m|^2\bigg]\dott
          \end{split}
      \end{equation}}
      \argument{\lref{arg4};\cref{estimate lim}}{that
      \begin{equation}\llabel{arg5}
          \liminf_{n\to \infty}\cL(\vartheta^n)<\displaystyle\frac 1M\bigg[\textstyle\sum\limits_{m=1}^M|\bfY^m-\Y^m|^2\bigg]=\displaystyle\liminf_{n\to \infty}\cL(\varTheta^n)\dott
      \end{equation}}
  \argument{\lref{arg5};}{that there exists $m\in \N$ such that \llabel{arg6} $\cL(\vartheta^m)<\liminf_{n\to\infty}\cL(\varTheta^n)$\dott}
  \argument{\lref{arg6};}{\lref{improve risk}\dott}
\end{aproof}
    \begin{athm}{lemma}{prop: improve risk_case2b1}
        Assume \cref{setting: improve risk}, assume $\bfl_1> 3$, assume $\inf_{n\in\N}(\# (\inact_{ \nnode}^{ \varTheta^n}))\geq 3$, assume $(\cup_{z=-1}^1\bfA_z)=\{1,2,\dots,M\}$, assume  $\min_{m\in \{1,2,\dots,M\}}\bfY^m< \max_{m\in \{1,2,\dots,M\}}\bfY^m$, and assume $(\cup_{z\in\{-1,1\}}(\bfA_{z}\cap\bfM_{z}))=\emptyset$. Then there exists $\vartheta\in \R^{\fd_\nnode}$ such that 
        \begin{equation}
            \cL(\vartheta)<\liminf_{n\to \infty}\cL(\varTheta^n).
        \end{equation}
    \end{athm}
\begin{aproof}
    Throughout this proof let $\T_1,\T_2,\delta,\varepsilon\in \R$ satisfy
    \begin{equation}\llabel{def: T_1}
         \T_1=\min\limits_{m\in \{1,2,\dots,M\}}\bfY^m,\qquad\T_2=\max\limits_{m\in \{1,2,\dots,M\}}\bfY^m,\qquad\delta=\displaystyle\frac 1M\bigg[\textstyle\sum\limits_{m=1}^M(\Y^m-\bfY^m)\bigg],
    \end{equation}
  \begin{equation}\llabel{def: delta_case2}
        \text{and}\qquad \varepsilon=-\bigg[\textstyle\sum\limits_{m=1}^M\bfY^m(\bfY^m-\Y^m)\bigg]\bigg[\textstyle\sum\limits_{m=1}^M(\bfY^m)^2\bigg]^{-1}\mathbbm 1_{\{0\}}(\delta),
    \end{equation}
   and for every $n\in \N$ let $\vartheta^{n}=(\vartheta^{n}_1,\dots,\vartheta^{n}_{\fd_\nnode})\in \R^{\fd_\nnode}$ satisfy for all $s\in \{1,2,\dots,\fd_\bfl\}$ that
   \begin{equation}\llabel{ARG1}
\vartheta^{n}_s=\bigl(1+\varepsilon \mathbbm 1_{(\sum_{h=1}^{\fL-1}\nnode_h(\nnode_{h-1}+1),\infty)}(s)\bigr)\varTheta^n_s+\delta\mathbbm 1_{\{\fd_\bfl\}}(s).
   \end{equation}
\startnewargseq
\argument{\cref{eq:empirical_risk_for_mini_batch-conj3:setting: improve risk};\lref{ARG1}; the fact that for all $x,y\in \R$ it holds that $(x+y)^2=x^2+2xy+y^2$}{that for all $n\in \N$ it holds that
\begin{equation}\llabel{eq: case 2b2pre0}
\begin{split}
\cL(\vartheta^{n})&=\displaystyle\frac 1M \bigg[\textstyle\sum\limits_{m=1}^M\big|\mathcal N_\bfl^{\fL,\vartheta^{n}}(\X^m)-\Y^m\big|^2\bigg]
=\displaystyle\frac 1M \bigg[\textstyle\sum\limits_{m=1}^M\big|(1+\varepsilon)\mathcal N_\bfl^{\fL,\varTheta^{n}}(\X^m)+\delta-\Y^m\big|^2\bigg]\\
&=\displaystyle\frac 1M \bigg[\textstyle\sum\limits_{m=1}^M\big|\bigl(\mathcal N_\bfl^{\fL,\varTheta^{n}}(\X^m)-\Y^m\bigr)+\bigl(\varepsilon\mathcal N_\bfl^{\fL,\varTheta^{n}}(\X^m)+\delta\bigr)\big|^2\bigg] \\
&=\displaystyle\frac 1M\bigg[\textstyle\sum
\limits_{m=1}^M\big|\mathcal N_\bfl^{\fL,\varTheta^n}(\X^m)-\Y^m\big|^2\bigg]+\displaystyle\frac 2M\bigg[\textstyle\sum\limits_{m=1}^M\varepsilon \mathcal N_\bfl^{\fL,\varTheta^n}(\X^m)\big[\mathcal N_\bfl^{\fL,\varTheta^n}(\X^m)-\Y^m\big]\bigg]\\
&+\displaystyle\frac 2M\bigg[\textstyle\sum\limits_{m=1}^M\delta \big[\mathcal N_\bfl^{\fL,\varTheta^n}(\X^m)-\Y^m\big]\bigg]+\displaystyle\frac 1M\bigg[\textstyle\sum\limits_{m=1}^M\big[\varepsilon\mathcal N_\bfl^{\fL,\varTheta^n}(\X^m)+\delta\big]^2\bigg]\dott
\end{split}
\end{equation}
}
\argument{\lref{eq: case 2b2pre0};\cref{estimate lim}; the fact that for all $m\in\{1,2,\dots,M\}$ it holds that $\bfY^m=\lim_{n\to \infty}\mathcal N_{\bfl}^{\fL,\varTheta^n}(\X^m)$}{that 
\begin{equation}\llabel{eq: case 2b2pre}
    \begin{split}
\liminf_{n\to\infty}\cL(\vartheta^{n})&=\liminf_{n\to\infty}\cL(\varTheta^{n})
+\displaystyle\frac{2\varepsilon}{M}\bigg[\textstyle\sum\limits_{m=1}^M \bfY^m(\bfY^m-\Y^m)\bigg]\\
&+\displaystyle\frac{2\delta}{M}\bigg[\textstyle\sum\limits_{m=1}^M (\bfY^m-\Y^m)\bigg]+\displaystyle\frac1M\bigg[\textstyle\sum\limits_{m=1}^M[\varepsilon\bfY^m+\delta]^2\bigg]\dott
\end{split}
\end{equation}}
\argument{\lref{def: delta_case2};}{that \llabel{arg0.5}$\delta\varepsilon=0$\dott}
\argument{\lref{arg0.5};\lref{def: T_1};\lref{def: delta_case2};\lref{eq: case 2b2pre}}{that
\begin{equation}\llabel{EQT1}
\begin{split}
&\liminf_{n\to\infty}\cL(\vartheta^n)\\
&=\liminf_{n\to\infty}\cL(\varTheta^n)+\displaystyle\frac{2\varepsilon}{M}\bigg[\textstyle\sum\limits_{m=1}^M\bfY^m(\bfY^m-\Y^m)\bigg]\bigg[\textstyle\sum\limits_{m=1}^M(\bfY^m)^2\bigg]^{-1}\bigg[\sum\limits_{m=1}^M(\bfY^m)^2\bigg]\mathbbm 1_{\{0\}}(\delta)\\
&-2\delta^2+\displaystyle\frac 1M\bigg[\textstyle\sum\limits_{m=1}^M[\varepsilon^2(\bfY^m)^2+\delta^2]\bigg]\dott
\end{split}
\end{equation}}
\argument{\lref{EQT1};\lref{def: delta_case2}}{that 
\begin{equation}\llabel{EQT2}
    \begin{split}
\liminf_{n\to\infty}\cL(\vartheta^n)
=\liminf_{n\to\infty}\cL(\varTheta^n)-\displaystyle\frac{2\varepsilon^2}{M}\bigg[\textstyle \sum\limits_{m=1}^M(\bfY^m)^2\bigg]-2\delta^2+\displaystyle\frac{\varepsilon^2}{M}\bigg[\textstyle\sum\limits_{m=1}^M(\bfY^m)^2\bigg]+\delta^2\dott
\end{split}
\end{equation}
}
\argument{\lref{EQT2};}{that
\begin{equation}\llabel{EQT3}
    \liminf_{n\to\infty}\cL(\vartheta^n)
=\liminf_{n\to\infty}\cL(\varTheta^n)-\displaystyle\frac{\varepsilon^2}{M}\bigg[\textstyle\sum\limits_{m=1}^M(\bfY^m)^2\bigg]-\delta^2\dott
\end{equation}}
In the following we prove that
\begin{equation}\llabel{improve risk}
\displaystyle\frac{\varepsilon^2}{M}\bigg[\textstyle\sum\limits_{m=1}^M(\bfY^m)^2\bigg]+\delta^2>0
\end{equation}
We prove \lref{improve risk} by contradiction. In the following we thus assume that
\begin{equation}\llabel{improve risk1}
    \displaystyle\frac{\varepsilon^2}{M}\bigg[\textstyle\sum\limits_{m=1}^M(\bfY^m)^2\bigg]+\delta^2=0
\end{equation}
\startnewargseq
\argument{the assumption that $\min_{m\in \{1,2,\dots,M\}}\bfY^m< \max_{m\in \{1,2,\dots,M\}}\bfY^m$}{that there exists $m\in\{1,2,\dots,M\}$ such that \llabel{Argt1} $\bfY^m\neq 0$\dott}
\argument{\lref{Argt1};}{that
\begin{equation}\llabel{Argt2}
    \sum_{m=1}^M (\bfY^m)^2>0\dott
\end{equation}}
\argument{\lref{Argt2};\lref{improve risk1}}{that 
\begin{equation}\llabel{argt3}
\varepsilon=\delta=0
\dott
\end{equation}}
\argument{\lref{argt3};\lref{def: T_1}; \lref{def: delta_case2};\lref{Argt2}} {that
    \begin{equation}\llabel{case2}
    \sum_{m=1}^M(\bfY^m-\Y^m)=  \sum\limits_{m=1}^M\bfY^m(\bfY^m-\Y^m)=0
    \end{equation}}
        \argument{\lref{case2};the fact that $\bfA_{-1}\cap\bfA_1=\emptyset$; the fact that $(\bfA_{-1}\cup\bfA_0\cup\bfA_1)=\{1,2,\dots,M\}$; the fact that for all $m\in \bfA_0$ it holds that $\bfY^m-\Y^m=0$} { that
   \begin{equation} \llabel{arg3}
   \begin{split}
 &\biggl[\textstyle\sum
 \limits_{m\in \bfA_{-1}}(\bfY^m-\Y^m)\biggr]+\biggl[\sum\limits_{m\in \bfA_{1}}(\bfY^m-\Y^m)\biggr]\\
 &=\biggl[\textstyle\sum\limits_{m\in \bfA_{-1}}\bfY^m(\bfY^m-\Y^m)\biggr]+\biggl[\textstyle\sum\limits_{m\in \bfA_{1}}\bfY^m(\bfY^m-\Y^m)\biggr]=0\dott
   \end{split}
   \end{equation}}
   \argument{\lref{arg3}; the fact that for all $m\in \bfA_{-1}$ it holds that $\bfY^m=T_1$; the fact that for all $m\in \bfA_1$ it holds that $\bfY^m=T_2$}{that
   \begin{equation}\llabel{arg4}
   \begin{split}
      &\biggl[\textstyle\sum\limits_{m\in \bfA_{-1}}(\bfY^m-\Y^m)\biggr]+\biggl[\textstyle\sum\limits_{m\in \bfA_{1}}(\bfY^m-\Y^m)\biggr]\\
      &=\biggl[\textstyle\sum\limits_{m\in \bfA_{-1}}T_1(\bfY^m-\Y^m)\biggr]+\biggl[\textstyle\sum\limits_{m\in \bfA_{1}}T_2(\bfY^m-\Y^m)\biggr]=0\dott
      \end{split}
   \end{equation}}
   \argument{\lref{arg4};}{that
   \begin{equation}\llabel{arg4.1}
       T_1\biggl[\textstyle\sum\limits_{m\in \bfA_{-1}}(\bfY^m-\Y^m)\biggr]+T_2\biggl[\textstyle\sum
       \limits_{m\in \bfA_{1}}(\bfY^m-\Y^m)\biggr]=0\dott
   \end{equation}}
   \argument{\lref{arg4.1};\lref{arg4}}{that
   \begin{equation}\llabel{arg4.2}
       (T_2-T_1)\biggl[\textstyle\sum\limits_{m\in \bfA_{1}}(\bfY^m-\Y^m)\biggr]=0\dott
   \end{equation}}
   \argument{\lref{arg4.2};\lref{arg4};the fact that $T_1<T_2$}{that
  \begin{equation}\llabel{Argt3}
       \sum_{m\in \bfA_{-1}}(\bfY^m-\Y^m)=\sum_{m\in \bfA_{1}}(\bfY^m-\Y^m)=0\dott
   \end{equation}}
   \argument{\cref{def: M};}{that for all $z\in\{-1,1\}$, $m\in\bfM_z$ it holds that \llabel{argtt1} $z\bfY^m>z\Y^m$\dott}
   \argument{\lref{argtt1};}{that for all $z\in\{-1,1\}$, $m\in\{1,2,\dots,M\}\backslash\bfM_z$ it holds that \llabel{argtt2} $z\bfY^m\leq z\Y^m$\dott}
\argument{\lref{argtt2}; the fact that for all $z\in\{-1,1\}$ it holds that $\bfA_z\backslash(\{1,2,\dots,M\}\backslash\bfM_z)=(\bfA_z\cap\bfM_z)=\emptyset$}{that for all $z\in\{-1,1\}$, $m\in \bfA_z$ it holds that \llabel{argtt3}$z\bfY^m\leq z\Y^m$\dott}
\argument{\lref{argtt3};}{ for all $p\in \bfA_{-1}$, $q\in \bfA_1$ that
    \begin{equation}\llabel{Argt1'}
        \bfY^p\geq \Y^p\qqandqq\bfY^q\leq \Y^q\dott
    \end{equation}}
   \argument{\lref{Argt3};\lref{Argt1'}}{for all $m\in(\bfA_{-1}\cup\bfA_1)$ that
   \begin{equation}\llabel{Argt4}
       \bfY^m=\Y^m\dott
   \end{equation}}
   \argument{\lref{Argt4};the fact that for all $m\in \bfA_0$ it holds that $\bfY^m=\Y^m$; the fact that $(\bfA_{-1}\cup\bfA_0\cup\bfA_1)=\{1,2,\dots,M\}$}{that 
   \begin{equation}\llabel{argt1'}
       \sum_{m=1}^M|\bfY^m-\Y^m|^2= 0\dott
   \end{equation}}
   \argument{\cref{estimate lim};\lref{argt1'}} {that \begin{equation}\llabel{argt1}
 0<\liminf_{n\to\infty}\cL(\varTheta^n)= \sum_{m=1}^M|\bfY^m-\Y^m|^2=0\dott
   \end{equation}}
   This contradiction shows \lref{improve risk}.
   \startnewargseq
   \argument{\lref{EQT3};\lref{improve risk}}{that \llabel{EQT7}$\liminf_{n\to\infty}\cL(\vartheta^n)
<\liminf_{n\to\infty}\cL(\varTheta^n)$\dott}
\argument{\lref{EQT7};}{there exists $m\in \N$ such that $\cL(\vartheta^m)<\liminf_{n\to\infty}\cL(\varTheta^n)$\dott}
\end{aproof}
\subsection{Improving empirical risks for general limiting output data}\label{subsec: improving general limiting data}
\begin{athm}{prop}{lem: improve risk}
Assume \cref{setting: improve risk}, assume $\bfl_1>3$, and assume $\inf_{n\in\N}(\# (\inact_{ \nnode}^{ \varTheta^n}))\geq 3$. Then there exists $\vartheta\in \R^{\fd_\nnode}$ such that 
\begin{equation}\label{conclude: lem: improve risk}
    \cL(\vartheta)<\liminf_{n\to\infty}\cL(\varTheta^n).
\end{equation}
\end{athm}
\begin{aproof}
    \argument{ \cref{prop: improve risk_case2b2}; \cref{prop: improve risk_case2a}; \cref{prop: improve risk_case2b3};\cref{prop: improve risk_case2b1}}  { \cref{conclude: lem: improve risk}\dott}
\end{aproof}
\begin{athm}{cor}{cor: improve risk}
    Assume \cref{setting:multilayer}, let $M\in \N$, for every $m\in\{1,2,\dots,M\}$ let $\X^m\in [a,b]^d$, $\Y^m \in \R$, assume $\#\{\X^1,\X^2,\dots,\X^M\}=M$, let $\fL\in \N\backslash\{1\}$, $\bfl=(\bfl_0,\bfl_1,\dots,\bfl_{\fL})\in \N^{\fL+1}$ satisfy $\bfl_0=d$, $\bfl_1\geq 3$, and $\bfl_\fL=1$, let
$ 
  \cL \colon \R^{ \fd_{ \nnode } } \to \R 
$
satisfy for all 
$ \theta \in \R^{ \fd_{ \nnode} }$
that
\begin{equation}
\label{eq:empirical_risk_for_mini_batch-conj3: cor}
  \cL( \theta) 
  = 
 \displaystyle \frac{ 1 }{ M } 
  \biggl[ \textstyle
  \sum\limits_{ m = 1 }^{ M}  
    \abs{
      \mN^{\fL,\theta}_{\nnode}(\X^m)
      - 
      \Y^m
    }^2
  \biggr]
  ,
\end{equation}
 and assume $\inf_{\theta\in \R^{\fd_{\bfl}}}\cL(\theta)>0$. Then 
    \begin{equation}\label{conclude: cor: improve risk}
       \inf_{\theta\in \R^{\fd_{\bfl}}}\cL(\theta) < \inf_{\theta\in \R^{\fd_{\bfl}},\ \#(\inact_{\bfl}^\theta)\geq 3}\cL(\theta).
    \end{equation}
\end{athm}
\begin{aproof}
    In the proof of \eqref{conclude: cor: improve risk} we distinguish between the case $\bfl_1=3$ and the case $\bfl_1>3$. We first prove \eqref{conclude: cor: improve risk} in the case
    \begin{equation}\llabel{case1}
        \bfl_1=3.
    \end{equation}
\argument{\cref{relization multi};\cref{eq:setting_definition_of_I_set};\cref{eq:empirical_risk_for_mini_batch-conj3: cor};\lref{case1}; the assumption that $\inf_{\theta\in \R^{\fd_\bfl}}\cL(\theta)>0$}{ that 
\begin{equation}\llabel{infisconstant}
\begin{split}
       0&< \inf_{\theta\in \R^{\fd_\bfl}}\cL(\theta)\leq \inf_{\theta\in \R^{\fd_{\bfl}},\ \#(\inact_{\bfl}^\theta)\geq 3}\cL(\theta)=\inf_{\theta\in \R^{\fd_{\bfl}},\ \inact_{\bfl}^\theta=\{1,2,3\}}\cL(\theta)\\
       &= \displaystyle\inf_{\theta\in \R^{\fd_{\bfl}},\ \inact_{\bfl}^\theta=\{1,2,3\}}\bigg(\frac{1}{M}\bigg[\textstyle\sum\limits_{m=1}^M|\mN^{\fL,\theta}_{\nnode}(\X^m)-Y^m|^2\bigg]\bigg)=\displaystyle\inf\limits_{Q\in \R} \bigg(\displaystyle\frac{1}{M}\bigg[\textstyle\sum\limits_{m=1}^M|Q-Y^m|^2\bigg]\bigg) \dott
       \end{split}
        \end{equation}}
        \argument{\lref{infisconstant};}{that there exist $\varTheta=(\varTheta^n)_{n\in \N}\colon \N\to\{\theta\in \R^{\fd_\bfl}\colon \inact^\theta_\bfl=\{1,2,3\}\}$ which satisfy for all $n,m\in \N$ that $\varTheta^n=\varTheta^m$ and
        \begin{equation}\llabel{infisconstant2}
            0<\displaystyle\inf\limits_{Q\in \R} \bigg(\displaystyle\frac{1}{M}\bigg[\textstyle\sum\limits_{m=1}^M|Q-Y^m|^2\bigg]\bigg)=\cL(\varTheta^n)\dott
        \end{equation}}
        \startnewargseq
\argument{\lref{infisconstant};\lref{infisconstant2};\cref{prop: improve risk_case2b2}}{that there exists $\vartheta\in \R^{\fd_\bfl}$ such that 
\begin{equation}\llabel{vartheta}
 \cL(\vartheta)<\liminf_{n\to\infty}\cL(\varTheta^n)=\inf_{\theta\in \R^{\fd_\bfl},\ \#(\inact_\bfl^\theta)\geq 3}\cL(\theta)\dott
\end{equation}}
\argument{\lref{vartheta};}{\eqref{conclude: cor: improve risk} in the case $\bfl_1=3$\dott}
    We now prove \eqref{conclude: cor: improve risk} in the case 
    \begin{equation}\llabel{case2}
        \bfl_1>3.
    \end{equation}
    \startnewargseq
    \argument{the fact that $\inf_{\theta\in \R^{\fd_{\bfl}},\ \#(\inact_{\bfl}^\theta)\geq 3}\cL(\theta)<\infty$}{that there exists $\varTheta=(\varTheta^k)_{k\in \N}\colon \N\to\{\theta\in \R^{\fd_\bfl}\colon \#(\inact_{\bfl}^\theta)\geq 3\}$ which satisfies 
    \begin{equation}\llabel{def: varTheta}
\limsup_{k\to\infty}\cL(\varTheta^k)=\liminf_{k\to\infty}\cL(\varTheta^k)=\inf_{\theta\in \R^{\fd_{\bfl}},\ \#(\inact_{\bfl}^\theta)\geq 3}\cL(\theta)<\infty \dott
    \end{equation}}
    \startnewargseq
    \argument{\lref{def: varTheta}; the assumption that $\inf_{\theta\in \R^{\fd_\bfl}}\cL(\theta)>0$}{that \begin{equation}\llabel{ARG1}
0<\inf_{\theta\in\R^{\fd_\bfl}}\cL(\theta)\leq
    \inf_{\theta\in\R^{\fd_\bfl},\,\#(\inact_\bfl^{\theta})\geq 3}\cL(\theta)=\liminf_{k\to\infty}\cL(\varTheta^k)=\limsup_{k\to\infty}\cL(\varTheta^k)\leq
    \sup_{k\in \N}\cL(\varTheta^k)<\infty\dott
    \end{equation}}
    \argument{\lref{ARG1}; \cref{eq:empirical_risk_for_mini_batch-conj3: cor}}{that \llabel{ARG2} $\sup_{k\in \N}(\sum_{m=1}^M\allowbreak|\mN^{\fL,\varTheta^k}_\bfl(\X^m)-\Y^m|^2<\infty)$\dott}
    \argument{\lref{ARG2};}{that
    \begin{equation}\llabel{ARG3}
        \sup_{k\in \N}\bigg(\textstyle\sum\limits_{m=1}^M|\mN^{\fL,\varTheta^k}_\bfl(\X^m)|^2\bigg)^{\!\nicefrac{1}{2}}<\infty\dott
    \end{equation}}
    \argument{\lref{ARG3};}{that there exist $\bfY^1,\bfY^2,\ldots,\bfY^M\in \R$ and a strictly increasing $k=(k_n)_{n\in \N}\colon \N\to\N$ which satisfy
    \begin{equation}\llabel{ARG4}
\limsup_{n\to\infty}\bigg(\textstyle\sum\limits_{m=1}^M |\mN^{\fL,\varTheta^{k_n}}_\bfl(\X^m)-\bfY^m|\bigg)=0\dott
    \end{equation}}
    \startnewargseq
    \argument{\lref{case2};\lref{def: varTheta};\lref{ARG1};\lref{ARG4};\cref{lem: improve risk};the assumption that $\#\{\X^1,\X^2,\allowbreak\ldots,\X^M\}=M$}{that there exists $\vartheta\in \R^{\fd_\bfl}$ such that
    \begin{equation}\llabel{ARG5}
        \inf_{\theta\in \R^{\fd_\bfl}}\cL(\theta)\leq\cL(\vartheta)<\liminf_{n\to\infty}\cL(\varTheta^{k_n})=\inf_{\theta\in \R^{\fd_{\bfl}},\ \#(\inact_{\bfl}^\theta)\geq 3}\cL(\theta)\dott
    \end{equation}}
    \argument{\lref{ARG5};}{\cref{conclude: cor: improve risk} in the case $\bfl_1>3$\dott}
\end{aproof}
\def\Reals{\mathbb{R} }
\newcommand{\Rr}{\Reals }
\newcommand{\A}{\mathscr{A} }
\renewcommand{\B}{\mathscr{B} }
\def\enne{\mathbb{N} }
\def\ip #1#2{\left<#1,#2\right>}
\def\mN{\mathcal N}
\def\mQ{\mathcal Q}
\def\mL{\mathcal L}
\def\mA{\mathcal A}
\def\mG{\mathcal G}
\def\mX{\mathcal X}
\def\mM{\mathcal{M} }
\def\mH{\mathcal H}
\renewcommand{\R}{\mathbb {A}}
\renewcommand{\neural}[3]{\mathcal{N}_{#1}^{#2,#3}}
\renewcommand{\neurali}[4]
{\mathcal{N}_{#1,#2}^{#3,#4}}
\renewcommand{\fG}{\mathcal{G}}
\renewcommand{\mG}{\mathcal{G}}
\section{Explicit representations for generalized gradients of the risk functions in data driven supervised deep learning (DL) problems}\label{sec: represent general grad}
Theorem 2.9 in \cite{MaArconvergenceproof} provides explicit representations 
for generalized gradients of the risk function describing the mean squared error 
between the realization of an \ANN\ and a given target function. 
The main result of this section, \cref{theo: explicit formula} in \cref{subsec: explicit representations for generalized gradient} below, generalizes Theorem 2.9 in 
\cite{MaArconvergenceproof} from the setup with a given target function 
to a general supervised learning problem with a (probability) measure describing the law 
of input and output data. We prove \cref{theo: explicit formula} through an application 
of Theorem 2.9 in \cite{MaArconvergenceproof} and 
through the well-known formulation of the target function as a factorization 
of the conditional expectation of the output datum given the input data. 
The arguments in this section are essentially well-known (cf., \eg, \cite{BerGroJent2020,2001OnTM,ArBePhi2024,klenkeprobability}) and only for completeness 
we recall in \cref{subsec: fratorization lemmas} the appropriate factorization lemmas that we employ 
in our proof of \cref{theo: explicit formula}.

\subsection{Mathematical description of risks and gradients for supervised DL problems}
\begin{setting}\label{setting: explicit}
Let 
$ \fd,L \in \N $,
$ \ell_0,\ell_1,\dots,\ell_L\in \N $
satisfy $ \fd = \sum_{ k = 1 }^L \ell_k( \ell_{k-1} + 1) $,
for every $k\in \{1,2,\dots,L\}$, $\theta = ( \theta_1, \ldots, \theta_{ \fd }) \in \Rr^{ \fd } $, $i\in \{1,2,\dots,\ell_k\}$ let $\fb^{k,\theta}_{i}, \fw^{k,\theta}_{i,1}, \fw^{k,\theta}_{i,2},\ldots, \fw^{k,\theta}_{i,\ell_{k-1}}\in \Rr$
satisfy for all
$ j \in \{ 1, 2,\ldots, \allowbreak\ell_{k-1}\allowbreak\} $ that
\begin{equation}
\label{wb}
\fb^{ k, \theta }_i =
  \theta_{\ell_k\ell_{k-1} + i+\sum_{h= 1 }^{ k-1} \ell_h( \ell_{h-1} + 1)} 
  \qqandqq
  \fw^{ k, \theta }_{ i, j } 
  =
  \theta_{ (i-1)\ell_{k-1} + j+\sum_{h= 1 }^{ k-1} \ell_h( \ell_{h-1} + 1)}
   \,,
\end{equation}
let $ \A \in (0,\infty) $, 
$ \B \in ( \A, \infty) $, let $\R_r\colon \mathbb{R} \rightarrow \mathbb{R}$, $r\in [1,\infty]$, 
satisfy for all $r\in [1,\infty)$, $x\in (-\infty,\scrA r^{-1}]\cup[\scrB r^{-1},\infty)$, $y\in \mathbb R$ that
\begin{equation}\label{limR_0}
    \R_r\in C^1(\Rr,\Rr),\quad \R_r(x)=x\mathbbm 1_{(\scrA r^{-1},\infty)}(x),\qandq 0\leq \R_r(y)\leq \R_\infty(y)=\max\{y,0\},
    \end{equation}
assume 
$
  \sup_{ r \in [1, \infty) }
  \sup_{ x \in \Reals } 
  | ( \R_r )'( x ) | < \infty 
$,
for every $r\in[1,\infty]$, $\theta\in\Rr^{\fd}$ let $\neural{r}{k}{\theta}=(\neurali{r}{1}{k}{\theta},\dots,\neurali{r}{\ell_k}{k}{\theta}) \allowbreak\colon\allowbreak \Rr^{ \ell_0 } \to \Rr^{ \ell_k} $, $k \in \{0,1,\dots,L\}$, satisfy for all $k\in \{0,1,\dots,L-1\}$, $x=(x_1,\dots, x_{\ell_0})\in \Rr^{\ell_0}$, $i\in\{1,2,\dots,\ell_{k+1}\}$ that
\begin{equation}\label{explicit relization}
  \neurali{r}{i}{k+1}{\theta}( x ) = \fb_i^{k+1,\theta}+\textstyle\sum\limits_{j=1}^{\ell_{k}}\fw_{i,j}^{k+1,\theta}\big(x_j\indicator{\{0\}}(k) 
  +\bA_{r^{1/\!\max\{k,1\}}}(\neurali{r}{j}{k}{\theta}(x))\indicator{\N}(k) 
    \big),
\end{equation}
let $ a \in \Reals $, $ b \in (a, \infty) $, 
let $\mu\colon \mathcal B([a,b]^{\ell_0}\times \Rr^{\ell_L})\to[0,\infty]$ be a measure,
for every $ r \in [1,\infty]$ 
let $ \mL_r\colon \Reals^{ \fd }\to \Reals $ satisfy\footnote{Note that for all $d \in \N$, $v = ( v_1, ..., v_d ) \in \Rr^d$ it holds that $\| v \| = [\sum_{ i = 1 }^d | v_i |^2]^{ 1 / 2 }$.}
for all $ \theta \in \Reals^{ \fd } $
that
\begin{equation}
\label{L_r}
  \displaystyle
  \mL_r( \theta)=\int_{[a,b]^{ \ell_0 } \times \Rr^{\ell_L}}
  \|\mN_r^{L, \theta }(x)-y\|^2\, \mu( \d x,\d y),
\end{equation}
and let $ \mG = ( \mG_1, \dots, \mG_{ \fd } ) \colon \Reals^{ \fd } \to \Reals^{ \fd } $ 
satisfy
for all $ \theta \in \{\vartheta \in \Reals^{ \fd }\colon(( \nabla\mL_r)( \vartheta))_{ r \in [1,\infty) }
\text{ is convergent} \} $ that
$ \mG( \theta)=\lim_{r\to \infty }( \nabla\mL_r)( \theta) $.
\end{setting}
\newcommand{\diml}{\mathbf{d}}
\DeclarePairedDelimiter{\pr}{ (}{)}
\subsection{Integrability properties}
\begin{athm}{lemma}{Lemma XX}
    Assume \cref{setting: explicit}. Then 
    \begin{enumerate}[label=(\roman*)]
        \item \llabel{item 1} it holds that $\mu([a,b]^{\ell_0}\times \Rr^{\ell_L})<\infty$,
        \item \llabel{item 2} it holds that
$\int_{[a,b]^{\ell_0}\times\Rr^{\ell_L}}\|y\|^2\, \mu(\d x,\d y)<\infty$,
\item \llabel{item 3} it holds that $\int_{[a,b]^{\ell_0}\times\Rr^{\ell_L}}\|x\|^2\, \mu(\d x,\d y)<\infty$, and
\item \llabel{item 4} it holds for all $r\in [1,\infty]$, $k\in \{1,2,\dots,L\}$, $\theta\in \Rr^{\fd}$ that
$\int_{[a,b]^{\ell_0}\times\Rr^{\ell_L}}\|\mathcal N_{r}^{k,\theta}(x)\|^2\, \mu(\d x,\d y)\allowbreak<\infty$.
    \end{enumerate}
\end{athm}
\begin{aproof}
    Throughout this proof let $e=(e_1,\dots,e_L)\in \Rr^{L}$ satisfy for all $i\in \{1,2,\dots,\allowbreak L-1\}$ that $e_i=0$ and $e_L=1$ and let $\theta=(\theta_1,\dots,\theta_L)\in \Rr^{L}$ satisfy for all $i\in \{1,2,\dots,\allowbreak \fd-1\}$ that $\theta_i=0$ and $\theta_\fd=1$.
    \argument{\cref{explicit relization};induction}{for all $x\in [a,b]^{\ell_0}$ that 
    \begin{equation}\llabel{eq1}
        \neural{\infty}{L}{\theta}(x)=e\qqandqq \neural{\infty}{L}{0}(x)=0\dott
    \end{equation}}
    \argument{\lref{eq1};the fact that $\cL_\infty(0),\cL_\infty(\theta)\in \Rr$}{that
    \begin{equation}\llabel{eq2}
        \int_{[a,b]^{ \ell_0 } \times \Rr^{\ell_L}}
  \|y-e\|^2\, \mu( \d x,\d y)<\infty \qqandqq \int_{[a,b]^{ \ell_0 } \times \Rr^{\ell_L}}
  \|y\|^2\, \mu( \d x,\d y)<\infty\dott
    \end{equation}}
    \argument{\lref{eq2}; the fact that for all $x,y\in \Rr^L$ it holds that $\|x+y\|^2\leq 2\|x\|^2+2\|y\|^2$}{that
    \begin{equation}\llabel{eq3}
        \begin{split}
             &\mu([a,b]^{\ell_0}\times \Rr^{\ell_L})=\int_{[a,b]^{ \ell_0 } \times \Rr^{\ell_L}}
  \|e\|^2\, \mu( \d x,\d y)=\int_{[a,b]^{ \ell_0 } \times \Rr^{\ell_L}}
  \|e-y+y\|^2\, \mu( \d x,\d y)\\
  &\leq \int_{[a,b]^{ \ell_0 } \times \Rr^{\ell_L}}
  2\|e-y\|^2\, \mu( \d x,\d y)+\int_{[a,b]^{ \ell_0 } \times \Rr^{\ell_L}}
  2\|y\|^2\, \mu( \d x,\d y) <\infty\dott
        \end{split}
    \end{equation}}
    \argument{\lref{eq3};}{\lref{item 1}\dott}
    \argument{\lref{eq2};}{\lref{item 2}\dott}
    \argument{\lref{item 1};}{that
    \begin{equation}\llabel{eq4}
\int_{[a,b]^{\ell_0}\times\Rr^{\ell_L}}\|x\|^2\, \mu(\d x,\d y)\leq \int_{[a,b]^{\ell_0}\times\Rr^{\ell_L}}\ell_0(\max\{|a|,|b|\})^2\, \mu(\d x,\d y)<\infty\dott
    \end{equation}}
    \argument{\lref{eq4};}[verbs=ep]{\lref{item 3}\dott}
    \argument{the fact that for all $r\in [1,\infty]$, $k\in \{1,2,\dots,L\}$, $\theta\in \Rr^{\fd}$ it holds that $\mathcal N_r^{k,\theta}$ is continuous}{that for all $r\in [1,\infty]$, $k\in \{1,2,\dots,L\}$, $\theta\in \Rr^{\fd}$ it holds that \llabel{argg1} $\mathcal N_r^{k,\theta}$ is bounded on $[a,b]^{\ell_0}$\dott}
    \argument{\lref{argg1};\lref{item 1}}{\lref{item 4}\dott}
\end{aproof}
\subsection{Factorization lemmas for conditional expectations}\label{subsec: fratorization lemmas}
In this section we recall well-known factorization lemmas (for conditional expectations). In particular, the following two lemmas, \cref{Lemma YY1} and \cref{Lemma YY2}, are, \eg, proved as Corollary 1.97 in \cite{klenkeprobability}.
\begin{athm}{lemma}{Lemma YY1}   
Let $\Omega$ be a set, let $(S,\mathcal S)$ be a measurable space, let $X\colon \Omega\to S$ be a function, and let $Y\colon \Omega\to [0,\infty]$ be $\sigma(X)/\mathcal B([0,\infty])$-measurable. Then there exists a measurable $f\colon S\to [0,\infty]$ such that $Y=f\circ X$.
\end{athm}
\begin{aproof}
\argument{\cite[Item (ii) in Theorem 1.96]{klenkeprobability};the fact that $Y$ is $\sigma(X)/\mathcal B([0,\infty])$-measurable}{that there exist $A=(A_n)_{n\in \N}\colon \N\to\sigma(X)$ and $\alpha=(\alpha_n)_{n\in \N}\colon \N\allowbreak\to [0,\infty)$ which satisfy for all $\omega\in \Omega$ that
\begin{equation}\llabel{eq1}
    Y(\omega)=\textstyle\sum\limits_{n=1}^\infty \alpha_n\mathbbm 1_{A_n}(\omega)\dott
\end{equation}}
\startnewargseq
\argument{the fact that for all $n\in \N$ it holds that $A_n\in \sigma(X)$;}{that there exists $\fA=(\fA_n)_{n\in \N} \colon \N\to \mathcal S$ which satisfies for all $n\in \N$ that
\begin{equation}\llabel{eq2}
    X^{-1}(\fA_n)=A_n.
\end{equation}}
In the following let $f\colon S\to [0,\infty]$ satisfy for all $s\in S$ that
\begin{equation}\llabel{def: f}
    f(s)= \textstyle\sum\limits_{n=1}^\infty \alpha_n\mathbbm 1_{\fA_n}(s)\dott
\end{equation}
\startnewargseq
\argument{\lref{eq1};\lref{eq2};\lref{def: f}}{that 
\begin{equation}\llabel{eq4}
    Y=f\circ X\dott
\end{equation}}
\argument{\lref{def: f}; the fact that for all $n\in\N$ it holds that $\fA_n\in \mathcal S$}{that \llabel{arg1} $f$ is measurable\dott}
\end{aproof}
\begin{athm}{lemma}{Lemma YY2}
    Let $\Omega$ be a set, let $(S,\mathcal S)$ be a measurable space, let $X\colon \Omega\to S$ be a function, and let $Y\colon \Omega\to [-\infty,\infty]$ be $\sigma(X)/\mathcal B([-\infty,\infty])$-measurable. Then there exists a measurable $f\colon S\to [-\infty,\infty]$ such that $Y=f\circ X$.
\end{athm}
\begin{aproof}
\argument{the assumption that $Y$ is $\sigma(X)/\mathcal B([-\infty,\infty])$-measurable}{that \llabel{argt1} $(\Omega\ni\omega\mapsto \max\{Y(\omega),0\}\in [0,\infty])$ and $(\Omega\ni\omega\mapsto \max\{-Y(\omega),0\}\in [0,\infty])$ are $\sigma(X)/\mathcal B([0,\infty])$-measurable\dott}
\argument{\cref{Lemma YY1};\lref{argt1}}{that there exist measurable $g_k\colon S \to [0,\infty]$, $k\in \{1,2\}$, which satisfy for all $\omega\in \Omega$ that
\begin{equation}\llabel{eq1}
    \max\{Y(\omega),0\}=(g_1\circ X)(\omega)\qqandqq
    \max\{-Y(\omega),0\}=
  ( g_2\circ X)(\omega)\dott
\end{equation}}
\startnewargseq
In the following let $f\colon S\to [-\infty,\infty]$ satisfy for all $s\in S$ that
\begin{equation}\llabel{def: f}
    f(s)=\begin{cases}
    0&\colon g_1(s)=g_2(s)=\infty\\
    g_1(s)-g_2(s)&\colon \text{else}.
    \end{cases}
\end{equation}
\startnewargseq
\argument{\lref{eq1};\lref{def: f}}{that for all $\omega\in \Omega$  it holds that 
\begin{equation}\llabel{eq4}
\begin{split}
    Y(\omega)&=\max\{Y(\omega),0\}+\min\{Y(\omega),0\}=\max\{Y(\omega),0\}-\max\{-Y(\omega),0\}\\
    &=(g_1\circ X)(\omega)-(g_2\circ X)(\omega)=g_1(X(\omega))-g_2(X(\omega))=f(X(\omega))\dott
    \end{split}
\end{equation}}
\argument{\lref{eq4};}{that $Y=f\circ X$\dott}
\argument{\lref{def: f}; the fact that for all $k\in \{1,2\}$ it holds that $g_k$ is measurable}{that \llabel{arg1} $f$ is measurable\dott}
\end{aproof}
\begin{athm}{lemma}{Lemma YY}
   Let $\Omega$ be a set, let $(S,\mathcal S)$ be a measurable space, let $X\colon \Omega\to S$ be a function, let $\delta\in \N$, let $D\in \mathcal B ([-\infty,\infty]^\delta)$ be non-empty, and let $Y\colon \Omega\to D$ be $\sigma(X)/\mathcal B(D)$-measurable. Then there exists a measurable $f\colon S\to D$ such that $Y=f\circ X$.
\end{athm}
\begin{aproof}
Throughout this proof let $y=(y_1,\dots,y_\delta)\colon \Omega\to [-\infty,\infty]^\delta$ satisfy for all $\omega\in \Omega$ that
    \begin{equation}\llabel{def: y}
        y(\omega)=Y(\omega),
    \end{equation} let $v\in D$, and let $\phi\colon [-\infty,\infty]^\delta\to D$ satisfy for all $x\in [-\infty,\infty]^\delta$ that
\begin{equation}\llabel{def: phi}
    \phi(x)=\begin{cases}
        x&\colon x\in D\\
        v&\colon x\notin D.
    \end{cases}
\end{equation}
\argument{\lref{def: y}; the assumption that $Y$ is $\sigma(X)/\mathcal B(D)$-measurable}{that for all $A\in \mathcal B([\infty,\infty]^\delta)$ it holds that \llabel{argt1} $y^{-1}(A)=y^{-1}(A\cap D)=Y^{-1}(A\cap D)\in \sigma(X)$\dott}
\argument{\lref{argt1};}{that \llabel{argt2} $y$ is $\sigma(X)/\mathcal B([-\infty,\infty]^\delta)$-measurable\dott}
\argument{\lref{argt2};}{for all $i\in \{1,2,\dots,\delta\}$ that \llabel{argt3} $y_i$ is $\sigma(X)/\mathcal B([-\infty,\infty])$-measurable\dott}
\argument{\lref{argt3};\cref{Lemma YY2}}{that there exist measurable $g_i\colon S \to [-\infty,\infty]$, $i\in\{1,2,\dots,\delta\}$, which satisfy for all $i\in \{1,2,\dots,\delta\}$ that
\begin{equation}\llabel{def: F_i}
    y_i=g_i\circ X\dott
\end{equation}}
\startnewargseq
In the following let $G\colon S\to [-\infty,\infty]^\delta$ and $f\colon S\to D$ satisfy for all $s\in S$ that
\begin{equation}\llabel{eq2}
    G(s)=(g_1(s),g_2(s),\dots,g_\delta(s))\qqandqq f(s)=\phi (G(s))\dott
\end{equation}
\startnewargseq
\argument{\lref{def: y};\lref{def: F_i};\lref{eq2} }{that for all $\omega\in \Omega$ it holds that
\begin{equation}\llabel{eq3}
\begin{split}
Y(\omega)&=y(\omega)=(y_1(\omega),y_2(\omega),\dots,y_\delta(\omega))\\
&= ((g_1\circ X)(\omega),(g_2\circ X)(\omega),\dots, (g_\delta\circ X)(\omega))=G(X(\omega))\dott
\end{split}
\end{equation}}
\argument{\lref{eq3};\lref{def: phi};\lref{eq2}}{that for all $\omega\in \Omega$ it holds that
\begin{equation}\llabel{eqt1}
    Y(\omega)=G(X(\omega))=\phi(G(X(\omega)))=f(X(\omega))\dott
\end{equation}}
\argument{\lref{eqt1};}{that $Y=f\circ X$\dott}
\argument{\lref{def: phi};\lref{eq2}; the fact that $G$ is measurable}{that \llabel{arg4} $f$ is measurable\dott}
\end{aproof}
\begin{athm}{cor}{Lemma YY'}
    Let $\Omega$ be a set,  let $(S,\mathcal S)$ be a measurable space, let $\delta\in \N$, let $D\in \mathcal B( [-\infty,\infty]^\delta)$ be non-empty, and let $X\colon \Omega\to S$ and $Y\colon \Omega\to D$ be functions. Then the following two statements are equivalent:
    \begin{enumerate}[label=(\roman*)]
        \item \label{item 1: aaaa} It holds that  $Y$ is $\sigma(X)/\mathcal B(D)$-measurable.
        \item \label{item 2: aaaa} There exists a measurable $f\colon S\to D$ such that $Y=f\circ X$.
    \end{enumerate}
\end{athm}
\begin{aproof}
    \argument{\cref{Lemma YY};}{that (\ref{item 1: aaaa}$\rightarrow$\ref{item 2: aaaa})\dott}
    \startnewargseq
    \argument{the fact that for every measurable $f\colon S\to D$ it holds that $f\circ X$ is $\sigma(X)/\mathcal B(D)$-measurable}{that (\ref{item 2: aaaa}$\rightarrow$\ref{item 1: aaaa})\dott}
\end{aproof}
Conditional expectations provide natural examples of the $\sigma(X)/\mathcal{B}(D)$-measurable random variable in \cref{item 1: aaaa} in \cref{Lemma YY'}. This class of examples is the subject of following auxiliary lemma, \cref{Lemma VV} below. We note that the random variables $X$ and $Y$ in \cref{Lemma VV} do not necessarily need to be integrable (cf., \eg, \cite[Item (i) in Lemma 3.12]{Dereichnonconvergence2024}).

\begin{athm}{lemma}{Lemma VV}
    Let $(\Omega,\mathcal F,\P)$ be a probability space, let $d,\delta\in \N$, and let $X=(X_1,\dots,X_d)\colon \Omega\to\Rr^d$ and $Y=(Y_1,\dots,Y_\delta)\colon \Omega\to\Rr^\delta$ be random variables which satisfy $\P$-a.s.\ that $\E[\|Y\|
     | X]<\infty$. Then there exists a measurable $f\colon \Rr^d\to\Rr^\delta$ such that it holds $\P$-a.s.\ that
    \begin{equation}\llabel{conclude}
        \E[Y|X]=f(X).
    \end{equation}
\end{athm}
\begin{aproof}
    \argument{the assumption that it holds $\P$-a.s.\ that $\E[\|Y\| |X]<\infty$ }{that there exists a $\sigma(X)/\mathcal B(\Rr^\delta)$-measurable $Z\colon \Omega\to \Rr^{\delta}$ which satisfies $\P$-a.s.\ that
    \begin{equation}\llabel{NR}
        \E[Y|X]=Z
    \end{equation}
    (cf., \eg,\ \cite[Item (i) in Lemma 3.11 and item (i) in Lemma 3.12]{Dereichnonconvergence2024})\dott}
    \startnewargseq
    \argument{\cref{Lemma YY'} (applied with $D\curvearrowleft \Rr^\delta$ in the notation of \cref{Lemma YY'});the fact that $Z$ is $\sigma(X)/\mathcal B(\Rr^\delta)$-measurable}{that there exists a measurable $f\colon \Rr^d\to\Rr^\delta$ which satisfies
    \begin{equation}\llabel{NR1}
        Z=f\circ X\dott
    \end{equation}}
    \startnewargseq
    \argument{\lref{NR};\lref{NR1}}{\lref{conclude}\dott}
\end{aproof}
\begin{athm}{cor}{Lemma WW}
    Let $(\Omega,\mathcal F,\P)$ be a probability space, let $d,\delta\in \N$, let $A\subseteq\Rr^d$ be a set, and let $X=(X_1,\dots,X_d)\colon \Omega\to A$ and $Y=(Y_1,\dots,Y_\delta)\colon \Omega\to \Rr^\delta$ be random variables which satisfy $\E[\|Y\|]<\infty$. Then there exists a measurable $f\colon A\to\Rr^\delta$ such that it holds $\P$-a.s.\ that
    \begin{equation}\llabel{conclude}
        \E[Y|X]=f(X).
    \end{equation}
\end{athm}
\begin{aproof}
    Throughout this proof let $\bfX\colon\Omega\to \Rr^d$ satisfy for all $\omega\in \Omega$ that
    \begin{equation}\llabel{def: bfX}
        \bfX(\omega)=X(\omega).
    \end{equation}
    \argument{\lref{def: bfX}; the fact that $X\colon \Omega\to A$ is measurable}{that for all $B\in \mathcal B(\Rr^d)$ it holds that
    \begin{equation}\llabel{NR0}
        \bfX^{-1}(B)=\bfX^{-1}(B\cap A)=X^{-1}(B\cap A)\in \mathcal F\dott
    \end{equation}}
    \argument{\lref{NR0};}{that \llabel{arg1} $\bfX$ is measurable\dott}
    \argument{\lref{arg1};\cref{Lemma VV} (applied with $\Omega\curvearrowleft\Omega$, $\mathcal{F}\curvearrowleft\mathcal{F}$, $\P\curvearrowleft\P$, $d\curvearrowleft d$, $\delta\curvearrowleft\delta$, $X\curvearrowleft\bfX$, $Y\curvearrowleft Y$ in the notation of \cref{Lemma VV});\lref{def: bfX}}{that there exists $F\colon \Rr^d\to\Rr^{\delta}$ such that it holds $\P$-a.s.\ that
    \begin{equation}\llabel{NR1}
        \E[Y|\bfX]=F(\bfX)\dott
    \end{equation}}
    In the following let $f\colon A\to \Rr^\delta$ satisfy for all $x\in A$ that
    \begin{equation}\llabel{NR2}
        f(x)=F(x).
    \end{equation}
    \startnewargseq
    \argument{\lref{NR2};}{that for all $B\in \mathcal B(\Rr^\delta)$ it holds that
    \begin{equation}\llabel{eq2}
    \begin{split}
        f^{-1}(B)&=\{x\in A\colon f(x)\in B\}=\{x\in A\colon F(x)\in B\}\\
        &=\{x\in \Rr^d\colon F(x)\in B\}\cap A=([F^{-1}(B)]\cap A)\in \mathcal B(A)\dott
        \end{split}
    \end{equation}}
    \argument{\lref{eq2};}{that $f$ is measurable\dott}
    \argument{\lref{NR0};}{that \llabel{arg3} $\sigma(\bfX)=\cup_{B\in \mathcal B(\Rr^d)}\{\bfX^{-1}(B)\}=\cup_{B\in \mathcal B(\Rr^d)}\{X^{-1}(B\cap A)\}=\sigma(X)$\dott}
    \argument{\lref{arg3};\lref{def: bfX};\lref{NR1};\lref{NR2}}{that it holds $\P$-a.s.\ that
    \begin{equation}\llabel{eq3}
        \begin{split}
\E[Y|X]=\E[Y|\sigma(X)]=\E[Y|\sigma(\bfX)]=\E[Y|\bfX]=F(\bfX)=f(X)\dott
        \end{split}
    \end{equation}}
    \argument{\lref{eq3};the fact that $f$ is measurable}{\lref{conclude}\dott}
\end{aproof}
\subsection{Supervised learning risk decomposition}
\begin{athm}{lemma}{Lemma ZZ}
    Assume \cref{setting: explicit}. Then there exists a measurable $f\colon [a,b]^{\ell_0}\to\Rr^{\ell_L}$ such that
    \begin{enumerate}[label=(\roman*)]
        \item \llabel{item 1} it holds that $\int_{[a,b]^{\ell_0}\times \Rr^{\ell_L}}\|f(x)\|^2\,\mu(\d x,\d y)<\infty$,
        \item \llabel{item 1.5} it holds for all $A\in\mathcal B([a,b]^{\ell_0})$ that
        \begin{equation}
            \int_{A\times \Rr^{\ell_L}} (f(x)-y)\,\mu(\d x,\d y)=0,
        \end{equation}
        \item \llabel{item 2} it holds for all measurable $g\colon[a,b]^{\ell_0}\to \Rr^{\ell_L}$ that
        \begin{equation}
            \int_{[a,b]^{\ell_0}\times \Rr^{\ell_L}}\|g(x)-y\|^2\, \mu(\d x,\d y)=\int_{[a,b]^{\ell_0}\times \Rr^{\ell_L}}\|g(x)-f(x)\|^2+\|f(x)-y\|^2\,\mu(\d x,\d y),
        \end{equation}
        and
        \item \llabel{item 3} it holds for all $r\in [1,\infty]$, $\theta\in \Rr^{\fd}$ that
        \begin{equation}
            \cL_r(\theta)=\int_{[a,b]^{ \ell_0 }\times \Rr^{\ell_L} }
  \|\mN_r^{L, \theta }(x)-f(x)\|^2+\|y-f(x)\|^2\, \mu( \d x,\d y).
        \end{equation}
    \end{enumerate}
    \end{athm}
    \renewcommand{\R}{\mathbb{R}}
\begin{aproof}
    Throughout this proof assume without loss of generality that $\mu([a,b]^{\ell_0}\times\R^{\ell_L})>0$, let $\Omega=[a,b]^{\ell_0}\times\Rr^{\ell_L}$, let $\mathcal F=\mathcal B(\Omega)$, let $\P\colon \mathcal F\to [0,1]$ satisfy for all $A\in \mathcal F$ that 
    \begin{equation}\llabel{NRZ}
       \P(A)=\frac{\mu(A)}{\mu(\Omega)}
    \end{equation}
    (cf.\ \cref{Lemma XX}),
     and let $X\colon \Omega\to [a,b]^{\ell_0}$ and $Y\colon \Omega\to \Rr^{\ell_L}$ satisfy for all $x\in [a,b]^{\ell_0}$, $y\in \Rr^{\ell_L}$ that
    \begin{equation}\llabel{eq1}
        X(x,y)=x\qqandqq Y(x,y)=y.
    \end{equation}
    \argument{\lref{NRZ};\lref{eq1}}{that for all measurable $\varphi\colon\Omega\to [0,\infty)$ it holds that
    \begin{equation}\llabel{eq2}
    \begin{split}
\E[\varphi(X,Y)]&=\int_{\Omega}\varphi(X(\omega),Y(\omega))\,\P(\d \omega)=\frac{1}{\mu(\Omega)}\int_{\Omega}\varphi(X(\omega),Y(\omega))\,\mu(\d\omega)\\
&=\frac{1}{\mu(\Omega)}\int_{\Omega}\varphi(x,y)\,\mu(\d x,\d y)\dott
\end{split}
    \end{equation}}
    \argument{\lref{eq2};\cref{Lemma XX}}{that
    \begin{equation}\llabel{NR: Integrability}
        \E[\|X\|^2+\|Y\|^2]=\frac{1}{\mu(\Omega)}\int_{\Omega}\|x\|^2+\|y\|^2\, \mu(\d x,\d y)<\infty\dott
    \end{equation}}
    \argument{the factorization lemma in \cref{Lemma WW};\lref{NR: Integrability}}{that there exists a measurable $f\colon [a,b]^{\ell_0}\allowbreak\to\R^{\ell_L}$ which satisfies $\P$-a.s.\ that
    \begin{equation}\llabel{eq3}
        \E[Y|X]=f(X)\dott
    \end{equation}}
    \argument{\lref{eq3};\lref{eq2};\lref{NR: Integrability}}{\lref{item 1, item 1.5}\dott}
    \startnewargseq
    \argument{\lref{eq2};\lref{NR: Integrability}; \lref{eq3}}{that for all measurable $g\colon [a,b]^{\ell_0}\to\Rr^{\ell_L}$ it holds that
    \begin{equation}\llabel{eq4}
        \begin{split}
            &\frac{1}{\mu(\Omega)}\int_{[a,b]^{\ell_0}\times\Rr^{\ell_L}}\|g(x)-y\|^2\,\mu(\d x,\d y)\\
            &=\E\bigl[\|g(X)-Y\|^2\bigl]=\E\bigl[\|(g(X)-f(X))+(f(X)-Y)\|^2\bigl]\\
            &=\E\bigl[g(X)-f(X)\|^2+2\spro{g(X)-f(X),f(X)-Y}+\|f(X)-Y\|^2\bigl]\\
            &=\E\bigl[\|g(X)-f(X)\|^2\bigl]+2\E\bigl[\spro{g(X)-f(X),f(X)-Y}\bigl]+\E\bigl[\|f(X)-Y\|^2\bigl]\\
            &=\E\bigl[\|g(X)-f(X)\|^2+\|f(X)-Y\|^2\bigl]+2\E\bigl[\E[\spro{g(X)-f(X),f(X)-Y}|X]\bigl]\\
            &=\E\bigl[\|g(X)-f(X)\|^2+\|f(X)-Y\|^2\bigl]+2\E\bigl[\spro{g(X)-f(X),\E[f(X)-Y|X]}\bigl]\\
            &=\E\bigl[\|g(X)-f(X)\|^2+\|f(X)-Y\|^2\bigl]+2\E\bigl[\spro{g(X)-f(X),f(X)-\E[Y|X]}\bigl]\\
            &=\E\bigl[\|g(X)-f(X)\|^2+\|f(X)-Y\|^2\bigl]\\
            &=\frac{1}{\mu(\Omega)}\int_{[a,b]^{\ell_0}\times\Rr^{\ell_L}}\|g(x)-f(x)\|^2+\|f(x)-y\|^2\,\mu(\d x,\d y)\dott
        \end{split}
    \end{equation}}
    \argument{\lref{eq4};}[verbs=ep]{\lref{item 2}\dott}
    \startnewargseq
    \argument{\lref{item 2};\cref{L_r}}[verbs=ep]{\lref{item 3}\dott}
\end{aproof}
\begin{athm}{lemma}{Lemma ZZZ}
    Assume \cref{setting: explicit}, for every $k\in \{1,2\}$ let $f_k\colon [a,b]^{\ell_0}\to \Rr^{\ell_L}$ be measurable, and assume for all $k\in\{1,2\}$, $A\in \mathcal B([a,b]^{\ell_0})$ that $\int_{A\times\Rr^{\ell_L}}\|f_k(x)\|\, \mu(\d x,\d y)<\infty$ and
    \begin{equation}\llabel{eq1}
        \int_{A\times \R^{\ell_L}}(f_k(x)-y)\,\mu(\d x,\d y)=0.
    \end{equation}
    Then \begin{equation}\llabel{conclude}
       \mu(\{(x,y)\in [a,b]^{\ell_0}\times\R^{\ell_L}\colon f_1(x)\neq f_2(x)\})=0.
    \end{equation}
\end{athm}
\begin{aproof}
  Throughout this proof assume without loss of generality that $\mu([a,b]^{\ell_0}\times\R^{\ell_L})>0$, let $\Omega=[a,b]^{\ell_0}\times\Rr^{\ell_L}$, let $\mathcal F=\mathcal B(\Omega)$, let $\mathcal H=\mathcal B([a,b]^{\ell_0})\otimes \{\emptyset,\Rr^{\ell_L}\}$, let $\P\colon \mathcal F\to [0,1]$ satisfy for all $A\in \mathcal F$ that 
    \begin{equation}\llabel{NRZ}
       \P(A)=\frac{\mu(A)}{\mu(\Omega)}
    \end{equation}
    (cf.\ \cref{Lemma XX}), and let $X\colon \Omega\to [a,b]^{\ell_0}$ and $Y\colon \Omega\to \Rr^{\ell_L}$ satisfy for all $x\in [a,b]^{\ell_0}$, $y\in \Rr^{\ell_L}$ that
    \begin{equation}\llabel{def: XY}
        X(x,y)=x\qqandqq Y(x,y)=y.
    \end{equation}
    \argument{\lref{eq1};\lref{NRZ};\lref{def: XY}}{that for all $k\in \{1,2\}$, $A\in \mathcal B([a,b]^{\ell_0})$ it holds that 
    \begin{equation}\llabel{eq3}
\begin{split}
\E\bigl[(Y-f_k(X))\mathbbm 1_{A\times\Rr^{\ell_L}}\bigr]&=\int_{\Omega} [Y(\omega)-f_k(X(\omega))]\mathbbm 1_{A\times\Rr^{\ell_L}}(\omega)\,\P(\d \omega)\\
&=\frac{1}{\mu(\Omega)}\int_{\Omega}[Y(\omega)-f_k(X(\omega))]\mathbbm 1_{A\times\Rr^{\ell_L}}(\omega)\,\mu(\d\omega)\\
&=\frac{1}{\mu(\Omega)}\int_{\Omega}(y-f_k(x))\mathbbm 1_{A\times\Rr^{\ell_L}}(x,y)\,\mu(\d x,\d y)\dott\\
&=\frac{1}{\mu(\Omega)}\int_{A\times\Rr^{\ell_L}}(y-f_k(x))\,\mu(\d x,\d y)=0\dott
\end{split}
    \end{equation}}
    \argument{\lref{def: XY};}{for all $A\in \mathcal B([a,b]^{\ell_0}$ that
    \begin{equation}\llabel{eq4}
    \begin{split}
        \{(x,y)\in [a,b]^{\ell_0}\times \Rr^{\ell_L}\colon X(x,y)\in A\}&=\{(x,y)\in [a,b]^{\ell_0}\times \Rr^{\ell_L}\colon x\in A\}\\
        &=\{x\in [a,b]^{\ell_0}\colon x\in A\}\times \Rr^{\ell_L}\in \mathcal H\dott
        \end{split}
    \end{equation}}
    \argument{\lref{eq3};\lref{eq4}}{that \llabel{arg1} for all $k\in \{1,2\}$ it holds a.s.\ that $\E[Y|\mathcal H]=f_k(X)$\dott}
    \argument{\lref{arg1};\cite[Proposition 3.9]{Dereichnonconvergence2024}}{that \llabel{arg2} it holds $\P$-a.s.\ that $f_1(X)=f_2(X)$\dott}
    \argument{\lref{arg2};\lref{NRZ};\lref{def: XY}}{\lref{conclude}\dott}
\end{aproof}
\renewcommand{\R}{\mathbb{A}}
\subsection{Explicit representations for generalized gradients of the risk functions}\label{subsec: explicit representations for generalized gradient}
\begin{athm}{prop}{theo: explicit formula1}
    Assume \cref{setting: explicit}, 
for every $ k \in \enne_0 $ 
let $ \diml_k \in \enne_0 $ satisfy 
$ \diml_k = \sum_{ n = 1 }^k \ell_n ( \ell_{ n - 1 } + 1 ) $,
 let $ \theta  \in \Rr^{ \fd } $, let $f=(f_1,\dots,f_{\ell_L})\colon [a,b]^{\ell_0}\to\Rr^{\ell_L}$ be measurable, and assume for all $A\in \mathcal B([a,b]^{\ell_0})$ that $\int_{A\times \Rr^{\ell_L}}\|f(x)\|^2\allowbreak\,\mu(\d x,\d y)<\infty$ and  $\int_{A\times \Rr^{\ell_L}}(f(x)-y)\,\mu(\d x,\d y)=0$. Then
\begin{enumerate}[label=(\roman*)]
\item \llabel{item 1}
it holds for all $ r \in [1,\infty) $ 
that $ \cL_r \in C^1( \Reals^{ \fd }, \Reals ) $,
\item \llabel{item 2}
it holds for all $ r \in [1,\infty) $, $ k \in \{ 1,2, \ldots, L \} $, 
$ i \in \{ 1,2, \ldots, \ell_k \} $,
$ j \in \{ 1,2, \ldots, \ell_{k-1} \} $ that
\begin{equation}
\begin{split}
& 
  \pr*{  \frac{\partial\cL_r}
   {\partial\theta_{ (i-1)\ell_{k-1} + j +\diml_{k-1} } }
    } ( \theta)
\\
&
  =
  \sum_{\substack{v_k,v_{k+1},  
  \ldots,v_L\in \enne, \\\forall w\in \enne
  \cap[k, L]\colon v_w\leq\ell_w} }
  \int_{[a,b]^{ \ell_0 }\times\Rr^{\ell_L} }2\,
  \Big[
  \R_{r^{1/ ( \max \cu{k - 1 , 1 } ) } } ( \mN^{ \max\{k-1,1\}, \theta }_{r,j}(x))
  \indicator{ (1, L]}(k)
  +x_j \indicator{\{1\} }(k)
  \Big]\\
  &\quad\cdot
  \Big[ \indicator{\{ i \} }(v_k)\Big]
  \Big[ \mN_{r,v_L}^{L, \theta }(x)-f_{v_L}(x)\Big] 
  \Big[
  \textstyle{\prod}_{n={k+1} }^{L} \big(
  \fw^{n, \theta }_{v_n, v_{n-1} }
  \big[( \R_{r^{1/(n - 1 ) } } ) ' ( \mN^{n-1, \theta }_{r,v_{n-1} }(x))\big]
  \big)
  \Big]\, \mu( \d x,\d y),
\end{split}
\end{equation}
\item \llabel{item 3}
   it holds for all $ r\in [1,\infty)  $, $ k \in \{ 1,2, \ldots, L\} $, 
  $ i \in \{ 1,2, \ldots, \ell_k \} $ that
  \begin{equation}
  \begin{split}
  & \pr*{  \frac{\partial\cL_r}
  {\partial\theta_{\ell_k\ell_{k-1} + i+ \diml_{k-1} } }
   } ( \theta)
  =
  \sum_{\substack{v_k,v_{k+1},  
  \ldots,v_L\in \enne, \\\forall w\in \enne
  \cap[k, L]\colon v_w\leq\ell_w} }
  \int_{[a,b]^{ \ell_0 }\times\Rr^{\ell_L}}2\,
  \Big[ \indicator{\{ i \} }(v_k)\Big]
  \\
  &\quad\cdot
  \Big[ \mN_{r,v_L}^{L, \theta }(x)-f_{v_L}(x)\Big]
  \Big[
  \textstyle{\prod}_{n={k+1} }^{L} \big(
  \fw^{n, \theta }_{v_n, v_{n-1} }
  \big[( \R_{r^{1 / ( n - 1 ) } } )'( \mN^{n-1, \theta }_{r,v_{n-1} }(x))\big]
  \big)
  \Big]\, \mu( \d x,\d y),
\end{split}
\end{equation}
\item \llabel{item 4}
it holds that $ \limsup\nolimits_{r\to \infty }  \pr*{  \abs{ \cL_r( \theta)-\cL_{ \infty }( \theta) } +
 \|( \nabla\cL_r)( \theta)-\cG( \theta)\|  }  =0$,
 \item \llabel{item 5}
 it holds for all $ k \in \{ 1,2, \ldots, L\} $,
  $ i \in \{ 1,2, \ldots, \ell_k \} $, 
  $ j \in \{ 1,2, \ldots, \ell_{k-1} \} $ that
  \begin{equation}
  \begin{split}
  &\cG_{ (i-1)\ell_{k-1} + j + \diml_{k - 1 } }( \theta)\\
  &=
  \sum_{\substack{v_k,v_{k+1},  
  \ldots,v_L\in \enne, \\\forall w\in \enne
  \cap[k, L]\colon v_w\leq\ell_w} }
  \int_{[a,b]^{ \ell_0 } \times\Rr^{\ell_L}}2\,
  \Big[
  \R_{ \infty }( \mN^{ \max\{k-1,1\}, \theta }_{ \infty,j}(x))
  \indicator{ (1, L]}(k)
  +x_j \indicator{\{1\} }(k)
  \Big]\\
  &\quad\cdot
  \Big[ \indicator{\{ i \} }(v_k)\Big]
  \Big[ \mN_{ \infty,v_L}^{L, \theta }(x)-f_{v_L}(x)\Big] 
  \Big[
  \textstyle{\prod}_{n={k+1} }^{L} \big(
  \fw^{n, \theta }_{v_n, v_{n-1} }
   \mathbbm 1_{(0,\infty)}(\neurali{\infty}{v_{n-1}}{n-1}{\theta}(x))
  \big)
  \Big]\, \mu( \d x,\d y),
\end{split}
\end{equation}
and
 \item\llabel{item 6}
 it holds for all $ k \in \{ 1,2, \ldots, L\} $,
  $ i \in \{ 1,2, \ldots, \ell_k \} $ that
  \begin{equation}
  \begin{split}
    &\cG_{\ell_k\ell_{k-1} + i+ \diml_{k-1 } }( \theta)
    =\sum_{\substack{v_k,v_{k+1},  
  \ldots,v_L\in \enne, \\\forall w\in \enne
  \cap[k, L]\colon v_w\leq\ell_w} }
  \int_{[a,b]^{ \ell_0 } \times\Rr^{\ell_L}}2\,
  \Big[ \indicator{\{ i \} }(v_k)\Big]
   \\
  &\quad\cdot
  \Big[ \mN_{ \infty,v_L}^{L, \theta }(x)-f_{v_L}(x)\Big]
  \Big[
  \textstyle{\prod}_{n={k+1} }^{L} \big(
  \fw^{n, \theta }_{v_n, v_{n-1} }
 \mathbbm 1_{(0,\infty)}(\neurali{\infty}{v_{n-1}}{n-1}{\theta}(x))
  \big)
  \Big]\, \mu( \d x,\d y).
\end{split}
\end{equation}
\end{enumerate}
\end{athm}

\begin{aproof}
Throughout this proof for every $r\in [1,\infty]$ let $\fL_r\colon \Rr^{\fd}\to \Rr$ satisfy for all $\theta\in \Rr^\fd$ that
\begin{equation}\llabel{def: L}
\fL_r(\theta)=\int_{[a,b]^{\ell_0}\times\Rr^{\ell_L}}\|\mathcal N_r^{L,\theta}(x)-f(x)\|^2\,\mu
(\d x,\d y),
\end{equation}
let $\lambda\colon \mathcal B([a,b]^{\ell_0})\to [0,\infty]$ satisfy for all $A\in \mathcal B([a,b]^d)$ that 
\begin{equation}\llabel{def: lambda}
    \lambda(A)=\mu(A\times\Rr^{\ell_L}),
\end{equation}
and let $ \mathfrak G = ( \mathfrak G_1, \dots, \mathfrak G_{ \fd } ) \colon \Reals^{ \fd } \to \Reals^{ \fd } $ 
satisfy for all $ \vartheta \in \{\varTheta \in \Reals^{ \fd }\colon(( \nabla\fL_r)( \varTheta))_{ r \in [1,\infty)}
\text{ is convergent} \} $ that
$ \mathcal G( \vartheta)=\lim_{r\to \infty }( \nabla \fL_r)( \vartheta) $.
\argument{\lref{def: lambda};}{that for all measurable $g\colon[a,b]^{\ell_0}\to\Rr^{
\ell_L}$ with 
$\min_{z\in \{-1,1\}}\int_{[a,b]^{\ell_0}}\max\{zg(x),0\}\,\lambda(\d x)<\infty$ it holds that 
\begin{equation}\llabel{NR}
\int_{[a,b]^{\ell_0}\times\Rr^{\ell_L}}g(x)\,\mu(\d x,\d y)=\int_{[a,b]^{\ell_0}}g(x)\,\lambda(\d x)\dott
\end{equation}}
\argument{\lref{NR};\lref{def: L}
}
{for all $r\in [1,\infty]$, $\theta\in \Rr^{\fd}$ that
\begin{equation}\llabel{eq1}
    \displaystyle
  \fL_r( \theta)=\int_{[a,b]^{ \ell_0 } \times \Rr^{\ell_L}}
  \|\mN_r^{L, \theta }(x)-f(x)\|^2\, \lambda( \d x).
\end{equation}}
   \argument{\lref{eq1};\cite[Theorem 2.9]{MaArconvergenceproof} (applied with 
 $L\curvearrowleft L$, $\fd\curvearrowleft\fd$, $\ell\curvearrowleft\ell$, $a\curvearrowleft a$, $b\curvearrowleft b$,  $\scrA\curvearrowleft\scrA$, $\scrB\curvearrowleft\scrB$,  $\fw\curvearrowleft\fw$, $\fb\curvearrowleft\fb$, 
$\scrR\curvearrowleft \mathbb A$,
$\mN\curvearrowleft\mN$, 
$f\curvearrowleft f$,
 $\mu\curvearrowleft \lambda$, $\cL\curvearrowleft\cL$,
$\mathcal G\curvearrowleft \mathfrak G$ in the notation of \cite[Theorem 2.9]{MaArconvergenceproof})}{that
\begin{enumerate}[label=(\Roman*)]
\item \llabel{Item 1} it holds for all $ r \in [1,\infty) $ 
that $ \fL_r \in C^1( \Reals^{ \fd }, \Reals ) $,
\item \llabel{Item 2}
it holds for all $ r \in [1,\infty) $, $ k \in \{ 1,2,\ldots, L \} $, 
$ i \in \{ 1,2, \ldots, \ell_k \} $,
$ j \in \{ 1,2, \ldots, \ell_{k-1} \} $ that
\begin{equation}
\begin{split}
& 
  \pr*{  \frac{\partial\fL_r}
   {\partial\theta_{ (i-1)\ell_{k-1} + j +\diml_{k-1} } }
    } ( \theta)
\\
&
  =
  \sum_{\substack{v_k,v_{k+1},  
  \ldots,v_L\in \enne, \\\forall w\in \enne
  \cap[k, L]\colon v_w\leq\ell_w} }
  \int_{[a,b]^{ \ell_0 } }2\,
  \Big[
  \R_{r^{1/ ( \max \cu{k - 1 , 1 } ) } } ( \mN^{ \max\{k-1,1\}, \theta }_{r,j}(x))
  \indicator{ (1, L]}(k)
  +x_j \indicator{\{1\} }(k)
  \Big]\\
  &\quad\cdot
  \Big[ \indicator{\{ i \} }(v_k)\Big]
  \Big[ \mN_{r,v_L}^{L, \theta }(x)-f_{v_L}(x)\Big] 
  \Big[
  \textstyle{\prod}_{n={k+1} }^{L} \big(
  \fw^{n, \theta }_{v_n, v_{n-1} }
  \big[( \R_{r^{1/(n - 1 ) } } ) ' ( \mN^{n-1, \theta }_{r,v_{n-1} }(x))\big]
  \big)
  \Big]\, \lambda( \d x),
\end{split}
\end{equation}
\item \llabel{Item 3}
   it holds for all $ r\in [1,\infty)  $, $ k \in \{ 1,2, \ldots, L\} $, 
  $ i \in \{ 1,2, \ldots, \ell_k \} $ that
  \begin{equation}
  \begin{split}
  & \pr*{  \frac{\partial\fL_r}
  {\partial\theta_{\ell_k\ell_{k-1} + i+ \diml_{k-1} } }
   } ( \theta)
  =
  \sum_{\substack{v_k,v_{k+1},  
  \ldots,v_L\in \enne, \\\forall w\in \enne
  \cap[k, L]\colon v_w\leq\ell_w} }
  \int_{[a,b]^{ \ell_0 }}2\,
  \Big[ \indicator{\{ i \} }(v_k)\Big]
  \\
  &\quad\cdot
  \Big[ \mN_{r,v_L}^{L, \theta }(x)-f_{v_L}(x)\Big]
  \Big[
  \textstyle{\prod}_{n={k+1} }^{L} \big(
  \fw^{n, \theta }_{v_n, v_{n-1} }
  \big[( \R_{r^{1 / ( n - 1 ) } } )'( \mN^{n-1, \theta }_{r,v_{n-1} }(x))\big]
  \big)
  \Big]\, \lambda( \d x),
\end{split}
\end{equation}
\item \llabel{Item 4}
it holds that $ \limsup\nolimits_{r\to \infty }  \pr*{  \abs{ \fL_r( \theta)-\fL_{ \infty }( \theta) } +
 \|( \nabla\fL_r)( \theta)-\mathcal G( \theta)\|  }  =0$,
 \item \llabel{Item 5}
 it holds for all $ k \in \{ 1,2, \ldots, L\} $,
  $ i \in \{ 1,2, \ldots, \ell_k \} $, 
  $ j \in \{ 1,2, \ldots, \ell_{k-1} \} $ that
  \begin{equation}
  \begin{split}
  &\mathfrak G_{ (i-1)\ell_{k-1} + j + \diml_{k - 1 } }( \theta)\\
  &=
  \sum_{\substack{v_k,v_{k+1},  
  \ldots,v_L\in \enne, \\\forall w\in \enne
  \cap[k, L]\colon v_w\leq\ell_w} }
  \int_{[a,b]^{ \ell_0 }}2\,
  \Big[
  \R_{ \infty }( \mN^{ \max\{k-1,1\}, \theta }_{ \infty,j}(x))
  \indicator{ (1, L]}(k)
  +x_j \indicator{\{1\} }(k)
  \Big]\\
  &\quad\cdot
  \Big[ \indicator{\{ i \} }(v_k)\Big]
  \Big[ \mN_{ \infty,v_L}^{L, \theta }(x)-f_{v_L}(x)\Big] 
  \Big[
  \textstyle{\prod}_{n={k+1} }^{L} \big(
  \fw^{n, \theta }_{v_n, v_{n-1} }
   \mathbbm 1_{(0,\infty)}(\neurali{\infty}{v_{n-1}}{n-1}{\theta}(x))
  \big)
  \Big]\, \lambda( \d x),
\end{split}
\end{equation}
and
 \item\llabel{Item 6}
 it holds for all $ k \in \{ 1,2, \ldots, L\} $,
  $ i \in \{ 1,2, \ldots, \ell_k \} $ that
  \begin{equation}
  \begin{split}
    &\mathfrak G_{\ell_k\ell_{k-1} + i+ \diml_{k-1 } }( \theta)
    =\sum_{\substack{v_k,v_{k+1},  
  \ldots,v_L\in \enne, \\\forall w\in \enne
  \cap[k, L]\colon v_w\leq\ell_w} }
  \int_{[a,b]^{ \ell_0 } }2\,
  \Big[ \indicator{\{ i \} }(v_k)\Big]
   \\
  &\quad\cdot
  \Big[ \mN_{ \infty,v_L}^{L, \theta }(x)-f_{v_L}(x)\Big]
  \Big[
  \textstyle{\prod}_{n={k+1} }^{L} \big(
  \fw^{n, \theta }_{v_n, v_{n-1} }
   \mathbbm 1_{(0,\infty)}(\neurali{\infty}{v_{n-1}}{n-1}{\theta}(x))
  \big)
  \Big]\, \lambda( \d x).
\end{split}
\end{equation}
\end{enumerate}
}
\argument{\cref{Lemma ZZ};}{that there exists a measurable $\bff\colon [a,b]^{\ell_0}\to\Rr^{\ell_L}$ which satisfies that
    \begin{enumerate}[label=(\Alph*)]
        \item \llabel{Ittem 1} it holds that $\int_{[a,b]^{\ell_0}\times \Rr^{\ell_L}}\|\bff(x)\|^2\mu(\d x,\d y)<\infty$,
        \item \llabel{Ittem 1.5} it holds for all $A\in\mathcal B([a,b]^{\ell_0})$ that
        \begin{equation}
            \int_{A\times \Rr^{\ell_L}} (\bff(x)-y)\,\mu(\d x,\d y)=0,
        \end{equation}
        and
        \item \llabel{Ittem 3} it holds for all $r\in [1,\infty]$, $\theta\in \Rr^{\fd}$ that
        \begin{equation}
            \cL_r(\theta)=\int_{[a,b]^{ \ell_0 }\times \Rr^{\ell_L} }
  \|\mN_r^{L, \theta }(x)-\bff(x)\|^2+\|y-\bff(x)\|^2\, \mu( \d x,\d y).
        \end{equation}
    \end{enumerate}}
    \startnewargseq
\argument{\cref{Lemma ZZZ}; the assumption that for all $A\in \mathcal B([a,b]^{\ell_0})$ it holds  that $\int_{A\times \Rr^{\ell_L}}\|f(x)\|\allowbreak\,\mu(\d x,\d y)<\infty$ and  $\int_{A\times \Rr^{\ell_L}}(f(x)-y)\,\mu(\d x,\d y)=0$;\lref{Ittem 1};\lref{Ittem 1.5}} 
{that \llabel{arrg1} $f=\bff$\dott}
\argument{\lref{arrg1};\lref{Ittem 3}}{that for all $r\in [1,\infty]$, $\theta\in \Rr^{\fd}$ it holds  that
        \begin{equation}\llabel{EQ1}
            \cL_r(\theta)=\int_{[a,b]^{ \ell_0 }\times \Rr^{\ell_L} }
  \|\mN_r^{L, \theta }(x)-f(x)\|^2+\|y-f(x)\|^2\, \mu( \d x,\d y).
        \end{equation}}
 \argument{\lref{def: L};\lref{EQ1}}{that for all $r\in [1,\infty)$, $\vartheta\in \Rr^{\fd}$ it holds that
 \begin{equation}\llabel{key}
     (\nabla \cL_r)(\vartheta)=(\nabla\fL_r)(\vartheta)\dott
 \end{equation}}
 \argument{\lref{key}; the fact that for all $ \vartheta \in \{\varTheta \in \Reals^{ \fd }\colon(( \nabla\fL_r)( \varTheta))_{ r \in [1,\infty)}
\text{ is convergent} \} $ it holds that
$ \mathfrak G( \vartheta)=\lim_{r\to \infty }\allowbreak( \nabla \fL_r)( \vartheta) $; the fact that for all $ \vartheta \in \{\varTheta \in \Reals^{ \fd }\colon(( \nabla\cL_r)( \varTheta))_{ r \in [1,\infty) }
\text{ is convergent} \} $ it holds that
$ \mG( \vartheta)=\lim_{r\to \infty }( \nabla \cL_r)( \vartheta) $;}{that for all $\vartheta\in \Rr^\fd$ it holds that
\begin{equation}\llabel{key2}
    \mathfrak G(\vartheta)=\mG(\vartheta)\dott
\end{equation}}
\argument{ \lref{NR};\lref{key};\lref{key2};\lref{Item 1}; \lref{Item 2}; \lref{Item 3};\lref{Item 4};\lref{Item 5};\lref{Item 6}; \cref{Lemma XX}}{\lref{item 1,item 2,item 3, item 4, item 5, item 6}\dott}
\end{aproof}
\begin{athm}{lemma}{Lemma ZZZZ}
    Assume \cref{setting: explicit}, let $v\in \Rr^{\ell_L}$, let $f\colon [a,b]^{\ell_0}\to\Rr^{\ell_L}$, $g\colon[a,b]^{\ell_0}\to\Rr$, and $h\colon[a,b]^{\ell_0}\to\Rr$ be measurable, assume $\int_{[a,b]^{\ell_0}\times\Rr^{\ell_L}}(\|f(x)\|^2+|g(x)|^2+|h(x)|^2)\,\mu(\d x,\d y)<\infty$, and assume for all  $A\in\mathcal B([a,b]^{\ell_0})$ that
$
            \int_{A\times \Rr^{\ell_L}} (f(x)-y)\,\mu(\d x,\d y)=0$ (cf.\ \cref{Lemma XX}). Then
    \begin{equation}\llabel{conclude}
        \int_{[a,b]^{\ell_0}\times\Rr^{\ell_L}}\bigl[g(x)-\spro{v,f(x)}\bigr]h(x)\,\mu(\d x, \d y)=\int_{[a,b]^{\ell_0}\times\Rr^{\ell_L}}\bigl[g(x)-\spro{v,y}\bigr]h(x)\,\mu(\d x,\d y).
    \end{equation}
\end{athm}
\begin{aproof}
    Throughout this proof assume without loss of generality that $\mu([a,b]^{\ell_0}\times\Rr^{\ell_L})>0$, let $\Omega=[a,b]^{\ell_0}\times\Rr^{\ell_L}$, let $\mathcal F=\mathcal B(\Omega)$, let $\P\colon \mathcal F\to [0,1]$ satisfy for all $A\in \mathcal F$ that 
    \begin{equation}\llabel{NRZ}
       \P(A)=\frac{\mu(A)}{\mu(\Omega)}
    \end{equation}
    (cf.\ \cref{Lemma XX}),
     and let $X\colon \Omega\to [a,b]^{\ell_0}$ and $Y\colon \Omega\to \Rr^{\ell_L}$ satisfy for all $x\in [a,b]^{\ell_0}$, $y\in \Rr^{\ell_L}$ that
    \begin{equation}\llabel{eq1}
        X(x,y)=x\qqandqq Y(x,y)=y.
    \end{equation}
    \argument{\lref{NRZ};\lref{eq1}}{that for all measurable $\varphi\colon\Omega\to [0,\infty)$ it holds that
    \begin{equation}\llabel{eq2}
    \begin{split}
\E[\varphi(X,Y)]&=\int_{\Omega}\varphi(X(\omega),Y(\omega))\,\P(\d \omega)=\frac{1}{\mu(\Omega)}\int_{\Omega}\varphi(X(\omega),Y(\omega))\,\mu(\d\omega)\\
&=\frac{1}{\mu(\Omega)}\int_{\Omega}\varphi(x,y)\,\mu(\d x,\d y)\dott
\end{split}
    \end{equation}}
    \argument{\lref{eq2};\cref{Lemma XX}}{that
    \begin{equation}\llabel{NR: Integrability}
        \E[\|X\|^2+\|Y\|^2]=\frac{1}{\mu(\Omega)}\int_{\Omega}\|x\|^2+\|y\|^2\, \mu(\d x,\d y)<\infty\dott
    \end{equation}}
    \argument{\lref{eq2};the fact that for all  $A\in\mathcal B([a,b]^{\ell_0})$ it holds that
$
            \int_{A\times \Rr^{\ell_L}} (f(x)-y)\,\mu(\d x,\d y)=0$}{that for all $A\in 
            \mathcal B([a,b]^{\ell_0})$ it holds that
            \begin{equation}\llabel{eq2.5}
            \begin{split}
                &\E[Y\mathbbm 1_{A\times\Rr^{\ell_L}}(X,Y)]=\frac{1}{\mu(\Omega)}\int_{\Omega}y\mathbbm 1_{A\times\Rr^{\ell_L}}(x,y)\,\mu(\d x,\d y)\\
                &=\frac{1}{\mu(\Omega)}\int_{A\times\Rr^{\ell_L}}y\mathbbm \,\mu(\d x,\d y)=\frac{1}{\mu(\Omega)}\int_{A\times\Rr^{\ell_L}}f(x)\mathbbm \,\mu(\d x,\d y)\\
                &=\frac{1}{\mu(\Omega)}\int_{\Omega}f(x)\mathbbm 1_{A\times\Rr^{\ell_L}}(x,y)\,\mu(\d x,\d y)=\E[f(X)\mathbbm 1_{A\times\Rr^{\ell_L}}(X,Y)]\dott
                \end{split}
            \end{equation}}
    \argument{\lref{eq2.5}; \lref{NR: Integrability};the fact that $\sigma(X)=\mathcal B([a,b]^{\ell_0})\times \Rr^{\ell_L}$} {that it holds $\P$-a.s.\ that
    \begin{equation}\llabel{eq3}
        \E[Y|X]=f(X)\dott
    \end{equation}}
    \argument{\lref{eq2};\cref{Lemma XX}}{that
    \begin{equation}\llabel{eq4}
    \begin{split}
&\E[|h(X)\spro{v,Y}|]
=\frac{1}{\mu(\Omega)}\int_{\Omega}|h(x)\spro{v,y}|\,\mu(\d x,\d y)\\
&\leq \frac{2}{\mu(\Omega)}\int_{\Omega}|h(x)|^2\,\mu(\d x,\d y)
+\frac{2}{\mu(\Omega)}\int_{\Omega}\|y\|^2\|v\|^2\,\mu(\d x,\d y)<\infty\dott
\end{split}
    \end{equation}}
    \argument{\lref{eq4};\lref{eq3}}{that it holds a.s.\ that
    \begin{equation}\llabel{eq5}  \E[h(X)\spro{v,Y}|X]=h(X)\E[\spro{v,Y}|X]=h(X)\spro{v,\E[Y|X]}=h(X)\spro{v,f(X)}\dott
    \end{equation}}
    \argument{\lref{eq5};}{that \llabel{ARGG1}$\E[h(X)\spro{v,Y}]=\E[h(X)\spro{v,f(X)}]$\dott}
    \argument{\lref{ARGG1};\lref{eq2}}{that
    \begin{equation}\llabel{eq6}
    \begin{split}
&\int_{[a,b]^{\ell_0}\times\Rr^{\ell_L}}h(x)\spro{v,f(x)}\,\mu(\d x, \d y)=
\E[h(X)\spro{v,f(X)}]\\
&=\E[h(X)\spro{v,Y}]=\int_{[a,b]^{\ell_0}\times\Rr^{\ell_L}}h(x)\spro{v,y}\,\mu(\d x,\d y)\dott
\end{split}
    \end{equation}}
    \argument{\lref{eq6};}{\lref{conclude}\dott}
\end{aproof}
\begin{athm}{cor}{theo: explicit formula}
    Assume \cref{setting: explicit}, 
for every $ k \in \enne_0 $ 
let $ \diml_k \in \enne_0 $ satisfy 
$ \diml_k = \sum_{ n = 1 }^k \ell_n ( \ell_{ n - 1 } + 1 ) $,
and let $ \theta  \in \Rr^{ \fd } $. 
Then 
\begin{enumerate}[label=(\roman *)]
\item
\label{item 1: main}
it holds for all $ r \in [1,\infty) $ 
that $ \mL_r \in C^1( \Reals^{ \fd }, \Reals ) $,
\item
\label{item 2: main}
it holds for all $ r \in [1,\infty) $, $ k \in \{ 1,2, \ldots, L \} $, 
$ i \in \{ 1,2, \ldots, \ell_k \} $,
$ j \in \{ 1,2,\ldots, \ell_{k-1} \} $ that
\begin{equation}
\begin{split}
& 
  \pr*{  \frac{\partial\mL_r}
   {\partial\theta_{ (i-1)\ell_{k-1} + j +\diml_{k-1} } }
    } ( \theta)
\\
&
  =
  \sum_{\substack{v_k,v_{k+1},  
  \ldots,v_L\in \enne, \\\forall w\in \enne
  \cap[k, L]\colon v_w\leq\ell_w} }
  \int_{[a,b]^{ \ell_0 }\times\Rr^{\ell_L} }2\,
  \Big[
  \R_{r^{1/ ( \max \cu{k - 1 , 1 } ) } } ( \mN^{ \max\{k-1,1\}, \theta }_{r,j}(x))
  \indicator{ (1, L]}(k)\\
  &+x_j \indicator{\{1\} }(k)
  \Big]
  \Big[ \indicator{\{ i \} }(v_k)\Big]
  \Big[ \mN_{r,v_L}^{L, \theta }(x)-y_{v_L}\Big] \\
  &\cdot\Big[
  \textstyle{\prod}_{n={k+1} }^{L} \big(
  \fw^{n, \theta }_{v_n, v_{n-1} }
  \big[( \R_{r^{1/(n - 1 ) } } ) ' ( \mN^{n-1, \theta }_{r,v_{n-1} }(x))\big]
  \big)
  \Big]\, \mu( \d x,\d y_1,\d y_2,\dots \d y_{\ell_L}),
\end{split}
\end{equation}
\item
\label{item 3: main}
   it holds for all $ r\in [1,\infty)  $, $ k \in \{ 1,2, \ldots, L\} $, 
  $ i \in \{ 1,2, \ldots, \ell_k \} $ that
  \begin{equation}
  \begin{split}
  & \pr*{  \frac{\partial\mL_r}
  {\partial\theta_{\ell_k\ell_{k-1} + i+ \diml_{k-1} } }
   } ( \theta)
  \\
  &=
  \sum_{\substack{v_k,v_{k+1},  
  \ldots,v_L\in \enne, \\\forall w\in \enne
  \cap[k, L]\colon v_w\leq\ell_w} }
  \int_{[a,b]^{ \ell_0 } \times\Rr^{\ell_L}}2\,
  \Big[ \indicator{\{ i \} }(v_k)\Big]
  \Big[ \mN_{r,v_L}^{L, \theta }(x)-y_{v_L}\Big]\\
  &\cdot\Big[
  \textstyle{\prod}_{n={k+1} }^{L} \big(
  \fw^{n, \theta }_{v_n, v_{n-1} }
  \big[( \R_{r^{1 / ( n - 1 ) } } )'( \mN^{n-1, \theta }_{r,v_{n-1} }(x))\big]
  \big)
  \Big]\, \mu( \d x,\d y_1,\d y_2,\dots \d y_{\ell_L}),
\end{split}
\end{equation}
\item 
\label{item 4: main}
it holds that $ \limsup\nolimits_{r\to \infty }  \pr*{  \abs{ \mL_r( \theta)-\mL_{ \infty }( \theta) } +
 \|( \nabla\mL_r)( \theta)-\mG( \theta)\|  }  =0$,
 \item  
 \label{item 5: main}
 it holds for all $ k \in \{ 1,2, \ldots, L\} $,
  $ i \in \{ 1,2, \ldots, \ell_k \} $, 
  $ j \in \{ 1,2, \ldots, \ell_{k-1} \} $ that
  \begin{align}
  &\mG_{ (i-1)\ell_{k-1} + j + \diml_{k - 1 } }( \theta)\nonumber\\
  &=
  \sum_{\substack{v_k,v_{k+1},  
  \ldots,v_L\in \enne, \\\forall w\in \enne
  \cap[k, L]\colon v_w\leq\ell_w} }
  \int_{[a,b]^{ \ell_0 }\times\Rr^{\ell_L} }2\,
  \Big[
  \R_{ \infty }( \mN^{ \max\{k-1,1\}, \theta }_{ \infty,j}(x))
  \indicator{ (1, L]}(k)
  +x_j \indicator{\{1\} }(k)
  \Big]\Big[ \indicator{\{ i \} }(v_k)\Big]\nonumber\\
  &\cdot
  \Big[ \mN_{ \infty,v_L}^{L, \theta }(x)-y_{v_L}\Big] 
  \Big[
  \textstyle{\prod}_{n={k+1} }^{L} \big(
  \fw^{n, \theta }_{v_n, v_{n-1} }
   \mathbbm 1_{(0,\infty)}(\neurali{\infty}{v_{n-1}}{n-1}{\theta}(x))
  \big)
  \Big]\,\mu( \d x,\d y_1,\d y_2,\dots \d y_{\ell_L}),
\end{align}
and
 \item  
 \label{item 6: main}
 it holds for all $ k \in \{ 1,2, \ldots, L\} $,
  $ i \in \{ 1,2, \ldots, \ell_k \} $ that
  \begin{equation}
  \begin{split}
    &\mG_{\ell_k\ell_{k-1} + i+ \diml_{k-1 } }( \theta)
    \\
    &=\sum_{\substack{v_k,v_{k+1},  
  \ldots,v_L\in \enne, \\\forall w\in \enne
  \cap[k, L]\colon v_w\leq\ell_w} }
  \int_{[a,b]^{ \ell_0 }\times\Rr^{\ell_L} }2\,
  \Big[ \indicator{\{ i \} }(v_k)\Big]
  \Big[ \mN_{ \infty,v_L}^{L, \theta }(x)-y_{v_L}\Big]\\
  &\cdot\Big[
  \textstyle{\prod}_{n={k+1} }^{L} \big(
  \fw^{n, \theta }_{v_n, v_{n-1} }
   \mathbbm 1_{(0,\infty)}(\neurali{\infty}{v_{n-1}}{n-1}{\theta}(x))
  \big)
  \Big]\, \mu( \d x,\d y_1,\d y_2,\dots \d y_{\ell_L}).
\end{split}
\end{equation}
\end{enumerate}
\end{athm}
\begin{aproof}
\argument{\cref{Lemma ZZ};}{that there exists a measurable $f\colon [a,b]^{\ell_0}\allowbreak\to\Rr^{\ell_L}$ which satisfies that
    \begin{enumerate}[label=(\Roman*)]
        \item \llabel{Item 1} it holds that $\int_{[a,b]^{\ell_0}\times \Rr^{\ell_L}}\|f(x)\|^2\,\mu(\d x,\d y)<\infty$ and
        \item \llabel{Item 1.5} it holds for all $A\in\mathcal B([a,b]^{\ell_0})$ that
        \begin{equation}
            \int_{A\times \Rr^{\ell_L}} (f(x)-y)\,\mu(\d x,\d y)=0.
        \end{equation}
    \end{enumerate}}
    \startnewargseq
    \argument{\lref{Item 1};\lref{Item 1.5};\cref{Lemma XX};\cref{theo: explicit formula1};\cref{Lemma ZZZZ}} [verbs=ep]{\cref{item 1: main,item 2: main,item 3: main,item 4: main,item 5: main,item 6: main}\dott}
\end{aproof}
\section{Non-convergence to global minimizers for
stochastic gradient descent (SGD) optimization methods} \label{sec: non convergence of SGD method}
\renewcommand{\B}{B }
\renewcommand{\R}{\mathbb{R}}
\renewcommand{\neural}[4]{\mathcal{N}_{#1,#2}^{#3,#4}}
\renewcommand{\neurali}[5]
{\mathcal{N}_{#1,#2,#3}^{#4,#5}}
\renewcommand{\mG}{\mathcal{G}}
In this section we present and prove \cref{conjecture: multilayer2} in \cref{subsec: non-convergence: general}, which is the main result of this work.
\cref{conjecture: multilayer2} shows for a large class of \SGD\ optimization methods (including the plain vanilla standard \SGD, the momentum \SGD, the Nesterov accelerated \SGD, the \Adagrad, the \RMSprop, the \Adam, the \Adamax, the AMSGrad, and the \Nadam\ optimizers; see \cref{sec: SGD optimization methods})
that in the training of deep \ReLU\ \ANNs\ we have that 
the empirical risk of the considered optimization method does with high probability not converge to 
the infimal value of the empirical risk function. In particular, this ensures that 
the considered optimization method does with high probability not converge to global minimizers 
of the optimization problem; see \cref{cor: conjecture: multilayer} in \cref{subsec: non convergence to globmin point} below. 
Moreover, in this section we also establish in \cref{lem: estimate proba of nonconvergence: general case: rate of convergence 2} in \cref{subsec: non-convergence: general} below 
lower bounds on the speed of convergence to zero of the probabilities to converge to 
the infimal value of the empirical risk function. 
In particular, it turns out that the probability to not converge to the infimal value of the empirical risk function 
converges at least exponentially quickly to zero as the width of the first hidden layer of the \ANN\ 
(the number of neurons on the first hidden layer) and 
the depth of the \ANN\ (the number of hidden layers), respectively, 
increase to infinity (see \cref{lem: estimate proba of nonconvergence: general case: rate of convergence 2} for details).
\subsection{Mathematical framework for SGD methods 
in the training of deep ANNs}
\begin{setting}\label{setting: SGD}
Let $\scrA\in (0,\infty)$, $\scrB\in (\scrA,\infty)$, let $\mathbb{A}_r\colon \mathbb{R} \rightarrow \mathbb{R}$, $r\in[1,\infty]$, 
satisfy for all $r\in [1,\infty)$, $x\in (-\infty,\scrA r^{-1}]\cup[\scrB r^{-1},\infty)$, $y\in \R$ that
\begin{equation}\label{A_r}
    \mathbb A_r\in C^1(\R,\R),\quad \mathbb A_r(x)=x\mathbbm 1_{(\scrA r^{-1},\infty)}(x),\qandq 0\leq \mathbb A_r(y)\leq \mathbb A_\infty(y)=\max\{y,0\},
    \end{equation}
assume $
  \sup_{ r \in [1,\infty) }
  \sup_{ x \in \R } 
  | ( \mathbb A_r )'( x ) | < \infty 
$, for every $ L\in \N$, $\ell=(\ell_0,\ell_1,\dots, \ell_L) \in \N^{L+1}$ let $\fd_\ell \in\N$ satisfy $ \fd_\ell= \sum_{ k = 1 }^{L}\ell_k ( \ell_{k-1} + 1 ) $,
for every $r\in[1,\infty]$, $L\in\N$, $\ell=(\ell_0,\ell_1,\dots,\ell_L)\in \N^{L+1}$, $\theta=(\theta_1,\dots,\theta_{\fd_\ell})\in\R^{\fd_\ell}$ let $\neural{r}{\ell}{k}{\theta}=(\neurali{r}{\ell}{1}{k}{\theta},\dots,\neurali{r}{\ell}{\ell_k}{k}{\theta}) \colon \R^{ \ell_0 } \to \R^{ \ell_k} $, $k \in \{0,1,\dots,L\}$, satisfy for all $k\in \{0,1,\dots,L-1\}$, $x=(x_1,\dots, x_{\ell_0})\in \R^{\ell_0}$, $i\in\{1,2,\dots,\ell_{k+1}\}$ that
\begin{multline}\label{relization}
  \neurali{r}{\ell}{i}{k+1}{\theta}( x ) = \theta_{\ell_{k+1}\ell_{k}+i+\sum_{h=1}^{k}\ell_h(\ell_{h-1}+1)}+\sum\limits_{j=1}^{\ell_{k}}\theta_{(i-1)\ell_{k}+j+\sum_{h=1}^{k}\ell_h(\ell_{h-1}+1)}\big(x_j\indicator{\{0\}}(k) \\ 
  +\bA_{r^{1/\!\max\{k,1\}}}(\neurali{r}{\ell}{j}{k}{\theta}(x))\indicator{\N}(k) 
    \big),
\end{multline}
let $d\in \N$, $a\in \R$, $b\in (a,\infty)$, let $ ( \Omega, \mathcal{F}, \P) $ be a probability space, for every $ m, n \in \N_0 $ 
let 
$ X^m_n \colon \Omega \to [a,b]^d $
and 
$ Y^m_n \colon \Omega \to \R $
be random variables, assume for all $i\in \N$, $j\in \N\backslash\{i\}$ that $\P( X_0^i=X_0^j)=0$, for every $\ell\in (\cup_{L=1}^\infty\N^L)$, $n\in \N_0$ let $M_n^\ell\in \N$,
for every $r\in [1,\infty]$, $L\in \N$, $\ell\in(\{d\}\times\N^L\times \{1\})$, $n\in \N_0$ let
$   \cLnri{n}{r}{\ell}\colon \R^{ \fd_{ \ell } } \times \Omega \to \R$
satisfy for all 
$ \theta \in \R^{ \fd_{ \ell} }$
that
\begin{equation}
\label{setting: multilayer: eq:empirical_risk_for_mini_batch}
  \displaystyle
  \cLnri{n}{ r}{\ell }( \theta) 
  = 
  \frac{ 1 }{ M^{ \ell }_n } 
  \biggl[ \textstyle
  \sum\limits_{ m = 1 }^{ M^{ \ell }_n} 
    \abs{
      \neural{r}{\ell}{L+1}{\theta}( \X^m_n )
      - 
      Y^m_n 
    }^2
  \biggr],
\end{equation}
 for every $\ell\in(\cup_{L=1}^\infty(\{d\}\times\N^L\times\{1\}))$, $n\in \N_0$ let 
$ 
  \fG_n^{ \ell } 
  = ( \fG_n^{ \ell, 1 }, \dots, \fG_n^{ \ell, \fd_{ \ell} } ) 
  \colon \R^{ \fd_{ \ell } } \times \Omega \to \R^{ \fd_{ \ell } } 
$ 
satisfy for all $\omega\in\Omega$, $\theta\in \{\vartheta\in \R^{\fd_\ell}\colon (\nabla_{\vartheta} \cLnri{n}{r}{\ell}(\vartheta,\omega))_{r\in [1,\infty)}$ is convergent$\}$
that
\begin{equation}
\label{setting: multilayer: eq:gradients_SGD4:}
  \fG^{\ell}_n( \theta,\omega) 
  = 
  \lim_{r\to\infty}\bigl[\nabla_\theta \cLnri{n}{r}{\ell}(\theta,\omega)\bigr]
  ,
\end{equation}
let 
	$
	\Theta_n^\ell 
	= 
	( \Theta_n^{ \ell , 1 }, \dots, \Theta_n^{ \ell , \fd_{ \ell } } ) 
	\colon \Omega  \to \R^{\fd_{ \ell } }
	$
	be a random variable, and let 
	$
	\Phi_n ^\ell
	= 
	( 
	\Phi^{\ell, 1 }_n, \dots, 
	\Phi^{ \ell , \fd_{ \ell} }_n 
	)
	\colon 
	\allowbreak
	( \R^{ \fd_{ \ell } } )^{ n + 1 }
	\allowbreak
	\to 
	\R^{ \fd_{ \ell } }
	$ 
	satisfy 
	for all 
	$
	g =
	( 
	( g_{ i, j } )_{ j \in \{ 1, 2, \dots, \fd_{ \ell } \} }
	)_{
		i \in \{ 0, 1, \dots, n \}
	}
	\in 
	(
	\R^{ 
		\fd_{ \ell }
	}
	)^{ n + 1 }
	$, 
	$ 
	j \in \{1,2,\dots,\fd_\ell\}  
	$
	with 
	$
	\sum_{ i = 0 }^n
	\abs{ g_{ i, j } }
	= 0
	$
	that 
	$
	\Phi^{\ell , j }_n( g ) = 0 
	$,
	 assume for all $\ell\in (\cup_{L=1}^\infty(\{d\}\times\N^L\times\{1\}))$, $n\in\N$ that 
	\begin{equation}
		\label{eq: setting: SGD_def process}
		\Theta_{ n  } ^\ell
		= 
		\Theta_{n-1} ^\ell - 
		\Phi_{n-1}^\ell \bigl(
		\fG_1^\ell( \Theta_0^\ell  ) ,
		\fG_2^\ell ( \Theta_1^\ell  ) ,
		\dots ,
		\fG_n^\ell ( \Theta_{n-1}^\ell )
		\bigr),
	\end{equation}
and for every $L\in \N$, $\ell=(\ell_0,\ell_1,\dots,\ell_L)\in \N^{L+1}$, $\theta\in\R^{\fd_\ell}$ let 
  $\inact_{ \ell}^{ \theta } 
  \subseteq \N 
$
satisfy
\begin{equation}
\label{eq:setting_definition_of_I_set1}
  \inact_{ \ell }^{ \theta }
  = 
  \bigl\{ 
    i \in \{1,2,\dots,\ell_1\}
    \colon 
    \bigl(
      \forall \, x \in [a,b]^{\ell_0} \colon  
        \neurali{\infty}{\ell}{i}{1}{\theta}(x)
      < 0
    \bigr)
  \bigr\}.
\end{equation}
\end{setting}

\renewcommand{\bfl}{\ell}

\subsection{Vanishing gradients and inactive neurons}\label{non convegence: vanishing gradients}
\begin{athm}{lemma}{lem: vanishing gradients}[Vanishing gradients]
    Assume \cref{setting: SGD} and let $L\in \N\backslash\{1\}$, $\bfl=(\bfl_0,\bfl_1,\dots,\bfl_L)\in (\{d\}\times\N^{L-1}\times\{1\})$. Then it holds for all $\theta\in \R^{\fd_\bfl}$, $n\in \N_0$, $\omega\in \Omega$, $i\in \inact^{\theta}_{\bfl}$, $ j \in ( \cup_{ k = 1 }^d \{ ( i - 1 ) d + k \} ) \cup \cu{ \bfl_1 d + i } $ that
    \begin{equation}\label{conclude: vanishing gradients}
        \fG^{\bfl,j}_n(\theta,\omega)=0.
    \end{equation}
\end{athm}
\begin{aproof}
Throughout this proof for every $i\in\{1,2,\dots,d\}$ let $\tau_i\colon \R^d\to\R$ satisfy for all $x=(x_1,\dots,x_d)\in \R^d$ that $\tau_i(x)=x_i$, for every $k\in\{1,2,\dots,L\}$, $\theta=(\theta_1,\dots,\theta_{\fd_\bfl})\in \R^{\fd_\bfl}$, $i\in \{1,2,\dots,\bfl_k\}$
let 
$  
    \fb_i^{ k, \theta }, \fw^{ k, \theta }_{ i, 1 } ,\fw^{ k, \theta }_{ i, 2 },\dots, \fw^{ k, \theta }_{ i, \bfl_{k-1} }
  \in \R $ 
satisfy for all 
$ j \in \{ 1,2, \ldots, \bfl_{ k - 1 } \} $ 
that
\begin{equation}
\llabel{wb}
\fb^{ k, \theta }_i 
  =
  \theta_{ \bfl_k \bfl_{ k - 1 } + i 
  + 
  \sum_{ h = 1 }^{ k - 1 } \bfl_h ( \bfl_{ h - 1 } + 1 ) }
  \qqandqq
  \fw^{ k, \theta }_{ i, j }
  = 
  \theta_{ ( i - 1 ) \bfl_{ k - 1 } + j 
  + 
  \sum_{ h = 1 }^{ k - 1 } \bfl_h ( \bfl_{ h - 1 } + 1 ) },
\end{equation}
and for every $n\in \N_0$, $\omega\in \Omega$ let $\mu_{n,\omega}\colon \mathcal B([a,b]^d\times\R)\to [0,\infty]$ satisfy for all $A\in \mathcal B([a,b]^d\times\R)$ that 
\begin{equation}\llabel{def: mu}
    \mu_{n,\omega}(A)=\frac{1}{M_n^\bfl}\biggl[\textstyle\sum\limits_{m=1}^{M_n^\bfl}\mathbbm 1_{A}\bigl((X_n^m(\omega),Y_n^m(\omega))\bigr)\biggr].
\end{equation}
\argument{\cref{setting: multilayer: eq:empirical_risk_for_mini_batch};\lref{def: mu}}{that for all $r\in [1,
\infty]$, $\theta\in \R^{\fd_\bfl}$, $n\in \N_0$, $\omega\in \Omega$ it holds that
\begin{equation}\llabel{eq0.5}
\cLnri{n}{r}{\bfl}(\theta,\omega)=\int_{[a,b]^d\times\R^{\bfl_L}}|\neural{r}{\ell}{L}{\theta}(x)-y|^2\,\mu_{n,\omega}(\d x,\d y)\dott
\end{equation}}
\argument{\cref{eq:setting_definition_of_I_set1};}{for all $\theta\in \R^{\fd_\bfl}$, $i\in \inact_\bfl^{\theta}$, $x\in[a,b]^d$ that
\begin{equation}\llabel{eq1}
    \neurali{\infty}{\bfl}{i}{1}{\theta}(x)<0\dott
\end{equation}}
\argument{\lref{eq1};}{for all $\theta\in \R^{\fd_\bfl}$, $i\in \inact_\bfl^{\theta}$, $x\in [a,b]^d$ that
\begin{equation}\llabel{eq2}
    \mathbbm 1_{(0,\infty)}(\neurali{\infty}{\bfl}{i}{1}{\theta}(x))=0\dott
\end{equation}}
\argument{\lref{eq2};}{for all $\theta\in \R^{\fd_\bfl}$, $v\in \{1,2,\dots,\bfl_1\}$, $i\in \inact_\bfl^{\theta}$, $x\in [a,b]^d$ that
\begin{equation}\llabel{eq3}
    \mathbbm 1_{\{i\}}(v)  \mathbbm 1_{(0,\infty)}(\neurali{\infty}{\bfl}{v}{1}{\theta}(x))=0\dott
\end{equation}}
\argument{\cref{item 5: main,item 6: main} in \cref{theo: explicit formula} (applied for every $n\in \N_0$, $\theta\in \R^{\fd_\bfl}$, $\omega\in \Omega$ with $\fd\curvearrowleft\fd_{\bfl}$, $L\curvearrowleft L$, $\ell\curvearrowleft\bfl$, $((\fb^{k,\theta}_i)_{ i\in\{1,2,\dots,\bfl_k\}})_{(k,\theta)\in \{1,2,\dots,L\}\times \R^{\fd_\bfl}}\curvearrowleft\allowbreak(\allowbreak(\fb^{k,\theta}_i\allowbreak)_{ i\in\{1,2,\dots,\bfl_k\}}\allowbreak)_{(k,\theta)\in \{1,2,\dots,L\}\times \R^{\fd_\bfl}}$, $((\fw_{i,j}^{k,\theta}\allowbreak)_{(i,j)\in \{1,2,\dots,\bfl_k\}\times\{1,2,\dots,\bfl_{k-1}\}}\allowbreak)_{(k,\theta)\in \{1,2,\dots,L\}\times \R^{\fd_\bfl}}\curvearrowleft ((\fw_{i,j}^{k,\theta}\allowbreak\allowbreak)_{(i,j)\in \{1,2,\dots,\bfl_k\}\times\{1,2,\dots,\bfl_{k-1}\}}\allowbreak)_{(k,\theta)\in \{1,2,\dots,L\}\times \R^{\fd_\bfl}}$,  $\scrA\curvearrowleft\scrA$, $\scrB\curvearrowleft\scrB$, 
$(\mathbb A_r)_{r\in [1,\infty)}\curvearrowleft (\mathbb A_r)_{r\in [1,\infty)}$, 
$(\mN^{k,\vartheta}_r)_{(r,\vartheta,k)\in [1,\infty]\times \R^{\fd}\times \{0,1,\dots,L\}}\allowbreak\curvearrowleft (\neural{r}{\bfl}{k}{\vartheta})_{(r,\vartheta,k)\in [1,\infty]\times \R^{\fd_\bfl}\times \{0,1,\dots,L\}}$, $a\curvearrowleft a$, $b\curvearrowleft b$,
 $\mu\curvearrowleft \mu_
{n,\omega}$, $(\cL_r)_{r\in [1,\infty]}\curvearrowleft(\cLnri{n}{r}{\bfl})_{r\in [1,\infty]}$,
$\mG\curvearrowleft \fG^\bfl_n$ in the notation of \cref{theo: explicit formula});\lref{def: mu};\lref{eq0.5}}{that for all $n\in \N_0$,
$ i \in \{ 1,2, \ldots, \bfl_1 \} $, 
$ j \in \{ 1,2, \ldots, d \} $, $\theta\in \R^{\fd_\bfl}$, $\omega\in \Omega$
it holds that
\begin{multline}
  \fG^{ 
    \bfl,( i - 1 ) d + j 
  }_n( \theta,\omega ) 
=\frac{1}{M_n^\bfl}\Biggl[
 \textstyle \sum\limits_{m=1}^{M_n^\bfl}\biggl[
      \sum\limits_{ 
        \substack{
          v_1,v_2, \ldots, v_L \in \N, 
        \\
          \forall \, w \in \N \cap [1,L] \colon 
          v_w \leq \bfl_w
        }
      }
    2
      \tau_j(X_n^m)
          \indicator{ \{ i \} }( v_1 )
        \Bigl(
          \neurali{\infty}{\bfl}{v_L}{L}{\theta}( X^m_n(\omega) ) 
          - 
          Y^m_n(\omega) 
        \Bigr)\\
        \cdot
        \biggl(
          \textstyle
          \prod\limits_{ k= 2}^L
            \bigl[\fw^{ k, \theta }_{ v_k, v_{ k - 1 } }
               \mathbbm 1_{(0,\infty)} 
            \bigl(\neurali{\infty}{\bfl}{v_{k-1}}{k-1}{\theta} (X_n^m (\omega))\bigr)
       \bigr] \biggr)
  \biggr]\Biggr] 
\end{multline}
and
\begin{multline}
\llabel{G_b}
  \fG^{\bfl, 
    \bfl_1d  + i 
  }_n( \theta,\omega )
  =\frac{1}{M_n^\bfl}\Biggl[
 \textstyle \sum\limits_{m=1}^{M_n^\bfl}\biggl[
      \sum\limits_{ 
        \substack{
          v_1,v_2, \ldots, v_L \in \N, 
        \\
          \forall \, w \in \N \cap [1,L] \colon 
          v_w \leq \bfl_w
        }
      }
    2
          \indicator{ \{ i \} }( v_1 )
        \Bigl(\neurali{\infty}{\bfl}{v_L}{L}{\theta}
          ( X^m_n(\omega) ) 
          - 
          Y^m_n (\omega)
        \Bigr)
        \\
        \qquad\cdot
        \biggl(
          \textstyle
          \prod\limits_{ k = 2}^L
            \bigl[\fw^{ k, \theta }_{ v_k, v_{ k - 1 } }
            \mathbbm 1_{(0,\infty)} 
            \bigl(\neurali{\infty}{\bfl}{v_{k-1}}{k-1}{\theta} (X_n^m (\omega))\bigr)
        \bigr]\biggr)
  \biggr]\Biggr].
\end{multline}}
\argument{\lref{eq3};\lref{G_b}; the fact that for all $m,n\in \N_0$, $\omega\in \Omega$ it holds that $X_n^m(\omega)\allowbreak\in\allowbreak[a,b]^d$}{that for all $n\in \N_0$,
$\theta\in \R^{\fd_\bfl}$, $ i \in \inact_\bfl^{\theta} $, 
$ j \in \{ 1,2, \ldots, d \} $, $\omega\in \Omega$
it holds that
    \begin{multline}\llabel{eq4}
\fG^{ 
    \bfl,( i - 1 ) d + j 
  }_n( \theta,\omega ) 
=\frac{1}{M_n^\bfl}\Biggl[
 \textstyle \sum\limits_{m=1}^{M_n^\bfl}\biggl[
      \sum\limits_{ 
        \substack{
          v_1,v_2 \ldots, v_L \in \N, 
        \\
          \forall \, w \in \N \cap [1,L] \colon 
          v_w \leq \bfl_w
        }
      }
    2
          \tau_j(X_n^m)
         \fw^{ 2, \theta }_{ v_2, v_{ 1} } \indicator{ \{ i \} }( v_1 ) 
           \mathbbm 1_{(0,\infty)}(\neurali{
         \infty}{\bfl}{v_1}{1}{\theta}(X_n^m (\omega)))\\
       \cdot \Bigl(
          \neurali{\infty}{\bfl}{v_L}{L}{\theta}( X^m_n(\omega) ) 
          - 
          Y^m_n(\omega) 
        \Bigr)
        \biggl(
          \textstyle
          \prod\limits_{ k = 3}^L
            \bigl[\fw^{ k, \theta }_{ v_k, v_{ k- 1 } }
               \mathbbm 1_{(0,\infty)} 
            \bigl(\neurali{\infty}{\bfl}{v_{k-1}}{k-1}{\theta} (X_n^m (\omega))\bigr)
        \bigr]\biggr)
  \biggr]\Biggr] =0
\end{multline}
and 
\begin{multline}\llabel{eq5}
  \fG^{ 
    \bfl,\bfl_1d+i
  }_n( \theta,\omega ) 
=\frac{1}{M_n^\bfl}\Biggl[
 \textstyle \sum\limits_{m=1}^{M_n^\bfl}\biggl[
      \sum\limits_{ 
        \substack{
          v_1,v_2, \ldots, v_L \in \N, 
        \\
          \forall \, w \in \N \cap [1,L] \colon 
          v_w \leq \bfl_w
        }
      }
    2 
         \fw^{ 2, \theta }_{ v_2, v_{ 1 } } \indicator{ \{ i \} }( v_1 ) 
            \mathbbm 1_{(0,\infty)}(\neurali{
         \infty}{\bfl}{v_1}{1}{\theta}(X_n^m (\omega)))\\
        \cdot\Bigl(
          \neurali{\infty}{\bfl}{v_L}{L}{\theta}( X^m_n(\omega) ) 
          - 
          Y^m_n(\omega) 
        \Bigr) 
        \biggl(
          \textstyle
          \prod\limits_{ k = 3}^L
            \bigl[\fw^{ k, \theta }_{ v_k, v_{ k- 1 } }
               \mathbbm 1_{(0,\infty)} 
            \bigl(\neurali{\infty}{\bfl}{v_{k-1}}{k-1}{\theta} (X_n^m (\omega))\bigr)
        \bigr]\biggr)
  \biggr]\Biggr] 
  =0\dott
\end{multline}}
\argument{\lref{eq5};}[verbs=ep]{\cref{conclude: vanishing gradients}\dott}
\end{aproof}

\begin{athm}{cor}{lem: estimate proba of nonconvergence: part 1}[Inactive neurons]
Assume \cref{setting: SGD} and let $L\in \N\backslash\{1\}$, $\bfl=(\bfl_0,\bfl_1,\dots,\bfl_L)\in (\{d\}\times\N^{L-1}\times\{1\})$. Then it holds for all $n\in \N_0$, $\omega\in \Omega$ that
\begin{equation}\label{conclude: estimate proba of nonconvergence: part 1}
\bigl(\inact^{\Theta^\bfl_0}_\bfl(\omega)\bigr)\subseteq\bigl(\inact^{\Theta^\bfl_n}_\bfl(\omega)\bigr).
\end{equation}
\begin{aproof}
 Throughout this proof for every $i\in \{1,2,\dots,\bfl_1\}$ let $ \fU_{ i } \subseteq \R^{\fd_\bfl}$ satisfy 
\begin{equation}
\llabel{eq:in_proof_introduction_of_U_set_esimate proba convergence}
\begin{split}
  \fU_{ i } 
& = 
  \{  
    \theta 
    \in \R^{\fd_{ \bfl } } 
    \colon  
    i\in \inact^{\theta}_\bfl
  \} 
  \dott
\end{split}
\end{equation}
\argument{\lref{eq:in_proof_introduction_of_U_set_esimate proba convergence}}{for all $\theta\in\R^{\fd_\bfl}$ that
\begin{equation}\llabel{NRZZZ}
   \inact^\theta_\bfl=\{i\in \{1,2,\dots,\bfl_1\}\colon\theta\in\fU_i\}\dott 
\end{equation}}
\argument{\cref{relization};\cref{eq:setting_definition_of_I_set1};\lref{eq:in_proof_introduction_of_U_set_esimate proba convergence}}{that for all $i\in \{1,2,\dots,\bfl_1\}$ it holds that
\begin{equation}\llabel{NR8}
    \begin{split}
        \fU_i&=\bigl\{\theta\in \R^{\fd_\bfl}\colon \bigl(\forall\, x\in [a,b]^d\colon \neurali{\infty}{\bfl}{i}{1}{\theta}(x)<0\bigr)\bigr\}\\
        &=\bigl\{\theta=(\theta_1,\dots,\theta_{\fd_\bfl})\in \R^{\fd_\bfl}\colon \bigl(\forall\, x_1,x_2,\dots,x_d\in[a,b]\colon \theta_{\bfl_1d+i}+\textstyle\sum_{j=1}^d \theta_{(i-1)d+j}x_j<0\bigr)\bigr\}\dott
    \end{split}
\end{equation}}
\argument{\cref{lem: vanishing gradients};\lref{eq:in_proof_introduction_of_U_set_esimate proba convergence}}{that for all $n\in \N_0$, $\omega\in \Omega$, $i\in \{1,2,\dots,\bfl_1\}$, $\theta\in \fU_i$, $j\in (\cup_{k=1}^d\{(i-1)d+k\})\cup\{\bfl_1d+i\}$ it holds that
\begin{equation}\llabel{eqtg1}
    \fG^{\bfl,j}_{n}(\theta,\omega)=0\dott
\end{equation}}
\argument{\lref{eqtg1};}{that for all $n\in \N_0$, $\omega\in \Omega$, $i\in \{1,2,\dots,\bfl_1\}$, $\theta_0,\theta_1,\dots,\theta_n\in \fU_i$, $j\in (\cup_{k=1}^d\{(i-1)d+k\})\cup\{\bfl_1d+i\}$ it holds that
\begin{equation}\llabel{NR_7}
    \sum_{v=0}^n |\fG^{\bfl,j}_{v+1}(\theta_v,\omega)|=0\dott
\end{equation}}
\argument{the fact that for all 
$ n \in \N_0 $, $\omega\in \Omega$,
 $
  g \allowbreak
  =
  ( 
    ( g_{ i, j }\allowbreak )_{ j \in \{ 1, 2, \dots, \fd_{ \ell } \} }
  )_{
    i \in \{ 0, 1,\allowbreak \dots, n \}
  }\allowbreak
  \in 
  (
    \R^{ 
      \fd_{ \ell }
    }
  )^{ n + 1 }
$, 
$ 
  j \in \{ 1, 2, \dots, \fd_{ \ell } \} 
$
with 
$
  \sum_{ i = 0 }^n
  \abs{ g_{ i, j } }
  = 0
$
it holds that 
$
  \Phi^{\ell , j }_n( g ) = 0 
$;}{that for all $n\in\N_0$, $g=\allowbreak((g_{v,j}\allowbreak)_{j\in \{1,2,\dots,\fd_\bfl\}}\allowbreak)_{v\in \{0,1,\dots,n\}}\in (\R^{\fd_\bfl})^{n+1}$, $i\in\{1,2,\dots,\bfl_1\}$, $j\in (\cup_{k=1}^d\{(i-1)d+k\})\cup\{\bfl_1d+i\}$ with $\sum_{v=0}^n|g_{v,j}|=0$ it holds that
\begin{equation}\llabel{eqtg3}
\Phi^{\bfl,j}_n(g)=0\dott
\end{equation}}
\argument{\lref{eqtg3};\lref{NR_7}}{that for all $n\in \N_0$, $\omega\in \Omega$, $i\in \{1,2,\dots,\bfl_1\}$, $\theta_0,\theta_1,\dots,\allowbreak\theta_n\in \fU_i$, $j\in (\cup_{k=1}^d\{(i-1)d+k\})\cup\{\bfl_1d+i\}$ it holds that
\begin{equation}\llabel{eqtg4}
    \Phi^{\bfl,j}_n\bigl(\fG^\bfl_1(\theta_0,\omega), \fG^\bfl_2(\theta_1,\omega),\dots, \fG^\bfl_{n+1}(\theta_n,\omega)\bigr)=0\dott
\end{equation}}
\argument{\lref{eqtg4};}{that for all $n\in \N_0$, $\omega\in \Omega$, $i\in\{1,2,\dots,\bfl_1\}$, $j\in (\cup_{k=1}^d\{(i-1)d+k\})\cup\{\bfl_1d+i\}$ with $(\cup_{k=0}^n \{\Theta^\bfl_k(\omega)\}\}\subseteq\fU_i$ it holds that
\begin{equation}\llabel{eqtg5}
 \Phi^{ \bfl , j }_n \bigl( \fG^\bfl _1 (  \Theta^\bfl _0 ( \omega ), \omega ) 
    ,
    \fG^\bfl _2( 
      \Theta^\bfl _1 ( \omega ), \omega 
    ) 
    ,
    \dots 
    ,
    \fG^\bfl _{n+1}( 
      \Theta^\bfl _{n} ( \omega ), \omega 
    ) \bigr)
  = 0 \dott
\end{equation}}
\argument{\lref{eqtg5};}{that for all $n\in \N$, $\omega\in \Omega$, $i\in \{1,2,\dots,\bfl_1\}$, $j\in (\cup_{k=1}^d\{(i-1)d+k\})\cup\{\bfl_1d+i\}$ with $(\cup_{k=0}^{n-1} \{\Theta^\bfl_k(\omega)\}\}\subseteq\fU_i$ it holds that
\begin{equation}\llabel{eqtg6}
    \Phi^{ \bfl , j }_{n-1} \bigl( \fG^\bfl _1 (  \Theta^\bfl _0 ( \omega ), \omega ) 
    ,
    \fG^\bfl _2( 
      \Theta^\bfl _1 ( \omega ), \omega 
    ) 
    ,
    \dots 
    ,
    \fG^\bfl _{n}( 
      \Theta^\bfl _{n-1} ( \omega ), \omega 
    ) \bigr)
  = 0 \dott
\end{equation}}
\argument{ \lref{eqtg6};\cref{eq: setting: SGD_def process}} 
{
for all $n\in \N$, $\omega\in \Omega$,
$ i \in \{1,2,\dots,\bfl_1\} $, 
$ j \in ( \cup_{ k = 1 }^d \{ ( i - 1 ) d + k \} ) \cup \cu{ \bfl_1  d + i } $
with 
$
  ( 
    \cup_{ k = 0 }^{n-1} \{ 
      \Theta^\bfl_k ( \omega )
    \} 
  )
  \subseteq 
  \fU_{ i }
$
that 
\begin{equation}\llabel{arg4}
  \Theta^{\bfl, j }_{ n  }( \omega ) 
  =
  \Theta^{ \bfl , j }_{n-1}( \omega ) 
  \dott
\end{equation}}
\argument{\lref{NR8};\lref{arg4}} 
{ for all $n\in \N$, $\omega\in \Omega$,
$ i \in \{1,2,\dots,\bfl_1\} $
with 
$
  ( 
    \cup_{ k = 0 }^{n-1} \{ 
      \Theta_k^\bfl ( \omega )
    \} 
  )
  \subseteq 
  \fU_{ i }
$
 that 
\begin{equation}\llabel{arg5}
  ( 
    \cup_{ k = 0 }^{ n } \{ 
      \Theta_k^\bfl ( \omega )
    \} 
  )
  \subseteq 
  \fU_{ i }
  \dott
\end{equation}}
\argument{\lref{arg5};induction} {\llabel{arg6}
for all 
$ n \in \N_0 $, $\omega\in \Omega$, 
$ i \in \{1,2,\dots,\bfl_1\} $
with 
$ 
  \Theta_0^\bfl ( \omega ) 
  \in \fU_{ i } 
$
that
$
  ( 
    \cup_{ k = 0 }^n
    \{ 
      \Theta_k^\bfl ( \omega ) 
    \} 
  ) 
  \subseteq 
  \fU_{ i }
$\dott}
\argument{\lref{arg6}}{
for all 
$\omega\in \Omega$, $ i \in \{1,2,\dots,\bfl_1\}$
with 
$ 
  \Theta_0^\bfl ( \omega ) 
  \in \fU_{ i } 
$
that 
\begin{equation}\llabel{arg7}
  ( 
    \cup_{ k = 0 }^{ \infty }
    \{ 
      \Theta_k^\bfl ( \omega ) 
    \} 
  ) 
  \subseteq 
  \fU_{ i }
  \dott
\end{equation}}
\argument{\lref{arg7};}{that for all $n\in \N_0$, $\omega\in \Omega$, $i\in \{1,2,\dots,\bfl_1\}$ with $\Theta^\bfl_0(\omega)\in \fU_i$ it holds that \llabel{arg7.1}$\Theta^\bfl_n(\omega)\in \fU_i$\dott}
\argument{\lref{arg7.1};\lref{NRZZZ}}{\cref{conclude: estimate proba of nonconvergence: part 1}\dott}
\end{aproof}
    \subsection{Non-covergence for very wide ANNs}\label{non-convergence: very wide}
\end{athm}
\begin{athm}{prop}{lem: estimate proba of nonconvergence}
    Assume \cref{setting: SGD} and let $L\in \N\backslash\{1\}$, $\bfl=(\bfl_0,\bfl_1,\dots,\bfl_L)\in (\{d\}\times\N^{L-1}\times\{1\})$.
 Then
    \begin{equation}\label{conclude: lem: estimate proba of nonconvergence}
    \begin{split}
      \P\biggl( \inf_{n\in \N_0} \cLnri{0}{\infty}{\bfl}(\NNelll^{\bfl}_n)\leq\inf_{\theta\in \R^{\fd_\bfl}}\cLnri{0}{\infty}{\bfl}(\theta)\biggr)
      \leq \P\Bigl(\# \bigl(\inact_{ \bfl }^{ \NNelll_0^{\bfl}}\bigr)< 3\Bigr)+ \P\Big(\inf_{\theta\in \R^{\fd_\bfl}}\cLnri{0}{\infty}{\bfl}(\theta)=0\Big).
      \end{split}
    \end{equation}
\end{athm}
\begin{aproof} 
Throughout this proof assume without loss of generality that $\bfl_1\geq 3$ (otherwise note that $\P\bigl(\inf_{n\in \N_0}   \cLnri{0}{\infty}{\bfl}(\NNelll^{\bfl}_n)\leq\inf_{\theta\in \R^{\fd_\bfl}}\cLnri{0}{\infty}{\bfl}(\theta)\bigr)\leq 1=\P(\# (\inact_{ \bfl }^{ \NNelll_0^{\bfl}})\leq\bfl_1< 3)\leq \P\bigl(\# (\inact_{ \bfl }^{ \NNelll_0^{\bfl}})< 3\bigr)+ \P(\inf_{\theta\in \R^{\fd_\bfl}}\cLnri{0}{\infty}{\bfl}(\theta)=0)$) and let $E\subseteq\Omega$ satisfy
\begin{equation}\llabel{def: E}
    E=\Bigl\{ \#\bigl(\inact_\bfl^{\Theta_0^\bfl}\bigr)< 3\Bigr\}\cup\Bigl\{\#\bigl\{\X_0^1,\X_0^2,\dots,\X_0^{M_0^\bfl}\bigr\}<M_0^\bfl\Big\}
    \cup\biggl\{ \inf_{\theta\in \R^{\fd_{\bfl}}}\cLnri{0}{\infty}{\bfl}(\theta)\leq 0\biggr\}.
\end{equation}
\argument{\cref{lem: estimate proba of nonconvergence: part 1};}{ for all $n\in\N_0$, $\omega\in \Omega$ that 
\begin{equation}\llabel{arg8}
\#\bigl(\inact_\bfl^{\Theta_n^\bfl(\omega)}\bigr)\geq \#\bigl(\inact_\bfl^{\Theta_0^\bfl(\omega)}\bigr)\dott
    \end{equation}}
    \argument{\lref{arg8};}{ that for all $\omega\in \Omega$ with $\#\big(\inact_\bfl^{\Theta_0^\bfl(\omega)}\big)\geq 3$ it holds that 
    \begin{equation}\llabel{arg12}
\inf\nolimits_{n\in\N_0}\bigl(\#\bigl(\inact_\bfl^{\Theta_n^\bfl(\omega)}\bigr)\bigr)\geq 3\dott
    \end{equation}}
\argument{\lref{arg12}; \cref{cor: improve risk}}{for all $\omega\in \Omega$ with
$\#\big(\inact_\bfl^{\Theta_0^\bfl(\omega)}\big)\geq 3$, $\#\big\{\X_0^1(\omega), \X_0^2(\omega),\dots,\allowbreak \X_0^{M_0^\bfl}(\omega)\big\}=M_0^\bfl$, and $\inf_{\theta\in \R^{\fd_{\bfl}}}\cLnri{0}{\infty}{\bfl}(\theta,\omega)>0$ that 
\begin{equation}\llabel{arg8.75}
    \inf_{n\in \N_0} \cLnri{0}{\infty}{\bfl}(\Theta_n^\bfl(\omega),\omega)\geq \inf_{\theta\in \R^{\fd_{\bfl}},\ \#(\inact_{\bfl}^{\theta})\geq 3}\cLnri{0}{\infty}{\bfl}(\theta,\omega)>  \inf_{\theta\in \R^{\fd_{\bfl}}}\cLnri{0}{\infty}{\bfl}(\theta,\omega)\dott
\end{equation}}
\argument{\lref{arg8.75};\lref{def: E}}{\llabel{arg8.8}that for all $\omega\in \Omega$ with $\inf_{n\in \N_0} \cLnri{0}{\infty}{\bfl}(\NNelll^{\bfl}_n(\omega),\omega)\leq\inf_{\theta\in \R^{\fd_\bfl}}\cLnri{0}{\infty}{\bfl}(\theta,\omega)$ it holds that $\omega\in E$\dott}
\argument{\lref{arg8.8};;\lref{def: E}}{that
\begin{equation}\llabel{NRZ}
\begin{split}
     &\P\biggl( \inf_{n\in \N_0}\cLnri{0}{\infty}{\bfl}(\NNelll^{\bfl}_n)\leq\inf_{\theta\in \R^{\fd_\bfl}}\cLnri{0}{\infty}{\bfl}(\theta)\biggr)\\
     &\leq\P(E)
     \leq \P\Big(\#\big(\inact_\bfl^{\Theta_0^\bfl}\big)< 3\Big)+\P\Big(\#\big\{\X_0^1,\X_0^2,\dots,\X_0^{M_0^\bfl}\big\}<M_0^\bfl\Big)+\P\bigg(\inf_{\theta\in \R^{\fd_{\bfl}}}\cLnri{0}{\infty}{\bfl}(\theta)\leq 0\bigg)\\
&\leq\P\Bigl(\#\big(\inact_\bfl^{\Theta_0^\bfl}\big)< 3\Bigr)+\P\Bigl(\exists\, \mathcal M\in \N\colon \#\big\{\X_0^1,\X_0^2,\dots,\X_0^{\mathcal M}\big\}<\mathcal M\Bigr)+\P\biggl(\inf_{\theta\in \R^{\fd_{\bfl}}}\cLnri{0}{\infty}{\bfl}(\theta)\leq 0\biggr)\dott
     \end{split}
\end{equation}}
\argument{ the assumption that for all $i\in \N$, $j\in \N\backslash\{i\}$ it holds that $\P(X_0^i=X_0^j )=0$}{that \llabel{arg9} $\P(\exists \, i\in \N, j\in \N\backslash\{i\}\colon \X^i_0=\X^j_0)=0$\dott}
\argument{\lref{arg9};}{that \llabel{arg10} $\P(\exists \,  n\in \N\colon \#\big\{\X_0^1,\X_0^2,\dots,\X_0^{n}\big\}<n)=0$\dott}
\argument{\lref{arg10};\lref{NRZ}}{\cref{conclude: lem: estimate proba of nonconvergence}\dott}
\end{aproof}
\renewcommand{\nnode}{\ell}
\begin{athm}{cor}{very wide 3}
    Assume \cref{setting: SGD}, for every $k\in \N$ let $L_k\in \N\backslash\{1\}$, $\nnode_k=(\nnode_k^0,\nnode_k^1,\dots,\nnode_k^{L_k})\in (\{d\}\times\N^{L_k-1}\times\{1\}) $, 
let $ \Dens \colon \R \to [0,\infty) $ be measurable, 
assume 
$\sup_{ \eta \in (0,\infty) } \bigl(  \eta ^{ - 1 } \int_{ - \eta }^{ \eta } 1_{ \R \backslash { \{0\} } }( \Dens( x ) ) \, \d x \bigr)\allowbreak \geq 2$,
 let $(\cki_{k,i})_{(k,i)\in \N^2}\subseteq(0,\infty)$ 
satisfy for all $k\in \N$,
$x_1,x_2,\dots,x_{\fd_{\bfl_k}} \in \R$ 
that
\begin{equation}
  \P\bigl( \cap_{i=1}^{\bfl_k^1d+\bfl_k^1}
    \bigl\{
    \NNelll^{\bfl_k,i}_0 
    < x_i\bigr\}
  \bigr)
  =
  \prod_{i=1}^{\bfl_k^1d+\bfl_k^1}\bigg[\int_{ - \infty }^{\cki_{k,i}x_i} \Dens(y) \, \d y\bigg]
,
\end{equation}
and assume $\sup_{k\in \N}\max_{i\in \{1,2,\dots,\bfl_k^1\}}\max_{ j\in\{1,2,\dots,d\}}\allowbreak\bigl((c_{k,(i-1)d+j})^{-1}c_{k,\bfl_k^1 d+i}\bigr)<\infty$.
Then there exists $\boundexp\in (0,\infty)$ which satisfies for all $k\in \N$ that
\begin{equation}\llabel{conclude}
\begin{split}
\textstyle
 &\P\biggl( \inf_{n\in \N_0}  \cLnri{0}{\infty}{\bfl_k}(\NNelll^{\bfl_k}_n)=\inf_{\theta\in \R^{\fd_{\bfl_k}}}\cLnri{0}{\infty}{\bfl_k}(\theta)\biggr)\leq \boundexp^{-1}\exp(-\boundexp\bfl_k^1)+\P\biggl(\textstyle\inf\limits_{\theta\in \R^{\fd_{\bfl_k}}}\cLnri{0}{\infty}{\bfl_k} (\theta)=0\biggr).
 \end{split}
\end{equation}
\end{athm}
\begin{aproof}
    \argument{\cref{cor:estimate dead neuron hidden layer 1.1} (applied with $\bfc\curvearrowleft 3$ in the notation of \cref{cor:estimate dead neuron hidden layer 1.1})}{that there exists $\boundexp\in (0,\infty)$ which satisfies for all $k\in \N$ that
\begin{equation}\llabel{eq1}
\begin{split}
    \P\Big(\# \bigl(\inact_{ \bfl_k }^{ \Theta_0^{\bfl_k}}\bigl)< 3\Big)\leq \boundexp^{-1}\exp(-\boundexp\bfl_k^1)\dott
    \end{split}
    \end{equation}}
    \startnewargseq
    \argument{ \cref{lem: estimate proba of nonconvergence};\lref{eq1}}{that for all $k\in \N$ it holds that
    \begin{equation}\llabel{eq10}
    \begin{split}
       & \P\biggl( \inf_{n\in \N_0} \cLnri{0}{\infty}{\bfl_k}(\NNelll^{\bfl_k}_n)\leq\inf\limits_{\theta\in \R^{\fd_{\bfl_k}}}\cLnri{0}{\infty}{\bfl_k}(\theta)\biggr)\\
        &\leq \P\Big(\# \big(\inact_{ \bfl_k }^{ \Theta_0^{\bfl_k}}\big)< 3\Big)
+\P\bigg(\inf_{\theta\in \R^{\fd_{\bfl_k}}}\cLnri{0}{\infty}{\bfl_k}(\theta)=0\bigg)\\
&  \leq \boundexp^{-1}\exp(-\boundexp\bfl_k^1)+ \P\bigg(\inf_{\theta\in \R^{\fd_{\bfl_k}}}\cLnri{0}{\infty}{\bfl_k}(\theta)=0\bigg)\dott
\end{split}
    \end{equation}}
    \argument{\lref{eq10};}{\lref{conclude}\dott}
\end{aproof}
\begin{athm}{cor}{cor: nonconvergence}
    Assume \cref{setting: SGD}, for every $k\in \N$ let $L_k\in \N\backslash\{1\}$, $\nnode_k=(\nnode_k^0,\nnode_k^1,\dots,\nnode_k^{L_k})\in (\{d\}\times\N^{L_k-1}\times\{1\})$, assume $\liminf_{k\to \infty}\bfl_k^1=\infty$ and $\liminf_{k\to \infty}\P\big(\inf_{\theta\in \R^{\fd_{\bfl_k}}}\cLnri{0}{\infty}{\bfl_k} \allowbreak(\allowbreak\theta\allowbreak)
    \allowbreak>0\big)=1$, 
let $ \Dens \colon \R \to [0,\infty) $ be measurable, 
assume 
$\sup_{ \eta \in (0,\infty) } \bigl(  \eta ^{ - 1 } \int_{ - \eta }^{ \eta } 1_{ \R \backslash { \{0\} } }( \Dens( x ) ) \, \d x \bigr) \geq 2$,
 let $(\cki_{k,i})_{(k,i)\in \N^2}\subseteq(0,\infty)$ 
satisfy for all $k\in \N$,
$x_1,x_2,\dots,x_{\fd_{\bfl_k}} \in \R$ 
that
\begin{equation}
  \P\bigl( \cap_{i=1}^{d\bfl_k^1+\bfl_k^1}
    \bigl\{
    \NNelll^{\bfl_k,i}_0 
    < x_i\bigr\}
  \bigr)
  =
  \prod_{i=1}^{d\bfl_k^1+\bfl_k^1}\bigg[\int_{ - \infty }^{\cki_{k,i}x_i} \Dens(y) \, \d y\bigg],
\end{equation}
and assume $\sup_{k\in \N}\max_{i\in \{1,2,\dots,\bfl_k^1\}}\max_{ j\in\{1,2,\dots,d\}}\allowbreak\bigl((c_{k,(i-1)d+j})^{-1}c_{k,\bfl_k^1 d+i}\bigr)<\infty$.
Then
\begin{equation}\label{conclusion:cor 4.16}
\textstyle
 \liminf\limits_{k\to\infty}\P\biggl( \inf\limits_{n\in \N_0}  \cLnri{0}{\infty}{\bfl_k}(\NNelll^{\bfl_k}_n)>\inf\limits_{\theta\in \R^{\fd_{\bfl_k}}}\cLnri{0}{\infty}{\bfl_k}(\theta)\biggr)= 1.
\end{equation}
\end{athm}
\begin{aproof}
    \argument{\cref{very wide 3}; the assumption that $\liminf_{k\to \infty}\bfl_k^1=\infty$; the assumption that $\liminf_{k\to \infty}\P\big(\inf_{\theta\in \R^{\fd_{\bfl_k}}}\cLnri{0}{\infty}{\bfl_k} \allowbreak(\allowbreak\theta\allowbreak)
    \allowbreak>0\big)=1$}{\cref{conclusion:cor 4.16}\dott}
\end{aproof}
\renewcommand{\bfl}{\mathbf{l}}
\subsection{Non-convergence for very deep ANNs}\label{subsec: non convegence: very deep}
\begin{athm}{lemma}{vanishing gradient: very deep}
     Assume \cref{setting: SGD}, let $L\in \N\backslash\{1\}$, $k\in \{2,3,\dots,L\}$, $\ell\in (\{d\}\times \N^{L}\times\{1\})$, let $\fJ\subseteq \R^{\fd_\ell}$ satisfy
\begin{equation}\llabel{def: J}
	\textstyle \fJ = \bigg\{ \NNel = (\NNel_1,\dots,\NNel_{\fd_\ell}) \in \R^{\fd_\ell} \colon \bigg[ \forall\, j \in \N \cap \bigg(\sum\limits_{i=1}^{k-1} \ell_i(\ell_{i-1}+1),\sum\limits_{i=1}^{k} \ell_i(\ell_{i-1}+1)\bigg] \colon \NNel_j < 0 \bigg] \bigg\}.
\end{equation}
    and let $\vartheta\in \fJ$.
    Then it holds for all $n\in\N_0$, $\omega\in \Omega$, $j\in \N\cap[1,\sum_{i=1}^k\ell_i(\ell_{i-1}+1)]$ that
    \begin{equation}\llabel{conclude}
        \fG^{\ell,j}_{n}(\vartheta,\omega)=0.
    \end{equation}
\end{athm}
\begin{aproof}
Throughout this proof for every $i\in\{1,2,\dots,d\}$ let $\tau_i\colon \R^d\to\R$ satisfy for all $x=(x_1,\dots,x_d)\in \R^d$ that $\tau_i(x)=x_i$, let $\bfl=(\bfl_0,\dots,\bfl_{L+1})\in \N^{L+2}$ satisfy $\bfl=\ell$, let $\theta=(\theta_1,\dots,\theta_{\fd_\bfl})\in \R^{\fd_\bfl}$ satisfy $\theta=\vartheta$,
let 
$ 
  \fw^{ m} = 
  ( \fw^{ m }_{ i, j } )_{ 
    (i,j) \in \{ 1, \ldots, \bfl_m \} \times \{ 1, \ldots, \bfl_{ m- 1 } \} 
  }
  \in \R^{ \bfl_m \times \bfl_{ m- 1 } }
$, 
$
  m \in \N 
$, 
and 
$
  \fb^{ m } 
  = 
  ( \fb^{ m }_1, \dots, \fb^{ m }_{ \bfl_m} )
  \in \R^{ \bfl_m } 
$,
$ m \in \N $, 
satisfy for all 
$ m \in \{ 1, \dots, L+2 \} $, 
$ i \in \{ 1, \ldots, \bfl_m \} $,
$ j \in \{ 1, \ldots, \bfl_{ m - 1 } \} $ 
that
\begin{equation}
\llabel{wb}
  \fw^{ m}_{ i, j }
  = 
  \theta_{ ( i - 1 ) \bfl_{ m- 1 } + j 
  + 
  \sum_{ h = 1 }^{ m- 1 } \bfl_h ( \bfl_{ h - 1 } + 1 ) }
\qqandqq
  \fb^{ m }_i 
  =
  \theta_{ \bfl_m\bfl_{ m - 1 } + i 
  + 
  \sum_{ h = 1 }^{ m - 1 } \bfl_h ( \bfl_{ h - 1 } + 1 ) } 
  ,
\end{equation}
and for every $n\in \N_0$, $\omega\in \Omega$ let $\mu_{n,\omega}\colon \mathcal B([a,b]^d\times\R)\to [0,\infty)$ satisfy for all $A\in \mathcal B([a,b]^d\times\R)$ that 
\begin{equation}\llabel{def: mu}
    \mu_{n,\omega}(A)=\frac{1}{M_n^\bfl}\biggl[\textstyle\sum\limits_{m=1}^{M_n^\bfl}\mathbbm 1_{A}(X_n^m(\omega),Y_n^m(\omega))\biggr].
\end{equation}
\argument{\cref{setting: multilayer: eq:empirical_risk_for_mini_batch};\lref{def: mu}}{that for all $r\in [1,
\infty]$, $n\in \N_0$, $\omega\in \Omega$, $\theta\in \R^{\fd_\bfl}$ it holds that
\begin{equation}\llabel{eq0.5}
\cLnri{n}{r}{\bfl}(\theta,\omega)=\int_{[a,b]^d\times\R^{\bfl_L}}|\neural{r}{\ell}{L}{\theta}(x)-y|^2\,\mu_{n,\omega}(\d x,\d y)\dott
\end{equation}}
    \argument{\lref{def: J};the fact that $\theta\in \fJ$}{that
    for all $j\in \N \cap(\sum_{i=1}^{k-1} \bfl_i(\bfl_{i-1}+1),\sum_{i=1}^{k} \bfl_i(\bfl_{i-1}+1)]$ it holds that
    \begin{equation}\llabel{eq1'}
    \theta_j<0\dott
    \end{equation}}
    \argument{\lref{eq1'};}{for all $i\in \{1,2,\dots,\bfl_k\}$, $j\in \{1,2,\dots,\bfl_{k-1}\}$ that
    \begin{equation}\llabel{eq2}
        \theta_{\bfl_{k}\bfl_{k-1}+i+\sum_{h=1}^{k-1}\bfl_h(\bfl_{h-1}+1)}<0\qqandqq \theta_{(i-1)\bfl_{k-1}+j+\sum_{h=1}^{k-1}\bfl_h(\bfl_{h-1}+1)} <0\dott
    \end{equation}}
\argument{\cref{relization multi};\lref{eq2}}{for all $i\in \{1,2,\dots,\bfl_k\}$, $x\in\R^d$ that
\begin{equation}\llabel{eq4}
\begin{split}
  \neurali{\infty}{\bfl}{i}{k}{\theta}( x ) 
  &= \theta_{\bfl_{k}\bfl_{k-1}+i+\sum_{h=1}^{k-1}\bfl_h(\bfl_{h-1}+1)}+\sum\limits_{j=1}^{\bfl_{k-1}}\theta_{(i-1)\bfl_{k-1}+j+\sum_{h=1}^{k-1}\bfl_h(\bfl_{h-1}+1)} 
  \max\!\big\{\neurali{\infty}{\bfl}{j}{k-1}{\theta}(x),0\big\}\\
  &\leq \theta_{\bfl_{k}\bfl_{k-1}+i+\sum_{h=1}^{k-1}\bfl_h(\bfl_{h-1}+1)}<0\dott
  \end{split}
\end{equation}}
\argument{\lref{eq4}}{for all $i\in \{1,2,\dots,\bfl_k\}$, $x\in \R^d$ that
\begin{equation}\llabel{eq5}
    \mathbbm 1_{(0,\infty)}(\neurali{\infty}{\bfl}{i}{k}{\theta}(x))=0\dott
\end{equation}}
\argument{\lref{eq5};}{for all $l\in\{1,2,\dots,k\}$, $v_l\in \{1,2,\dots,\bfl_l\}$, $v_{l+1}\in\{1,2,\dots,\bfl_{l+1}\}$, $\dots$, $v_{L+1}\in \{1,2,\dots,\bfl_{L+1}\}$, $x\in \R^d$ that
\begin{equation}\llabel{eq6}
\begin{split}
\textstyle
& \prod\limits_{ p = l + 1 }^{L+1}\bigl(
            \mathbbm 1_{(0,\infty)}(\neurali{\infty}{\bfl}{v_{p-1}}{p-1}{\theta}( x ))\bigr)\\
            &=  \indicator{(0,\infty)}
             (\neurali{\infty}{\bfl}{v_k}{k}{\theta}
            ( x ))\biggl(\textstyle\prod\limits_{ p \in \{l+1,l+2,\dots,L+1\}\backslash\{k+1\} }
             \bigl(\mathbbm 1_{(0,\infty)}(\neurali{\infty}{\bfl}{v_{p-1}}{p-1}{\theta}( x ))\bigr)\biggr)=0\dott
             \end{split}
\end{equation}}
\argument{\lref{eq6};}{for all $l\in\{1,2,\dots,k\}$, $v_l\in \{1,2,\dots,\bfl_l\}$, $v_{l+1}\in\{1,2,\dots,\bfl_{l+1}\}$, $\dots$, $v_{L+1}\in \{1,2,\dots,\bfl_{L+1}\}$, $x\in \R^d$ that
\begin{equation}\llabel{eq7}
\begin{split}
\textstyle
&\prod\limits_{ p = l + 1 }^{L+1}
          \bigl(
            \fw^{ p }_{ v_p, v_{ p - 1 } }
           \mathbbm 1_{(0,\infty)}(\neurali{\infty}{\bfl}{v_{p-1}}{p-1}{\theta}( x ))
          \bigr)\\
          &=\biggl( \prod\limits_{ p = l + 1 }^{L+1}
            \fw^{ p }_{ v_p, v_{ p - 1 } }\biggr)\biggl(\prod\limits_{ p = l + 1 }^{L+1}\mathbbm 1_{(0,\infty)}(\neurali{\infty}{\bfl}{v_{p-1}}{p-1}{\theta}( x ))\biggr)=0\dott
            \end{split}
\end{equation}}
\argument{\cref{item 5: main,item 6: main} in \cref{theo: explicit formula} (applied for every $n\in \N_0$, $\theta\in \R^{\fd_\bfl}$, $\omega\in \Omega$ with $\fd\curvearrowleft\fd_{\bfl}$, $L\curvearrowleft L$, $\ell\curvearrowleft\bfl$, $((\fb^{k,\theta}_i)_{ i\in\{1,2,\dots,\bfl_k\}})_{(k,\theta)\in \{1,2,\dots,L\}\times \R^{\fd_\bfl}}\curvearrowleft\allowbreak(\allowbreak(\fb^{k,\theta}_i\allowbreak)_{ i\in\{1,2,\dots,\bfl_k\}}\allowbreak)_{(k,\theta)\in \{1,2,\dots,L\}\times \R^{\fd_\bfl}}$, $((\fw_{i,j}^{k,\theta}\allowbreak)_{(i,j)\in \{1,2,\dots,\bfl_k\}\times\{1,2,\dots,\bfl_{k-1}\}}\allowbreak)_{(k,\theta)\in \{1,2,\dots,L\}\times \R^{\fd_\bfl}}\curvearrowleft ((\fw_{i,j}^{k,\theta}\allowbreak)_{(i,j)\in \{1,2,\dots,\bfl_k\}\times\{1,2,\dots,\bfl_{k-1}\}}\allowbreak)_{(k,\theta)\in \{1,2,\dots,L\}\times \R^{\fd_\bfl}}$,  $\scrA\curvearrowleft\scrA$, $\scrB\curvearrowleft\scrB$, 
$(\mathbb A_r)_{r\in [1,\infty)}\curvearrowleft (\mathbb A_r)_{r\in [1,\infty)}$, 
$(\mN^{k,\vartheta}_r)_{(r,\vartheta,k)\in [1,\infty]\times \R^{\fd}\times \{0,1,\dots,L\}}\allowbreak\curvearrowleft (\neural{r}{\bfl}{k}{\vartheta})_{(r,\vartheta,k)\in [1,\infty]\times \R^{\fd_\bfl}\times \{0,1,\dots,L\}}$, $a\curvearrowleft a$, $b\curvearrowleft b$,
 $\mu\curvearrowleft \mu_
{n,\omega}$, $(\cL_r)_{r\in [1,\infty]}\curvearrowleft(\cLnri{n}{r}{\bfl})_{r\in [1,\infty]}$,
$\mG\curvearrowleft \fG^\bfl_n$ in the notation of \cref{theo: explicit formula});\lref{def: mu};\lref{eq0.5}}{for all $n\in \N_0$, $l\in \{2,3,\dots,L+1\}$, $i\in \{1,2,\dots,\bfl_l\}$, $j\in \{1,2,\dots,\bfl_{l-1}\}$, $\theta\in \R^{\fd_\bfl}$, $\omega\in \Omega$ that
\begin{align}
&
  \fG_n^{\bfl, 
    ( i - 1 ) \bfl_{ l - 1 } + j 
    + 
    \sum_{ h = 1 }^{ l - 1 } \bfl_h ( \bfl_{ h - 1 } + 1 ) 
  }( \theta,\omega ) \notag
\\
&
  =\frac{1}{M_n^\bfl}\Biggl[\textstyle\sum\limits_{m=1}^{M_n^\bfl}\biggl[
      \sum\limits_{ 
        \substack{
          v_l, v_{ l + 1 }, \ldots, v_L \in \N, 
        \\
          \forall \, w \in \N \cap [l,L+1] \colon 
          v_w \leq \bfl_w
        }
      }
    2
    \Bigl(\bigl(\mathbb A_\infty\bigl(
        \neurali{\infty}{\bfl}{j}{l-1}{\theta}( X_n^m )\bigr)\bigr)\mathbbm 1_{(1,L+1]}(l)+\tau_j(X_n^m)\mathbbm 1_{\{1\}}(l)\Bigr)\indicator{ \{ i \} }( v_l ) \notag
\\
&
       \qquad \cdot
        \Bigl(
          \neurali{\infty}{\bfl}{v_{L+1}}{L+1}{\theta}( X^m_n(\omega)) 
          - 
          Y^m_n(\omega)
        \Bigr)
        \Bigl(
          \textstyle
          \prod_{ p = l + 1 }^{L+1}
            \bigl[\fw^{ p }_{ v_p, v_{ p - 1 } }
             \mathbbm 1_{(0,\infty)}\bigl(\neurali{\infty}{\bfl}{v_{p-1}}{p-1}{\theta}( X_n^m (\omega))\bigr)\bigr]
        \Bigr)
      \biggr]
    \Biggr] \label{G_w}
\end{align}
and
\begin{equation}\llabel{G_b}
\begin{split}
&
  \fG_n^{\bfl, 
    \bfl_l\bfl_{l-1} + i 
    + 
    \sum_{ h = 1 }^{ l - 1 } \bfl_h ( \bfl_{ h - 1 } + 1 ) 
  }( \theta,\omega ) 
\\
&
  =\frac{1}{M_n^\bfl}\Biggl[\textstyle\sum\limits_{m=1}^{M_n^\bfl}\biggl[
      \sum\limits_{ 
        \substack{
          v_l, v_{ l + 1 }, \ldots, v_L \in \N, 
        \\
          \forall \, w \in \N \cap [l,L+1] \colon 
          v_w \leq \bfl_w
        }
      }
    2\indicator{ \{ i \} }( v_l )
        \Bigl(
         \neurali{\infty}{\bfl}{v_{L+1}}{L+1}{\theta}( X^m_n(\omega)) 
          - 
          Y^m_n(\omega)
        \Bigr)
\\
&
       \qquad \cdot
        \Bigl(
          \textstyle
          \prod_{ p = l + 1 }^{L+1}
            \bigl[\fw^{ p }_{ v_p, v_{ p - 1 } }
            \mathbbm 1_{(0,\infty)}\bigl(\neurali{\infty}{\bfl}{v_{p-1}}{p-1}{\theta}( X_n^m (\omega))\bigr)\bigr]
        \Bigr)
      \biggr]
    \Biggr] \dott
\end{split}
\end{equation}
}
\argument{\lref{G_b};\lref{eq7}}{for all $l\in \{1,2,\dots,k\}$, $i\in \{1,2,\dots,\bfl_l\}$, $j\in \{1,2,\dots,\bfl_{l-1}\}$, $\omega\in \Omega$ that
\begin{equation}\llabel{eq8}
    \fG_n^{\bfl, 
    ( i - 1 ) \bfl_{ l - 1 } + j 
    + 
    \sum_{ h = 1 }^{ l - 1 } \bfl_h ( \bfl_{ h - 1 } + 1 ) 
  }( \theta ,\omega) = \fG_n^{\bfl, 
     \bfl_l\bfl_{l-1} + i  
    + 
    \sum_{ h = 1 }^{ l - 1 } \bfl_h ( \bfl_{ h - 1 } + 1 ) 
  }( \theta,\omega ) =0\dott
\end{equation}}
\argument{\lref{eq8};}{\lref{conclude}\dott}
\end{aproof}

\renewcommand{\bfl}{\ell}

\begin{athm}{lemma}{very deep: auxillary 1}
     Assume \cref{setting: SGD}, let $L\in \N\backslash\{1\}$, $\bfl=(\bfl_1,\dots,\bfl_{L+1})\in (\{d\}\times \N^{L}\times\{1\})$, and let $\InAct\subseteq\R^{\fd_\bfl}$ satisfy
\begin{equation}\llabel{def I}
    \InAct= \bigcup_{k=2}^L \bigg\{ \NNel = (\NNel_1,\dots,\NNel_{\fd_\bfl}) \in \R^{\fd_\bfl} \colon \bigg[ \forall\, j \in \N \cap \bigg(\textstyle\sum\limits_{i=1}^{k-1} \nnode_i(\nnode_{i-1}+1),\sum\limits_{i=1}^{k} \nnode_i(\nnode_{i-1}+1)\bigg] \colon \NNel_j < 0 \bigg] \bigg\}.
    \end{equation}
    Then
    \begin{equation}\llabel{conclude}
       \P\bigl(\Theta^\bfl_0\in \InAct\bigr)\leq  \P
        \bigl(\forall\, n\in \N\colon \Theta^\bfl_n\in \InAct\bigr).
    \end{equation}
\end{athm}
\begin{aproof}
Throughout this proof for every $k\in\{2,3,\dots,L\}$ let $\InActL{k}\subseteq \R^{\fd_\bfl}$ satisfy
\begin{equation}\llabel{def: J}
	\InActL{k} = \bigg\{ \NNel = (\NNel_1,\dots,\NNel_{\fd_\bfl}) \in \R^{\fd_\bfl} \colon \bigg[ \forall\, j \in \N \cap \bigg(\sum\limits_{i=1}^{k-1} \nnode_i(\nnode_{i-1}+1),\sum\limits_{i=1}^{k} \nnode_i(\nnode_{i-1}+1)\bigg] \colon \NNel_j < 0 \bigg] \bigg\}.
 \end{equation}
 \argument{\lref{def I}; \lref{def: J};}{that
 \begin{equation}\llabel{argg1}
     \InAct=(\cup_{k=2}^L\InActL{k})\dott
 \end{equation}}
 \argument{\cref{vanishing gradient: very deep};\lref{def: J}}{that for all $k\in\{2,3,\dots,L\}$, $m\in \N_0$, $\theta\in \InActL{k}$, $j\in (\textstyle\sum_{i=1}^{k-1} \nnode_i(\nnode_{i-1}+1),\sum_{i=1}^{k} \nnode_i(\nnode_{i-1}+1)]$, $\omega\in \Omega$ it holds that
\begin{equation}\llabel{eqtg1}
    \fG^{\bfl,j}_{m}(\theta,\omega)=0\dott
\end{equation}}
    \argument{\lref{eqtg1};}{that for all $k\in\{2,3,\dots,L\}$, $m\in \N_0$, $\theta_0,\theta_1,\dots,\theta_m\in \InActL{k}$, $j\in (\textstyle\sum_{i=1}^{k-1} \nnode_i(\nnode_{i-1}+1),\sum_{i=1}^{k} \nnode_i(\nnode_{i-1}+1)]$, $\omega\in \Omega$ it holds that
\begin{equation}\llabel{NR_7}
    \sum_{v=0}^m |\fG^{\bfl,j}_{v+1}(\theta_v,\omega)|=0\dott
\end{equation}}
\argument{the fact that for all $\ell\in (\cup_{\fL=1}^\infty (\{d\}\times\N^{\fL}))$,
$ m \in \N_0 $,
 $
  g \allowbreak
  =
  ( 
    ( g_{ i, j }\allowbreak )_{ j \in \{ 1, 2, \dots, \fd_{ \ell } \} }
  )_{
    i \in \{ 0, 1,\allowbreak \dots, m \}
  }\allowbreak
  \in 
  (
    \R^{ 
      \fd_{ \ell }
    }
  )^{ m + 1 }
$, 
$ 
  j \in \{ 1, 2, \dots, \fd_{ \ell } \} 
$
with 
$
  \sum_{ i = 0 }^m
  \abs{ g_{ i, j } }
  = 0
$
it holds that 
$
  \Phi^{\ell , j }_m( g ) = 0 
$;}{that for all $m\in\N_0$, $g=\allowbreak((g_{v,j}\allowbreak)_{j\in \{1,2,\dots,\fd_\bfl\}}\allowbreak)_{v\in \{0,1,\dots,m\}}\in (\R^{\fd_\bfl})^{m+1}$, $j\in (\textstyle\sum_{i=1}^{k-1} \nnode_i(\nnode_{i-1}+1),\sum_{i=1}^{k} \nnode_i(\nnode_{i-1}+1)]$ with $\sum_{v=0}^m|g_{v,j}|=0$ it holds that
\begin{equation}\llabel{eqtg3}
\Phi^{\bfl,j}_m(g)=0\dott
\end{equation}}
\argument{\lref{eqtg3};\lref{NR_7}}{that for all $k\in\{2,3,\dots,L\}$, $m\in \N_0$, $\theta_0,\theta_1,\dots,\theta_m\in \InActL{k}$, $j\in (\textstyle\sum_{i=1}^{k-1} \nnode_i(\nnode_{i-1}+1),\sum_{i=1}^{k} \nnode_i(\nnode_{i-1}+1)]$, $\omega\in \Omega$ it holds that
\begin{equation}\llabel{eqtg4}
    \Phi^{\bfl,j}_m(\fG^\bfl_1(\theta_0,\omega), \fG^\bfl_2(\theta_1,\omega),\dots, \fG^\bfl_{m+1}(\theta_m,\omega))=0\dott
\end{equation}}
\argument{\lref{eqtg4};}{that for all $k\in\{2,3,\dots,L\}$, $m\in \N_0$, $j\in (\textstyle\sum_{i=1}^{k-1} \nnode_i(\nnode_{i-1}+1),\sum_{i=1}^{k} \nnode_i(\nnode_{i-1}+1)]$, $\omega\in \Omega$ with $(\cup_{q=0}^m \{\Theta^\bfl_q(\omega)\}\}\subseteq\InActL{k}$ it holds that
\begin{equation}\llabel{eqtg5}
 \Phi^{ \bfl , j }_m \bigl( \fG^\bfl _1 (  \Theta^\bfl _0 ( \omega ), \omega ) 
    ,
    \fG^\bfl _2( 
      \Theta^\bfl _1 ( \omega ), \omega 
    ) 
    ,
    \dots 
    ,
    \fG^\bfl _{m+1}( 
      \Theta^\bfl _{m} ( \omega ), \omega 
    ) \bigr)
  = 0 \dott
\end{equation}}
\argument{\lref{eqtg5};}{that for all $k\in \{2,3,\dots,L\}$, $m\in \N$, $j\in (\textstyle\sum_{i=1}^{k-1} \nnode_i(\nnode_{i-1}+1),\sum_{i=1}^{k} \nnode_i(\nnode_{i-1}+1)]$, $\omega\in \Omega$ with $(\cup_{q=0}^{m-1} \{\Theta^\bfl_q(\omega)\}\}\subseteq\fU_i$ it holds that
\begin{equation}\llabel{eqtg6}
    \Phi^{ \bfl , j }_{m-1} \bigl( \fG^\bfl _1 (  \Theta^\bfl _0 ( \omega ), \omega ) 
    ,
    \fG^\bfl _2( 
      \Theta^\bfl _1 ( \omega ), \omega 
    ) 
    ,
    \dots 
    ,
    \fG^\bfl _{m}( 
      \Theta^\bfl _{m-1} ( \omega ), \omega 
    ) \bigr)
  = 0 \dott
\end{equation}}
\argument{ \lref{eqtg6};\cref{eq: setting: SGD_def process}} 
{
for all $k\in \{2,3,\dots,L\}$, $m\in \N$,
$j\in (\textstyle\sum_{i=1}^{k-1} \nnode_i(\nnode_{i-1}+1),\sum_{i=1}^{k} \nnode_i(\nnode_{i-1}+1)]$, $\omega\in\Omega$
with 
$
  ( 
    \cup_{ q = 0 }^{m-1} \{ 
      \Theta^\bfl_q ( \omega )
    \} 
  )
  \subseteq 
  \InActL{k}
$
that 
\begin{equation}\llabel{arg4}
  \Theta^{\bfl, j }_{ m  }( \omega ) 
  =
  \Theta^{ \bfl , j }_{m-1}( \omega ) 
  \dott
\end{equation}}
\argument{\lref{arg4}} 
{for all $k\in \{2,3,\dots,L\}$, $m\in \N$, $\omega\in\Omega$
with 
$
  ( 
    \cup_{ q = 0 }^{m-1} \{ 
      \Theta_q^\bfl ( \omega )
    \} 
  )
  \subseteq 
  \InActL{k}
$
 that 
\begin{equation}\llabel{arg5}
  ( 
    \cup_{ q = 0 }^{ m } \{ 
      \Theta_q^\bfl ( \omega )
    \} 
  )
  \subseteq 
 \InActL{k}
  \dott
\end{equation}}
\argument{\lref{arg5};induction} {\llabel{arg6} for all
$k\in \{2,3,\dots,L\}$, $m\in \N$, $\omega\in\Omega$ with 
$ 
  \Theta_0^\bfl ( \omega ) 
  \in \InActL{k}
$
that
$
  ( 
    \cup_{ q= 0 }^m 
    \{ 
      \Theta_q^\bfl ( \omega ) 
    \} 
  ) 
  \subseteq 
  \InActL{k}
$\dott}
\argument{\lref{arg6}}{
for all 
$k\in\{2,3,\dots,L\}$, $\omega\in \Omega$
with 
$ 
  \Theta_0^\bfl ( \omega ) 
  \in \InActL{k}
$
that 
\begin{equation}\llabel{arg7}
  ( 
    \cup_{ q= 0 }^{ \infty }
    \{ 
      \Theta_q^\bfl ( \omega ) 
    \} 
  ) 
  \subseteq 
  \InActL{k}
  \dott
\end{equation}}
\argument{\lref{arg7};\lref{argg1}}{for all $\omega\in \Omega$ with $\Theta_0^\bfl\in \InAct$ that
\begin{equation}\llabel{arg8}
    ( 
    \cup_{ q= 0 }^{ \infty }
    \{ 
      \Theta_q^\bfl ( \omega ) 
    \} 
  ) 
  \subseteq 
  \InAct
  \dott
\end{equation}}
\argument{\lref{arg8};}{\lref{conclude}\dott}
\end{aproof}

\begin{athm}{lemma}{very deep: auxillary 2}
    Assume \cref{setting: SGD}, let $L\in \N\backslash\{1\}$, $l\in \N$, $\bfl=(\bfl_0,\bfl_1,\dots,\bfl_{L+1})\in (\{d\}\times\{1,2,\dots,l\}^{L}\times \{1\})$,  $c_{1},c_2,\dots,c_{\fd_{\bfl}}\in(0,\infty)$, assume that
$
  c_i\Theta^{ \bfl, i }_0 
$, 
$ i \in \{ 1, 2, \dots, \fd_\bfl \} $, 
are \iid, and let $\InAct\subseteq\R^{\fd_\bfl}$ satisfy
\begin{equation}\llabel{def: I}
    \InAct= \bigcup_{k=2}^L \bigg\{ \NNel = (\NNel_1,\dots,\NNel_{\fd_\bfl}) \in \R^{\fd_\bfl} \colon \bigg[ \forall\, j \in \N \cap \bigg(\textstyle\sum\limits_{i=1}^{k-1} \nnode_i(\nnode_{i-1}+1),\sum\limits_{i=1}^{k} \nnode_i(\nnode_{i-1}+1)\bigg] \colon \NNel_j < 0 \bigg] \bigg\},
    \end{equation}
    Then 
    \begin{equation}\label{eq: very deep auxillary concule}
        \P
        \big(\forall\, n\in \N\colon \Theta_n^\bfl\in \InAct\big)\geq 1-\big[1-[\P(\Theta_0^{\bfl,1}<0)]^{l(l+1)}\big]^{L-1}.
    \end{equation}
\end{athm}
\begin{aproof}
    Throughout this proof for every $k\in\{2,3,\dots,L\}$ let $\InActL{k}\subseteq \R^{\fd_\bfl}$ satisfy
\begin{equation}\llabel{def: J}
	\InActL{k} = \bigg\{ \NNel = (\NNel_1,\dots,\NNel_{\fd_\bfl}) \in \R^{\fd_\bfl} \colon \bigg[ \forall\, j \in \N \cap \bigg(\sum\limits_{i=1}^{k-1} \nnode_i(\nnode_{i-1}+1),\sum\limits_{i=1}^{k} \nnode_i(\nnode_{i-1}+1)\bigg] \colon \NNel_j < 0 \bigg] \bigg\}.
\end{equation}
\argument{ \lref{def: I};\lref{def: J}}{that
\begin{equation}\llabel{eq1}
    \InAct=(\cup_{k=2}^L \InActL{k})\dott
\end{equation}}
\argument{the assumption that $
  c_i\Theta^{ \bfl, i }_0 
$, 
$ i \in \{ 1, 2, \dots, \fd_\bfl \} $, 
are \iid }{that \llabel{arg1} $\Theta^{ \bfl, i }_0$, $i\in \{1,2,\dots,\fd_\bfl\}$, are independent\dott}
\argument{\lref{arg1};\lref{def: J}}{that \llabel{arg2} the events $\Theta_0^{\bfl}\in \InActL{k}$, $k\in \{2,3,\dots,L\}$, are independent\dott}
\argument{\lref{arg2};\lref{eq1}}{that
\begin{equation}\llabel{eq2}
\begin{split}
\P\bigl(\Theta^\bfl_0\in\InAct\bigr)&=\P\bigl(\Theta^\bfl_0\in(\cup_{k=2}^L\InActL{k})\bigr)=1-\P\bigl(\Theta^\bfl_0\notin(\cup_{k=2}^L\InActL{k})\bigr)\\
&=1-\P\bigl(\Theta^\bfl_0\in(\cap_{k=2}^L(\R^{\fd_\bfl}\backslash\InActL{k})))
=1-\textstyle\prod\limits_{k=2}^L \P\bigl(\Theta^\bfl_0\in(\R^{\fd_\bfl}\backslash\InActL{k})\bigr)\\
&=1-\biggl[\textstyle\prod\limits_{k=2}^L \Bigl[1-\P\bigl(\Theta^\bfl_0\in\InActL{k}\bigr)\Bigr]\biggr]\dott
\end{split}
\end{equation}}
\argument{\lref{def: J}; the fact that for all $i\in \{1,2,\dots,\fd_\bfl\}$ it holds that $c_i>0$; the assumption that $
  c_i\Theta^{ \bfl, i }_0 
$, 
$ i \in \{ 1, 2, \dots, \fd_\bfl \} $, 
are \iid}{that for all $k\in \{2,3,\dots,L\}$ it holds that
\begin{equation}\llabel{eq3}
\begin{split}
    \P\bigl(\Theta^\bfl_0\in\InActL{k}\bigr)&= \P\biggl(\forall\, j \in \N \cap \bigg(\textstyle\sum\limits_{i=1}^{k-1} \nnode_i(\nnode_{i-1}+1),\sum\limits_{i=1}^{k} \nnode_i(\nnode_{i-1}+1)\bigg] \colon \Theta^{\bfl,j}_0< 0\biggr)\\
    &=\P\biggl(\forall\, j \in \N \cap \bigg(\textstyle\sum\limits_{i=1}^{k-1} \nnode_i(\nnode_{i-1}+1),\sum\limits_{i=1}^{k} \nnode_i(\nnode_{i-1}+1)\bigg] \colon c_j\Theta_0^{\bfl,j} < 0\biggr)\\
&=\prod\nolimits_{j=\textstyle\sum_{i=1}^{k-1} \nnode_i(\nnode_{i-1}+1)}^{\sum_{i=1}^{k} \nnode_i(\nnode_{i-1}+1)}\P\bigl(c_1\Theta_0^{\bfl,1}<0\bigr)\\
&=\bigl[\P(c_1\Theta_0^{\bfl,1}<0)\bigr]^{\bfl_k(\bfl_{k-1}+1)}=\bigl[\P(\Theta_0^{\bfl,1}<0)\bigr]^{\bfl_k(\bfl_{k-1}+1)}\dott
    \end{split}
\end{equation}}
\argument{\lref{eq3}; the fact that for all $k\in\{1,2,\dots,L\}$ it holds that $\bfl_k\leq l$}{ for all $k\in \{2,3,\dots,L\}$ that
\begin{equation}\llabel{eq4}
     \P\bigl(\Theta^\bfl_0\in\InActL{k}\bigr)
\geq\bigl[\P(\Theta_0^{\bfl,1}<0)\bigr]^{l(l+1)}\dott
\end{equation}}
\argument{\lref{eq4};\lref{eq2}}{that
\begin{equation}\llabel{eq5}
     \P
        \big( \Theta_0^\bfl\in \InAct\big)\geq 1-\big[1-[\P(\Theta_0^{\bfl,1}<0)]^{l(l+1)}\big]^{L-1}\dott
\end{equation}}
\argument{\lref{eq5};\cref{very deep: auxillary 1}}{
\cref{eq: very deep auxillary concule}\dott}
\end{aproof}
\begin{athm}{lemma}
{lem: estimate proba of nonconvergence: L to infty}
    Assume \cref{setting: SGD}, let $L\in \N\backslash\{1\}$, $l\in \N$, $\bfl\in (\{d\}\times\{1,2,\dots,l\}^{L}\times \{1\})$, $c_{1},c_2,\dots,c_{\fd_{\bfl}}\in(0,\infty)$, and assume that
$
  c_i\Theta^{ \bfl, i }_0 
$, 
$ i \in \{ 1, 2, \dots, \fd_\bfl \} $, 
are \iid \ Then
    \begin{equation}\label{conclude: L to infty}
    \begin{split}
     \P\biggl( \inf_{n\in \N_0} \cLnri{0}{\infty}{\bfl}(\NNelll^{\bfl}_n)\leq\inf_{\theta\in \R^{\fd_\bfl}}\cLnri{0}{\infty}{\bfl}(\theta)\biggr)&\leq \big[1-[\P(\Theta_0^{\bfl,1}<0)]^{l(l+1)}\big]^{L-1}+\P\Big(\inf_{\theta\in \R^{\fd_\bfl}}\cLnri{0}{\infty}{\bfl}(\theta)=0\Big).
      \end{split}
    \end{equation}
\end{athm}
\renewcommand{\bfl}{\mathbf{l}}
\renewcommand{\nnode}{\mathbf{l}}
\begin{aproof}
Throughout this proof let $\bfl=(\bfl_0,
\bfl_1,\dots,\bfl_L)\in \N^{L+1}$ satisfy $\bfl=\ell$, for every $k\in\{2,3,\dots,L\}$ let $\InActL{k}\subseteq \R^{\fd_\bfl}$ satisfy
\begin{equation}\llabel{def: J}
	 \InActL{k}= \bigg\{ \NNel = (\NNel_1,\dots,\NNel_{\fd_\bfl}) \in \R^{\fd_\bfl} \colon \bigg[ \forall\, j \in \N \cap \bigg(\textstyle\sum\limits_{i=1}^{k-1} \nnode_i(\nnode_{i-1}+1),\sum\limits_{i=1}^{k} \nnode_i(\nnode_{i-1}+1)\bigg] \colon \NNel_j < 0 \bigg] \bigg\},
\end{equation}
and let $\InAct\subseteq \R^{\fd_\bfl}$ satisfy
 $\InAct = \bigcup_{k \in \{2,3,\dots,L\}} \InActL{k}$. 
\argument{\cref{prop: improve risk_case2b2}; the assumption that for all $i\in \N$, $j\in \N\backslash\{i\}$ it holds that $\P(X_0^i=X_0^j )=0$} {that
\begin{equation}\llabel{not global min: constant relization}
    \P\Bigg(\Biggl\{\inf_{\theta\in \R^{\fd_\bfl}}\cLnri{0}{\infty}{\bfl}(\theta)=\inf _{Q\in \R}\biggl(\displaystyle\frac{ 1 }{ M^{ \bfl }_0 } 
  \biggl[ 
  \textstyle\sum\limits_{ m = 1 }^{ M^{ \bfl }_0} 
    \abs{
      Q
      - 
      Y^m_0 
    }^2
  \biggr]\biggr)\Biggr\}\cap\Bigl\{\displaystyle\inf_{\theta\in \R^{\fd_\bfl}}\cLnri{0}{\infty}{\bfl}(\theta)>0\Bigr\}\Bigg)=0\dott
  \end{equation}}
\argument{\lref{def: J}; the fact that $\InAct = \bigcup_{k \in \{2,3,\dots,L\}} \InActL{j}$}{\llabel{arg1} for all $\omega\in\Omega$ with $(\forall\, n\in\N\colon\Theta^\bfl_n(\omega)\in \InAct)$ that there exists $(Q_n)_{n\in \N}\subseteq\R$ such that for all $n\in \N$, $x\in \R^{d}$ it holds that $\mathcal N_{\infty,\bfl}^{L+1,\Theta_n^\bfl(\omega)}(x)=Q_n$\dott}
 \argument{\lref{arg1};}{for all $\omega\in \Omega$ with $(\forall\, n\in \N\colon\Theta^\bfl_n(\omega)\in \InAct)$ and $\inf_{n\in \N_0} \cLnri{0}{\infty}{\bfl}(\NNelll^{\bfl}_n(\omega),\omega)\leq\inf_{\theta\in \R^{\fd_\bfl}}\cLnri{0}{\infty}{\bfl}(\theta,\omega)$ that there exists $Q\in \R$ such that \llabel{arg2} $\inf_{\theta\in \R^{\fd_\bfl}}\cLnri{0}{\infty}{\bfl}(\theta,\omega)\allowbreak= \frac{ 1 }{ M^{ \bfl }_0 } 
  \bigl[ 
  \sum\nolimits_{ m = 1 }^{ M^{ \bfl }_0} 
    \abs{
      Q
      - 
      Y^m_0 (\omega)
    }^2
  \bigr]$\dott} 
  \argument{\lref{arg2}; \lref{not global min: constant relization}} { that 
  \begin{equation}\llabel{arg3}
  \begin{split}
      &\P\bigg(\biggl\{ \inf_{n\in \N_0}  \cLnri{0}{\infty}{\bfl}(\NNelll^{\bfl}_n)\leq\inf_{\theta\in \R^{\fd_\bfl}}\cLnri{0}{\infty}{\bfl}(\theta)\biggr\}\cap\{\forall \, n\in \N\colon \Theta_\bfl^n\in \InAct\}\bigg)\\
      &\leq \P\Biggl(\exists\, Q\in \R\colon \inf_{\theta\in \R^{\fd_\bfl}}\cLnri{0}{\infty}{\bfl}(\theta)=\displaystyle\frac{ 1 }{ M^{ \bfl }_0 } 
  \biggl[ 
  \textstyle\sum\limits_{ m = 1 }^{ M^{ \bfl }_0}  
    \abs{
      Q
      - 
      Y^m_0 
    }^2
  \biggr]\Biggr)\\
      &= \P\Biggl(\inf_{\theta\in \R^{\fd_\bfl}}\cLnri{0}{\infty}{\bfl}(\theta)=\inf _{Q\in \R}\biggl(\displaystyle\frac{ 1 }{ M^{ \bfl }_0 } 
  \biggl[ 
  \textstyle\sum\limits_{ m = 1 }^{ M^{ \bfl }_0} 
    \abs{
      Q
      - 
      Y^m_0 
    }^2
  \biggr]\biggr)\Biggr)\\
      &\leq \P\Big(\inf_{\theta\in \R^{\fd_\bfl}}\cLnri{0}{\infty}{\bfl}(\theta)=0\Big)\dott
      \end{split}
  \end{equation}}
 \argument{\lref{arg3}; the assumption that $\bfl=(\bfl_0,\bfl_1,\dots,\bfl_L)\in (\{d\}\times\{1,2,\dots,l\}^{L}\times \{1\})$; \cref{very deep: auxillary 2}}{that
  \begin{equation}\llabel{eq2}
      \begin{split}
           &\P\biggl( \inf_{n\in \N_0} \cLnri{0}{\infty}{\bfl}(\NNelll^{\bfl}_n)\leq\inf_{\theta\in \R^{\fd_\bfl}}\cLnri{0}{\infty}{\bfl}(\theta)\biggr)\\
           &\leq 1-\P
        \big(\forall n\in \N\colon \Theta_\bfl^n\in \InAct\big)+\P\Big(\inf_{\theta\in \R^{\fd_\bfl}}\cLnri{0}{\infty}{\bfl}(\theta)=0\Big)\\
        &\leq \big[1-[\P(\Theta_0^{\bfl,1}<0)]^{l(l+1)}\big]^{L-1}+\P\Big(\inf_{\theta\in \R^{\fd_\bfl}}\cLnri{0}{\infty}{\bfl}(\theta)=0\Big)\dott
      \end{split}
  \end{equation}}
\end{aproof}
\renewcommand{\bfl}{\ell}
\renewcommand{\nnode}{\ell}
\subsection{Non-convergence for general ANNs}\label{subsec: non-convergence: general}
\begin{athm}{cor}
{lem: estimate proba of nonconvergence: general case}
    Assume \cref{setting: SGD}, let $L\in \N\backslash\{1\}$, $l\in \N$, $\bfl\in (\{d\}\times\{1,2,\dots,l\}^{L}\times \{1\})$, $c_{1},c_2,\dots,c_{\fd_{\bfl}}\in(0,\infty)$, and assume that
$
  c_i\Theta^{ \bfl, i }_0 
$, 
$ i \in \{ 1, 2, \dots, \fd_\bfl \} $, 
are \iid \ Then
    \begin{equation}\label{conclude: lem: estimate proba of nonconvergence: general case}
    \begin{split}
      & \P\biggl( \inf_{n\in \N_0}  \cLnri{0}{\infty}{\bfl}(\NNelll^{\bfl}_n)\leq\inf_{\theta\in \R^{\fd_\bfl}}\cLnri{0}{\infty}{\bfl}(\theta)\biggr)\\
      &\leq \P\Big(\inf_{\theta\in \R^{\fd_\bfl}}\cLnri{0}{\infty}{\bfl}(\theta)=0\Big)+\min\!\Big\{\P\Bigl(\# \bigl(\inact_{ \bfl }^{ \NNelll_0^{\bfl}}\bigr)< 3\Bigr),\big[1-[\P(\Theta_0^{\bfl,1}<0)]^{l(l+1)}\big]^{L-1}\Big\}.
      \end{split}
    \end{equation}
\end{athm}
\begin{aproof}
  \argument{ \cref{lem: estimate proba of nonconvergence};\cref{lem: estimate proba of nonconvergence: L to infty}} {\eqref{conclude: lem: estimate proba of nonconvergence: general case}\dott}
\end{aproof}
\begin{athm}{cor}
{lem: estimate proba of nonconvergence: general case: rate of convergence 2}
   Assume \cref{setting: SGD}, for every $k\in \N$ let $l_k\in \N$, $L_k\in \N\backslash\{1\}$, $\nnode_k=(\nnode_k^0,\nnode_k^1,\dots,\allowbreak\nnode_k^{L_k})\allowbreak\in (\{d\}\times\{1,2,\dots,l_k\}^{L_k}\times\{1\}) $, 
   assume for all $k\in \N$ that $\P\big(\inf_{\theta\in \R^{\fd_{\bfl_k}}}\cLnri{0}{\infty}{\bfl_k} (\theta)>0\big)=1$, let $ \Dens \colon \R \to [0,\infty) $ be measurable, 
assume $\sup_{ \eta \in (0,\infty) } \bigl(  \eta ^{ - 1 } \int_{ - \eta }^{ \eta } \mathbbm1_{ \R \backslash { \{0\} } }( \Dens( x ) ) \, \d x \bigr) \geq 2$, let $(\cki_{k,i})_{(k,i)\in \N^2}\allowbreak\subseteq(0,\infty)$ 
satisfy for all $k\in \N
$,
$x_1,x_2,\dots,x_{\fd_{\bfl_k}} \in \R$
that
  \begin{equation}
      \P \bigl( \cap_{i=1}^{\fd_{\bfl_k}}
    \big\{
    \NNelll^{\bfl_k,i}_0 
    < x_i\big\}
  \bigr)
  =
  \prod_{i=1}^{\fd_{\bfl_k}}\biggl[\int_{ - \infty }^{\cki_{k,i}x_i} \Dens(y) \, \d y\biggr],
   \end{equation}
  and assume $\sup_{k\in \N}\max_{i\in \{1,2,\dots,\bfl_k^1\}}\allowbreak\max_{ j\in\{1,2,\dots,d\}}\allowbreak\bigl((c_{k,(i-1)d+j})^{-1}c_{k,\bfl_k^1 d+i}\bigr)\allowbreak<\infty$. Then there exists $\boundexp\in(0,\infty)$ which satisfies for all $k\in \N$ that
  \begin{equation}\llabel{conclude}
    \begin{split}
      & \P\biggl( \inf_{n\in \N_0} \cLnri{0}{\infty}{\bfl_k}(\NNelll^{\bfl_k}_n)\leq\inf_{\theta\in \R^{\fd_{\bfl_k}}}\cLnri{0}{\infty}{\bfl_k}(\theta)\biggr)\\
      &\leq \min\Bigl\{\boundexp^{-1}\exp\bigl(-\boundexp \bfl_k^{1}\bigr),\Bigl[1-\bigl[\textstyle\int_{-\infty}^0\Dens(y)\,\d y\bigr]^{l_k(l_k+1)}\Bigr]^{L_k-1}\Bigr\}.
      \end{split}
    \end{equation}
\end{athm}
\begin{aproof}
\argument{\cref{Lemma X};the assumption that for all $k\in\N$, $x_1,x_2,\dots,x_{\fd_{\bfl_k}} \in \R$
it holds that
  $\P \bigl( \cap_{i=1}^{\fd_{\bfl_k}}
    \big\{
    \NNelll^{\bfl_k,i}_0 
    < x_i\big\}
  \bigr)
  =
  \prod_{i=1}^{\fd_{\bfl_k}}\big(\int_{ - \infty }^{\cki_{k,i}x_i} \Dens(y) \, \d y\big)$}{that \llabel{arg1} 
\begin{enumerate}[label=(\roman*)]
     \item \llabel{item 1} it holds that $\int_{-\infty}^{\infty}\Dens(y)\,\d y=1$,
    \item \llabel{item 2} it holds for all $k\in \N$, $m\in \{1,2,\dots,\fd_{\bfl_k}\}$, $x\in \R$ that $\P(\Theta_0^{\bfl_k,m}<x)=\int_{-\infty}^{\cki_{k,m}x}\Dens(y)\,\d y$, and
    \item \llabel{item 3} it holds for all $k\in \N$ that $c_{k,i}\Theta_0^{\bfl_k,i}$, $i\in \{1,2,\dots,\fd_{\bfl_k}\}$, are \iid\ 
\end{enumerate}}
\startnewargseq
\argument{\lref{item 2, item 3};}{that for all $k\in \N$,
$ x_1,x_2,\dots,x_{\bfl_k^1d+\bfl_k^1} \in \R$ 
it holds that
\begin{equation}\llabel{eq0}
  \P\bigl( \cap_{i=1}^{\bfl_k^1d+\bfl_k^1}
    \{
    \NNelll^{\bfl_k,i}_0 
    < x_i\}
  \bigr)
  =
\prod_{i=1}^{\bfl_k^1d+\bfl_k^1}\biggl[\int_{ - \infty }^{\cki_{k,i}x_i} \Dens(y) \, \d y\biggr]\dott
\end{equation}}
\argument{\lref{eq0};\cref{cor:estimate dead neuron hidden layer 1.1}}{that there exists $\boundexp\in (0,\infty)$ which satisfies for all $k\in \N$ that
\begin{equation}\llabel{eq0.5}
    \P\Bigl(\# \bigl(\inact_{ \bfl_k }^{ \NNelll_0^{\bfl_k}}\bigr)< 3\Bigr)\leq \boundexp^{-1}\exp\bigl(-\boundexp \bfl_k^{1}\bigr)\dott
\end{equation}}
\startnewargseq
\argument{\lref{item 3}; \lref{eq0.5};\cref{lem: estimate proba of nonconvergence: general case}}{that for all $k\in \N$ it holds that
\begin{equation}\llabel{eq1}
    \begin{split}
      & \P\biggl( \inf_{n\in \N_0} \cLnri{0}{\infty}{\bfl_k}(\NNelll^{\bfl_k}_n)\leq\inf_{\theta\in \R^{\fd_{\bfl_k}}}\cLnri{0}{\infty}{\bfl_k}(\theta)\biggr)\\
      &\leq \P\Big(\inf_{\theta\in \R^{\fd_{\bfl_k}}}\cLnri{0}{\infty}{\bfl_k}(\theta)=0\Big)+\min\!\Big\{\boundexp^{-1}\exp\bigl(-\boundexp \bfl_k^{1}\bigr),\big[1-[\P(\Theta_0^{\bfl_k,1}<0)]^{l_k(l_k+1)}\big]^{L_k-1}\Big\}.
      \end{split}
\end{equation}}
    \argument{\lref{eq1};\lref{item 2}; the assumption that for all $k\in \N$ it holds that $\P\big(\inf_{\theta\in \R^{\fd_{\bfl_k}}}\cLnri{0}{\infty}{\bfl_k} (\theta)\allowbreak>0\big)=1$}{\lref{conclude}\dott}
\end{aproof}
\begin{athm}{prop}{conjecture: multilayer2'}
Assume \cref{setting: SGD},
 let $\f\colon\N\to \N$ be a function, for every $k\in \N$ let $L_k\in \N$, $\bfl_k =(\bfl_k^0,\bfl_k^1\dots,\bfl_k^{L_k}) \in (\{d\}\times\N^{L_k-1}\times\{1\})$ satisfy $\max\{\bfl_k^1,\bfl_k^2,\dots,\bfl_k^{L_k}\}\leq \f(\bfl_k^1)$,
 assume 
 \begin{equation}
 \textstyle
     \liminf_{k\to \infty} \fd_{\bfl_k}=\infty \qqandqq \liminf_{k \to \infty}\P\big(\inf_{\theta\in \R^{\fd_{\bfl_k}}}\cLnri{0}{\infty}{\bfl_k}(\theta) >0\big)=1,
 \end{equation} 
 let $\Dens\colon \R\to [0,\infty)$ be measurable, let $(\cki_{k,i})_{(k,i)\in \N^2}\allowbreak\subseteq(0,\infty)$ 
satisfy for all $k\in \N
$,
$x_1,x_2,\dots,x_{\fd_{\bfl_k}} \in \R$
that
  $\P \bigl( \cap_{i=1}^{\fd_{\bfl_k}}
    \big\{
    \NNelll^{\bfl_k,i}_0 
    < x_i\big\}
  \bigr)
  =
  \prod_{i=1}^{\fd_{\bfl_k}}\big(\int_{ - \infty }^{\cki_{k,i}x_i} \Dens(y) \, \d y\big)$, assume $\sup_{ \eta \in (0,\infty) } \bigl(  \eta ^{ - 1 } \int_{ - \eta }^{ \eta } \mathbbm 1_{ \R \backslash { \{0\} } }\allowbreak( \Dens( x ) )\allowbreak \, \d x \bigr) \geq 2$,
  and assume $\sup_{k\in \N}\max_{i\in \{1,2,\dots,\bfl_k^1\}}\max_{ j\in\{1,2,\dots,d\}}\bigl((c_{k,(i-1)d+j})^{-1}c_{k,\bfl_k^1 d+i}\bigr)<\infty$. Then
\begin{equation}\label{conclusion:conjecture4.1.2'}
\textstyle
\liminf\limits_{ k \to \infty } \P\biggl(\inf\limits_{n\in \N_0}  \cLnri{0}{\infty}{\bfl_k}(\NNelll^{\bfl_k}_n)>\inf\limits_{\theta\in \R^{\fd_{\bfl_k}}}\cLnri{0}{\infty}{\bfl_k}(\theta)\biggr)=1.
\end{equation}
\end{athm}
\begin{aproof}
  Throughout this proof let $F\colon\N\to\N$ satisfy for all $n\in \N$ that 
  \begin{equation}\llabel{def: F}
      \F(n)=\max_{m\in \{1,2,\dots,n\}}f(m).
  \end{equation} 
  \argument{\lref{def: F};}{that $\F$ is non-decreasing\dott}
  \argument{\lref{def: F}}{for all $n\in \N$ that \llabel{eqtg1} $f(n)\leq \F(n)$\dott}
  \argument{\lref{eqtg1}; the assumption that for all $k\in \N$ it holds that $\max\{\bfl_k^1,\bfl_k^2,\dots,\bfl_k^{L_k}\}\leq \f(\bfl_k^1)$}{that for all $k\in \N$ it holds that
  \begin{equation}\llabel{property: F}
      \max\{\bfl_k^1,\bfl_k^2,\dots,\bfl_k^{L_k}\}\leq \F(\bfl_k^1)\dott
  \end{equation}}
   \argument{\lref{property: F};the fact that for all $v,w\in \N$ it holds that $vw\geq \max\{v,w\}$;the assumption that $\liminf_{k\to\infty}\fd_{\bfl_k}=\infty$}{ that
   \begin{equation}\llabel{arg'1}
   \begin{split}
   &\liminf_{k\to\infty}\bigl(L_k\F(\bfl_k^1)\bigr)\geq \liminf_{k\to\infty}\max\{L_k,\F(
   \bfl_k^1
   )\} \\
   &\geq \liminf_{k\to\infty}\max\{L_k,\max\{\bfl_k^1,\bfl_k^2,\ldots,\bfl_k^{L_k}\}\}=\liminf_{k\to\infty}\max\{L_k,\bfl_k^1,\bfl_k^2,\ldots,\bfl_k^{L_k}\} =\infty\dott
   \end{split}
   \end{equation}}
   \argument{the assumption that for all $k\in \N$ it holds that $\ell_k^{L_k}=1$;}{that for all $k\in \N$ with $L_kF(\ell_k^1)> F(1)$ it holds that \llabel{argtt1} $L_k>1$\dott}
   \argument{\lref{argtt1};\lref{arg'1}}{that there exists $\kappa\in\N$ which satisfies for all $k\in \N\cap[\kappa,\infty)$ that
   \begin{equation}\llabel{def M}
       L_k>1\dott
   \end{equation}}
   \startnewargseq
   \argument{\cref{lemma X: item 2} in \cref{Lemma X};the assumption that for all $k\in \N
$,
$ x_1,x_2,\dots,x_{\fd_{\bfl_k}} \in \R$ 
it holds that
$
  \P \bigl( \cap_{i=1}^{\fd_{\bfl_k}}
    \{
    \NNelll^{\bfl_k,i}_0 
    < x_i\}
  \bigr)
  =
  \prod_{i=1}^{\fd_{\bfl_k}}\bigl(\int_{ - \infty }^{\cki_{k,i}x_i} \Dens(y) \, \d y\bigr)
$}{ that for all $k\in \N$, $i\in \{1,2,\dots,\fd_{\bfl_k}\}$ it holds that 
\begin{equation}\llabel{arg2.5}
    \P\big(\Theta_0^{\bfl_k,i}<0\big)=\int_{ - \infty }^{0} \Dens(y) \, \d y\dott
\end{equation}}
\argument{\lref{arg2.5}; the assumption that $\sup_{ \eta \in (0,\infty) } \bigl( \eta ^{ - 1 } \int_{ - \eta }^{ \eta } \mathbbm 1_{ \R \backslash {\{ 0\} } }( \Dens( x ) ) \, \d x \bigr) \geq 2$; \cref{lemma X: item 1} in \cref{Lemma X}}{ that there exists $p\in(0,1)$ which satisfies for all $k\in \N$, $i\in \{1,2,\dots,\fd_{\bfl_k}\}$ that 
\begin{equation}\llabel{arg2.75}\P\big(\Theta_0^{\bfl_k,i}<0\big)=p\dott
\end{equation}} 
\argument{\cref{cor:estimate dead neuron hidden layer 1};\lref{def M}}{that there exists $N=(N_\varepsilon)_{\varepsilon\in (0,\infty)}\colon\allowbreak(0,\infty)\to\{4,5,6,\dots\}$ which satisfies for all $\varepsilon\in (0,\infty)$, $k\in \N\cap [\kappa,\infty)$ with $\bfl_k^1>N_{\varepsilon}$ that \begin{equation}\llabel{arg0.5}
\P\Big(\# \bigl(\inact_{ \bfl_k }^{ \NNelll_0^{\bfl_k}}\bigr)< 3\Big)<\varepsilon\dott
 \end{equation}}
 \startnewargseq
 \argument{\lref{arg'1};\lref{def M}; \lref{arg2.75}; the assumption that $\liminf_{k \to \infty}\P\big(\inf_{\theta\in \R^{\fd_{\bfl_k}}}\cLnri{0}{\infty}{\bfl_k}(\theta) >0\big)=1$}{that 
there exist $K=(K_\varepsilon)_{\varepsilon\in (0,\infty)}\colon (0,\infty)\to \N$ and $P=(P_\varepsilon)_{\varepsilon\in (0,\infty)}\colon (0,\infty)\allowbreak\to (\N\cap[\kappa,\infty))$ which satisfy for all $\varepsilon\in (0,\infty)$ that
\begin{equation}\llabel{def:P}
\begin{split}
  \big(1-p^{\F(N_\varepsilon)(\F(N_\varepsilon)+1)}\big)^{\frac{K_\varepsilon}{\F(N_\varepsilon)}-2}<\varepsilon, \qquad\inf_{k\in \N\cap(P_\varepsilon,\infty)}\bigl(L_k\F(\bfl_k^1)\bigr)>K_\varepsilon\geq 4F(N_\varepsilon),
 \end{split}
 \end{equation}
 \begin{equation}\llabel{NRZZZ}
     \begin{split}
\text{and}\qquad \inf_{k\in\N\cap(P_\varepsilon,\infty)}\P\Big(\inf\nolimits_{\theta\in \R^{\fd_{\bfl_k}}}\cLnri{0}{\infty}{\bfl_k}(\theta)>0\Big)> 1-\varepsilon\dott
 \end{split}
 \end{equation}}
 \startnewargseq
\argument{\lref{def M};\lref{arg0.5};\lref{def:P};\lref{NRZZZ};\cref{lem: estimate proba of nonconvergence}} {that for all $\varepsilon\in (0,\infty)$, $k\in \N\cap(P_\varepsilon,\infty)$ with $\bfl_k^1>N_\varepsilon$ it holds that 
  \begin{equation}\llabel{case 1}
  \begin{split}
  &\P\biggl(\inf_{n\in \N_0}  \cLnri{0}{\infty}{\bfl_k}(\NNelll^{\bfl_k}_n)\leq\inf\limits_{\theta\in \R^{\fd_{\bfl_k}}}\cLnri{0}{\infty}{\bfl_k}(\theta)\biggr) 
\\&\leq \P\Big(\inf_{\theta\in \R^{\fd_{\bfl_k}}}\cLnri{0}{\infty}{\bfl_k}(\theta)=0\Big)+\P\Big(\# \bigl(\inact_{ \bfl_k }^{ \NNelll_0^{\bfl_k}}\bigr)< 3\Big)<2\varepsilon\dott
   \end{split}
  \end{equation}}
  \argument{the fact that $\F$ is non-decreasing; \lref{def:P}}{that for all $\varepsilon\in (0,\infty)$, $k\in\N\cap(P_\varepsilon,\infty)$ with $\bfl_k^1\leq N_\varepsilon$ it holds that 
  \begin{equation}\llabel{arg2}
  L_k\geq \frac{K_\varepsilon}{\F(\bfl_k^1)}\geq\frac {K_\varepsilon}{\F(N_\varepsilon)}\geq 4\dott
  \end{equation}} 
  \argument{\lref{def M};\lref{arg2.75};\lref{arg2};\cref{lem: estimate proba of nonconvergence: general case};\cref{lemma X: item 3} in \cref{Lemma X}; \lref{def:P};\lref{NRZZZ}; the fact that $\F$ is non-decreasing}{\llabel{arg3} that for all $\varepsilon\in (0,\infty)$, $k\in \N\cap(P_\varepsilon,\infty)$ with $\bfl_k^1\leq N_\varepsilon$ it holds that 
  \begin{equation}\llabel{case 2}
  \begin{split}
  &\P\biggl(\inf_{n\in \N_0}  \cLnri{0}{\infty}{\bfl_k}(\NNelll^{\bfl_k}_n)\leq\inf\limits_{\theta\in \R^{\fd_{\bfl_k}}}\cLnri{0}{\infty}{\bfl_k}(\theta)\bigg) \\
  &\leq\P\Big(\inf_{\theta\in \R^{\fd_{\bfl_k}}}\cLnri{0}{\infty}{\bfl_k}(\theta)=0\Big)+\min\!\Big\{\big[1-[\P(\Theta_0^{\bfl_k,1}<0)]^{\F(\bfl_k^1)(\F(\bfl_k^1)+1)}\big]^{L_k-2},\P\Big(\# \bigl(\inact_{ \bfl_k }^{ \NNelll_0^{\bfl
_k}}\bigr)< 3\Big)\Big\}\\
  &\leq \P\Big(\inf_{\theta\in \R^{\fd_{\bfl_k}}}\cLnri{0}{\infty}{\bfl_k}(\theta)=0\Big)+\big[1-[\P(\Theta_0^{\bfl_k,1}<0)]^{\F(\bfl_k^1)(\F(\bfl_k^1)+1)}\big]^{L_k-2}\\
  &\leq \varepsilon+ [1-p^{\F(N_\varepsilon)(\F(N_\varepsilon)+1)}]^{L_k-2}\leq \varepsilon+ [1-p^{\F(N_\varepsilon)(\F(N_\varepsilon)+1)})]^{\frac{K_\varepsilon}{\F(N_\varepsilon)}-2}<2\varepsilon\dott
   \end{split}
  \end{equation}}
  \argument{\lref{case 1};\lref{case 2}}{that for all $\varepsilon\in (0,\infty)$, $k\in \N\cap(P_\varepsilon,\infty)$ it holds that 
  \begin{equation}\llabel{P<epsilon}
       \P\biggl(\inf_{n\in \N_0}  \cLnri{0}{\infty}{\bfl_k}(\NNelll^{\bfl_k}_n)\leq\inf\limits_{\theta\in \R^{\fd_{\bfl_k}}}\cLnri{0}{\infty}{\bfl_k}(\theta)\biggr)<2\varepsilon\dott
  \end{equation}}
  \argument{\lref{P<epsilon}}[verbs=eps]{ \eqref{conclusion:conjecture4.1.2'}\dott}
  \end{aproof}
  \begin{athm}{lemma}{lemma f}
     Let $(a_n)_{n\in \N} \subseteq \R$ and $(b_n)_{n\in\N} \subseteq\R$ satisfy $\liminf_{n\to\infty} \aaa_n=\infty$. Then there exists a non-decreasing $f\colon \R\to \N$ such that for all $n\in \N$ it holds that 
     \begin{equation}\llabel{conclude}
         \bb_n\leq f(\aaa_n).
     \end{equation}
  \end{athm}
  \begin{aproof}
  \argument{the assumption that $\liminf_{n\to\infty}\aaa_n=\infty$}{that there exists $m\in\N$ which satisfies
  \begin{equation}\llabel{eq0}
      \aaa_m=\inf_{n\in \N}\aaa_n=\min_{n\in \N}\aaa_n\dott
  \end{equation}}
  \startnewargseq
      \argument{the assumption that $\liminf_{n\to\infty}\aaa_n=\infty$;}{that for all $x\in [\aaa_m,\infty)$ it holds that \llabel{arg1} $0<\#(\{n\in \N\colon \aaa_n\leq x\})<\infty$\dott}
      \argument{\lref{arg1};}{that there exists $f\colon \R\to\N$ which satisfies for all $x\in \R$ that
      \begin{equation}\llabel{eq1}
    f(x)=\begin{cases}
        f(\aaa_m)&\colon x<\aaa_m\\
        \max\bigl(\cup_{v\in \{n\in \N\colon \aaa_n\leq x\}}\{\lfloor|\bb_v|+1\rfloor\}\bigr)&\colon x\geq \aaa_m.
    \end{cases}
    \end{equation}}
    \startnewargseq
      \argument{\lref{eq0};\lref{eq1}; the fact that for all $x\in \R$ it holds that $\lfloor|x|+1\rfloor\geq |x|\geq x$}[verbs=spd]{ that for all $N\in \N$ it holds that
      \begin{equation}\llabel{eq2}
      \begin{split}
          f(\aaa_N)=\max\bigl(\cup_{v\in \{n\in \N\colon \aaa_n\leq \aaa_N\}}\{\lfloor|\bb_v|+1\rfloor\}\bigr)\geq \max\bigl(\cup_{v\in \{n\in \N\colon \aaa_n\leq \aaa_N\}}\{\bb_v\}\bigr)\geq \bb_N\dott
          \end{split}
      \end{equation}}
      \argument{\lref{eq1}; the fact that for all $x,y\in \R$ with $x\leq y$ it holds that $\{n\in \N\colon b_n\leq x\}\subseteq \{n\in \N\colon b_n\leq y\}$}{that for all $x,y\in \R$ with $b_m\leq x\leq y$ it holds that
      \begin{equation}\llabel{eq3}
      \begin{split}
          &f(b_m)\leq f(x)=\max\bigl(\cup_{v\in \{n\in \N\colon \aaa_n\leq x\}}\{\lfloor|\bb_v|+1\rfloor\}\bigr)\\
          &\leq\max\bigl(\cup_{v\in \{n\in \N\colon \aaa_n\leq y\}}\{\lfloor|\bb_v|+1\rfloor\}\bigr)=f(y) \dott
          \end{split}
      \end{equation}}
      \argument{\lref{eq3};\lref{eq1}}{that $f$ is non-decreasing \llabel{arg2}\dott} 
\argument{\lref{arg2};\lref{eq2}}[verbs=ep]{\lref{conclude}\dott}
  \end{aproof}
  \begin{athm}{theorem}{conjecture: multilayer2}
Let $d\in\N$,
$ a \in \R $, 
$ b \in (a, \infty)  $, let $ ( \Omega, \mathcal{F}, \P) $ be a probability space, for every $ m, n \in \N_0 $ 
let 
$ X^m_n \colon \Omega \to [a,b]^d $
and 
$ Y^m_n \colon \Omega \to \R $
be random variables, assume for all $i\in \N$, $j\in \N\backslash\{i\}$ that $\P( X_0^i=X_0^j)=0$, let $\f\colon\N\to \N$ be a function, for every $k\in \N_0$ let $\fd_k,L_k\in \N$, $\bfl_k =(\bfl_k^0,\bfl_k^1\dots,\bfl_k^{L_k}) \in \N^{L_k+1}$ satisfy $\bfl_k^0=d$, $\bfl_k^{L_k}=1$, $\fd_k=\sum_{ i = 1 }^{L}\bfl_k^i ( \bfl_k^{i-1} + 1 )$, and $\max\{\bfl_k^1,\bfl_k^2,\dots,\bfl_k^{L_k}\}\leq \f(\bfl_k^1)$, let $\scrA\in (0,\infty)$, $\scrB\in (\scrA,\infty)$, let $\mathbb{A}_r\colon \mathbb{R} \rightarrow \mathbb{R}$, $r \in [1,\infty]$, satisfy for all $r\in [1,\infty)$, $x\in (-\infty,\scrA r^{-1}]\cup[\scrB r^{-1},\infty)$, $y\in \R$ that
\begin{equation}\label{A_r1}
    \mathbb A_r\in C^1(\R,\R),\quad \mathbb A_r(x)=x\mathbbm 1_{(\scrA r^{-1},\infty)}(x),\qandq 0\leq \mathbb A_r(y)\leq \mathbb A_\infty(y)=\max\{y,0\},
    \end{equation}
assume $
  \sup_{ r \in [1,\infty) }
  \sup_{ x \in \R } 
  | ( \mathbb A_r )'( x ) | < \infty 
$, for every $r\in [1,\infty]$, $k\in\N_0$, $\theta=(\theta_1,\dots,\theta_{\fd_{k}})\in\R^{\fd_k}$ let $\mN^{ v, \theta }_{r,k}=(\mN^{ v, \theta }_{r,k,1},\dots,\mN^{ v, \theta }_{r,k,\bfl_k^v}) \colon \R^{ d } \to \R^{ \bfl_k^v} $, $v \in \{0,1,\dots,L_k\}$, satisfy for all $v\in \{0,1,\dots,L_k-1\}$, $x=(x_1,\dots, x_{d})\in \R^{d}$, $i\in\{1,2,\dots,\bfl_k^{v+1}\}$ that
\begin{multline}
  \mN^{v+1, \theta }_{r,k,i}( x ) = \theta_{\bfl_k^{v+1}\bfl_k^{v}+i+\sum_{h=1}^{v}\bfl_k^h(\bfl_k^{h-1}+1)}+\sum\limits_{j=1}^{\bfl_k^{v}}\theta_{(i-1)\bfl_k^{v}+j+\sum_{h=1}^{v}\bfl_k^h(\bfl_k^{h-1}+1)}\big(x_j\indicator{\{0\}}(v) \\ 
  +\mathbb A_{r^{1/\!\max\{v,1\}}}(\mN^{v,\theta}_{r,k,j}(x))\indicator{\N}(v) 
    \big),
\end{multline}
for every $k,n\in \N_0$ let $ M^{ k }_n \in  \N $, for every $r\in [1,\infty]$, $k,n\in \N_0$ let
$ 
  \cLnri{n}{r}{k} \colon \R^{ \fd_{ k } } \times \Omega \to \R 
$
satisfy for all 
$ \theta \in \R^{ \fd_{ k} }$
that
\begin{equation}
\displaystyle
  \cLnri{n}{r}{k}( \theta) 
  = 
  \frac{ 1 }{ M^{ k }_n } 
  \biggl[ \textstyle
  \sum\limits_{ m = 1 }^{ M^{ k }_n}  
    \abs{
      \mN^{ L_k,\theta }_{r,k}( X^m_n )
      - 
      Y^m_n 
    }^2
  \biggr]
  ,
\end{equation}
 for every $k,n\in \N_0$ let 
$ 
  \fG_n^{ k }  
  \colon \R^{ \fd_{ k} } \times \Omega \to \R^{ \fd_{ k } } 
$ 
satisfy for all $\omega\in \Omega$, $\theta\in \{\vartheta\in \R^{\fd_k}\colon (\nabla_{\vartheta} \cLnri{n}{r}{k}(\vartheta,\omega))_{r\in [1,\infty)}$ is convergent$\}$
that
\begin{equation}
\label{eq:gradients_SGD4}
  \fG^{k}_n( \theta,\omega) 
  = 
  \lim_{r\to\infty}\bigl[\nabla_\theta \cLnri{n}{r}{k}(\theta,\omega)\bigr],
\end{equation}
let 
	$
	\Theta_n^k =(\Theta_n^{k,1},\dots,\Theta_n^{k,\fd_{k}})
	\colon \Omega  \to \R^{\fd_{ k } }
	$
	be a random variable, and let 
	$
	\Phi_n ^k
	= 
	( 
	\Phi^{k, 1 }_n, \dots, 
	\Phi^{ k , \fd_{ k} }_n 
	)
	\colon 
	\allowbreak
	( \R^{ \fd_{k } } )^{ n + 1 }
	\allowbreak
	\to 
	\R^{ \fd_{k } }
	$ 
	satisfy 
	for all 
	$
	g =
	( 
	( g_{ i, j } )_{ j \in \{ 1, 2, \dots, \fd_{ k } \} }
	)_{
		i \in \{ 0, 1, \dots, n \}
	}
	\in 
	(
	\R^{ 
		\fd_{ k }
	}
	)^{ n + 1 }
	$, 
	$ 
	j \in \{1,2,\dots,\fd_k\}  
	$
	with 
	$
	\sum_{ i = 0 }^n
	\abs{ g_{ i, j } }
	= 0
	$
	that 
	$
	\Phi^{k , j }_n( g ) = 0 
	$,
	 assume for all $k,n\in \N$ that 
	\begin{equation}
		\label{eq: setting: SGD_def process_main theorem 2}
		\Theta_{ n  } ^k
		= 
		\Theta_{n-1} ^k - 
		\Phi_{n-1}^k \bigl(
		\fG_1^k( \Theta_0^k  ) ,
		\fG_2^k ( \Theta_1^k  ) ,
		\dots ,
		\fG_n^k ( \Theta_{n-1}^k )
		\bigr),
	\end{equation} 
  assume $\liminf_{k\to \infty} \fd_{k}=\infty$ and $\liminf_{k \to \infty}\allowbreak \P\big(\inf_{\theta\in \R^{\fd_{k}}}\cL_0^{\infty,k}(\theta) >0\big)=1$, let $\Dens\colon \R\to [0,\infty)$ be measurable, assume $\sup_{ \eta \in (0,\infty) } \bigl(  \eta ^{ - 1 } \int_{ - \eta }^{ \eta } \mathbbm 1_{ \R \backslash { \{0\} } }\allowbreak( \Dens( x ) ) \, \d x \bigr) \geq 2$, let $(\cki_{k,i})_{(k,i)\in \N^2}\subseteq(0,\infty)$ 
satisfy for all $k\in \N
$,
$x_1,x_2,\dots,x_{\fd_{k}} \in \R$ 
that
$
  \P \bigl( \cap_{i=1}^{\fd_{k}}
    \big\{
    \NNelll^{k,i}_0 
    < x_i\big\}
  \bigr)
  =
  \prod_{i=1}^{\fd_{k}}\big(\int_{ - \infty }^{\cki_{k,i}x_i} \Dens(y) \, \d y\big)
$, and assume $\sup_{k\in \N}\max_{i\in \{1,2,\dots,\bfl_k^1\}}\max_{ j\in\{1,2,\dots,d\}}\bigl((c_{k,(i-1)d+j})^{-1}c_{k,\bfl_k^1 d+i}\bigr)<\infty$. Then
\begin{equation}\label{conclusion:conjecture4.1}
\textstyle
\liminf\limits_{ k \to \infty } \P\biggl(\inf\limits_{n\in \N_0}  \cLnri{0}{\infty}{k}(\NNelll^{k}_n)>\inf\limits_{\theta\in \R^{\fd_k}}\cLnri{0}{\infty}{k}(\theta)\biggr)=1.
\end{equation}
\end{athm}
\begin{aproof}
    \argument{\cref{conjecture: multilayer2'};}{\cref{conclusion:conjecture4.1}\dott}
\end{aproof}
\subsection{Non-convergence to global minimum points}\label{subsec: non convergence to globmin point}
 \begin{athm}{cor}{cor: conjecture: multilayer}
Assume \cref{setting: SGD},
 let $f\colon\N\to \N$ be a function, for every $k\in\N$ let $L_k\in \N$, $\bfl_k =(\bfl_k^0,\bfl_k^1,\dots,\bfl_k^{L_k}) \in \N^{L_k+1}$ satisfy $\bfl_k^0=d$, $\bfl_k^{L_k}=1$, and $\max_{i\in \{1,2,\dots,L_k\}}\bfl_k^i\leq f(\bfl_k^1)$ and let $\Xi_k\colon \Omega \rightarrow \mathbb{R}^{\fd_{\bfl_k}}$ be a random variable which satisfies
\begin{equation}\llabel{def: Xi}
\Xi_k \in \operatorname{argmin}_{\theta \in \mathbb{R}^{\fd_{\bfl_k}}}\cLnri{0}{\infty}{\bfl_k} (\theta),
\end{equation}
 assume $\liminf_{k\to \infty} \fd_{\bfl_k}=\infty$ and $\liminf_{k \to \infty}\P(\cLnri{0}{\infty}{\bfl_k} (\Xi_k)>0)=1$, let $\Dens\colon \R\to [0,\infty)$ be measurable, assume $\sup_{ \eta \in (0,\infty) } \bigl(  \eta ^{ - 1 } \int_{ - \eta }^{ \eta } \mathbbm 1_{ \R \backslash {\{ 0\} } }( \Dens( x ) ) \, \d x \bigr) \geq 2$, let $(\cki_{k,i})_{(k,i)\in \N^2}\subseteq(0,\infty)$ 
satisfy for all $k\in \N
$,
$x_1,x_2,\dots,x_{\fd_{\bfl_k}} \in \R$
that
$
  \P \bigl( \cap_{i=1}^{\fd_{\bfl_k}}
    \big\{
    \NNelll^{\bfl_k,i}_0 
    <x_i\big\}
  \bigr)
  =
  \prod_{i=1}^{\fd_{\bfl_k}}\big(\int_{ - \infty }^{ \cki_{k,i}x_i} \Dens(y) \, \d y\big)
$, and assume $\sup_{k\in \N}\max_{i\in \{1,2,\dots,\bfl_k^1\}}\max_{ j\in\{1,2,\dots,d\}}\bigl((c_{k,(i-1)d+j})^{-1}c_{k,\bfl_k^1 d+i}\bigr)<\infty$.
Then 
\begin{equation}\label{conclusion:cor:conjecture4.1}
\textstyle
\limsup\limits_{ k \to \infty } \P\bigg(
  \limsup\limits_{ n\to\infty } \| \Theta^{\bfl_k}_n-\Xi_k \|= 0
  \bigg) = 0.
\end{equation}
\end{athm}
\begin{aproof}
    \argument{the fact that for all $\omega\in \Omega$, $k\in \N$ it holds that $\R^{\fd_{\bfl_k}}\ni\theta\mapsto\cLnri{0}{\infty}{\bfl_k}(\theta,\omega)\in \R$ is continuous;}{ that for all $k\in \N$ it holds that
    \begin{equation}\llabel{arg1}
    \begin{split}
       &  \P\bigg(
  \limsup\limits_{ n\to\infty } \| \Theta^{\bfl_k}_n-\Xi_k \|= 0
  \bigg)
  \leq \P\bigg(\textstyle
  \limsup\limits_{ n\to\infty }|\cLnri{0}{\infty}{\bfl_k}(\NNelll^{\bfl_k}_n)-\cLnri{0}{\infty}{\bfl_k}(\Xi_k)|= 0\bigg)\\
&\leq \P\biggl(\liminf\limits_{ n\to\infty }  \cLnri{0}{\infty}{\bfl_k}(\NNelll^{\bfl_k}_n)\leq\cLnri{0}{\infty}{\bfl_k}(\Xi_k)\biggr)\leq \P\biggl(\inf\limits_{n\in \N_0}  \cLnri{0}{\infty}{\bfl_k}(\NNelll^{\bfl_k}_n)\leq\cLnri{0}{\infty}{\bfl_k}(\Xi_k)\biggr)\dott
  \end{split}
    \end{equation}}
    \argument{\cref{conjecture: multilayer2'}; \lref{def: Xi}}
    {that
    \begin{equation}\llabel{arg2}
    \begin{split}
    &\limsup_{k\to\infty}\P\biggl(\inf_{n\in \N_0}  \cLnri{0}{\infty}{\bfl_k}(\NNelll^{\bfl_k}_n)\leq\cLnri{0}{\infty}{\bfl_k}(\Xi_k)\biggr)\\
    &=
        \limsup_{k\to\infty}\P\biggl(\inf_{n\in \N_0}  \cLnri{0}{\infty}{\bfl_k}(\NNelll^{\bfl_k}_n)\leq\inf\limits_{\theta\in \R^{\fd_{\bfl_k}}}\cLnri{0}{\infty}{\bfl_k}(\theta)\biggr)=0\dott
        \end{split}
    \end{equation}}
\argument{\lref{arg2}; \lref{arg1}}
    {\cref{conclusion:cor:conjecture4.1}\dott}
\end{aproof}
\section{SGD optimization methods}\label{sec: SGD optimization methods}
In this section we show that the non-convergence result for general \SGD\ optimization methods in \cref{conjecture: multilayer2} in \cref{sec: non convergence of SGD method} above (cf.\ \cref{eq:gradients_SGD4}--\cref{eq: setting: SGD_def process_main theorem 2} in \cref{conjecture: multilayer2} and \cref{setting: multilayer: eq:gradients_SGD4:}--\cref{eq: setting: SGD_def process} in \cref{setting: SGD}), in particular, applies
\begin{itemize}
\item to the momentum \SGD\ optimizer (see \cref{draf: momentum} in \cref{subsec: momentum SGD}),
\item  to the plain vanilla standard \SGD\ optimizer (see \cref{draf: standard SGD} in \cref{subsec: standard SGD}),  
\item to the Nesterov accelerated \SGD\ optimizer (see \cref{draf: nesterov} in \cref{subsec: nesterov}),
\item to the \Adagrad\ optimizer (see \cref{draf adagrad} in \cref{subsec: Adagrad}),
\item to the \RMSprop\ optimizer (see \cref{draf RMS prop} in \cref{subsec: RMSprop}),
\item to the \Adam\ optimizer (see \cref{draf adam} in \cref{subsec: Adam}),
\item to the \Adamax\ optimizer (see \cref{draf adamax} in \cref{subsec: Adamax}),
\item to the AMSGrad optimizer (see \cref{draf amsgrad} in \cref{subsec: AMSgrad}), and
\item to the \Nadam\ optimizer (see \cref{draf nadam} in \cref{subsec: Nadam}).
\end{itemize}
In the case of the momentum \SGD\ optimization method (see \cref{draf: momentum}) and in the case of the \Adam\ optimizer (see \cref{draf adam}), 
these facts have (in a slightly weaker form) already been established
in \cite[Lemma 4.8 and Lemma 4.9]{ArAd2024}.

In \cref{subsec: draf genenral} we first consider a general class of \SGD\ optimization methods and link one-step recursions for such methods with full history recursive descriptions of such methods (cf.\ \cref{general SGD pre} in \cref{subsec: draf genenral} and, \eg, \cite[Section 6]{Beckerlearning2023}) and also provide sufficient conditions that ensure that the assumption in \cref{conjecture: multilayer2} on the functions providing the full history recursive formulations of the considered \SGD\ methods (cf.\ \cref{eq:gradients_SGD4}--\cref{eq: setting: SGD_def process_main theorem 2} in \cref{conjecture: multilayer2} and \cref{setting: multilayer: eq:gradients_SGD4:}--\cref{eq: setting: SGD_def process} in \cref{setting: SGD}) is satisfied in the context of the methods considered in \cref{subsec: draf genenral}.

We also would like to point out that even though \cref{conjecture: multilayer2} and the framework 
in \cref{setting: SGD}, respectively, apply to a large class of \SGD\ optimization methods 
and, in particular, to each of the above named popular optimizers, there are also some 
well-known optimizers that are not covered by \cref{conjecture: multilayer2} and \cref{setting: SGD}, respectively. 
Specifically, \cref{conjecture: multilayer2} and \cref{setting: SGD}, respectively, do not apply to the popular Adam-W optimizer (cf.\ \cite{Loshchilov2017DecoupledWD}) 
and also not to the Adadelta optimizer (cf.\ \cite{zeiler2012adadeltaadaptivelearningrate}).

\subsection{General SGD optimization methods}\label{subsec: draf genenral}
In the elementary result in \cref{general SGD pre} below we link one-step recursions 
of a general class of \SGD\ optimization methods with full history recursive descriptions 
of such schemes (cf., \eg, \cite[Section 6]{Beckerlearning2023}) and, thereafter, in the elementary result in \cref{general SGD} below 
we provide sufficient conditions to ensure that the assumption in \cref{conjecture: multilayer2} on the functions 
providing the full history recursive descriptions of the considered \SGD\ optimization methods (cf.\ \cref{eq:gradients_SGD4}--\cref{eq: setting: SGD_def process_main theorem 2} in \cref{conjecture: multilayer2} and \cref{setting: multilayer: eq:gradients_SGD4:}--\cref{eq: setting: SGD_def process} in \cref{setting: SGD}) is fulfilled.

\begin{athm}{prop}{general SGD pre}
    Let $\fd,\numf\in \N$, let $\fnormal_n \colon \R^{\numf}\times \R^\fd\to \R^\numf$, $n\in \N$, and $\fcapital_n \colon (\R^{\fd})^{n}\to \R^\numf$, $n\in \N$, satisfy for all $n\in \N\backslash\{1\}$, $g=(g_1,\dots,g_n)\in (\R^{\fd})^{n}$ that 
     \begin{equation}\llabel{def: F}
     \begin{split}
\fcapital_1(g_1)=\fnormal_1(0,g_1) \qqandqq \fcapital_{n}(g)=\fnormal_{n}\bigl(\fcapital_{n-1}(g_1,g_2,\dots,g_{n-1}),g_n\bigr),
    \end{split}
     \end{equation}
      let $(\Omega,\mathcal F,\P)$ be a probability space, for every $n\in \N$ let $\fG_n\colon \R^\fd\times \Omega\to \R^\fd$, $\fbold_n\colon \R^{\numf}\times\R^\fd\to \R^\fd$, and $\Phi_n\colon (\R^\fd)^{n}\to\R^\fd$ satisfy for all $g=(g_1,\dots,g_{n}) 
	\in 
	(
	\R^{ 
		\fd
	}
	)^{ n  }
	$ that
 \begin{equation}\llabel{def: Phi}
     \Phi_n(g)=\fbold_{n}(\fcapital_{n}(g),g_{n}),
 \end{equation}
and let $\bfM\colon \N_0\times\Omega\to\R^\numf$ and $\Theta\colon \N_0\times\Omega\to \R^\fd$ satisfy for all $n\in \N$ that
\begin{equation}\llabel{eq2}
    \bfM_0=0, \qquad  \bfM_n=\fcapital_{n}(\fG_1(\Theta_0),\fG_2(\Theta_1),\dots,\fG_n(\Theta_{n-1})),
\end{equation}
\begin{equation}\llabel{def: M}
    \text{and}\qquad \Theta_{n} = \Theta_{n-1} - \Phi_{n} ( \fG_1 ( \Theta_0 ) , \fG_2 ( \Theta_1 ) , \ldots, \fG_n ( \Theta_{n-1} )).
\end{equation}
  Then
   it holds for all 
	$n \in \N$
	that 
	 \begin{equation}\llabel{conclude}
  \begin{split}
   \bfM_0 = 0,\quad
	 			\bfM_n  = \fnormal_{n}(\bfM_{n-1},\fG_n(\Theta_{n-1})),\quad
	 	\text{and}\quad &\Theta_{n} = \Theta_{n-1} -\fbold_{n}(\bfM_{n},\fG_n(\Theta_{n-1})). 
     \end{split}
     \end{equation}
\end{athm}
\begin{aproof}
    \argument{\lref{def: F};\lref{eq2};\lref{def: M}}{for all $n\in \N\backslash\{1\}$ that
 \begin{equation}\llabel{EQ5}
 \begin{split}
\bfM_n&=\fcapital_{n}(\fG_1(\Theta_0),\fG_2(\Theta_1),\dots,\fG_n(\Theta_{n-1}))
     \\
     &=\fnormal_{n}\bigl(\fcapital_{n-1}\bigl(\fG_1(\Theta_0),\fG_2(\Theta_1),\dots,\fG_{n-1}(\Theta_{n-2})\bigr),\fG_n(\Theta_{n-1})\bigr)
     \\&= \fnormal_{n}(\bfM_{n-1},\fG_n(\Theta_{n-1}))
     \end{split}
 \end{equation}
 and
 \begin{equation}\llabel{EQ6}
   \bfM_1 =\fcapital_1(\fG_1(\Theta_0))= \fnormal_1(0,\fG_1(\Theta_0))=\fnormal_1\bigl(\bfM_0,\fG_{1}(\Theta_0)\bigr)\dott
 \end{equation}}
 \argument{\lref{def: Phi};\lref{eq2};\lref{def: M}}{for all $n\in \N$ that
 \begin{equation}\llabel{EQ7}
 \begin{split}
     &\Theta_{n} = \Theta_{n-1} - \Phi_{n} ( \fG_1 ( \Theta_0 ) , \fG_2 ( \Theta_1 ) , \ldots, \fG_n ( \Theta_{n-1} ))\\
     &=\Theta_{n-1}-\fbold_{n}(\fcapital_{n}(\fG_1(\Theta_0),\fG_2(\Theta_1),\dots,\fG_{n}(\Theta_{n-1})),\fG_{n}(\Theta_{n-1}))\\
     &=\Theta_{n-1} -\fbold_{n}(\bfM_{n},\fG_{n}(\Theta_{n-1}))\dott
      \end{split}
 \end{equation}}
 \argument{\lref{EQ7};\lref{eq2};\lref{EQ5};\lref{EQ6}}{\lref{conclude}\dott}
\end{aproof}
\begin{athm}{cor}{general SGD}
    Let $\fd,\numf\in \N$, for every $n,m\in \N_0$ let $\fnormal_n^m=(\fnormal_{n,1}^m,\dots,\fnormal_{n,\fd}^m)\colon (\R^{\fd})^{\numf+1}\to \R^{\fd}$ satisfy for all  $g=((g_{i,j})_{j\in \{1,2,\dots,\fd\}})_{i\in \{0,1,\dots,\numf\}}\in (\R^\fd)^{\numf+1}$, $j\in \{1,2,\dots,\fd\}$ with $\sum_{i=0}^{\numf}|g_{i,j}|=0$ that 
    \begin{equation}\llabel{def: f}
    \fnormal_{n,j}^m(g)=0,
    \end{equation}
     let $\fcapital_n^m\colon (\R^{\fd})^{n+1}\to \R^\fd$, $n,m
     \in \N_0$, satisfy for all $n\in \N$, $m\in \{1,2,\dots,\numf\}$, $x=(x_1\dots,x_{n})\in (\R^{\fd})^{n}$, $y\in \R^\fd$ that $\fcapital_0^m(y)=\fnormal_0^m(0,0,\dots, 0,y)$ and
     \begin{equation}\llabel{def: F}
     \begin{split}
   \fcapital_n^{m}(x_1,x_2,\dots,x_n,y)=\fnormal_n^m\bigl(\fcapital_{n-1}^{1}(x),\fcapital_{n-1}^{2}(x),\dots,
    \fcapital_{n-1}^{\numf}(x),y\bigr),
    \end{split}
     \end{equation}
     let $(\Omega,\mathcal F,\P)$ be a probability space, for every $n\in \N_0$ let $\fbold_n=(\fbold_{n,1},\dots,\fbold_{n,\fd})\colon (\R^{\fd})^{\numf+1}\to \R^\fd$ satisfy for all  $g =
	( 
	( g_{ i, j } )_{ j \in \{ 1, 2, \dots, \fd \} }
	\allowbreak)_{
		i \in \{ 0, 1, \dots, \numf \}
	}
	\in 
	(
	\R^{ 
		\fd
	}
	)^{ \delta+ 1 }
	$, 
	$ 
	j \in \{1,2,\dots,\fd\}  
	$ with $\sum_{i=0}^{\numf}|g_{i,j}|=0$ that
        
     \begin{equation}\llabel{def: bff}
         \fbold_{n,j}(g)=0,
     \end{equation} 
      let $\Phi_n=(\Phi_n^1,\dots,\Phi_n^\fd)\colon (\R^\fd)^{n+1}\to\R^\fd$ satisfy for all $g=(g_0,g_1,\dots,g_{n}) 
	\in 
	(
	\R^{ 
		\fd
	}
	)^{ n + 1 }
	$ that
 \begin{equation}\llabel{def: Phi}
     \Phi_n(g)=\fbold_{n}(\fcapital_n^1(g),\fcapital_n^2(g),\dots,\fcapital_n^\numf(g),g_{n}),
 \end{equation}
and let $\fG_n=(\fG_n^1,\ldots,\fG_n^\fd)\colon \R^\fd\times \Omega\to \R^\fd$ be a function, and let $\Theta\colon \N_0\times\Omega\to \R^\fd$ satisfy for all $n\in \N$ that
\begin{equation}\llabel{eq2}
    \Theta_{n} = \Theta_{n-1} - \Phi_{n-1} ( \fG_1 ( \Theta_0 ) , \fG_2 ( \Theta_1 ) , \ldots, \fG_n ( \Theta_{n-1} )).
\end{equation}
  Then
    \begin{enumerate}[label=(\roman*)]
        \item \llabel{item 1} it holds for all $n\in\N_0$, $
	g =
	( 
	( g_{ i, j } )_{ j \in \{ 1, 2, \dots, \fd \} }
	)_{
		i \in \{ 0, 1, \dots, n \}
	}
	\in 
	(
	\R^{ 
		\fd
	}
	)^{ n + 1 }
	$, 
	$ 
	j \in \{1,2,\dots,\fd\}  
	$ with $\sum_{i=0}^n|g_{i,j}|=0$ that
        $\Phi_n^j(g)=0$ and
        \item \llabel{item 2} it holds that there exist $\bfM^m =(\bfM^m_n)_{n\in \N_0}\colon \N_0 \times \Omega \to \R^\fd$, $m\in\N$,
	which satisfy for all $m\in \{1,2,\dots,\numf\}$,
	$n \in \N$
	 that 
	 \begin{equation}
  \begin{split}
   \bfM_0^m = 0,\qquad
	 			\bfM_n^{m}  = \fnormal_{n-1}^{m}(\bfM_{n-1}^{1},\bfM_{n-1}^{2},\dots,\bfM_{n-1}^{\numf},\fG_n(\Theta_{n-1})),
     \end{split}
     \end{equation}
     \begin{equation}
         \begin{split}
	 	\text{and}\qquad &\Theta_{n} = \Theta_{n-1} -\fbold_{n-1}(\bfM^1_{n},\bfM^2_n,\dots,\bfM^\numf_n,\fG_{n}(\Theta_{n-1})). 
     \end{split}
	 \end{equation}
    \end{enumerate}
\end{athm}
\begin{aproof}
    Throughout this proof for every $m\in \{1,2,\dots,\numf\}$ let $\bfM^m=(\bfM^m_n)_{n\in \N_0}\allowbreak\colon \allowbreak\N_0\times \Omega\to\R^\fd$ satisfy for all $n\in \N$ that
    \begin{equation}\llabel{def: M}
\bfM_0^m=0\qqandqq\bfM_n^m=\fcapital_{n-1}^m(\fG_1(\Theta_0),\fG_2(\Theta_1),\dots,\fG_n(\Theta_{n-1}))
    \end{equation}
    and for every $n,m\in \N_0$ let $\bfF_{n}^m=(\bfF_{n,1}^m,\dots,\bfF_{n,\fd}^m)\colon (\R^\fd)^{n+1}\to \R^\fd$ satisfy for all $g\in (\R^\fd)^{n+1}$ that $\bfF_n^m(g)=\fcapital_n^m(g)$.
    In the following we prove that for all $n\in\N_0$, $m\in \{1,2,\dots,\numf\} $, $
	g =
	( 
	( g_{ i, j } )_{ j \in \{ 1, 2, \dots, \fd \} }
	)_{
		i \in \{ 0, 1, \dots, n \}
	}
	\in 
	(
	\R^{ 
		\fd
	}
	)^{ n + 1 }
	$, 
	$ 
	j \in \{1,2,\dots,\fd\}  
	$ with $\sum_{i=0}^n|g_{i,j}|=0$ it holds that 
 \begin{equation}\llabel{need to prove}
     \bfF_{n,j}^m(g)=0.
 \end{equation}
 We prove \lref{need to prove} by induction on $n$. For the base case $n=0$ observe that \lref{def: f} and \lref{def: F} implies that for all $m\in \{1,2,\dots,\numf\}$, $
	g =
	( 
	 g_{1},\dots,g_{\fd}
	)
	\in 
	\R^{ 
		\fd
	}
	$, 
	$ 
	j \in \{1,2,\dots,\fd\}  
	$ with $g_j=0$ it holds that 
 \begin{equation}\llabel{base}
     \bfF_{0,j}^m(g)=\fnormal_{0,j}^m(0,0,\dots,0,g)=0.
 \end{equation}
 This establishes \lref{need to prove} in the base case $n=0$.
For the induction step we assume that there exists $n\in \N_0$ which satisfies for all $m\in \{1,2,\dots,\numf\} $, $
	g =
	( 
	( g_{ i, j } )_{ j \in \{ 1, 2, \dots, \fd \} }
	)_{
		i \in \{ 0, 1, \dots, n \}
	}
	\in 
	(
	\R^{ 
		\fd
	}
	)^{ n + 1 }
	$, 
	$ 
	j \in \{1,2,\dots,\fd\}  
	$ with $\sum_{i=0}^n|g_{i,j}|=0$ that
 \begin{equation}\llabel{assumption}
      \bfF_{n,j}^m(g)=0.
 \end{equation}
 \argument{\lref{assumption}}{that for all $
	g =
	( 
	( g_{ i, j } )_{ j \in \{ 1, 2, \dots, \fd \} }
	)_{
		i \in \{ 0, 1, \dots, n+1 \}
	}
	\in 
	(
	\R^{ 
		\fd
	}
	)^{ n + 2 }
	$, 
	$ 
	j \in \{1,2,\dots,\fd\}  
	$ with $\sum_{i=0}^{n+1}|g_{i,j}|=0$ it holds that
 \begin{equation}\llabel{EQ1}
 \begin{split}
\textstyle\biggl[\sum\limits_{m=1}^{\numf}\bfF_{n,j}^m(g_0,g_1,\dots,g_{n})\biggr]+g_{n+1,j}=0\dott
\end{split}
 \end{equation}}
 \argument{\lref{EQ1};\lref{def: f};\lref{def: F}}{for all $m\in \{1,2,\dots,\numf\} $, $
	g =
	( 
	( g_{ i, j } \allowbreak)_{ j \in \{ 1, 2, \dots, \fd \} }
	\allowbreak)_{
		i \in \{ 0, 1, \dots, n+1 \}
	}
	\in 
	(
	\R^{ 
		\fd
	}
	)^{ n + 2 }
	$, 
	$ 
	j \in \{1,2,\dots,\fd\}  
	$ with $\sum_{i=0}^{n+1}|g_{i,j}|=0$  that
 \begin{equation}\llabel{EQ2}
\begin{split}
&\bfF_{n+1,j}^m(g)=\bfF_{n+1,j}^m(g_0,g_1,\dots,g_{n+1})\\
&=\fnormal_{n+1,j}^m\bigl(\bfF_n^1(g_0,g_1,\dots,g_{n}), \bfF_n^2(g_0,g_1,\dots,g_{n}),\dots,\bfF_n^\numf(g_0,g_1,\dots,g_{n}),g_{n+1}\bigr)=0\dott
\end{split}
 \end{equation}}
 \argument{\lref{EQ2};\lref{base};induction}{\lref{need to prove}\dott}
 \startnewargseq
 \argument{\lref{need to prove};}{that for all $n\in \N_0$, $
	g =
	( 
	( g_{ i, j } \allowbreak)_{ j \in \{ 1, 2, \dots, \fd \} }
	\allowbreak)_{
		i \in \{ 0, 1, \dots, n \}
	}
	\in 
	(
	\R^{ 
		\fd
	}
	)^{ n + 1 }
	$, 
	$ 
	j \in \{1,2,\dots,\fd\}  
	$ with $\sum_{i=0}^{n}|g_{i,j}|=0$ that
 \begin{equation}\llabel{EQ3}
     \biggl[\textstyle\sum_{m=1}^\numf|\bfF_{n,j}^m(g)|\biggr]+|g_{n,j}|=0\dott
 \end{equation}}
 \argument{\lref{EQ3};\lref{def: bff};\lref{def: Phi}}{for all $n\in\N_0$, $
	g =
	( 
	( g_{ i, j } )_{ j \in \{ 1, 2, \dots, \fd \} }
	)_{
		i \in \{ 0, 1, \dots, n \}
	}
	\in 
	(
	\R^{ 
		\fd
	}
	)^{ n + 1 }
	$, 
	$ 
	j \in \{1,2,\dots,\fd\}  
	$ with $\sum_{i=0}^n|g_{i,j}|=0$ that
 \begin{equation}\llabel{EQ4}
\Phi_{n}^j(g)=\fbold_{n,j}\bigl(\bfF_n^1(g),\bfF_n^2(g),\ldots,\bfF_n^\numf(g),g_n\bigr)=0\dott
 \end{equation}}
 \argument{\lref{EQ4};}{\lref{item 1}\dott}
 \argument{\cref{general SGD pre} (applied with $\fd\curvearrowleft\fd$, $\numf\curvearrowleft\fd\numf$, $(\fnormal_{n+1})_{n\in \N_0}\curvearrowleft((\fnormal_{n}^m)_{m\in \{1,2,\dots,\numf\}})_{n\in \N_0}$, $(\fcapital_{n+1})_{n\in \N_0}\curvearrowleft((\fcapital_{n}^m)_{m\in \{1,2,\dots,\numf\}})_{n\in \N_0}$, $(\fbold_{n+1})_{n\in \N_0}\curvearrowleft((\fbold_{n,m})_{m\in \{1,2,\dots,\fd\}})_{n\in \N_0}$, $(\Phi_{n+1})_{n\in \N_0}\curvearrowleft((\Phi_{n}^m)_{m\in \{1,2,\dots,\fd\}})_{n\in \N_0}$, $(\fG_{n})_{n\in \N_0}\curvearrowleft((\fG_{n}^m)_{m\in \{1,2,\dots,\fd\}}\allowbreak)_{n\in \N_0}$, $\Theta\curvearrowleft\Theta$, $(\bfM_n)_{n\in \N_0}\curvearrowleft((\bfM^m_n)_{m\in\{1,2,\dots,\numf\}})_{n\in \N_0}$ in the notation of \cref{general SGD pre});}{\lref{item 2}\dott}
\end{aproof}
\subsection{Momentum SGD optimizer}  \label{subsec: momentum SGD}
In this subsection we show in the elementary result in \cref{draf: momentum} below that the momentum \SGD\ optimizer (cf.\ \cite{Polyak1964SomeMO}) 
satisfies the assumptions of \cref{conjecture: multilayer2} (cf.\ \cref{eq:gradients_SGD4}--\cref{eq: setting: SGD_def process_main theorem 2} in \cref{conjecture: multilayer2} and \cref{setting: multilayer: eq:gradients_SGD4:}--\cref{eq: setting: SGD_def process} in \cref{setting: SGD}). \cref{draf: momentum} is a slightly strengthened version of \cite[Lemma 4.8]{ArAd2024} and we also refer, \eg, to \cite[Lemma 6.3]{Beckerlearning2023} for a special case of \cref{item 2: momentum} in \cref{draf: momentum}.

\begin{athm}{prop}{draf: momentum}
    Let $\fd\in \N$, $(\alpha_n)_{n\in \N}\subseteq\R$, $(\gamma_n)_{n\in \N}\subseteq\R$, for every $n\in \N_0$ let $\Phi_n=(\Phi_n^1,\dots,\Phi_n^\fd)\colon (\R^\fd)^{n+1}\to \R^\fd$ satisfy for all $
	g =
	( 
	( g_{ i, j } )_{ j \in \{ 1, 2, \dots, \fd \} }
	)_{
		i \in \{ 0, 1, \dots, n \}
	}
	\in 
	(
	\R^{ 
		\fd
	}
	)^{ n + 1 }
	$ that
    \begin{equation}\llabel{eq1}
    \textstyle
\Phi_n(g)=\gamma_{n+1}\bigl[\sum_{k=0}^{n}\bigl[(1-\alpha_{k+1})\textstyle\bigl(\prod_{l=k+1}^{n}\alpha_{l+1}\bigr)g_k\bigr]\bigr],
    \end{equation}
    let $(\Omega,\mathcal F,\P)$ be a probability space, for every $n\in \N_0$ let $\fG_n=(\fG_n^1,\ldots,\fG_n^\fd)\colon \R^\fd\times \Omega\to \R^\fd$ be a function, and let $\Theta\colon \N_0\times\Omega\to \R^\fd$ satisfy for all $n\in \N$ that
\begin{equation}\llabel{def: Phi}
    \Theta_{n} = \Theta_{n-1} - \Phi_{n-1} ( \fG_1 ( \Theta_0 ) , \fG_2 ( \Theta_1 ) , \ldots, \fG_n ( \Theta_{n-1} )).
\end{equation}
    Then
    \begin{enumerate}[label=(\roman*)]
        \item \label{item 1: momentum} it holds for all $n\in\N_0$, $
	g =
	( 
	( g_{ i, j } )_{ j \in \{ 1, 2, \dots, \fd \} }
	)_{
		i \in \{ 0, 1, \dots, n \}
	}
	\in 
	(
	\R^{ 
		\fd
	}
	)^{ n + 1 }
	$, 
	$ 
	j \in \{1,2,\dots,\fd\}  
	$ with $\sum_{i=0}^n|g_{i,j}|=0$ that
        $\Phi_n^j(g)=0$ and
        \item \label{item 2: momentum} it holds that there exists $\bfm \colon \N_0 \times \Omega \to \R^\fd$
	which satisfies for all
	$n \in \N$
	  that 
	 \begin{equation}
   \bfm_0 = 0,\quad
	 			\bfm_{n}  = \alpha_{n} \bfm_{n-1} + (1 - \alpha_{n} ) \fG_n ( \Theta_{n-1} ), 
	 			\qandq 	\Theta_{n} = \Theta_{n-1} - \gamma_{n} \bfm_{n}. 
	 \end{equation}
    \end{enumerate}
\end{athm}
\begin{aproof}
Throughout this proof for every $n\in \N_0$ let $\fnormal_n=(\fnormal_{n,1},\dots,\fnormal_{n,\fd})\colon \allowbreak(\R^\fd)^2\allowbreak\to\R^\fd$ satisfy for all $g=(g_0,g_1)\in (\R^\fd)^2$ that 
\begin{equation}\llabel{def: f}
    \fnormal_n(g)=\alpha_{n+1} g_0+(1-\alpha_{n+1})g_1,
\end{equation} 
let $\fcapital_n\colon (\R^{\fd})^{n+1}\to\R^\fd$ satisfy for all $g=(g_0,g_1,\dots,g_n)\in (\R^{\fd})^{n+1}$ that 
\begin{equation}\llabel{def: F}
    \fcapital_n(g)=\sum_{k=0}^{n}\bigl[(1-\alpha_{k+1})\textstyle\bigl(\prod_{l=k+1}^{n}\alpha_{l+1}\bigr)g_k\bigr],
\end{equation}
and let $\fbold_n=(\fbold_{n,1},\dots,\fbold_{n,\fd})\colon (\R^\fd)^2\to\R^\fd$ satisfy for all $g=(g_0,g_1)\in (\R^\fd)^2$  that 
\begin{equation}\llabel{def: bff}
    \fbold_n(g)=\gamma_{n+1} g_0.
\end{equation}
\argument{\lref{def: f};}{that for all $n\in \N_0$, $g=((g_{i,j})_{j\in \{1,2,\dots,\fd\}})_{i\in\{1,2\}}\in (\R^\fd)^2$, $j\in \{1,2,\dots,\fd\}$ with $\sum_{i=0}^1|g_{i,j}|=0$ it holds that
\begin{equation}\llabel{vr1}
    \fnormal_{n,j}(g)=\alpha _{n+1} g_{0,j}+(1-\alpha_{n+1}) g_{1,j}=0\alpha_{n+1}+0(1-\alpha_{n+1})=0\dott
\end{equation}}
\argument{\lref{def: f};\lref{def: F}}{for all $n\in \N\backslash\{1\}$, $
	g =
	( 
	( g_{ i, j } )_{ j \in \{ 1, 2, \dots, \fd \} }
	\allowbreak)_{
		i \in \{ 0, 1, \dots, n \}
	}
	\in 
	(
	\R^{ 
		\fd
	}
	)^{ n + 1 }
	$, $y\in \R^\fd$ that
 \begin{equation}\llabel{vr2}
     \begin{split}
    \fcapital_0(y)=(1-\alpha_1)y =\fnormal_0(0,y)
    \end{split}
    \end{equation}
and
\begin{equation}\llabel{vr3}
\begin{split}
\textstyle
   & \fcapital_n(g_0,g_1\dots,g_n)=\sum_{k=0}^{n}\bigl[(1-\alpha_{k+1})\textstyle\bigl(\prod_{l=k+1}^{n}\alpha_{l+1}\bigr)g_k\bigr]\\
   &=\biggl[\textstyle\sum\limits_{k=0}^{n-1}\bigl[(1-\alpha_{k+1})\textstyle\bigl(\prod_{l=k+1}^{n}\alpha_{l+1}\bigr)\bigr]g_k\biggr]+(1-\alpha_{n+1})g_{n}\\
&=\alpha_{n+1}\biggl[\textstyle\sum\limits_{k=0}^{n-1}\bigl[(1-\alpha_{k+1})\textstyle\bigl(\prod_{l=k+1}^{n-1}\alpha_{l+1}\bigr)g_k\bigr]\biggr]+(1-\alpha_{n+1})g_{n}\\
&=\alpha_{n+1} \fcapital_{n-1}(g_0,g_1,\dots,g_{n-1})+(1-\alpha_{n+1})g_n=
   \fnormal_n\bigl(\fcapital_{n-1}(g_0,g_1,\dots,g_{n-1}),g_n\bigr).
    \end{split}
     \end{equation}}
    \argument{\lref{def: bff}}{for all $g=((g_{i,j})_{j\in\{1,2,\dots,\fd\}})_{i\in \{0,1\}}\in (\R^\fd)^2$, $j\in \{1,2,\dots,\fd\}$ with $\sum_{i=0}^1|g_{i,j}|=0$ that
    \begin{equation}\llabel{vr4}
        \fbold_{n,j}(g)=\gamma_{n+1}g_{0,j}=0\dott
    \end{equation}}
    \argument{\lref{def: Phi};\lref{def: F};\lref{def: bff}}{for all $n\in \N_0$, $g=(g_0,g_1\dots,g_{n})\in (\R^\fd)^{n+1}$ that
    \begin{equation}\llabel{vr5}
\Phi_n(g)=\gamma_{n+1}\fcapital_n(g)=\fbold_n(\fcapital_n(g),g_{n})\dott
    \end{equation}}
\argument{\lref{vr1};\lref{vr2};\lref{vr3};\lref{vr4};\lref{vr5};\cref{general SGD} (applied with $\fd\curvearrowleft\fd$, $\numf\curvearrowleft 1$, $((\fnormal_{n}^m)_{m\in \{1,2,\dots,\numf\}}\allowbreak)_{n\in \N_0}\curvearrowleft((\fnormal_{n}^m)_{m\in \{1,2,\dots,\numf\}}\allowbreak)_{n\in \N_0}$, $((\fcapital_{n}^m)_{m\in \{1,2,\dots,\numf\}})_{n\in \N_0}\curvearrowleft((\fcapital_{n}^m)_{m\in \{1,2,\dots,\numf\}}\allowbreak)_{n\in \N_0}$, $((\fbold_{n,m}\allowbreak)_{m\in \{1,2,\dots,\fd\}}\allowbreak)_{n\in \N_0}\curvearrowleft((\fbold_{n,m}\allowbreak)_{m\in \{1,2,\dots,\fd\}}\allowbreak)_{n\in \N_0}$, $((\Phi_{n}^m)_{m\in \{1,2,\dots,\fd\}}\allowbreak)_{n\in \N_0}\curvearrowleft\allowbreak(\allowbreak(\Phi_{n}^m\allowbreak)_{m\in \{1,2,\dots,\fd\}}\allowbreak)_{n\in \N_0}$, $((\fG_{n}^m)_{m\in \{1,2,\dots,\fd\}}\allowbreak)_{n\in \N_0}\curvearrowleft((\fG_{n}^m)_{m\in \{1,2,\dots,\fd\}}\allowbreak)_{n\in \N_0}$, $\Theta\curvearrowleft\Theta$, $((\bfM_n^m)_{m\in \{1,2,\dots,\numf\}})_{n\in \N_0}\curvearrowleft(\bfm_n\allowbreak)_{n\in \N_0}$ in the notation of \cref{general SGD})}{\cref{item 1: momentum,item 2: momentum}\dott}
\end{aproof}
\subsection{Standard SGD optimizer}\label{subsec: standard SGD}
In this subsection we show in the elementary result in \cref{draf: standard SGD} below that the standard \SGD\ optimizer  
satisfies the assumptions of \cref{conjecture: multilayer2} (cf.\ \cref{eq:gradients_SGD4}--\cref{eq: setting: SGD_def process_main theorem 2} in \cref{conjecture: multilayer2} and \cref{setting: multilayer: eq:gradients_SGD4:}--\cref{eq: setting: SGD_def process} in \cref{setting: SGD}). We refer, \eg, to \cite[Lemma 6.2]{Beckerlearning2023} for a special case of \cref{item 2: standard SGD} in \cref{draf: standard SGD}.
\begin{athm}{prop}{draf: standard SGD}
    Let $\fd\in \N$, let $(\gamma_n)_{n\in \N}\subseteq\R$, for every $n\in \N_0$ let $\Phi_n=(\Phi_n^1,\dots,\Phi_n^\fd)\colon (\R^\fd)^{n+1}\allowbreak\to \R^\fd$ satisfy for all $
	g =(g_0,g_1,\dots,g_n)
	\in 
	(
	\R^{ 
		\fd
	}
	)^{ n + 1 }
	$ that
    \begin{equation}\llabel{eq1}
    \textstyle
\Phi_n(g)=\gamma_{n+1}g_n,
    \end{equation}
    let $(\Omega,\mathcal F,\P)$ be a probability space, for every $n\in \N_0$ let $\fG_n\colon \R^\fd\times \Omega\to \R^\fd$ be a function, and let $\Theta\colon \N_0\times\Omega\to \R^\fd$ satisfy for all $n\in \N$ that
\begin{equation}\llabel{def: Phi}
    \Theta_{n} = \Theta_{n-1} - \Phi_{n-1} ( \fG_1 ( \Theta_0 ) , \fG_2 ( \Theta_1 ) , \ldots, \fG_n ( \Theta_{n-1} )).
\end{equation}
    Then
    \begin{enumerate}[label=(\roman*)]
        \item \label{item 1: standard SGD} it holds for all $n\in\N_0$, $
	g =
	( 
	( g_{ i, j } )_{ j \in \{ 1, 2, \dots, \fd \} }
	)_{
		i \in \{ 0, 1, \dots, n \}
	}
	\in 
	(
	\R^{ 
		\fd
	}
	)^{ n + 1 }
	$, 
	$ 
	j \in \{1,2,\dots,\fd\}  
	$ with $\sum_{i=0}^n|g_{i,j}|=0$ that
        $\Phi_n^j(g)=0$ and
        \item \label{item 2: standard SGD} it holds for all $n\in\N$ that 
	 \begin{equation}
  \Theta_{n} = \Theta_{n-1} - \gamma_{n} \fG_n(\Theta_{n-1}). 
	 \end{equation}
    \end{enumerate}
\end{athm}
\begin{aproof}
\argument{\cref{draf: momentum} (applied with $\fd\curvearrowleft\fd$, $(\alpha_n)_{n\in\N}\curvearrowleft (0)_{n\in \N}$, $(\gamma_n)_{n \in \N  }\curvearrowleft(\gamma_n)_{n \in \N  }$, $((\Phi_{n}^m)_{m\in \{1,2,\dots,\fd\}})_{n\in \N_0}\curvearrowleft(\Phi_{n})_{n\in \N_0}$, $((\fG_{n}^m)_{m\in \{1,2,\dots,\fd\}}\allowbreak)_{n\in \N_0}\curvearrowleft(\fG_{n})\allowbreak_{n\in \N_0}$, $\Theta\curvearrowleft\Theta$ in the notation of \cref{draf: momentum})}{that
\begin{enumerate}[label=(\Roman*)]\llabel{item 2'}
        \item it holds for all $n\in\N_0$, $
	g =
	( 
	( g_{ i, j } )_{ j \in \{ 1, 2, \dots, \fd \} }
	)_{
		i \in \{ 0, 1, \dots, n \}
	}
	\in 
	(
	\R^{ 
		\fd
	}
	)^{ n + 1 }
	$, 
	$ 
	j \in \{1,2,\dots,\fd\}  
	$ with $\sum_{i=0}^n|g_{i,j}|=0$ that
        $\Phi_n^j(g)=0$ and
        \item it holds that there exists $\bfm \colon \N_0 \times \Omega \to \R^\fd$
	which satisfies for all
	$n \in \N$
	  that 
	 \begin{equation}
   \bfm_0 = 0,\qquad
	 			\bfm_{n}  = \fG_n ( \Theta_{n-1} ), 
	 			\qqandqq 	\Theta_{n} = \Theta_{n-1} - \gamma_{n} \bfm_{n}. 
	 \end{equation}
    \end{enumerate}}
    \argument{\lref{item 2'};}{\cref{item 1: standard SGD,item 2: standard SGD}\dott}
\end{aproof}
\subsection{Nesterov accelerated SGD optimizer}\label{subsec: nesterov}
In this subsection we show in the elementary result in \cref{draf: nesterov} below that the Nesterov accelerated \SGD\ optimizer (cf.\ \cite{Nesterov1983AMF}) 
satisfies the assumptions of \cref{conjecture: multilayer2} (cf.\ \cref{eq:gradients_SGD4}--\cref{eq: setting: SGD_def process_main theorem 2} in \cref{conjecture: multilayer2} and \cref{setting: multilayer: eq:gradients_SGD4:}--\cref{eq: setting: SGD_def process} in \cref{setting: SGD}).
\begin{athm}{prop}{draf: nesterov}
    Let $\fd\in \N$, $(\alpha_n)_{n\in \N}\subseteq\R$, $(\gamma_n)_{n\in \N}\subseteq\R$, for every $n\in \N_0$ let $\Phi_n=(\Phi_n^1,\dots,\Phi_n^\fd)\colon (\R^\fd)^{n+1}\to \R^\fd$ satisfy for all $
	g =
	( 
	( g_{ i, j } )_{ j \in \{ 1, 2, \dots, \fd \} }
	)_{
		i \in \{ 0, 1, \dots, n \}
	}
	\in 
	(
	\R^{ 
		\fd
	}
	)^{ n + 1 }
	$ that
    \begin{equation}\llabel{eq1}
    \textstyle
\Phi_n(g)=\gamma_{n+2}\alpha_{n+2}\bigl[\sum_{k=0}^{n}\bigl[(1-\alpha_{k+1})\textstyle\bigl(\prod_{l=k+1}^{n}\alpha_{l+1}\bigr)g_k\bigr]\bigr]+\gamma_{n+1}(1-\alpha_{n+1})g_n,
    \end{equation}
    let $(\Omega,\mathcal F,\P)$ be a probability space, for every $n\in \N_0$ let $\fG_n=(\fG_n^1,\ldots,\fG_n^\fd)\colon \R^\fd\times \Omega\to \R^\fd$ be a function, and let $\Theta\colon \N_0\times\Omega\to \R^\fd$ satisfy for all $n\in \N$ that
\begin{equation}\llabel{def: Phi}
    \Theta_{n} = \Theta_{n-1} - \Phi_{n-1} ( \fG_1 ( \Theta_0 ) , \fG_2 ( \Theta_1 ) , \ldots, \fG_n ( \Theta_{n-1} )).
\end{equation}
    Then
    \begin{enumerate}[label=(\roman*)]
        \item \llabel{item 1} it holds for all $n\in\N_0$, $
	g =
	( 
	( g_{ i, j } )_{ j \in \{ 1, 2, \dots, \fd \} }
	)_{
		i \in \{ 0, 1, \dots, n \}
	}
	\in 
	(
	\R^{ 
		\fd
	}
	)^{ n + 1 }
	$, 
	$ 
	j \in \{1,2,\dots,\fd\}  
	$ with $\sum_{i=0}^n|g_{i,j}|=0$ that
        $\Phi_n^j(g)=0$ and
        \item \llabel{item 2} it holds that there exists $\bfm \colon \N_0 \times \Omega \to \R^\fd$
	which satisfies for all
	$n \in \N$
	  that 
	 \begin{equation}
   \bfm_0 = 0,\qquad
	 			\bfm_{n}  = \alpha_{n} \bfm_{n-1} + (1 - \alpha_{n} ) \fG_n ( \Theta_{n-1} ),
     \end{equation}
	 \begin{equation}		
     \text{and}\qquad \Theta_{n} = \Theta_{n-1} - \gamma_{n+1}\alpha_{n+1} \bfm_{n}-\gamma_n(1-\alpha_n)\fG_n(\Theta_{n-1}). 
	 \end{equation}
    \end{enumerate}
\end{athm}
\begin{aproof}
Throughout this proof for every $n\in \N_0$ let $\fnormal_n=(\fnormal_{n,1},\dots,\fnormal_{n,\fd})\colon \allowbreak(\R^\fd)^2\allowbreak\to\R^\fd$ satisfy for all $g=(g_0,g_1)\in (\R^\fd)^2$ that 
\begin{equation}\llabel{def: f}
    \fnormal_n(g)=\alpha_{n+1} g_0+(1-\alpha_{n+1})g_1,
\end{equation} 
let $\fcapital_n\colon (\R^{\fd})^{n+1}\to\R^\fd$ satisfy for all $g=(g_0,g_1,\dots,g_n)\in (\R^{\fd})^{n+1}$ that 
\begin{equation}\llabel{def: F}
    \fcapital_n(g)=\sum_{k=0}^{n}\bigl[(1-\alpha_{k+1})\textstyle\bigl(\prod_{l=k+1}^{n}\alpha_{l+1}\bigr)g_k\bigr],
\end{equation}
and let $\fbold_n=(\fbold_{n,1},\dots,\fbold_{n,\fd})\colon (\R^\fd)^2\to\R^\fd$ satisfy for all $g=(g_0,g_1)\in (\R^\fd)^2$  that 
\begin{equation}\llabel{def: bff}
    \fbold_n(g)=\gamma_{n+2}\alpha_{n+2} g_0+\gamma_{n+1}(1-\alpha_{n+1})g_1.
\end{equation}
\argument{\lref{def: f};}{that for all $n\in \N_0$, $g=((g_{i,j})_{j\in \{1,2,\dots,\fd\}})_{i\in\{1,2\}}\in (\R^\fd)^2$, $j\in \{1,2,\dots,\fd\}$ with $\sum_{i=0}^1|g_{i,j}|=0$ it holds that
\begin{equation}\llabel{vr1}
    \fnormal_{n,j}(g)=\alpha _{n+1} g_{0,j}+(1-\alpha_{n+1}) g_{1,j}=0\alpha_{n+1}+0(1-\alpha_{n+1})=0\dott
\end{equation}}
\argument{\lref{def: f};\lref{def: F}}{for all $n\in \N\backslash\{1\}$, $
	g =
	( 
	( g_{ i, j } )_{ j \in \{ 1, 2, \dots, \fd \} }
	\allowbreak)_{
		i \in \{ 0, 1, \dots, n \}
	}
	\in 
	(
	\R^{ 
		\fd
	}
	)^{ n + 1 }
	$, $y\in \R^\fd$ that
 \begin{equation}\llabel{vr2}
     \begin{split}
    \fcapital_0(y)=(1-\alpha_1)y =\fnormal_0(0,y)
    \end{split}
    \end{equation}
and
\begin{equation}\llabel{vr3}
\begin{split}
\textstyle
   & \fcapital_n(g_0,g_1\dots,g_n)=\sum_{k=0}^{n}\bigl[(1-\alpha_{k+1})\textstyle\bigl(\prod_{l=k+1}^{n}\alpha_{l+1}\bigr)g_k\bigr]\\
   &=\biggl[\textstyle\sum\limits_{k=0}^{n-1}\bigl[(1-\alpha_{k+1})\textstyle\bigl(\prod_{l=k+1}^{n}\alpha_{l+1}\bigr)\bigr]g_k\biggr]+(1-\alpha_{n+1})g_{n}\\
&=\alpha_{n+1}\biggl[\textstyle\sum\limits_{k=0}^{n-1}\bigl[(1-\alpha_{k+1})\textstyle\bigl(\prod_{l=k+1}^{n-1}\alpha_{l+1}\bigr)g_k\bigr]\biggr]+(1-\alpha_{n+1})g_{n}\\
&=\alpha_{n+1} \fcapital_{n-1}(g_0,g_1,\dots,g_{n-1})+(1-\alpha_{n+1})g_n=
   \fnormal_n\bigl(\fcapital_{n-1}(g_0,g_1,\dots,g_{n-1}),g_n\bigr).
    \end{split}
     \end{equation}}
    \argument{\lref{def: bff}}{for all $g=((g_{i,j})_{j\in\{1,2,\dots,\fd\}})_{i\in \{0,1\}}\in (\R^\fd)^2$, $j\in \{1,2,\dots,\fd\}$ with $\sum_{i=0}^1|g_{i,j}|=0$ that
    \begin{equation}\llabel{vr4}
        \fbold_{n,j}(g)=\gamma_{n+2}\alpha_{n+2} g_{0,j}+\gamma_{n+1}(1-\alpha_{n+1})g_{1,j}=0\dott
    \end{equation}}
    \argument{\lref{def: Phi};\lref{def: F};\lref{def: bff}}{for all $n\in \N_0$, $g=(g_0,g_1\dots,g_{n})\in (\R^\fd)^{n+1}$ that
    \begin{equation}\llabel{vr5}
    \begin{split}
\Phi_n(g)&=\gamma_{n+2}\alpha_{n+2}\bigl[\textstyle\sum_{k=0}^{n}\bigl[(1-\alpha_{k+1})\textstyle\bigl(\prod_{l=k+1}^{n}\alpha_{l+1}\bigr)g_k\bigr]\bigr]+\gamma_{n+1}(1-\alpha_{n+1})g_n\\
&=\fbold_n(\fcapital_n(g),g_{n})\dott
\end{split}
    \end{equation}}
\argument{\lref{vr1};\lref{vr2};\lref{vr3};\lref{vr4};\lref{vr5};\cref{general SGD} (applied with $\fd\curvearrowleft\fd$, $\numf\curvearrowleft 1$, $((\fnormal_{n}^m)_{m\in \{1,2,\dots,\numf\}}\allowbreak)_{n\in \N_0}\curvearrowleft((\fnormal_{n}^m)_{m\in \{1,2,\dots,\numf\}}\allowbreak)_{n\in \N_0}$, $((\fcapital_{n}^m)_{m\in \{1,2,\dots,\numf\}})_{n\in \N_0}\curvearrowleft((\fcapital_{n}^m)_{m\in \{1,2,\dots,\numf\}}\allowbreak)_{n\in \N_0}$, $((\fbold_{n,m}\allowbreak)_{m\in \{1,2,\dots,\fd\}}\allowbreak)_{n\in \N_0}\curvearrowleft((\fbold_{n,m}\allowbreak)_{m\in \{1,2,\dots,\fd\}}\allowbreak)_{n\in \N_0}$, $((\Phi_{n}^m)_{m\in \{1,2,\dots,\fd\}}\allowbreak)_{n\in \N_0}\curvearrowleft\allowbreak(\allowbreak(\Phi_{n}^m\allowbreak)_{m\in \{1,2,\dots,\fd\}}\allowbreak)_{n\in \N_0}$, $((\fG_{n}^m)_{m\in \{1,2,\dots,\fd\}}\allowbreak)_{n\in \N_0}\curvearrowleft((\fG_{n}^m)_{m\in \{1,2,\dots,\fd\}}\allowbreak)_{n\in \N_0}$, $\Theta\curvearrowleft\Theta$, $((\bfM_n^m)_{m\in \{1,2,\dots,\numf\}})_{n\in \N_0}\curvearrowleft(\bfm_n\allowbreak)_{n\in \N_0}$ in the notation of \cref{general SGD})}{\lref{item 1, item 2}\dott}
\end{aproof}
\subsection{Adaptive gradient SGD (Adagrad) optimizer}\label{subsec: Adagrad}
In this subsection we show in the elementary result in \cref{draf adagrad} below that the \Adagrad\ optimizer (cf.\ \cite{JMLR:v12:duchi11a}) 
satisfies the assumptions of \cref{conjecture: multilayer2} (cf.\ \cref{eq:gradients_SGD4}--\cref{eq: setting: SGD_def process_main theorem 2} in \cref{conjecture: multilayer2} and \cref{setting: multilayer: eq:gradients_SGD4:}--\cref{eq: setting: SGD_def process} in \cref{setting: SGD}). We refer, \eg, to \cite[Lemma 6.4]{Beckerlearning2023} for a special case of \cref{item 2: adagrad} in \cref{draf adagrad}.
\begin{athm}{prop}{draf adagrad}
    Let $\fd\in \N$, $\varepsilon\in (0,\infty)$, let $(\gamma_n)_{n\in \N}\subseteq\R$, for every $n\in \N_0$ let $\Phi_n=(\Phi_n^1,\dots,\Phi_n^\fd)\colon (\R^\fd)^{n+1}\to \R^\fd$ satisfy for all $
	g =
	( 
	( g_{ i, j } )_{ j \in \{ 1, 2, \dots, \fd \} }
	)_{
		i \in \{ 0, 1, \dots, n \}
	}
	\in 
	(
	\R^{ 
		\fd
	}
	)^{ n + 1 }
	$, $j\in\{1,2,\dots,\fd\}$ that
    \begin{equation}\llabel{eq1}
    \textstyle
\Phi_n^j(g)=\gamma_{n+1}\bigl[\varepsilon+\bigl[\sum_{k=0}^{n}|g_{k,j}|^2\bigr]\bigr]^{-1/2}g_{n,j},
    \end{equation}
    let $(\Omega,\mathcal F,\P)$ be a probability space, for every $n\in \N_0$ let $\fG_n=(\fG_n^1,\ldots,\fG_n^\fd)\colon \R^\fd\times \Omega\to \R^\fd$ be a function, and let $\Theta\colon \N_0\times\Omega\to \R^\fd$ satisfy for all $n\in \N$ that
\begin{equation}\llabel{def: Phi}
    \Theta_{n} = \Theta_{n-1} - \Phi_{n-1} ( \fG_1 ( \Theta_0 ) , \fG_2 ( \Theta_1 ) , \ldots, \fG_n ( \Theta_{n-1} )).
\end{equation}
    Then
    \begin{enumerate}[label=(\roman*)]
        \item \label{item 1: adagrad} it holds for all $n\in\N_0$, $
	g =
	( 
	( g_{ i, j } )_{ j \in \{ 1, 2, \dots, \fd \} }
	)_{
		i \in \{ 0, 1, \dots, n \}
	}
	\in 
	(
	\R^{ 
		\fd
	}
	)^{ n + 1 }
	$, 
	$ 
	j \in \{1,2,\dots,\fd\}  
	$ with $\sum_{i=0}^n|g_{i,j}|=0$ that
        $\Phi_n^j(g)=0$ and
        \item \label{item 2: adagrad} it holds for all
	$n \in \N$, $j\in \{1,2,\dots,\fd\}$
	  that 
	 \begin{equation}
   \Theta_n^{j}=\Theta_{n-1}^{j}-\gamma_{n}\biggl[\varepsilon+\biggl[\textstyle\sum\limits_{k=0}^{n-1}|\fG_{k+1}^j\left(\Theta_{k}\right)|^2\biggr]\biggr]^{-1 / 2} \fG_n^j(\Theta_{n-1}). 
	 \end{equation}
    \end{enumerate}
\end{athm}
\begin{aproof}
    Throughout this proof for every $n\in \N_0$ let $\fnormal_n=(\fnormal_{n,1},\dots,\fnormal_{n,\fd})\colon \allowbreak(\R^\fd)^2\allowbreak\to\R^\fd$ satisfy for all $g=(g_0,g_1)\in (\R^\fd)^2$, $j\in \{1,2,\dots,\fd\}$ that 
\begin{equation}\llabel{def: f}
    \fnormal_{n,j}(g)= g_{0,j}+|g_{1,j}|^2,
\end{equation} 
let $\fcapital_n=(\fcapital_{n,1},\dots,\fcapital_{n,\fd})\colon (\R^{\fd})^{n+1}\to\R^\fd$ satisfy for all $g=(g_0,g_1,\dots,g_n)\in (\R^{\fd})^{n+1}$, $j\in \{1,2,\dots,\fd\}$ that 
\begin{equation}\llabel{def: F}
    \fcapital_{n,j}(g)=\sum_{k=0}^{n}|g_{k,j}|^2,
\end{equation}
and let $\fbold_n=(\fbold_{n,1},\dots,\fbold_{n,\fd})\colon (\R^\fd)^2\to\R^\fd$ satisfy for all $g=(g_0,g_1)\in (\R^\fd)^2$, $j\in \{1,2,\dots,\fd\}$  that 
\begin{equation}\llabel{def: bff}
    \fbold_{n,j}(g)=\gamma_{n+1}\bigl[\varepsilon+|g_{0,j}|^2\bigr]^{-1 / 2} g_{1,j}.
\end{equation}
\argument{\lref{def: f};}{that for all $n\in \N_0$, $g=((g_{i,j})_{j\in \{1,2,\dots,\fd\}})_{i\in\{1,2\}}\in (\R^\fd)^2$, $j\in \{1,2,\dots,\fd\}$ with $\sum_{i=0}^1|g_{i,j}|=0$ it holds that
\begin{equation}\llabel{vr1}
    \fnormal_{n,j}(g)= g_{0,j}+|g_{1,j}|^2=0+0=0\dott
\end{equation}}
\argument{\lref{def: f};\lref{def: F}}{for all $n\in \N\backslash\{1\}$, $
	g =
	( 
	( g_{ i, j } )_{ j \in \{ 1, 2, \dots, \fd \} }
	\allowbreak)_{
		i \in \{ 0, 1, \dots, n \}
	}
	\in 
	(
	\R^{ 
		\fd
	}
	)^{ n + 1 }
	$, $j\in\{1,2,\dots,\fd\}$, $y\in \R^\fd$ that
 \begin{equation}\llabel{vr2}
     \begin{split}
    \fcapital_{0,j}(y)=|g_{0,j}|^2 =\fnormal_{0,j}(0,y)
    \end{split}
    \end{equation}
and
\begin{equation}\llabel{vr3}
\begin{split}
\textstyle
   & \fcapital_{n,j}(g_0,g_1\dots,g_n)=\sum_{k=0}^{n}|g_{k,j}|^2=\biggl[\textstyle\sum\limits_{k=0}^{n-1}|g_{k,j}|^2\biggr]+|g_{n,j}|^2\\
   &=\fcapital_{n-1,j}(g_0,g_1,\dots,g_{n-1})+|g_{n,j}|^2=\fnormal_n\bigl(\fcapital_{n-1}(g_0,g_1,\dots,g_{n-1}),g_n\bigr).
    \end{split}
     \end{equation}}
    \argument{\lref{def: bff}}{for all $g=((g_{i,j})_{j\in\{1,2,\dots,\fd\}})_{i\in \{0,1\}}\in (\R^\fd)^2$, $j\in \{1,2,\dots,\fd\}$ with $\sum_{i=0}^1|g_{i,j}|=0$ that
    \begin{equation}\llabel{vr4}
        \fbold_{n,j}(g)=\gamma_{n+1}\bigl[\varepsilon+|g_{0,j}|^2\bigr]^{-1 / 2} g_{1,j}=0\dott
    \end{equation}}
    \argument{\lref{def: Phi};\lref{def: F};\lref{def: bff}}{for all $n\in \N_0$, $g=(g_0,g_1\dots,g_{n})\in (\R^\fd)^{n+1}$, $j\in \{1,2,\dots,\fd\}$ that
    \begin{equation}\llabel{vr5}
\Phi_n^j(g)=\gamma_{n+1}\biggl[\varepsilon+\biggl[\textstyle\sum\limits_{k=0}^{n}|g_{k,j}|^2\biggr]\biggr]^{-1/2}g_{n,j}=\gamma_{n+1}\bigl[\varepsilon+\fcapital_{n,j}(g)\bigr]^{-1/2}g_{n,j}=\fbold_{n,j}(\fcapital_n(g),g_{n})\dott
    \end{equation}}
\argument{\lref{vr1};\lref{vr2};\lref{vr3};\lref{vr4};\lref{vr5};\cref{general SGD} (applied with $\numf\curvearrowleft 1$, $\fnormal\curvearrowleft \fnormal$, $\fcapital\curvearrowleft \fcapital$, $\fbold\curvearrowleft\fbold$, $\Phi\curvearrowleft \Phi$, $\fG\curvearrowleft\fG$, $\Theta\curvearrowleft\Theta$ in the notation of \cref{general SGD})}{\cref{item 1: adagrad} and that there exists $\bfM=(\bfM^1,\dots,\bfM^\fd) \colon \N_0 \times \Omega \to \R^\fd$
	which satisfies for all
	$n \in \N$, $j\in \{1,2,\dots,\fd\}$
	  that 
	 \begin{equation}\llabel{eqq1}
	 		\bfM_0 = 0,\qquad	\bfM_{n}^j  =  \bfM_{n-1}^j +  |\fG_n^j ( \Theta_{n-1} )|^2,
    \end{equation}
    and
    \begin{equation}\llabel{eqq1.1}
	 				\Theta_n^{j}=\Theta_{n-1}^{j}-\gamma_{n}\bigl[\varepsilon+\bfM_{n}^j\bigr]^{-1 / 2} \fG_n^j(\Theta_{n-1}). 
	 \end{equation}}
\startnewargseq
\argument{\lref{eqq1};induction }{that for all $n\in \N$, $j\in \{1,2,\dots,\fd\}$ it holds that
\begin{equation}\llabel{eqq2}
    \bfM_{n}^j=\sum_{k=0}^{n-1}|\fG_{k+1}^j(\Theta_{k})|^2\dott
\end{equation}}
\argument{\lref{eqq2};\lref{eqq1}}{\cref{item 2: adagrad}\dott}
\end{aproof}
\subsection{Root mean square propagation SGD (RMSprop) optimizer }\label{subsec: RMSprop}
In this subsection we show in the elementary result in \cref{draf RMS prop} below that the \RMSprop\ optimizer (cf.\ \cite{Hinton24_RMSprop}) 
satisfies the assumptions of \cref{conjecture: multilayer2} (cf.\ \cref{eq:gradients_SGD4}--\cref{eq: setting: SGD_def process_main theorem 2} in \cref{conjecture: multilayer2} and \cref{setting: multilayer: eq:gradients_SGD4:}--\cref{eq: setting: SGD_def process} in \cref{setting: SGD}). We refer, \eg, to \cite[Lemma 6.5]{Beckerlearning2023} for a special case of \cref{item 2: rmsprop} in \cref{draf RMS prop}.
\begin{athm}{prop}{draf RMS prop}
   Let $\fd\in \N$, $(\gamma_n)_{n \in \N} \subseteq [0 , \infty )$,
	$(\beta_n)_{n \in \N } \subseteq [0 , 1 ]$,
	$\varepsilon \in (0 , \infty ) $, for every $n\in \N_0$ let $\Phi_n=(\Phi_n^1,\dots,\Phi_n^\fd)\colon (\R^\fd)^{n+1}\to \R^\fd$ satisfy for all $
	g =
	( 
	( g_{ i, j } )_{ j \in \{ 1, 2, \dots, \fd \} }
	)_{
		i \in \{ 0, 1, \dots, n \}
	}
	\in 
	(
	\R^{ 
		\fd
	}
	)^{ n + 1 }
	$, $j\in \{1,2,\dots,\fd\}$ that
    \begin{equation}\llabel{def: Phi}
    \begin{split}
    \textstyle
&\Phi_n^j(g)=\gamma_{n+1}\biggl(\varepsilon+\textstyle\sum\limits_{k=0}^{n}\bigl[(1-\beta_{k+1})\textstyle\bigl(\prod_{l=k+1}^{n}\beta_{l+1}\bigr)|g_{k,j}|^2\bigr]\biggr)^{-\nicefrac{1}{2}}g_{n,j},
\end{split}
    \end{equation}
    let $(\Omega,\mathcal F,\P)$ be a probability space, for every $n\in \N_0$ let $\fG_n=(\fG_n^1,\ldots,\fG_n^\fd)\colon \R^\fd\times \Omega\to \R^\fd$ be a function, and let $\Theta\colon \N_0\times\Omega\to \R^\fd$ satisfy for all $n\in \N$ that
\begin{equation}\llabel{eq2}
    \Theta_{n} = \Theta_{n-1} - \Phi_{n-1} ( \fG_1 ( \Theta_0 ) , \fG_2 ( \Theta_1 ) , \ldots, \fG_n ( \Theta_{n-1} )).
\end{equation}
    Then
    \begin{enumerate}[label=(\roman*)]
        \item \label{item 1: rmsprop} it holds for all $n\in\N_0$, $
	g =
	( 
	( g_{ i, j } )_{ j \in \{ 1, 2, \dots, \fd \} }
	)_{
		i \in \{ 0, 1, \dots, n \}
	}
	\in 
	(
	\R^{ 
		\fd
	}
	)^{ n + 1 }
	$, 
	$ 
	j \in \{1,2,\dots,\fd\}  
	$ with $\sum_{i=0}^n|g_{i,j}|=0$ that
        $\Phi_n^j(g)=0$ and
        \item \label{item 2: rmsprop} it holds that there exist $
	  \mathbb{M} = (\mathbb{M}^{1},\ldots,\mathbb{M}^{\defaultParamDim}) \colon \N_0\times \Omega \to \R^\defaultParamDim
    $
   which satisfy for all $n \in \N_0$, $j \in \{1,2,\ldots,\defaultParamDim\}$ that 
   \begin{equation}
   \mathbb{M}_0 = 0, \qquad \mathbb{M}_{n}^{j} 
= \beta_{n}\mathbb{M}_{n-1}^{j} + (1-\beta_{n})| \fG_n^j(\Theta_{n-1})|^2,
 \end{equation}
 and
\begin{equation}
	\Theta_{n}^{j}
= 
	\Theta_{n-1}^{j} - \frac{\gamma_{n}}{ [\varepsilon +\mathbb{M}_{n}^{j}]^{\nicefrac{1}{2}}} \fG_n^j(\Theta_{n-1}). 
\end{equation}
    \end{enumerate}
\end{athm}
\begin{aproof}
      Throughout this proof for every $n\in \N_0$ let $\fnormal_n=(\fnormal_{n,1},\dots,\fnormal_{n,\fd})\colon (\R^\fd)^2\allowbreak \to\R^\fd$ satisfy for all $g=((g_{i,j})_{j\in\{1,2,\dots,\fd\}})_{i\in\{0,1\}}\in (\R^\fd)^2$, $j\in \{1,2,\dots,\fd\}$ that 
\begin{equation}\llabel{def: f}
    \fnormal_{n,j}(g)=\beta_{n+1} g_{0,j}+(1-\beta_{n+1})|g_{1,j}|^2,
\end{equation} 
let $\fcapital_n=(\fcapital_{n,1},\ldots,\fcapital_{n,\fd})\colon (\R^{\fd})^{n+1}\to\R^\fd$ satisfy for all $g=((g_{i,j})_{j\in \{1,2,\dots,\fd\}}\allowbreak)_{i\in \{0,1,\dots,n\}}\in (\R^{\fd})^{n+1}$, $j\in \{1,2,\dots,\fd\}$ that 
\begin{equation}\llabel{def: F}
\begin{split}
   \fcapital_{n,j}(g)=\sum_{k=0}^{n}\bigl[(1-\beta_{k+1})\textstyle\bigl(\prod_{l=k+1}^{n}\beta_{l+1}\bigr)|g_{k,j}|^2\bigr],
    \end{split}
\end{equation}
and let $\fbold_n=(\fbold_{n,1},\dots,\fbold_{n,\fd})\colon (\R^\fd)^2\to\R^\fd$ satisfy for all $g=((g_{i,j})_{j\in \{1,2,\dots,\fd\}})_{i\in\{0,1\}}\in (\R^\fd)^2$, $j\in \{1,2,\dots,\fd\}$ that 
\begin{equation}\llabel{def: bff}
    \fbold_{n,j}(g)=\gamma_{n+1} \rbr*{ \varepsilon + g_{0,j} } ^{ \nicefrac{-1}{2} }
	g_{1,j}.
\end{equation}
\argument{\lref{def: f};}{that for all $n\in \N_0$, $g=((g_{i,j})_{j\in \{1,2,\dots,\fd\}})_{i\in\{0,1\}}\in (\R^\fd)^2$, $j\in \{1,2,\dots,\fd\}$ with $\sum_{i=0}^1|g_{i,j}|=0$ it holds that
\begin{equation}\llabel{vr1}
\begin{split}
   \fnormal_{n,j}(g)=\beta_{n+1}g_{0,j}+(1-\beta_{n+1})|g_{1,j}|^2=0\beta_{n+1}+0(1-\beta_{n+1})=0
    \dott
    \end{split}
\end{equation}}
\argument{\lref{def: f};\lref{def: F}}{for all $n\in \N\backslash\{1\}$, $
	g =
	( 
	( g_{ i, j } )_{ j \in \{ 1, 2, \dots, \fd \} }
	\allowbreak)_{
		i \in \{ 0, 1, \dots, n \}
	}
	\in 
	(
	\R^{ 
		\fd
	}
	)^{ n + 1 }
	$, $j\in \{1,2,\dots,\fd\}$, $y=(y_1,\dots,y_\fd)\in \R^\fd$ that
 \begin{equation}\llabel{vr2}
     \begin{split}
   \fcapital_{0,j}(y)=(1-\beta_1)|y_j|^2 =\fnormal_{0,j}(0,y_j)
    \end{split}
    \end{equation}
and
\begin{equation}\llabel{vr3}
\begin{split}
    \textstyle
   & \fcapital_{n,j}(g_0,g_1\dots,g_n)=\textstyle\sum\limits_{k=0}^{n}\bigl[(1-\beta_{k+1})\textstyle\bigl(\prod_{l=k+1}^{n}\beta_{l+1}\bigr)|g_{k,j}|^2\bigr]\\
   &=\biggl[\textstyle\sum\limits_{k=0}^{n-1}\bigl[(1-\beta_{k+1})\textstyle\bigl(\prod_{l=k+1}^{n}\beta_{l+1}\bigr)\bigr]|g_{k,j}|^2\biggr]+(1-\beta_{n+1})|g_{n,j}|^2\\
&=\beta_{n+1}\biggl[\textstyle\sum\limits_{k=0}^{n-1}\bigl[(1-\alpha_{k+1})\textstyle\bigl(\prod_{l=k+1}^{n-1}\beta_{l+1}\bigr)|g_{k,j}|^2\bigr]\biggr]+(1-\beta_{n+1})|g_{n,j}|^2\\
&=\beta_{n+1} \fcapital_{n-1,j}(g_1,g_2,\dots,g_{n-1})+(1-\beta_{n+1})|g_{n,j}|^2\\
&=
   \fnormal_{n,j}\bigl(\fcapital_{n-1}(g_1,g_2,\dots,g_{n-1}),g_n\bigr).
   \end{split}
\end{equation}}
    \argument{\lref{def: bff}}{for all $g=((g_{i,j})_{j\in \{1,2,\dots,\fd\}}\allowbreak)_{i\in \{0,1\}}\in (\R^\fd)^2$, $j\in \{1,2,\dots,\fd\}$ with $\sum_{i=0}^1|g_{i,j}|=0$ that
    \begin{equation}\llabel{vr4}
        \fbold_{n,j}(g)=\gamma_{n+1} \rbr*{ \varepsilon + g_{0,j} } ^{ \nicefrac{-1}{2} }
	g_{1,j}=0\dott
    \end{equation}}
    \argument{\lref{def: Phi};\lref{def: F};\lref{def: bff}}{for all $n\in \N_0$, $
	g =
	( 
	( g_{ i, j } )_{ j \in \{ 1, 2, \dots, \fd \} }
	\allowbreak)_{
		i \in \{ 0, 1, \dots, n \}
	}
	\in 
	(
	\R^{ 
		\fd
	}
	)^{ n + 1 }
	$, $j\in \{1,2,\dots,\fd\}$ that
    \begin{equation}\llabel{vr5}
    \begin{split}
       \Phi_n^j(g)&= \gamma_{n+1}\biggl(\varepsilon+\textstyle\sum\limits_{k=0}^{n}\bigl[(1-\beta_{k+1})\textstyle\bigl(\prod_{l=k+1}^{n}\beta_{l+1}\bigr)|g_{k,j}|^2\bigr]\biggr)^{-\nicefrac{1}{2}}g_{n,j}\\
&=\gamma_{n+1}\bigl(\varepsilon+\fcapital_{n,j}(g)\bigr)^{-\nicefrac{1}{2}}g_{n,j}
=\fbold_{n,j}(\fcapital_n,g_n)\dott
    \end{split}
    \end{equation}}
\argument{\lref{vr1};\lref{vr2};\lref{vr3};\lref{vr4};\lref{vr5};\cref{general SGD} (applied with $\fd\curvearrowleft\fd$, $\numf\curvearrowleft 1$, $((\fnormal_{n}^m)_{m\in \{1,2,\dots,\numf\}}\allowbreak)_{n\in \N_0}\curvearrowleft((\fnormal_{n}^m)_{m\in \{1,2,\dots,\numf\}}\allowbreak)_{n\in \N_0}$, $((\fcapital_{n}^m)_{m\in \{1,2,\dots,\numf\}})_{n\in \N_0}\curvearrowleft((\fcapital_{n}^m)_{m\in \{1,2,\dots,\numf\}}\allowbreak)_{n\in \N_0}$, $((\fbold_{n,m}\allowbreak)_{m\in \{1,2,\dots,\fd\}}\allowbreak)_{n\in \N_0}\curvearrowleft((\fbold_{n,m}\allowbreak)_{m\in \{1,2,\dots,\fd\}}\allowbreak)_{n\in \N_0}$, $((\Phi_{n}^m)_{m\in \{1,2,\dots,\fd\}}\allowbreak)_{n\in \N_0}\curvearrowleft\allowbreak(\allowbreak(\Phi_{n}^m\allowbreak)_{m\in \{1,2,\dots,\fd\}}\allowbreak)_{n\in \N_0}$, $((\fG_{n}^m)_{m\in \{1,2,\dots,\fd\}}\allowbreak)_{n\in \N_0}\curvearrowleft((\fG_{n}^m)_{m\in \{1,2,\dots,\fd\}}\allowbreak)_{n\in \N_0}$, $\Theta\curvearrowleft\Theta$, $((\bfM_n^m)_{m\in \{1,2,\dots,\numf\}})_{n\in \N_0}\curvearrowleft(\bbM_n\allowbreak)_{n\in \N_0}$ in the notation of \cref{general SGD})}{\cref{item 1: rmsprop,item 2: rmsprop}\dott}
\end{aproof}
\subsection{Adaptive moment estimation SGD (Adam) optimizer}\label{subsec: Adam}
In this subsection we show in the elementary result in \cref{draf adam} below that the \Adam\ optimizer (cf.\ \cite{kingma2017adammethodstochasticoptimization}) 
satisfies the assumptions of \cref{conjecture: multilayer2} (cf.\ \cref{eq:gradients_SGD4}--\cref{eq: setting: SGD_def process_main theorem 2} in \cref{conjecture: multilayer2} and \cref{setting: multilayer: eq:gradients_SGD4:}--\cref{eq: setting: SGD_def process} in \cref{setting: SGD}). \Cref{draf adam} is a slightly strengthened version of \cite[Lemma 6.8]{Beckerlearning2023} and we also refer to, \eg, to \cite[Lemma 6.8]{Beckerlearning2023} for a special case of \cref{item 2: adam} in \cref{draf adam}.

\begin{athm}{prop}{draf adam}
   Let $\fd\in \N$, $(\gamma_n)_{n \in \N} \subseteq [0 , \infty )$,
	$(\alpha_n)_{n \in \N } \subseteq [0 , 1 ]$,
	$(\beta_n)_{n \in \N } \subseteq [0 , 1 ]$,
	$\varepsilon \in (0 , \infty ) $ satisfy $\max \cu{ \alpha_1, \beta_1 } < 1$, for every $n\in \N_0$ let $\Phi_n=(\Phi_n^1,\dots,\Phi_n^\fd)\colon (\R^\fd)^{n+1}\to \R^\fd$ satisfy for all $
	g =
	( 
	( g_{ i, j } )_{ j \in \{ 1, 2, \dots, \fd \} }
	)_{
		i \in \{ 0, 1, \dots, n \}
	}
	\in 
	(
	\R^{ 
		\fd
	}
	)^{ n + 1 }
	$, $j\in \{1,2,\dots,\fd\}$ that
    \begin{equation}\llabel{def: Phi}
    \begin{split}
    \textstyle
\Phi_n^j(g)&=\gamma_{n+1}\biggl(\varepsilon+\biggl[\frac{\sum_{k=0}^{n}\bigl[(1-\beta_{k+1})\textstyle\bigl(\prod_{l=k+1}^{n}\beta_{l+1}\bigr)|g_{k,j}|^2\bigr]}{1-\prod_{l=0}^n\beta_{l+1}}\biggr]^{\nicefrac{1}{2}}\biggr)^{-1}\\
&\cdot\frac{\sum_{k=0}^{n}\bigl[(1-\alpha_{k+1})\textstyle\bigl(\prod_{l=k+1}^{n}\alpha_{l+1}\bigr)g_{k,j}\bigr]}{1-\prod_{l=0}^{n}\alpha_{l+1}},
\end{split}
    \end{equation}
    let $(\Omega,\mathcal F,\P)$ be a probability space, for every $n\in \N_0$ let $\fG_n=(\fG_n^1,\ldots,\fG_n^\fd)\colon \R^\fd\times \Omega\to \R^\fd$ be a function, and let $\Theta\colon \N_0\times\Omega\to \R^\fd$ satisfy for all $n\in \N$ that
\begin{equation}\llabel{eq2}
    \Theta_{n} = \Theta_{n-1} - \Phi_{n-1} ( \fG_1 ( \Theta_0 ) , \fG_2 ( \Theta_1 ) , \ldots, \fG_n ( \Theta_{n-1} )).
\end{equation}
    Then
    \begin{enumerate}[label=(\roman*)]
        \item \label{item 1: adam} it holds for all $n\in\N_0$, $
	g =
	( 
	( g_{ i, j } )_{ j \in \{ 1, 2, \dots, \fd \} }
	)_{
		i \in \{ 0, 1, \dots, n \}
	}
	\in 
	(
	\R^{ 
		\fd
	}
	)^{ n + 1 }
	$, 
	$ 
	j \in \{1,2,\dots,\fd\}  
	$ with $\sum_{i=0}^n|g_{i,j}|=0$ that
        $\Phi_n^j(g)=0$ and
        \item \label{item 2: adam} it holds that there exist $\bfm = (\bfm^1 , \ldots , \bfm ^{ \fd } )  \colon \N_0 \times \Omega \to \R^\fd$,
	$\bbM = (\bbM^1 , \ldots , \bbM^\fd )  \colon \N_0 \times \Omega \to \R^\fd$ which
	satisfy for all
	$n \in \N$,
	$j \in \{1,2,\dots,\fd\}$ that
	\begin{equation}
  \bfm_0 = \bbM_0 = 0,\qquad
		\bfm_{n}  = \alpha_{n} \bfm_{n-1} + (1 - \alpha_{n} ) \fG_{n} ( \Theta_{n-1} ) , 
  \end{equation}
  \begin{equation}
	\bbM^j_{n}  = \beta_{n} \bbM^j_{n-1} + (1 - \beta_{n} ) \abs{ \fG^j_{n} ( \Theta_{n-1} ) } ^2, 
 \end{equation}
 \begin{equation}
	\text{and}\qquad\Theta_{n}^j = \Theta_{n-1} ^j - \gamma_{n} \rbr*{ \varepsilon + \br*{ \tfrac{ \bbM_{n}^j }{1 - \prod_{l=1}^{n} \beta_l } } ^{ \nicefrac{1}{2}} } ^{ - 1 }
	\frac{\bfm_{n} ^j }{1 - \prod_{l=1}^{n} \alpha_l }.
 \end{equation}
    \end{enumerate}
\end{athm}
\begin{aproof}
     Throughout this proof for every $n\in \N_0$ let $\fnormal_n^1=(\fnormal_{n,1}^1,\dots,\fnormal_{n,\fd}^1)$, $\fnormal_n^2=(\fnormal_{n,1}^2,\dots,\fnormal_{n,\fd}^2)\colon (\R^\fd)^3\to\R^\fd$ satisfy for all $g=((g_{i,j})_{j\in\{1,2,\dots,\fd\}})_{i\in\{0,1,2\}}\in (\R^\fd)^3$, $j\in \{1,2,\dots,\fd\}$ that 
\begin{equation}\llabel{def: f}
    \fnormal_n^1(g)=\alpha_{n+1} g_0+(1-\alpha_{n+1})g_2 \qqandqq \fnormal_{n,j}^2(g)=\beta_{n+1} g_{1,j}+(1-\beta_{n+1})|g_{2,j}|^2,
\end{equation} 
let $\fcapital_n^1\colon (\R^{\fd})^{n+1}\to\R^\fd$ and $\fcapital_n^2=(\fcapital_{n,1}^2,\dots,\fcapital_{n,\fd}^2)\colon (\R^{\fd})^{n+1}\to\R^\fd$ satisfy for all $g=((g_{i,j})_{j\in \{1,2,\dots,\fd\}}\allowbreak)_{i\in \{0,1,\dots,n\}}\in (\R^{\fd})^{n+1}$, $j\in \{1,2,\dots,\fd\}$ that 
\begin{equation}\llabel{def: F}
    \fcapital_n^1(g)=\sum_{k=0}^{n}\bigl[(1-\alpha_{k+1})\bigl(\textstyle\prod_{l=k+1}^{n}\alpha_{l+1}\bigr)g_k\bigr]
    \end{equation}
    \begin{equation}
    \llabel{def: F2}
    \text{and}\qquad\fcapital_{n,j}^2(g)=\sum_{k=0}^{n}\bigl[(1-\beta_{k+1})\textstyle\bigl(\prod_{l=k+1}^{n}\beta_{l+1}\bigr)\bigr)|g_{k,j}|^2\bigr],
\end{equation}
and let $\fbold_n=(\fbold_{n,1},\dots,\fbold_{n,\fd})\colon (\R^\fd)^3\to\R^\fd$ satisfy for all $g=((g_{i,j})_{j\in \{1,2,\dots,\fd\}})_{i\in\{0,1,2\}}\in (\R^\fd)^3$, $j\in \{1,2,\dots,\fd\}$ that 
\begin{equation}\llabel{def: bff}
    \fbold_{n,j}(g)=\gamma_{n+1} \rbr*{ \varepsilon + \br*{ \tfrac{g_{1,j}}{1 - \prod_{l=1}^{n+1} \beta_l } } ^{ \nicefrac{1}{2}} } ^{ - 1 }
	\frac{g_{0,j} }{1 - \prod_{l=0}^{n}\alpha_{l+1} }.
\end{equation}
\argument{\lref{def: f};}{that for all $n\in \N_0$, $g=((g_{i,j})_{j\in \{1,2,\dots,\fd\}})_{i\in\{0,1,2\}}\in (\R^\fd)^3$, $j\in \{1,2,\dots,\fd\}$ with $\sum_{i=0}^2|g_{i,j}|=0$ it holds that
\begin{equation}\llabel{vr1}
    \fnormal_{n,j}^1(g)=\alpha _{n+1} g_{0,j}+(1-\alpha_{n+1}) g_{2,j}=0\alpha_{n+1}+0(1-\alpha_{n+1})=0
    \end{equation}
    and
    \begin{equation}\llabel{vr1.1}
 \fnormal_{n,j}^2(g)=\beta_{n+1}g_{1,j}+(1-\beta_{n+1})|g_{2,j}|^2=0\beta_{n+1}+0(1-\beta_{n+1})=0
    \dott
\end{equation}}
\argument{\lref{def: f};\lref{def: F};\lref{def: F2}}{for all $n\in \N\backslash\{1\}$, $
	g =
	( 
	( g_{ i, j } )_{ j \in \{ 1, 2, \dots, \fd \} }
	\allowbreak)_{
		i \in \{ 0, 1, \dots, n \}
	}
	\in 
	(
	\R^{ 
		\fd
	}
	)^{ n + 1 }
	$, $j\in \{1,2,\dots,\fd\}$, $y=(y_1,\dots,y_\fd)\in \R^\fd$ that
 \begin{equation}\llabel{vr2}
     \begin{split}
    \fcapital_0^1(y)=(1-\alpha_1)y =\fnormal_0^1(0,0,y),\qquad \fcapital_{0,j}^2(y)=(1-\beta_1)|y_j|^2 =\fnormal_{0,j}^2(0,0,y_j),
    \end{split}
    \end{equation}
\begin{equation}\llabel{vr3}
\begin{split}
\textstyle
   & \fcapital_n^{1}(g_0,g_1\dots,g_n)=\textstyle\sum\limits_{k=0}^{n}\bigl[(1-\alpha_{k+1})\textstyle\bigl(\prod_{l=k+1}^{n}\alpha_{l+1}\bigr)g_k\bigr]\\
   &=\biggl[\textstyle\sum\limits_{k=0}^{n-1}\bigl[(1-\alpha_{k+1})\textstyle\bigl(\prod_{l=k+1}^{n}\alpha_{l+1}\bigr)\bigr]g_k\biggr]+(1-\alpha_{n+1})g_{n}\\
&=\alpha_{n+1}\biggl[\textstyle\sum\limits_{k=0}^{n-1}\bigl[(1-\alpha_{k+1})\textstyle\bigl(\prod_{l=k+1}^{n-1}\alpha_{l+1}\bigr)g_k\bigr]\biggr]+(1-\alpha_{n+1})g_{n}\\
&=\alpha_{n+1} \fcapital_{n-1}^1(g_1,g_2,\dots,g_{n-1})+(1-\alpha_{n+1})g_n\\
&=
   \fnormal_n^1\bigl(\fcapital_{n-1}^1(g_1,g_2,\dots,g_{n-1}),\fcapital_{n-1}^2(g_1,g_2,\dots,g_{n-1}),g_n\bigr),
    \end{split}
    \end{equation}
and
\begin{equation}\llabel{vr3.5}
\begin{split}
    \textstyle
   & \fcapital_{n,j}^{2}(g_0,g_1\dots,g_n)=\textstyle\sum\limits_{k=0}^{n}\bigl[(1-\beta_{k+1})\textstyle\bigl(\prod_{l=k+1}^{n}\beta_{l+1}\bigr)|g_{k,j}|^2\bigr]\\
   &=\biggl[\textstyle\sum\limits_{k=0}^{n-1}\bigl[(1-\beta_{k+1})\textstyle\bigl(\prod_{l=k+1}^{n}\beta_{l+1}\bigr)\bigr]|g_{k,j}|^2\biggr]+(1-\beta_{n+1})|g_{n,j}|^2\\
&=\beta_{n+1}\biggl[\textstyle\sum\limits_{k=0}^{n-1}\bigl[(1-\alpha_{k+1})\textstyle\bigl(\prod_{l=k+1}^{n-1}\beta_{l+1}\bigr)|g_{k,j}|^2\bigr]\biggr]+(1-\beta_{n+1})|g_{n,j}|^2\\
&=\beta_{n+1} \fcapital_{n-1,j}^2(g_1,g_2,\dots,g_{n-1})+(1-\beta_{n+1})|g_{n,j}|^2\\
&=
   \fnormal_{n,j}^2\bigl(\fcapital_{n-1}^1(g_1,g_2,\dots,g_{n-1}),\fcapital_{n-1}^2(g_1,g_2,\dots,g_{n-1}),g_n\bigr).
   \end{split}
\end{equation}}
    \argument{\lref{def: bff}}{for all $g=((g_{i,j})_{j\in \{1,2,\dots,\fd\}}\allowbreak)_{i\in \{0,1,2\}}\in (\R^\fd)^3$, $j\in \{1,2,\dots,\fd\}$ with $\sum_{i=0}^2|g_{i,j}|=0$ that
    \begin{equation}\llabel{vr4}
        \fbold_{n,j}(g)=\gamma_{n+1} \rbr*{ \varepsilon + \br*{ \tfrac{g_{1,j}}{1 - \prod_{l=0}^{n} \beta_{l+1} } } ^{ \nicefrac{1}{2}} } ^{ - 1 }
	\frac{g_{0,j} }{1 - \prod_{l=0}^{n}\alpha_{l+1} }=0\dott
    \end{equation}}
    \argument{\lref{def: Phi};\lref{def: F};\lref{def: F2};\lref{def: bff}}{for all $n\in \N_0$, $
	g =
	( 
	( g_{ i, j } )_{ j \in \{ 1, 2, \dots, \fd \} }
	\allowbreak)_{
		i \in \{ 0, 1, \dots, n \}
	}
	\in 
	(
	\R^{ 
		\fd
	}
	)^{ n + 1 }
	$, $j\in \{1,2,\dots,\fd\}$ that
    \begin{equation}\llabel{vr5}
    \begin{split}
       \Phi_n^j(g)&= \gamma_{n+1}\biggl(\varepsilon+\biggl[\frac{\sum_{k=0}^{n}\bigl[(1-\beta_{k+1})\textstyle\bigl(\prod_{l=k+1}^{n}\beta_{l+1}\bigr)|g_{k,j}|^2\bigr]}{1-\prod_{l=0}^n\beta_{l+1}}\biggr]^{\nicefrac{1}{2}}\biggr)^{-1}\\
&\quad\cdot\frac{\sum_{k=0}^{n}\bigl[(1-\alpha_{k+1})\textstyle\bigl(\prod_{l=k+1}^{n}\alpha_{l+1}\bigr)g_{k,j}\bigr]}{1-\prod_{l=0}^{n}\alpha_{l+1}}\\
&=\gamma_{n+1} \rbr*{ \varepsilon + \br*{ \tfrac{\fcapital_{n,j}^2(g)}{1 - \prod_{l=0}^{n} \beta_{l+1} } } ^{ \nicefrac{1}{2}} } ^{ - 1 }
	\frac{\fcapital_{n,j}^1(g)}{1 - \prod_{l=0}^{n}\alpha_{l+1} }
=\fbold_{n,j}(\fcapital_n^1(g),\fcapital_n^2(g),g_n)\dott
    \end{split}
    \end{equation}}
\argument{\lref{vr1};\lref{vr1.1};\lref{vr2};\lref{vr3};\lref{vr3.5};\lref{vr4};\lref{vr5};\cref{general SGD} (applied with $\fd\curvearrowleft\fd$, $\numf\curvearrowleft 2$, $((\fnormal_{n}^m)_{m\in \{1,2,\dots,\numf\}}\allowbreak)_{n\in \N_0}\curvearrowleft((\fnormal_{n}^m)_{m\in \{1,2,\dots,\numf\}}\allowbreak)_{n\in \N_0}$, $((\fcapital_{n}^m)_{m\in \{1,2,\dots,\numf\}})_{n\in \N_0}\curvearrowleft((\fcapital_{n}^m)_{m\in \{1,2,\dots,\numf\}}\allowbreak)_{n\in \N_0}$, $((\fbold_{n,m}\allowbreak)_{m\in \{1,2,\dots,\fd\}}\allowbreak)_{n\in \N_0}\curvearrowleft((\fbold_{n,m}\allowbreak)_{m\in \{1,2,\dots,\fd\}}\allowbreak)_{n\in \N_0}$, $((\Phi_{n}^m)_{m\in \{1,2,\dots,\fd\}}\allowbreak)_{n\in \N_0}\curvearrowleft\allowbreak(\allowbreak(\Phi_{n}^m\allowbreak)_{m\in \{1,2,\dots,\fd\}}\allowbreak)_{n\in \N_0}$, $((\fG_{n}^m)_{m\in \{1,2,\dots,\fd\}}\allowbreak)_{n\in \N_0}\curvearrowleft((\fG_{n}^m)_{m\in \{1,2,\dots,\fd\}}\allowbreak)_{n\in \N_0}$, $\Theta\curvearrowleft\Theta$, $((\bfM_n^m)_{m\in \{1,2,\dots,\numf\}})_{n\in \N_0}\curvearrowleft(\bfm_n,\allowbreak\bbM_n\allowbreak)_{n\in \N_0}$ in the notation of \cref{general SGD})}{\cref{item 1: adam,item 2: adam}\dott}
\end{aproof}
\subsection{Adaptive moment estimation maximum SGD (Adamax) optimizer}\label{subsec: Adamax}
In this subsection we show in the elementary result in \cref{draf adamax} below that the \Adamax\ optimizer (cf.\ \cite{kingma2017adammethodstochasticoptimization}) 
satisfies the assumptions of \cref{conjecture: multilayer2} (cf.\ \cref{eq:gradients_SGD4}--\cref{eq: setting: SGD_def process_main theorem 2} in \cref{conjecture: multilayer2} and \cref{setting: multilayer: eq:gradients_SGD4:}--\cref{eq: setting: SGD_def process} in \cref{setting: SGD}). We refer, \eg, to \cite[Lemma 6.7]{Beckerlearning2023} for a special case of \cref{item 2: adamax} in \cref{draf adamax}.

\begin{athm}{prop}{draf adamax}
   Let $\fd\in \N$, $(\gamma_n)_{n \in \N} \subseteq [0 , \infty )$,
	$(\alpha_n)_{n \in \N } \subseteq [0 , 1 ]$,
	$(\beta_n)_{n \in \N } \subseteq [0 , 1 ]$,
	$\varepsilon \in (0 , \infty ) $ satisfy $\alpha_1 < 1$, for every $n\in \N_0$ let $\Phi_n=(\Phi_n^1,\dots,\Phi_n^\fd)\colon (\R^\fd)^{n+1}\to \R^\fd$ satisfy for all $
	g =
	( 
	( g_{ i, j } )_{ j \in \{ 1, 2, \dots, \fd \} }
	)_{
		i \in \{ 0, 1, \dots, n \}
	}
	\in 
	(
	\R^{ 
		\fd
	}
	)^{ n + 1 }
	$, $j\in \{1,2,\dots,\fd\}$ that
    \begin{equation}\llabel{def: Phi}
    \begin{split}
    \textstyle
&\Phi_n^j(g)\\
&=\gamma_{n+1}\Bigl(\varepsilon+\textstyle\bigl[\max_{k\in \{0,1,\dots,n\}}\bigl(\textstyle\prod_{l=k+1}^{n} \beta_{l+1}\bigr)|g_{k,j}|\bigr]\Bigr)^{-1}\frac{\sum_{k=0}^{n}\bigl[(1-\alpha_{k+1})\textstyle\bigl(\prod_{l=k+1}^{n}\alpha_{l+1}\bigr)g_{k,j}\bigr]}{1-\prod_{l=0}^{n}\alpha_{l+1}},
\end{split}
    \end{equation}
    let $(\Omega,\mathcal F,\P)$ be a probability space, for every $n\in \N_0$ let $\fG_n=(\fG_n^1,\ldots,\fG_n^\fd)\colon \R^\fd\times \Omega\to \R^\fd$ be a function, and let $\Theta\colon \N_0\times\Omega\to \R^\fd$ satisfy for all $n\in \N$ that
\begin{equation}\llabel{eq2}
    \Theta_{n} = \Theta_{n-1} - \Phi_{n-1} ( \fG_1 ( \Theta_0 ) , \fG_2 ( \Theta_1 ) , \ldots, \fG_n ( \Theta_{n-1} )).
\end{equation}
    Then
    \begin{enumerate}[label=(\roman*)]
        \item \label{item 1: adamax} it holds for all $n\in\N_0$, $
	g =
	( 
	( g_{ i, j } )_{ j \in \{ 1, 2, \dots, \fd \} }
	)_{
		i \in \{ 0, 1, \dots, n \}
	}
	\in 
	(
	\R^{ 
		\fd
	}
	)^{ n + 1 }
	$, 
	$ 
	j \in \{1,2,\dots,\fd\}  
	$ with $\sum_{i=0}^n|g_{i,j}|=0$ that
        $\Phi_n^j(g)=0$ and
        \item \label{item 2: adamax} it holds that there exist $\bfm = (\bfm^1 , \ldots , \bfm ^{ \fd } )  \colon \N_0 \times \Omega \to \R^\fd$,
	$\bbM = (\bbM^1 , \ldots , \bbM^\fd )  \colon \N_0 \times \Omega \to \R^\fd$ which
	satisfy for all
	$n \in \N$,
	$j \in \{1,2,\dots,\fd\}$ that
	\begin{equation}
  \bfm_0 = \bbM_0 = 0,\qquad
		\bfm_{n}  = \alpha_{n} \bfm_{n-1} + (1 - \alpha_{n} ) \fG_{n} ( \Theta_{n-1} ) , 
  \end{equation}
  \begin{equation}
	\bbM^j_{n}  = \max\{\beta_{n} \bbM^j_{n-1},\abs{ \fG^j_{n} ( \Theta_{n-1} ) }\}, 
 \end{equation}
 \begin{equation}
	\text{and}\qquad\Theta_{n}^j = \Theta_{n-1} ^j - \gamma_{n} \rbr*{ \varepsilon +  \bbM_{n}^j  } ^{ - 1 }
	\frac{\bfm_{n} ^j }{1 - \prod_{l=1}^{n} \alpha_l }.
 \end{equation}
    \end{enumerate}
\end{athm}
\begin{aproof}
     Throughout this proof for every $n\in \N_0$ let $\fnormal_n^1=(\fnormal_{n,1}^1,\dots,\fnormal_{n,\fd}^1)$, $\fnormal_n^2=(\fnormal_{n,1}^2,\dots,\fnormal_{n,\fd}^2)\colon (\R^\fd)^3\to\R^\fd$ satisfy for all $g=((g_{i,j})_{j\in\{1,2,\dots,\fd\}})_{i\in\{0,1,2\}}\in (\R^\fd)^3$, $j\in \{1,2,\dots,\fd\}$ that 
\begin{equation}\llabel{def: f}
    \fnormal_n^1(g)=\alpha_{n+1} g_0+(1-\alpha_{n+1})g_2 \qqandqq \fnormal_{n,j}^2(g)=\max\{|g_{2,j}|,\beta_{n+1} |g_{1,j}|\},
\end{equation} 
let $\fcapital_n^1\colon (\R^{\fd})^{n+1}\to\R^\fd$ and $\fcapital_n^2=(\fcapital_{n,1}^2,\dots,\fcapital_{n,\fd}^2)\colon (\R^{\fd})^{n+1}\to\R^\fd$ satisfy for all $g=((g_{i,j})_{j\in \{1,2,\dots,\fd\}}\allowbreak)_{i\in \{0,1,\dots,n\}}\in (\R^{\fd})^{n+1}$, $j\in \{1,2,\dots,\fd\}$ that 
\begin{equation}\llabel{def: F}
    \fcapital_n^1(g)=\sum_{k=0}^{n}\bigl[(1-\alpha_{k+1})\bigl(\textstyle\prod_{l=k+1}^{n}\alpha_{l+1}\bigr)g_k\bigr]
    \end{equation}
    \begin{equation}
    \llabel{def: F2}
    \text{and}\qquad\fcapital_{n,j}^2(g)=\max_{k\in \{0,1,\dots,n\}}\bigl(\textstyle\prod_{l=k+1}^{n} \beta_{l+1}\bigr)|g_{k,j}|,
\end{equation}
and let $\fbold_n=(\fbold_{n,1},\dots,\fbold_{n,\fd})\colon (\R^\fd)^3\to\R^\fd$ satisfy for all $g=((g_{i,j})_{j\in \{1,2,\dots,\fd\}})_{i\in\{0,1,2\}}\in (\R^\fd)^3$, $j\in \{1,2,\dots,\fd\}$ that 
\begin{equation}\llabel{def: bff}
    \fbold_{n,j}(g)=\gamma_{n+1} \rbr*{ \varepsilon + g_{1,j}} ^{ - 1 }
	\frac{g_{0,j} }{1 - \prod_{l=0}^{n}\alpha_{l+1} }.
\end{equation}
\argument{\lref{def: f};}{that for all $n\in \N_0$, $g=((g_{i,j})_{j\in \{1,2,\dots,\fd\}})_{i\in\{0,1,2\}}\in (\R^\fd)^3$, $j\in \{1,2,\dots,\fd\}$ with $\sum_{i=0}^2|g_{i,j}|=0$ it holds that
\begin{equation}\llabel{vr1}
    \fnormal_{n,j}^1(g)=\alpha _{n+1} g_{0,j}+(1-\alpha_{n+1}) g_{2,j}=0\alpha_{n+1}+0(1-\alpha_{n+1})=0
    \end{equation}
    and
    \begin{equation}\llabel{vr1.1}
 \fnormal_{n,j}^2(g)=\max\{|g_{2,j}|,\beta_{n+1}|g_{1,j}|\}=\max\{0,0\beta_{n+1}\}=0
    \dott
\end{equation}}
\argument{\lref{def: f};\lref{def: F};\lref{def: F2}}{for all $n\in \N\backslash\{1\}$, $
	g =
	( 
	( g_{ i, j } )_{ j \in \{ 1, 2, \dots, \fd \} }
	\allowbreak)_{
		i \in \{ 0, 1, \dots, n \}
	}
	\in 
	(
	\R^{ 
		\fd
	}
	)^{ n + 1 }
	$, $j\in \{1,2,\dots,\fd\}$, $y=(y_1,\dots,y_\fd)\in \R^\fd$ that
 \begin{equation}\llabel{vr2}
     \begin{split}
    \fcapital_0^1(y)=(1-\alpha_1)y =\fnormal_0^1(0,0,y),\qquad \fcapital_{0,j}^2(y)=|y_j|=\fnormal_{0,j}^2(0,0,y_j),
    \end{split}
    \end{equation}
\begin{equation}\llabel{vr3}
\begin{split}
\textstyle
   & \fcapital_n^{1}(g_0,g_1\dots,g_n)=\textstyle\sum\limits_{k=0}^{n}\bigl[(1-\alpha_{k+1})\textstyle\bigl(\prod_{l=k+1}^{n}\alpha_{l+1}\bigr)g_k\bigr]\\
   &=\biggl[\textstyle\sum\limits_{k=0}^{n-1}\bigl[(1-\alpha_{k+1})\textstyle\bigl(\prod_{l=k+1}^{n}\alpha_{l+1}\bigr)\bigr]g_k\biggr]+(1-\alpha_{n+1})g_{n}\\
&=\alpha_{n+1}\biggl[\textstyle\sum\limits_{k=0}^{n-1}\bigl[(1-\alpha_{k+1})\textstyle\bigl(\prod_{l=k+1}^{n-1}\alpha_{l+1}\bigr)g_k\bigr]\biggr]+(1-\alpha_{n+1})g_{n}\\
&=\alpha_{n+1} \fcapital_{n-1}^1(g_1,g_2,\dots,g_{n-1})+(1-\alpha_{n+1})g_n\\
&=
   \fnormal_n^1\bigl(\fcapital_{n-1}^1(g_1,g_2,\dots,g_{n-1}),\fcapital_{n-1}^2(g_1,g_2,\dots,g_{n-1}),g_n\bigr),
    \end{split}
    \end{equation}
and
\begin{equation}\llabel{vr3.5}
\begin{split}
   &\textstyle\fcapital_{n,j}^{2}(g_0,g_1\dots,g_n)=\max_{k\in \{0,1,\dots,n\}}\bigl(\textstyle\prod_{l=k+1}^{n} \beta_{l+1}\bigr)|g_{k,j}|\\
   &=\textstyle\max\bigl\{\bigl[\max_{k\in \{0,1,\dots,n-1\}}\bigl(\textstyle\prod_{l=k+1}^{n} \beta_{l+1}\bigr)|g_{k,j}|\bigr],|g_{n,j}|\bigr\}\\
   &=\textstyle\max\{\beta_{n+1}\bigl[\max_{k\in \{0,1,\dots,n-1\}}\bigl(\textstyle\prod_{l=k+1}^{n-1} \beta_{l+1}\bigr)|g_{k,j}|\bigr],|g_{n,j}|\}\\
   &=\max\{\beta_{n+1}\fcapital_{n-1,j}^2(g_1,g_2,\dots,g_{n-1}),|g_{n,j}|\}\\
&=\fnormal_{n,j}^2\bigl(\fcapital_{n-1}^1(g_1,g_2,\dots,g_{n-1}),\fcapital_{n-1}^2(g_1,g_2,\dots,g_{n-1}),g_n\bigr).
   \end{split}
\end{equation}}
    \argument{\lref{def: bff}}{for all $g=((g_{i,j})_{j\in \{1,2,\dots,\fd\}}\allowbreak)_{i\in \{0,1,2\}}\in (\R^\fd)^3$, $j\in \{1,2,\dots,\fd\}$ with $\sum_{i=0}^2|g_{i,j}|=0$ that
    \begin{equation}\llabel{vr4}
        \fbold_{n,j}(g)=\gamma_{n+1} \rbr*{ \varepsilon + g_{1,j}} ^{ - 1 }
	\frac{g_{0,j} }{1 - \prod_{l=0}^{n}\alpha_{l+1} }=0\dott
    \end{equation}}
    \argument{\lref{def: Phi};\lref{def: F};\lref{def: F2};\lref{def: bff}}{for all $n\in \N_0$, $
	g =
	( 
	( g_{ i, j } )_{ j \in \{ 1, 2, \dots, \fd \} }
	\allowbreak)_{
		i \in \{ 0, 1, \dots, n \}
	}
	\in 
	(
	\R^{ 
		\fd
	}
	)^{ n + 1 }
	$, $j\in \{1,2,\dots,\fd\}$ that
    \begin{equation}\llabel{vr5}
    \begin{split}
      & \Phi_n^j(g)\\
      &=\gamma_{n+1}\Bigl(\varepsilon+\textstyle\bigl[\max_{k\in \{0,1,\dots,n\}}\bigl(\textstyle\prod_{l=k+1}^{n} \beta_{l+1}\bigr)|g_{k,j}|\bigr]\Bigr)^{-1}\frac{\sum_{k=0}^{n}\bigl[(1-\alpha_{k+1})\textstyle\bigl(\prod_{l=k+1}^{n}\alpha_{l+1}\bigr)g_{k,j}\bigr]}{1-\prod_{l=0}^{n}\alpha_{l+1}}\\
&=\gamma_{n+1} \rbr*{ \varepsilon +  \fcapital_{n,j}^2(g)} ^{ - 1 }
	\frac{\fcapital_{n,j}^1(g)}{1 - \prod_{l=0}^{n}\alpha_{l+1} }
=\fbold_{n,j}(\fcapital_n^1(g),\fcapital_n^2(g),g_n)\dott
    \end{split}
    \end{equation}}
\argument{\lref{vr1};\lref{vr1.1};\lref{vr2};\lref{vr3};\lref{vr3.5};\lref{vr4};\lref{vr5};\cref{general SGD} (applied with $\fd\curvearrowleft\fd$, $\numf\curvearrowleft 2$, $((\fnormal_{n}^m)_{m\in \{1,2,\dots,\numf\}}\allowbreak)_{n\in \N_0}\curvearrowleft((\fnormal_{n}^m)_{m\in \{1,2,\dots,\numf\}}\allowbreak)_{n\in \N_0}$, $((\fcapital_{n}^m)_{m\in \{1,2,\dots,\numf\}})_{n\in \N_0}\curvearrowleft((\fcapital_{n}^m)_{m\in \{1,2,\dots,\numf\}}\allowbreak)_{n\in \N_0}$, $((\fbold_{n,m}\allowbreak)_{m\in \{1,2,\dots,\fd\}}\allowbreak)_{n\in \N_0}\curvearrowleft((\fbold_{n,m}\allowbreak)_{m\in \{1,2,\dots,\fd\}}\allowbreak)_{n\in \N_0}$, $((\Phi_{n}^m)_{m\in \{1,2,\dots,\fd\}}\allowbreak)_{n\in \N_0}\curvearrowleft\allowbreak(\allowbreak(\Phi_{n}^m\allowbreak)_{m\in \{1,2,\dots,\fd\}}\allowbreak)_{n\in \N_0}$, $((\fG_{n}^m)_{m\in \{1,2,\dots,\fd\}}\allowbreak)_{n\in \N_0}\curvearrowleft((\fG_{n}^m)_{m\in \{1,2,\dots,\fd\}}\allowbreak)_{n\in \N_0}$, $\Theta\curvearrowleft\Theta$, $((\bfM_n^m)_{m\in \{1,2,\dots,\numf\}})_{n\in \N_0}\curvearrowleft(\bfm_n,\allowbreak\bbM_n\allowbreak)_{n\in \N_0}$ in the notation of \cref{general SGD})}{\cref{item 1: adamax,item 2: adamax}\dott}
\end{aproof}
\subsection{AMSGrad optimizer}\label{subsec: AMSgrad}
In this subsection we show in the elementary result in \cref{draf amsgrad} below that the AMSGrad optimizer (cf.\ \cite{ReddiKale2019}) 
satisfies the assumptions of \cref{conjecture: multilayer2} (cf.\ \cref{eq:gradients_SGD4}--\cref{eq: setting: SGD_def process_main theorem 2} in \cref{conjecture: multilayer2} and \cref{setting: multilayer: eq:gradients_SGD4:}--\cref{eq: setting: SGD_def process} in \cref{setting: SGD}).
\begin{athm}{prop}{draf amsgrad}
   Let $\fd\in \N$, $(\gamma_n)_{n \in \N  } \subseteq [0 , \infty )$,
	$(\alpha_n)_{n \in \N } \subseteq [0 , 1 ]$,
	$(\beta_n)_{n \in \N } \subseteq [0 , 1 ]$,
	$\varepsilon \in (0 , \infty ) $ satisfy $\max \cu{ \alpha_1, \beta_1 } < 1$, for every $n\in \N_0$ let $\Phi_n=(\Phi_n^1,\dots,\Phi_n^\fd)\colon (\R^\fd)^{n+1}\to \R^\fd$ satisfy for all $
	g =
	( 
	( g_{ i, j } )_{ j \in \{ 1, 2, \dots, \fd \} }
	)_{
		i \in \{ 0, 1, \dots, n \}
	}
	\in 
	(
	\R^{ 
		\fd
	}
	)^{ n + 1 }
	$, $j\in \{1,2,\dots,\fd\}$ that
    \begin{equation}\llabel{def: Phi}
    \begin{split}
    \textstyle
\Phi_n^j(g)
&=\gamma_{n+1}\biggl(\varepsilon+\max_{m\in \{0,1,\dots,n\}}\biggl[\textstyle\sum\limits_{k=0}^{m}\bigl[(1-\beta_{k+1})\textstyle\bigl(\prod_{l=k+1}^{m}\beta_{l+1}\bigr)|g_{k,i}|^2\bigr]\biggr]^{\nicefrac{1}{2}}\biggr)^{-1}\\
&\quad\cdot \biggl[\textstyle\sum\limits_{k=0}^{n}\bigl[(1-\alpha_{k+1})\textstyle\bigl(\prod_{l=k+1}^{n}\alpha_{l+1}\bigr)g_{k,j}\bigr]\biggr],
\end{split}
    \end{equation}
    let $(\Omega,\mathcal F,\P)$ be a probability space, for every $n\in \N_0$ let $\fG_n=(\fG_n^1,\ldots,\fG_n^\fd)\colon \R^\fd\times \Omega\to \R^\fd$ be a function, and let $\Theta\colon \N_0\times\Omega\to \R^\fd$ satisfy for all $n\in \N$ that
\begin{equation}\llabel{eq2}
    \Theta_{n} = \Theta_{n-1} - \Phi_{n-1} ( \fG_1 ( \Theta_0 ) , \fG_2 ( \Theta_1 ) , \ldots, \fG_n ( \Theta_{n-1} )).
\end{equation}
    Then
    \begin{enumerate}[label=(\roman*)]
        \item \llabel{item 1} it holds for all $n\in\N_0$, $
	g =
	( 
	( g_{ i, j } )_{ j \in \{ 1, 2, \dots, \fd \} }
	)_{
		i \in \{ 0, 1, \dots, n \}
	}
	\in 
	(
	\R^{ 
		\fd
	}
	)^{ n + 1 }
	$, 
	$ 
	j \in \{1,2,\dots,\fd\}  
	$ with $\sum_{i=0}^n|g_{i,j}|=0$ that
        $\Phi_n^j(g)=0$ and
        \item \llabel{item 2} it holds that there exist $\bfm = (\bfm^1 , \ldots , \bfm ^{ \fd } )  \colon \N_0 \times \Omega \to \R^\fd$, 
	$\bbM = (\bbM^1 , \ldots , \bbM^\fd )  \colon \N_0 \times \Omega \to \R^\fd$, and $\fM=(\fM^{1}, \ldots, \fM^{\fd})\colon \N_0\times \Omega \rightarrow \mathbb{R}^{\fd}$ which
	satisfy for all
	$n \in \N$,
	$j \in \{1,2,\dots,\fd\}$ that
	\begin{equation}
		\begin{split} 
  \bfm_0 = \bbM_0 =\fM_0 =0,\qquad
\mathbf{m}_{n}=\alpha_{n} \mathbf{m}_{n-1}+(1-\alpha_{n}) \fG_n(\Theta_{n-1}),
\end{split}
\end{equation}
\begin{equation}
    \begin{split}
\mathbb{M}_{n}^{j}=\beta_{n} \mathbb{M}_{n-1}^{j}+(1-\beta_{n})|\fG_n^j(\Theta_{n})|^2, \qquad
\fM_n^j=\max\{\fM_{n-1}^j,\bbM_n^j\},
\end{split}
\end{equation}
\begin{equation}
\text { and } \qquad \Theta_n^{j}=\Theta_{n-1}^{j}-\gamma_{n}\bigl[\varepsilon+(\fM_n^{j})^{1 / 2}\bigr]^{-1} \bfm_n^{j}.
 \end{equation}
    \end{enumerate}
\end{athm}
\begin{aproof}
     Throughout this proof for every $n\in \N_0$ let $\fnormal_n^1=(\fnormal_{n,1}^1,\dots,\fnormal_{n,\fd}^1)$, $\fnormal_n^2=(\fnormal_{n,1}^2,\dots,\fnormal_{n,\fd}^2), \fnormal_n^3=(\fnormal_{n,1}^3,\dots,\fnormal_{n,\fd}^3)\colon (\R^\fd)^4\to\R^\fd$ satisfy for all $g=((g_{i,j})_{j\in\{1,2,\dots,\fd\}})_{i\in\{0,1,2,3\}}\allowbreak\in (\R^\fd)^4$, $j\in \{1,2,\dots,\fd\}$ that 
     \begin{equation}\llabel{def: f}
     \fnormal_n^1(g)=\alpha_{n+1} g_0+(1-\alpha_{n+1})g_3,\qquad \fnormal_{n,j}^2(g)=\beta_{n+1} g_{1,j}+(1-\beta_{n+1})|g_{3,j}|^2,
     \end{equation} 
     and
\begin{equation}\llabel{def: f1}
    \fnormal_{n,j}^3(g)=\max\{g_2,\beta_{n+1}g_{1,j}+(1-\beta_{n+1})|g_{3,j}|^2\},
\end{equation} 
let $\fcapital_n^1=(\fcapital_{n,1}^1,\ldots,\fcapital_{n,\fd}^1)\colon (\R^{\fd})^{n+1}\to\R^\fd$, $\fcapital_n^2=(\fcapital_{n,1}^2,\dots,\fcapital_{n,\fd}^2)\colon (\R^{\fd})^{n+1}\to\R^\fd$, $\fcapital_n^3=(\fcapital_{n,1}^3,\dots,\fcapital_{n,\fd}^3)\colon (\R^{\fd})^{n+1}\to\R^\fd$ satisfy for all $g=((g_{i,j})_{j\in \{1,2,\dots,\fd\}}\allowbreak)_{i\in \{0,1,\dots,n\}}\in (\R^{\fd})^{n+1}$, $j\in \{1,2,\dots,\fd\}$ that 
\begin{equation}\llabel{def: F}
    \fcapital_n^1(g)=\sum_{k=0}^{n}\bigl[(1-\alpha_{k+1})\textstyle\bigl(\prod_{l=k+1}^{n}\alpha_{l+1}\bigr)g_k\bigr],
    \end{equation}
    \begin{equation}
     \fcapital_{n,j}^2(g)=\sum_{k=0}^{n}\bigl[(1-\beta_{k+1})\textstyle\bigl(\prod_{l=k+1}^{n}\beta_{l+1}\bigr)|g_{k,j}|^2\bigr],
     \end{equation}
     and
     \begin{equation}
     \fcapital_{n,j}^3(g)=\max_{m\in \{0,1,\dots,n\}} \biggl[\textstyle\sum\limits_{k=0}^{m}\bigl[(1-\beta_{k+1})\textstyle\bigl(\prod_{l=k+1}^{m}\beta_{l+1}\bigr)|g_{k,j}|^2\bigr]\biggr],
\end{equation}
and let $\fbold_n=(\fbold_{n,1},\dots,\fbold_{n,\fd})\colon (\R^\fd)^4\to\R^\fd$ satisfy for all $g=((g_{i,j})_{j\in \{1,2,\dots,\fd\}})_{i\in\{0,1,2,3\}}\in (\R^\fd)^4$, $j\in \{1,2,\dots,\fd\}$ that 
\begin{equation}\llabel{def: bff}
    \fbold_{n,j}(g)=\gamma_{n+1}\bigl[\textstyle\varepsilon+(g_{2,j})^{1 / 2}\bigr]^{-1} g_{0,j}.
\end{equation}
\argument{\lref{def: f};}{that for all $n\in \N_0$, $g=((g_{i,j})_{j\in \{1,2,\dots,\fd\}})_{i\in\{0,1,2,3\}}\in (\R^\fd)^4$, $j\in \{1,2,\dots,\fd\}$ with $\sum_{i=0}^3|g_{i,j}|=0$ it holds that
\begin{equation}\llabel{vr1}
    \fnormal_{n,j}^1(g)=\alpha _{n+1} g_{0,j}+(1-\alpha_{n+1}) g_{3,j}=0\alpha_{n+1}+0(1-\alpha_{n+1})=0,
    \end{equation}
    \begin{equation}
   \fnormal_{n,j}^2(g)=\beta_{n+1}g_{1,j}+(1-\beta_{n+1})|g_{3,j}|^2=0\beta_{n+1}+0(1-\beta_{n+1})=0,
    \end{equation}
    \begin{equation}
   \text{and}\qquad \fnormal_{n,j}^3(g)= \max\{g_2,\beta_{n+1}g_{1,j}+(1-\beta_{n+1})|g_{3,j}|^2\}=0
    \dott
\end{equation}}
\argument{\lref{def: f};\lref{def: F}}{for all $n\in \N\backslash\{1\}$, $
	g =
	( 
	( g_{ i, j } )_{ j \in \{ 1, 2, \dots, \fd \} }
	\allowbreak)_{
		i \in \{ 0, 1, \dots, n \}
	}
	\in 
	(
	\R^{ 
		\fd
	}
	)^{ n + 1 }
	$, $j\in \{1,2,\dots,\fd\}$, $y=(y_1,\dots,y_\fd)\in \R^\fd$ that
 \begin{equation}\llabel{vr2}
\fcapital_0^1(y)=(1-\alpha_1)y =\fnormal_0^1(0,0,0,y),\qquad \fcapital_{0,j}^2(y)=(1-\beta_1)|y_j|^2 =\fnormal_{0,j}^2(0,0,0,y_j),
\end{equation}
\begin{equation}\llabel{vr2.1}
    \fcapital_0^3(y)=(1-\beta_1)|y_j|^2=\fnormal_{0}^3(0,0,0,y),
    \end{equation}
\begin{equation}\llabel{vr3}
\begin{split}
\textstyle
   & \fcapital_n^{1}(g_0,g_1\dots,g_n)=\textstyle\sum\limits_{k=0}^{n}\bigl[(1-\alpha_{k+1})\textstyle\bigl(\prod_{l=k+1}^{n}\alpha_{l+1}\bigr)g_k\bigr]\\
   &=\biggl[\textstyle\sum\limits_{k=0}^{n-1}\bigl[(1-\alpha_{k+1})\textstyle\bigl(\prod_{l=k+1}^{n}\alpha_{l+1}\bigr)\bigr]g_k\biggr]+(1-\alpha_{n+1})g_{n}\\
&=\alpha_{n+1}\biggl[\textstyle\sum\limits_{k=0}^{n-1}\bigl[(1-\alpha_{k+1})\textstyle\bigl(\prod_{l=k+1}^{n-1}\alpha_{l+1}\bigr)g_k\bigr]\biggr]+(1-\alpha_{n+1})g_{n}\\
&=\alpha_{n+1} \fcapital_{n-1}^1(g_1,g_2,\dots,g_{n-1})+(1-\alpha_{n+1})g_n\\
&=
   \fnormal_n^1\bigl(\fcapital_{n-1}^1(g_1,g_2,\dots,g_{n-1}),\fcapital_{n-1}^2(g_1,g_2,\dots,g_{n-1}),\fcapital_{n-1}^3(g_1,g_2,\dots,g_{n-1}),g_n\bigr),
    \end{split}
    \end{equation}
    \begin{equation}\llabel{vr3.25}
    \begin{split}
&\fcapital_{n,j}^{2}(g_0,g_1\dots,g_n)=\textstyle\sum\limits_{k=0}^{n}\bigl[(1-\beta_{k+1})\textstyle\bigl(\prod_{l=k+1}^{n}\beta_{l+1}\bigr)|g_{k,j}|^2\bigr]\\
   &=\biggl[\textstyle\sum\limits_{k=0}^{n-1}\bigl[(1-\beta_{k+1})\textstyle\bigl(\prod_{l=k+1}^{n}\beta_{l+1}\bigr)\bigr]|g_{k,j}|^2\biggr]+(1-\beta_{n+1})|g_{n,j}|^2\\
&=\beta_{n+1}\biggl[\textstyle\sum\limits_{k=0}^{n-1}\bigl[(1-\alpha_{k+1})\textstyle\bigl(\prod_{l=k+1}^{n-1}\beta_{l+1}\bigr)|g_{k,j}|^2\bigr]\biggr]+(1-\beta_{n+1})|g_{n,j}|^2\\
&=\beta_{n+1} \fcapital_{n-1,j}^2(g_1,g_2,\dots,g_{n-1})+(1-\beta_{n+1})|g_{n,j}|^2\\
&=
   \fnormal_{n,j}^2\bigl(\fcapital_{n-1}^1(g_1,g_2,\dots,g_{n-1}),\fcapital_{n-1}^2(g_1,g_2,\dots,g_{n-1}),\fcapital_{n,j}^{3}(g_0,g_1\dots,g_n),g_n\bigr),
   \end{split}
    \end{equation}
and
\begin{equation}\llabel{vr3.5}
\begin{split}
    \textstyle
   & \fcapital_{n,j}^{3}(g_0,g_1\dots,g_n)\\
   &=\max_{m\in \{0,1,\dots,n\}} \biggl[\textstyle\sum\limits_{k=0}^{m}\bigl[(1-\beta_{k+1})\textstyle\bigl(\prod_{l=k+1}^{m}\beta_{l+1}\bigr)|g_{k,j}|^2\bigr]\biggr]\\
   &=\max_{m\in\{0,1,\dots,n\}}\fcapital_{m,j}^2(g_0,g_1,\dots,g_m)\\
   &=\max\biggl\{\biggl[\max_{m\in\{0,1,\dots,n-1\}}\fcapital_{m,j}^2(g_0,g_1,\dots,g_m)\biggr],\fcapital_{n,j}^2(g_0,g_1,\dots,g_n)\biggr\}\\
   &=\max\biggl\{\Psi_{n-1,j}^3(g_0,g_1,\dots,g_{n-1}),\beta_{n+1} \fcapital_{n-1,j}^2(g_1,g_2,\dots,g_{n-1})+(1-\beta_{n+1})|g_{n,j}|^2\biggr\}\\
   &= \fnormal_{n,j}^3(\fcapital_{n-1}^1(g_1,g_2,\dots,g_{n-1}),\fcapital_{n-1}^2(g_1,g_2,\dots,g_{n-1}),\fcapital_{n,j}^{3}(g_0,g_1\dots,g_n),g_n\bigr).
   \end{split}
\end{equation}}
    \argument{\lref{def: bff}}{for all $g=((g_{i,j})_{j\in \{1,2,\dots,\fd\}}\allowbreak)_{i\in \{0,1,2,3\}}\in (\R^\fd)^4$, $j\in \{1,2,\dots,\fd\}$ with $\sum_{i=0}^3|g_{i,j}|=0$ that
    \begin{equation}\llabel{vr4}
        \fbold_{n,j}(g)=\gamma_{n+1}\bigl[\textstyle\varepsilon+(g_{2,j})^{1 / 2}\bigr]^{-1} g_{0,j}=0\gamma_{n}\varepsilon^{-1} =0\dott
    \end{equation}}
    \argument{\lref{def: Phi};\lref{def: F};\lref{def: bff}}{for all $n\in \N_0$, $
	g =
	( 
	( g_{ i, j } )_{ j \in \{ 1, 2, \dots, \fd \} }
	\allowbreak)_{
		i \in \{ 0, 1, \dots, n \}
	}
	\in 
	(
	\R^{ 
		\fd
	}
	)^{ n + 1 }
	$, $j\in \{1,2,\dots,\fd\}$ that
    \begin{equation}\llabel{vr5}
    \begin{split}
       &\Phi_n^j(g)= \gamma_{n+1}\biggl(\varepsilon+\max_{m\in \{0,1,\dots,n\}}\biggl[\textstyle\sum\limits_{k=0}^{m}\bigl[(1-\beta_{k+1})\textstyle\bigl(\prod_{l=k+1}^{m}\beta_{l+1}\bigr)|g_{k,i}|^2\bigr]\biggr]^{\nicefrac{1}{2}}\biggr)^{-1}\\
&\quad\cdot\biggl[\textstyle\sum\limits_{k=0}^{n}\bigl[(1-\alpha_{k+1})\textstyle\bigl(\prod_{l=k+1}^{n}\alpha_{l+1}\bigr)g_{k,j}\bigr]\biggr]\\
&=\gamma_{n+1}\bigl[\textstyle\varepsilon+(\fcapital_{n,j}^3(g))^{1 / 2}\bigr]^{-1} \fcapital_{n,j}^1(g)
=\fbold_{n,j}(\fcapital_n^1(g),\fcapital_n^2(g),\fcapital_n^3(g),g_n)\dott
    \end{split}
    \end{equation}}
\argument{\lref{vr1};\lref{vr2};\lref{vr2.1};\lref{vr3};\lref{vr3.25};\lref{vr3.5};\lref{vr4};\lref{vr5};\cref{general SGD} (applied with $\numf\curvearrowleft 3$, $\fnormal\curvearrowleft \fnormal$, $\fcapital\curvearrowleft \fcapital$, $\fbold\curvearrowleft\fbold$, $\Phi\curvearrowleft \Phi$, $\fG\curvearrowleft\fG$, $\Theta\curvearrowleft\Theta$, $\bfM^1\curvearrowleft \bfm$, $\bfM^2\curvearrowleft\bbM$, $\bfM^3\curvearrowleft\fM$ in the notation of \cref{general SGD})}{\lref{item 1} and that there exist $\bfm = (\bfm^1 , \ldots , \bfm ^{ \fd } )  \colon \N_0 \times \Omega \to \R^\fd$,
	 $\bbM = (\bbM^1 , \ldots , \bbM^\fd )  \colon \N_0 \times \Omega \to \R^\fd$, and $\fM=(\fM^{1}, \ldots, \fM^{\fd})\colon \N_0\times \Omega \rightarrow \mathbb{R}^{\fd}$ which
	satisfy for all
	$n \in \N$,
	$j \in \{1,2,\dots,\fd\}$ that
	\begin{equation}\llabel{eqt}
\bfm_0 = \bbM_0 =\fM_0 =0,\qquad \mathbf{m}_{n}=\alpha_{n} \mathbf{m}_{n-1}+(1-\alpha_{n}) \fG_n(\Theta_{n-1}), 
\end{equation}
\begin{equation}\llabel{eqt'}
\mathbb{M}_{n}^{j}=\beta_{n} \mathbb{M}_{n-1}^{j}+(1-\beta_{n})|\fG_n^j(\Theta_{n})|^2,
\end{equation}
\begin{equation}\llabel{eqt'1}
\fM_n^j=\max\{\fM_{n-1}^j,\beta_{n} \mathbb{M}_{n-1}^{j}+(1-\beta_{n})|\fG_n^j(\Theta_{n})|^2\},
\end{equation}
\begin{equation}\llabel{eqt''}
\text { and } \qquad \Theta_n^{j}=\Theta_{n-1}^{j}-\gamma_{n}\bigl[\varepsilon+(\fM_n^{j})^{1 / 2}\bigr]^{-1} \bfm_n^{j}.
 \end{equation}}
 \startnewargseq
 \argument{\lref{eqt};\lref{eqt'};\lref{eqt'1};\lref{eqt''}}{that for all $n\in \N$, $i\in \{1,2,\dots,\fd\}$ it holds that \begin{equation}
		\begin{split} 
  \bfm_0 = \bbM_0 =\fM_0 =0,\qquad
\mathbf{m}_{n}=\alpha_{n} \mathbf{m}_{n-1}+(1-\alpha_{n}) \fG_n(\Theta_{n-1}),
\end{split}
\end{equation}
\begin{equation}
    \begin{split}
\mathbb{M}_{n}^{j}=\beta_{n} \mathbb{M}_{n-1}^{j}+(1-\beta_{n})|\fG_n^j(\Theta_{n})|^2, \qquad
\fM_n^j=\max\{\fM_{n-1}^j,\bbM_n^j\},
\end{split}
\end{equation}
\begin{equation}\llabel{eqt2}
\text { and } \qquad \Theta_n^{j}=\Theta_{n-1}^{j}-\gamma_{n}\bigl[\varepsilon+(\fM_n^{j})^{1 / 2}\bigr]^{-1} \bfm_n^{j}.
 \end{equation}}
 \argument{\lref{eqt2};}{\lref{item 2}\dott}
\end{aproof}
\subsection{Nesterov accelerated adaptive moment estimation SGD (Nadam) optimizer}\label{subsec: Nadam}
In this subsection we show in the elementary result in \cref{draf nadam} below that the \Nadam\ optimizer (cf.\ \cite{Dozat2015IncorporatingNM}) 
satisfies the assumptions of \cref{conjecture: multilayer2} (cf.\ \cref{eq:gradients_SGD4}--\cref{eq: setting: SGD_def process_main theorem 2} in \cref{conjecture: multilayer2} and \cref{setting: multilayer: eq:gradients_SGD4:}--\cref{eq: setting: SGD_def process} in \cref{setting: SGD}).
\begin{athm}{prop}{draf nadam}
   Let $\fd\in \N$, $(\gamma_n)_{n \in \N } \subseteq [0 , \infty )$,
	$(\alpha_n)_{n \in \N} \subseteq [0 , 1 ]$,
	$(\beta_n)_{n \in \N } \subseteq [0 , 1 ]$,
	$\varepsilon \in (0 , \infty ) $ satisfy $\max \cu{ \alpha_1, \beta_1 } < 1$, for every $n\in \N_0$ let $\Phi_n=(\Phi_n^1,\dots,\Phi_n^\fd)\colon (\R^\fd)^{n+1}\to \R^\fd$ satisfy for all $
	g =
	( 
	( g_{ i, j } )_{ j \in \{ 1, 2, \dots, \fd \} }
	)_{
		i \in \{ 0, 1, \dots, n \}
	}
	\in 
	(
	\R^{ 
		\fd
	}
	)^{ n + 1 }
	$, $j\in \{1,2,\dots,\fd\}$ that
    \begin{equation}\llabel{def: Phi}
    \begin{split}
    \textstyle
&\Phi_n^j(g)=\gamma_{n+1}\biggl(\varepsilon+\biggl[\frac{\sum_{k=0}^{n}\bigl[(1-\beta_{k+1})\textstyle\bigl(\prod_{l=k+1}^{n}\beta_{l+1}\bigr)|g_{k,i}|^2\bigr]}{1-\prod_{l=0}^n\beta_{l+1}}\biggr]^{\nicefrac{1}{2}}\biggr)^{-1}\\
&\quad\cdot\biggl[\frac{\alpha_{n+2}\alpha_{n+1}\mathbbm 1_{(0,\infty)}(n)\bigl[\sum_{k=0}^{n-1}\bigl[(1-\alpha_{k+1})\textstyle\bigl(\prod_{l=k+1}^{n-1}\alpha_{l+1}\bigr)g_{k,j}\bigr]\bigr]}{1-\prod_{l=0}^{n+1}\alpha_{l+1}}\\
&\quad +\Bigl[\frac{1-\alpha_{n+1}}{1-\prod_{l=0}^{n}\alpha_{l+1}}+\frac{\alpha_{n+2}(1-\alpha_{n+1})}{1-\prod_{l=0}^{n+1}\alpha_{l+1}}\Bigr]g_{n,j}\bigg],
\end{split}
    \end{equation}
    let $(\Omega,\mathcal F,\P)$ be a probability space, for every $n\in \N_0$ let $\fG_n=(\fG_n^1,\ldots,\fG_n^\fd)\colon \R^\fd\times \Omega\to \R^\fd$ be a function, and let $\Theta\colon \N_0\times\Omega\to \R^\fd$ satisfy for all $n\in \N$ that
\begin{equation}\llabel{eq2}
    \Theta_{n} = \Theta_{n-1} - \Phi_{n-1} ( \fG_1 ( \Theta_0 ) , \fG_2 ( \Theta_1 ) , \ldots, \fG_n ( \Theta_{n-1} )).
\end{equation}
    Then
    \begin{enumerate}[label=(\roman*)]
        \item \llabel{item 1} it holds for all $n\in\N_0$, $
	g =
	( 
	( g_{ i, j } )_{ j \in \{ 1, 2, \dots, \fd \} }
	)_{
		i \in \{ 0, 1, \dots, n \}
	}
	\in 
	(
	\R^{ 
		\fd
	}
	)^{ n + 1 }
	$, 
	$ 
	j \in \{1,2,\dots,\fd\}  
	$ with $\sum_{i=0}^n|g_{i,j}|=0$ that
        $\Phi_n^j(g)=0$ and
        \item \llabel{item 2} it holds that there exist $\bfm = (\bfm^1 , \ldots , \bfm ^{ \fd } )  \colon \N_0 \times \Omega \to \R^\fd$,
 $\mathfrak{m}=(\mathfrak{m}^{1}, \ldots, \mathfrak{m}^{\fd})\colon \N_0\times \Omega \rightarrow \mathbb{R}^{\fd}$, and
	$\bbM = (\bbM^1 , \ldots , \bbM^\fd )  \colon \N_0 \times \Omega \to \R^\fd$ which
	satisfy for all
	$n \in \N$,
	$j \in \{1,2,\dots,\fd\}$ that
	\begin{equation}
\bfm_0 = \bbM_0 = 0,\qquad \mathbf{m}_{n}=\alpha_{n-1} \mathbf{m}_{n}+(1-\alpha_{n}) \fG_n(\Theta_{n-1}), 
\end{equation}
\begin{equation}
\mathfrak{m}_{n}=\left[\frac{1-\alpha_{n}}{1-\prod_{l=0}^{n-1} \alpha_{l+1}}\right] \fG_n(\Theta_{n-1})+\bigg[\frac{\alpha_{n+1}}{1-\prod_{l=0}^{n} \alpha_{l+1}}\bigg] \mathbf{m}_{n},
\end{equation}
\begin{equation}
\mathbb{M}_{n}^{j}=\beta_{n} \mathbb{M}_{n}^{j}+(1-\beta_{n})|\fG_n^j(\Theta_{n-1})|^2, 
\end{equation}
\begin{equation}
\text { and } \qquad \Theta_n^{j}=\Theta_{n-1}^{j}-\gamma_{n}\left[\textstyle\varepsilon+\left[\frac{\mathbb{M}_n^{j}}{\left(1-\prod_{l=1}^{n} \beta_l\right)}\right]^{1 / 2}\right]^{-1} \mathfrak{m}_n^{j}.
 \end{equation}
    \end{enumerate}
\end{athm}
\begin{aproof}
     Throughout this proof for every $n\in \N_0$ let $\fnormal_n^1=(\fnormal_{n,1}^1,\dots,\fnormal_{n,\fd}^1)$, $\fnormal_n^2=(\fnormal_{n,1}^2,\dots,\fnormal_{n,\fd}^2), \fnormal_n^3=(\fnormal_{n,1}^3,\dots,\fnormal_{n,\fd}^3)\colon (\R^\fd)^4\to\R^\fd$ satisfy for all $g=((g_{i,j})_{j\in\{1,2,\dots,\fd\}})_{i\in\{0,1,2,3\}}\allowbreak\in (\R^\fd)^4$, $j\in \{1,2,\dots,\fd\}$ that 
     \begin{equation}\llabel{def: f}
         \fnormal_n^1(g)=\alpha_{n+1} g_0+(1-\alpha_{n+1})g_3,\qquad \fnormal_{n,j}^3(g)=\beta_{n+1} g_{2,j}+(1-\beta_{n+1})|g_{3,j}|^2,
     \end{equation}
     and
\begin{equation}\llabel{def: f1}
    \fnormal_{n}^2(g)=\Bigl[\frac{1-\alpha_{n+1}}{1-\prod_{l=0}^{n}\alpha_{l+1}}+\frac{\alpha_{n+2}(1-\alpha_{n+1})}{1-\prod_{l=0}^{n+1}\alpha_{l+1}}\Bigr]g_3+\frac{\alpha_{n+2}\alpha_{n+1}}{1-\prod_{l=0}^{n+1}\alpha_{l+1}}g_0,
\end{equation} 
let $\fcapital_n^1=(\fcapital_{n,1}^1,\ldots,\fcapital_{n,\fd}^1)\colon (\R^{\fd})^{n+1}\to\R^\fd$, $\fcapital_n^2=(\fcapital_{n,1}^2,\dots,\fcapital_{n,\fd}^2)\colon (\R^{\fd})^{n+1}\to\R^\fd$, $\fcapital_n^3=(\fcapital_{n,1}^3,\dots,\fcapital_{n,\fd}^3)\colon (\R^{\fd})^{n+1}\to\R^\fd$ satisfy for all $g=((g_{i,j})_{j\in \{1,2,\dots,\fd\}}\allowbreak)_{i\in \{0,1,\dots,n\}}\in (\R^{\fd})^{n+1}$, $j\in \{1,2,\dots,\fd\}$ that 
\begin{equation}\llabel{def: F}
    \fcapital_n^1(g)=\sum_{k=0}^{n}\bigl[(1-\alpha_{k+1})\textstyle\bigl(\prod_{l=k+1}^{n}\alpha_{l+1}\bigr)g_k\bigr],
    \end{equation}
    \begin{equation}\llabel{def: F1}
    \begin{split}
    \fcapital_{n}^2(g)&= \frac{\alpha_{n+2}\alpha_{n+1}\mathbbm 1_{(0,\infty)}(n)\bigl[\sum_{k=0}^{n-1}\bigl[(1-\alpha_k)\textstyle\bigl(\prod_{l=k+1}^{n-1}\alpha_{l+1}\bigr)g_{k,i}\bigr]\bigr]}{1-\prod_{l=0}^{n+1}\alpha_{l+1}}\\
    &\quad+\Bigl[\frac{1-\alpha_{n+1}}{1-\prod_{l=0}^{n}\alpha_{l+1}}+\frac{\alpha_{n+2}(1-\alpha_{n+1})}{1-\prod_{l=0}^{n+1}\alpha_{l+1}}\Bigr]g_{n},
    \end{split}
     \end{equation}
     \begin{equation}\llabel{def: F2}
     \text{and}\qquad\fcapital_{n,j}^3(g)=\sum_{k=0}^{n}\bigl[(1-\beta_{k+1})\textstyle\bigl(\prod_{l=k+1}^{n}\beta_{l+1}\bigr)|g_{k,j}|^2\bigr],
    \end{equation}
and let $\fbold_n=(\fbold_{n,1},\dots,\fbold_{n,\fd})\colon (\R^\fd)^4\to\R^\fd$ satisfy for all $g=((g_{i,j})_{j\in \{1,2,\dots,\fd\}})_{i\in\{0,1,2,3\}}\in (\R^\fd)^4$, $j\in \{1,2,\dots,\fd\}$ that 
\begin{equation}\llabel{def: bff}
    \fbold_{n,j}(g)=\gamma_{n+1}\biggl[\textstyle\varepsilon+\left[\frac{g_{2,j}}{\left(1-\prod_{l=0}^{n} \beta_{l+1}\right)}\right]^{1 / 2}\biggr]^{-1} g_{1,j}.
\end{equation}
\argument{\lref{def: f};\lref{def: f1}}{that for all $n\in \N_0$, $g=((g_{i,j})_{j\in \{1,2,\dots,\fd\}})_{i\in\{0,1,2,3\}}\in (\R^\fd)^4$, $j\in \{1,2,\dots,\fd\}$ with $\sum_{i=0}^3|g_{i,j}|=0$ it holds that
\begin{equation}\llabel{vr1}
    \fnormal_{n,j}^1(g)=\alpha _{n+1} g_{0,j}+(1-\alpha_{n+1}) g_{3,j}=0\alpha_{n+1}+0(1-\alpha_{n+1})=0,
    \end{equation}
    \begin{equation}
    \begin{split}
    \fnormal_{n,j}^2(g)&=\Bigl[\frac{1-\alpha_{n+1}}{1-\prod_{l=0}^{n}\alpha_{l+1}}+\frac{\alpha_{n+2}(1-\alpha_{n+1})}{1-\prod_{l=0}^{n+1}\alpha_{l+1}}\Bigr]g_{3,j}+\frac{\alpha_{n+2}\alpha_{n+1}}{1-\prod_{l=0}^{n+1}\alpha_{l+1}}g_{0,j}\\
   &=0\Bigl[\frac{1-\alpha_{n}}{1-\prod_{l=0}^{n}\alpha_{l+1}}+\frac{\alpha_{n+2}(1-\alpha_{n+1})}{1-\prod_{l=0}^{n+1}\alpha_{l+1}}\Bigr]+0\Bigl[\frac{\alpha_{n+2}\alpha_{n+1}}{1-\prod_{l=0}^{n+1}\alpha_{l+1}}\Bigr]=0,
   \end{split}
   \end{equation}
   and
   \begin{equation}
   \fnormal_{n,j}^3(g)=\beta_{n+1}g_{2,j}+(1-\beta_{n+1})|g_{3,j}|^2=0\beta_{n+1}+0(1-\beta_{n+1})=0
    \dott
\end{equation}}
\argument{\lref{def: f};\lref{def: F};\lref{def: F2}}{for all $n\in \N\backslash\{1\}$, $
	g =
	( 
	( g_{ i, j } )_{ j \in \{ 1, 2, \dots, \fd \} }
	\allowbreak)_{
		i \in \{ 0, 1, \dots, n \}
	}
	\in 
	(
	\R^{ 
		\fd
	}
	)^{ n + 1 }
	$, $j\in \{1,2,\dots,\fd\}$, $y=(y_1,\dots,y_\fd)\in \R^\fd$ that
 \begin{equation}\llabel{vr2}
    \fcapital_0^1(y)=(1-\alpha_1)y =\fnormal_0^1(0,0,0,y),
    \end{equation}
    \begin{equation}\llabel{vr2.1}
    \fcapital_0^2(y)=\Bigl[\frac{1-\alpha_1}{1-\alpha_1}+\frac{\alpha_1(1-\alpha_1)}{1-\alpha_2\alpha_1}\Bigr]y=\fnormal_{0}^2(0,0,0,y),
    \end{equation}
    \begin{equation}\llabel{vr2.2}
        \fcapital_{0,j}^3(y)=(1-\beta_1)|y_j|^2 =\fnormal_{0,j}^3(0,0,0,y_j),
    \end{equation}
\begin{equation}\llabel{vr3}
\begin{split}
\textstyle
   & \fcapital_n^{1}(g_0,g_1\dots,g_n)=\textstyle\sum\limits_{k=0}^{n}\bigl[(1-\alpha_{k+1})\textstyle\bigl(\prod_{l=k+1}^{n}\alpha_{l+1}\bigr)g_k\bigr]\\
   &=\biggl[\textstyle\sum\limits_{k=0}^{n-1}\bigl[(1-\alpha_{k+1})\textstyle\bigl(\prod_{l=k+1}^{n}\alpha_{l+1}\bigr)\bigr]g_k\biggr]+(1-\alpha_{n+1})g_{n}\\
&=\alpha_{n+1}\biggl[\textstyle\sum\limits_{k=0}^{n-1}\bigl[(1-\alpha_{k+1})\textstyle\bigl(\prod_{l=k+1}^{n-1}\alpha_{l+1}\bigr)g_k\bigr]\biggr]+(1-\alpha_{n+1})g_{n}\\
&=\alpha_{n+1} \fcapital_{n-1}^1(g_1,g_2,\dots,g_{n-1})+(1-\alpha_{n+1})g_n\\
&=
   \fnormal_n^1\bigl(\fcapital_{n-1}^1(g_1,g_2,\dots,g_{n-1}),\fcapital_{n-1}^2(g_1,g_2,\dots,g_{n-1}),\fcapital_{n-1}^3(g_1,g_2,\dots,g_{n-1}),g_n\bigr),
    \end{split}
    \end{equation}
    \begin{equation}\llabel{vr3.25}
    \begin{split}
&\fcapital_n^{2}(g_0,g_1\dots,g_n)\\
&=\frac{\alpha_{n+2}\alpha_{n+1}\bigl[\sum_{k=0}^{n-1}\bigl[(1-\alpha_{k+1})\textstyle\bigl(\prod_{l=k+1}^{n-1}\alpha_{l+1}\bigr)g_{k,i}\bigr]\bigr]}{1-\prod_{l=0}^{n+1}\alpha_{l+1}}\\
    &\quad+\Bigl[\frac{1-\alpha_{n+1}}{1-\prod_{l=0}^{n}\alpha_{l+1}}+\frac{\alpha_{n+2}(1-\alpha_{n+1})}{1-\prod_{l=0}^{n+1}\alpha_{l+1}}\Bigr]g_{n}\\
    &=\Bigl[\frac{1-\alpha_{n+1}}{1-\prod_{l=0}^{n}\alpha_{l+1}}+\frac{\alpha_{n+2}(1-\alpha_{n+1})}{1-\prod_{l=0}^{n+1}\alpha_{l+1}}\Bigr]g_n+\frac{\alpha_{n+2}\alpha_{n+1}}{1-\prod_{l=0}^{n+1}\alpha_{l+1}}\fcapital_n^1(g_1,g_2,\dots,g_{n-1})\\
    &=\fnormal_n^2\bigl(\fcapital_{n-1}^1(g_1,g_2,\dots,g_{n-1}),\fcapital_{n-1}^2(g_1,g_2,\dots,g_{n-1}),\fcapital_{n-1}^3(g_1,g_2,\dots,g_{n-1}),g_n\bigr),
   \end{split}
    \end{equation}
and
\begin{equation}\llabel{vr3.5}
\begin{split}
    \textstyle
   & \fcapital_{n,j}^{3}(g_0,g_1\dots,g_n)=\textstyle\sum\limits_{k=0}^{n}\bigl[(1-\beta_{k+1})\textstyle\bigl(\prod_{l=k+1}^{n}\beta_{l+1}\bigr)|g_{k,j}|^2\bigr]\\
   &=\biggl[\textstyle\sum\limits_{k=0}^{n-1}\bigl[(1-\beta_{k+1})\textstyle\bigl(\prod_{l=k+1}^{n}\beta_{l+1}\bigr)\bigr]|g_{k,j}|^2\biggr]+(1-\beta_{n+1})|g_{n,j}|^2\\
&=\beta_{n+1}\biggl[\textstyle\sum\limits_{k=0}^{n-1}\bigl[(1-\alpha_{k+1})\textstyle\bigl(\prod_{l=k+1}^{n-1}\beta_{l+1}\bigr)|g_{k,j}|^2\bigr]\biggr]+(1-\beta_{n+1})|g_{n,j}|^2\\
&=\beta_{n+1} \fcapital_{n-1,j}^2(g_1,g_2,\dots,g_{n-1})+(1-\beta_{n+1})|g_{n,j}|^2\\
&=
   \fnormal_{n,j}^3\bigl(\fcapital_{n-1}^1(g_1,g_2,\dots,g_{n-1}),\fcapital_{n-1}^2(g_1,g_2,\dots,g_{n-1}),\fcapital_{n,j}^{3}(g_0,g_1\dots,g_n),g_n\bigr).
   \end{split}
\end{equation}}
    \argument{\lref{def: bff}}{for all $g=((g_{i,j})_{j\in \{1,2,\dots,\fd\}}\allowbreak)_{i\in \{0,1,2,3\}}\in (\R^\fd)^4$, $j\in \{1,2,\dots,\fd\}$ with $\sum_{i=0}^3|g_{i,j}|=0$ that
    \begin{equation}\llabel{vr4}
        \fbold_{n,j}(g)=\gamma_{n+1}\biggl[\textstyle\varepsilon+\left[\frac{g_{2,j}}{\left(1-\prod_{l=0}^{n} \beta_{l+1}\right)}\right]^{1 / 2}\biggr]^{-1} g_{1,j}=0\dott
    \end{equation}}
    \argument{\lref{def: Phi};\lref{def: F};\lref{def: F1};\lref{def: F2};\lref{def: bff}}{for all $n\in \N_0$, $
	g =
	( 
	( g_{ i, j } )_{ j \in \{ 1, 2, \dots, \fd \} }
	\allowbreak)_{
		i \in \{ 0, 1, \dots, n \}
	}
	\in 
	(
	\R^{ 
		\fd
	}
	)^{ n + 1 }
	$, $j\in \{1,2,\dots,\fd\}$ that
    \begin{equation}\llabel{vr5}
    \begin{split}
       \Phi_n^j(g)&= \gamma_{n+1}\biggl(\varepsilon+\biggl[\frac{\sum_{k=0}^{n}\bigl[(1-\beta_{k+1})\textstyle\bigl(\prod_{l=k+1}^{n}\beta_{l+1}\bigr)|g_{k,j}|^2\bigr]}{1-\prod_{l=0}^n\beta_{l+1}}\biggr]^{\nicefrac{1}{2}}\biggr)^{-1}\\
&\quad\cdot\biggl[\frac{\alpha_{n+2}\alpha_{n+1}\mathbbm 1_{(0,\infty)}(n)\bigl[\sum_{k=0}^{n-1}\bigl[(1-\alpha_k)\textstyle\bigl(\prod_{l=k+1}^{n-1}\alpha_{l+1}\bigr)g_{k,j}\bigr]\bigr]}{1-\prod_{l=0}^{n+1}\alpha_{l+1}}\\
&\quad+\Bigl[\frac{1-\alpha_{n+1}}{1-\prod_{l=0}^{n}\alpha_{l+1}}+\frac{\alpha_{n+2}(1-\alpha_{n+1})}{1-\prod_{l=0}^{n+1}\alpha_{l+1}}\Bigr]g_{n,i}\bigg]\\
&=\gamma_{n+1}\biggl[\textstyle\varepsilon+\left[\frac{\fcapital_{n,j}^3(g)}{\left(1-\prod_{l=0}^{n} \beta_{l+1}\right)}\right]^{1 / 2}\biggr]^{-1} \fcapital_{n,j}^2(g)
=\fbold_{n,j}(\fcapital_n^1(g),\fcapital_n^2(g),\fcapital_n^3(g),g_n)\dott
    \end{split}
    \end{equation}}
\argument{\lref{vr1};\lref{vr2};\lref{vr2.1};\lref{vr2.2};\lref{vr3};\lref{vr3.25};\lref{vr3.5};\lref{vr4};\lref{vr5};\cref{general SGD} (applied with $\numf\curvearrowleft 3$, $\fnormal\curvearrowleft \fnormal$, $\fcapital\curvearrowleft \fcapital$, $\fbold\curvearrowleft\fbold$, $\Phi\curvearrowleft \Phi$, $\fG\curvearrowleft\fG$, $\Theta\curvearrowleft\Theta$, $\bfM^1\curvearrowleft \bfm$, $\bfM^2\curvearrowleft\mathfrak{m}$, $\bfM^3\curvearrowleft\mathbb M$ in the notation of \cref{general SGD})}{\lref{item 1} and that there exist $\bfm = (\bfm^1 , \ldots , \bfm ^{ \fd } )  \colon \N_0 \times \Omega \to \R^\fd$,
 $\mathfrak{m}=(\mathfrak{m}^{1}, \ldots, \mathfrak{m}^{\fd})\colon \N_0\times \Omega \rightarrow \mathbb{R}^{\fd}$, and
	$\bbM = (\bbM^1 , \ldots , \bbM^\fd )  \colon \N_0 \times \Omega \to \R^\fd$ which
	satisfy for all
	$n \in \N$,
	$j \in \{1,2,\dots,\fd\}$ that
	\begin{equation}\llabel{eqt} 
\bfm_0 = \mathfrak{m}_0=\bbM_0 = 0,\qquad \mathbf{m}_{n}=\alpha_{n-1} \mathbf{m}_{n}+(1-\alpha_{n}) \fG_n(\Theta_{n-1}),
\end{equation}
\begin{equation}\llabel{eqt'}
\mathfrak{m}_{n}=\Bigl[\frac{1-\alpha_{n}}{1-\prod_{l=0}^{n-1}\alpha_{l+1}}+\frac{\alpha_{n+1}(1-\alpha_{n})}{1-\prod_{l=0}^{n}\alpha_{l+1}}\Bigr]\fG_n(\Theta_{n-1})+\frac{\alpha_{n+1}\alpha_{n}}{1-\prod_{l=0}^{n}\alpha_{l+1}}\mathbf{m}_{n-1},
\end{equation}
\begin{equation}\llabel{eqt''}
\mathbb{M}_{n}^{j}=\beta_{n} \mathbb{M}_{n-1}^{j}+(1-\beta_{n})|\fG_n^j(\Theta_{n-1})|^2, 
\end{equation}
\begin{equation}\llabel{eqt''1}
\text { and } \qquad \Theta_n^{j}=\Theta_{n-1}^{j}-\gamma_{n}\left[\textstyle\varepsilon+\left[\frac{\mathbb{M}_n^{j}}{\left(1-\prod_{l=0}^{n-1} \beta_l\right)}\right]^{1 / 2}\right]^{-1} \mathfrak{m}_n^{j}.
 \end{equation}}
 \startnewargseq
 \argument{\lref{eqt};\lref{eqt'};\lref{eqt''};\lref{eqt''1}}{that for all $n\in \N$, $i\in \{1,2,\dots,\fd\}$ it holds that
	\begin{equation}\llabel{eqt2} 
\bfm_0 = \mathfrak{m}_0=\bbM_0 = 0,\qquad \mathbf{m}_{n}=\alpha_{n} \mathbf{m}_{n-1}+(1-\alpha_{n}) \fG_n(\Theta_{n-1}),
\end{equation}
\begin{equation}\llabel{eqt3}
\mathfrak{m}_{n}=\left[\frac{1-\alpha_{n}}{1-\prod_{l=0}^{n-1} \alpha_{l+1}}\right] \fG_n(\Theta_{n-1})+\bigg[\frac{\alpha_{n+1}}{1-\prod_{l=0}^{n} \alpha_{l+1}}\bigg] \mathbf{m}_{n},
\end{equation}
\begin{equation}\llabel{eqt4}
\mathbb{M}_{n}^{j}=\beta_{n} \mathbb{M}_{n-1}^{j}+(1-\beta_{n})|\fG_n^j(\Theta_{n-1})|^2,
\end{equation}
\begin{equation}\llabel{eqt5}
\text{and}\qquad\Theta_n^{j}=\Theta_{n-1}^{j}-\gamma_{n}\left[\textstyle\varepsilon+\left[\frac{\mathbb{M}_n^i}{\left(1-\prod_{l=0}^{n-1} \beta_{l+1}\right)}\right]^{1 / 2}\right]^{-1} \mathfrak{m}_n^{j}\dott
 \end{equation}}
 \argument{\lref{eqt5};}{\lref{item 2}\dott}
\end{aproof}
\subsection*{Acknowledgements}
This work has been partially funded by the European Union (ERC, MONTECARLO, 101045811). The views and the opinions expressed in this work are however those of the authors only and do not necessarily reflect those of the European Union or the European Research Council (ERC). Neither the European Union nor the granting authority can be held responsible for them. Moreover, we gratefully acknowledge the Cluster of Excellence EXC 2044-390685587, Mathematics Münster: Dynamics-Geometry-Structure funded by the Deutsche Forschungsgemeinschaft (DFG, German Research Foundation).
\bibliographystyle{acm}
\bibliography{bibfile}
\end{document}